%% file: main.tex
\documentclass[en,oneside,onehalfspacing,phd]{risethesis}

\usepackage{babel}
\usepackage[utf8]{inputenc}
\usepackage[linesnumbered,vlined,ruled,boxed,commentsnumbered]{algorithm2e}
\usepackage{algpseudocode}
\usepackage{bm} 
\usepackage{amsmath,amssymb,amsfonts}

\usepackage{commath}
\usepackage{float}
\usepackage[flushleft]{threeparttable}
\usepackage{colortbl}
\usepackage{color}
\usepackage[table]{xcolor}
\usepackage{microtype}
\usepackage{bibentry}
\usepackage{subfig}
\usepackage{multirow}
\usepackage{rotating}
\usepackage{booktabs}
\usepackage{pdfpages}
\usepackage{lipsum}
\usepackage{scalefnt}
\let\counterwithout\relax

\usepackage{chngcntr}
\usepackage[tocflat,toctextentriesindented]{tocstyle}
\deactivatetocstyle[lof]
\deactivatetocstyle[lot]
\deactivatetocstyle[loa]
\usetocstyle{allwithdot}
\settocfeature[toc][0]{entryhook}{\bfseries\uppercase}
\settocfeature[toc][1]{entryhook}{\uppercase}
\settocfeature[toc][2]{entryhook}{\bfseries}
\settocfeature[toc][3]{entryhook}{\itshape}
\settocfeature{spaceafternumber}{3em}
\makeatletter
\renewcommand*\l@figure{\@dottedtocline{1}{1.5em}{5.9em}}
\renewcommand*\l@table{\@dottedtocline{1}{1.5em}{5em}}
\makeatother
\usepackage{titlesec}
\titleformat{\section}{\normalfont\large}{\thesection}{1em}{\MakeUppercase}
\titleformat{\subsection}{\normalfont\large}{\bfseries\thesubsection}{1em}{\bfseries}
\titleformat{\subsubsection}{\normalfont\large}{\itshape\thesubsubsection}{1em}{\itshape}

\DeclareCaptionLabelSeparator{colon}{ - }

\newcolumntype{L}[1]{>{\raggedright\let\newline\\\arraybackslash\hspace{0pt}}m{#1}}
\newcolumntype{C}[1]{>{\centering\let\newline\\\arraybackslash\hspace{0pt}}m{#1}}
\newcolumntype{R}[1]{>{\raggedleft\let\newline\\\arraybackslash\hspace{0pt}}m{#1}}

\newglossarystyle{modsuper}{%
  \setglossarystyle{super}%
  
  {
    \begin{longtable}
        {L{0.3\textwidth}L{0.7\textwidth}}}%
    {\end{longtable}
    \addtocounter{table}{-1}
  }
}

\usepackage[linkcolor=black,
            citecolor=blue,
            colorlinks=true,
            backref=page,
            pdfpagelabels,
            pdftitle={PhD Thesis - David Lopes de Macêdo},
            pdfauthor={David Lopes de Macêdo}]{hyperref}

\usepackage[colorlinks=true,citecolor=blue]{hyperref}%

\address{Recife}

\universitypt{Universidade Federal de Pernambuco}
\universityen{Federal University of Pernambuco}

\departmentpt{Centro de Informática}
\departmenten{Center of Informatics}

\programpt{Pós-graduação em Ciência da Computação}
\programen{Graduate in Computer Science}

\majorfieldpt{Ciência da Computação}
\majorfielden{Computer Science}

\concentrationareapt{Inteligência Computational}
\concentrationareaen{Computational Intelligence}

\title{Towards Robust Deep Learning}

\date{2022}

\author{David Lopes de Macêdo}
\adviser{Teresa Bernarda Ludermir}
\coadviser{Cleber Zanchettin}

\def\x{\checkmark}

\input{default_content/acronyms}
\input{default_content/list_symbols}
\makenoidxglossaries

\counterwithout{figure}{chapter}
\counterwithout{table}{chapter}

\usepackage{centernot}
\usepackage{comment}
\usepackage{tabularx}
\usepackage{lipsum}
\newcolumntype{Y}{>{\centering\arraybackslash}X}
\usepackage{ragged2e}

\usepackage{siunitx}
\usepackage{bibunits}
\usepackage[font=small]{caption} 
\renewcommand{\small}{\fontsize{9pt}{9pt}\selectfont}
\renewcommand{\footnotesize}{\fontsize{7pt}{7pt}\selectfont}

\usepackage{etoolbox}
\gappto{\UrlBreaks}{\UrlOrds}
\usepackage{pgfgantt}

\usepackage{makecell}

\usepackage{listings}

\usepackage{soul}
\sethlcolor{white} 
\DeclareRobustCommand{\hln}[1]{{\sethlcolor{white}\hl{#1}}}
\DeclareRobustCommand{\hlf}[1]{{\sethlcolor{white}\hl{#1}}}

\usepackage{framed}
\colorlet{shadecolor}{lightgray}
\usepackage[modulo]{lineno}

{\end{shaded}}

\newcommand\blfootnote[1]{%
  \begingroup
  \renewcommand\thefootnote{}\footnote{#1}%
  \addtocounter{footnote}{-1}%
  \endgroup
}

\usepackage{tikz}
\usetikzlibrary{backgrounds}
\usetikzlibrary{arrows,shapes}
\usetikzlibrary{tikzmark}
\usetikzlibrary{calc}
\usepackage{amsmath}
\usepackage{amsthm}
\usepackage{amssymb}
\usepackage{mathtools, nccmath}
\usepackage{wrapfig}

\usepackage{blindtext}

\usepackage{graphicx}

\usepackage{xspace}

\usepackage{array}
\usepackage{ragged2e}
\newcolumntype{P}[1]{>{\RaggedRight\hspace{0pt}}p{#1}}

\usepackage{tcolorbox}

\usepackage{tikz}
\usetikzlibrary{arrows,shapes,positioning,shadows,trees,mindmap}
\usepackage[edges]{forest}
\usetikzlibrary{arrows.meta}
\colorlet{linecol}{black!75}
\usepackage{xkcdcolors} 

\usepackage{tikz}
\usetikzlibrary{backgrounds}
\usetikzlibrary{arrows,shapes}
\usetikzlibrary{tikzmark}
\usetikzlibrary{calc}
\newcommand{\highlight}[2]{\colorbox{#1!17}{$\displaystyle #2$}}

\renewcommand{\highlight}[2]{\colorbox{#1!17}{#2}}

\usepackage{tikz}
\usepackage{pgfplots}
\usepackage{tikz-3dplot}
\tdplotsetmaincoords{60}{115}
\pgfplotsset{compat=newest}

\DeclarePairedDelimiterX{\infdivx}[2]{(}{)}{%
  #1\;\delimsize|\delimsize|\;#2%
}
\newcommand{\kld}[2]{\ensuremath{D_{KL}\infdivx{#1}{#2}}\xspace}

\begin{document}

\frontmatter
\includepdf[pages=-]{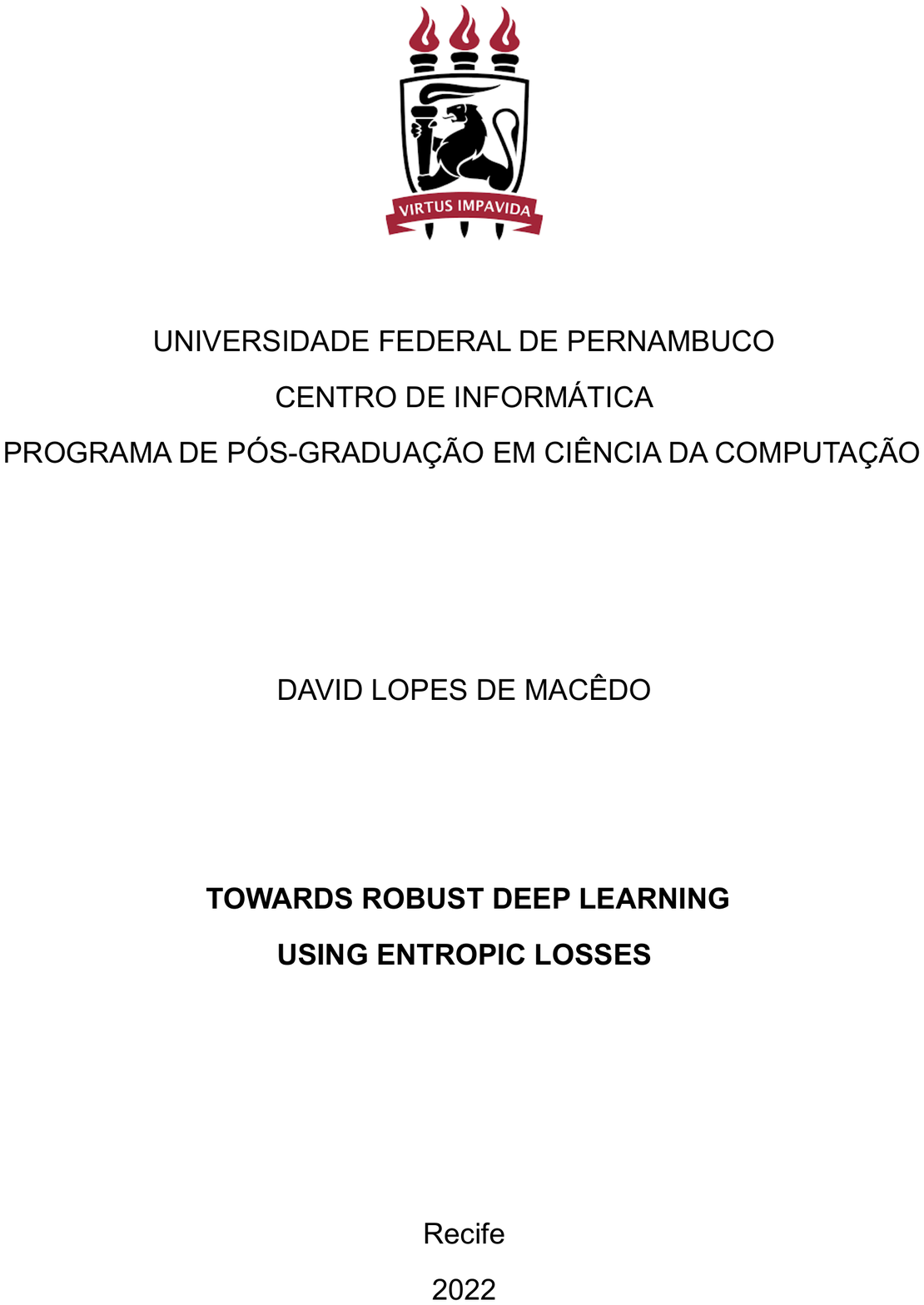}


\presentationpage


\includepdf[pages=-]{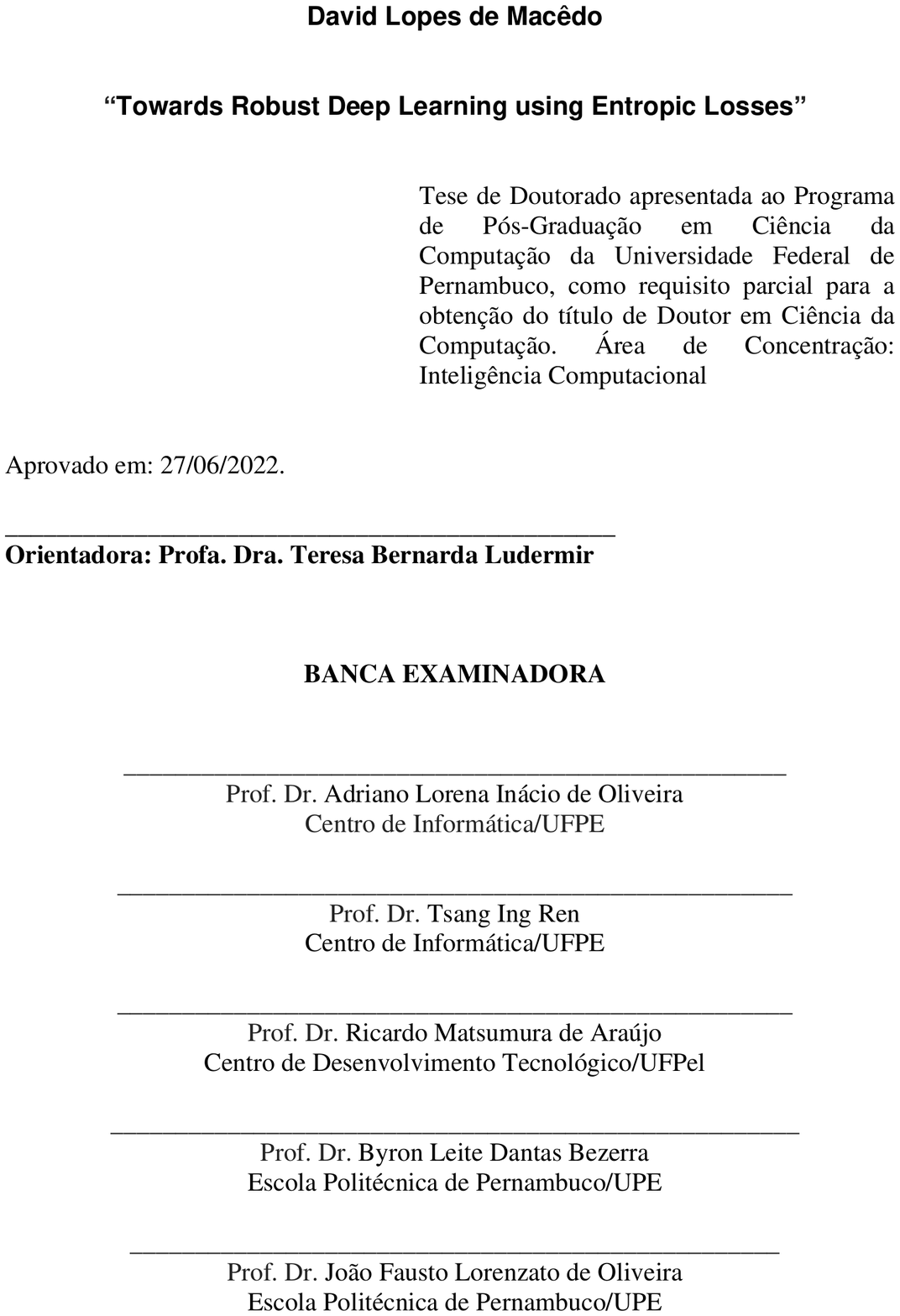}

\input{default_content/dedicatory}

\input{default_content/acknowledgments}

\input{default_content/epigraph}

\input{default_content/abstract}

\input{default_content/resumo}

{
\let\oldnumberline\numberline
\newcommand{\fignumberline}[1]{\figurename~#1~\enspace--~\enspace}
\renewcommand{\numberline}{\fignumberline}
\listoffigures
}

{
\let\oldnumberline\numberline
\newcommand{\tabnumberline}[1]{\tablename~#1~\enspace--~\enspace}
\renewcommand{\numberline}{\tabnumberline}
\listoftables
}

\listofacronyms



\tableofcontents

\mainmatter


\everypar{\looseness=-1}
\input{chapters/1.introduction}
\input{chapters/2.background}
\input{chapters/3.proposal}
\input{chapters/4.experiments}
\input{chapters/5.conclusion}


\titleformat{\chapter}[display]
{\bfseries\centering}
{\chapnumfont\textcolor{chapnumcol}{\MakeUppercase{\thechapter}}}
{-24pt}
{\Large\MakeUppercase}

\include{appendix/entropic_out-of-distribution_detection_ijcnn2021}
\include{appendix/entropic_out-of-distribution_detection_ieee2021}
\include{appendix/enhanced_isotropy_maximization_loss}
\include{appendix/distinction_maximization_loss}

\end{document}

%% file: default_content/acronyms.tex
\newacronym{ood}{OOD}{Out-of-Distribution}
\newacronym{id}{ID}{In-Distribution}
\newacronym{odin}{ODIN}{Out-of-DIstribution Detector for Neural networks}
\newacronym{godin}{GODIN}{Generalized Out-of-DIstribution Detector for Neural networks}
\newacronym{acet}{ACET}{Adversarial Confidence Enhancing Training}
\newacronym{isomax}{IsoMax}{Isotropy Maximization Loss}
\newacronym{isomax+}{IsoMax+}{Enhanced Isotropy Maximization Loss}
\newacronym{dismax}{DisMax}{Distinction Maximization Loss}
\newacronym{mps}{MPS}{Maximum Probability Score}
\newacronym{es}{ES}{Entropic Score}
\newacronym{mds}{MDS}{Minimum Distance Score}
\newacronym{mmles}{MMLES}{Maximum-Mean Logit Entropy Score}
\newacronym{logit+}{logit+}{enhanced logit}
\newacronym{tnr}{TNR@TPR95}{True Negative Rate at 95\% True Positive Rate}
\newacronym{dtacc}{DTACC}{Detection Accuracy}
\newacronym{auroc}{AUROC}{Area Under the Receiver Operating Characteristic Curve}
\newacronym{aupr}{AUPR}{Area Under the Precision-Recall Curve}
\newacronym{acc}{ACC}{Classification Accuracy}
\newacronym{ece}{ECE}{Expected Calibration Error}
\newacronym{fpr}{FPR}{Fractional Probability Regularization}
\newacronym{ue}{UE}{Uncertainty Estimation}
\newacronym{sc}{SC}{Scaled Cosine}
\newacronym{de}{DE}{Deep Ensembles}
\newacronym{duq}{DUQ}{Deterministic Uncertainty Quantification}
\newacronym{sngp}{SNGP}{Spectral-Normalized Neural Gaussian Process}

%% file: default_content/list_symbols.tex
\newglossaryentry{updatestep}{
  name = $\Delta$,
  description = Gradient
}

%% file: default_content/dedicatory.tex
\begin{dedicatory}
To my family.
\end{dedicatory}

%% file: default_content/acknowledgments.tex
\acknowledgements
This work would not have been possible without the support of many. I thank and dedicate this doctoral thesis to the following people:

I thank my advisor, Teresa Ludermir. Teresa is an exceptional researcher and professor, and her support was fundamental to motivating me throughout this research. I also thank my co-advisor, Cleber Zanchettin, for reviewing this work.

There is no doubt they contributed fundamentally to broadening my knowledge\footnote{\url{https://dlmacedo.com}} and to allowing me to become a Top Conferences (NeurIPS, ICLR, ICML) and IEEE Reviewer. Moreover, the work we did together led me to be a Member of the Program Committees of the NeurIPS Workshop on Machine Learning Safety and the ICML Workshop on Uncertainty \& Robustness in Deep Learning, in addition to an ICLR Highlighted Reviewer\footnote{\url{https://iclr.cc/Conferences/2022/Reviewers}}.

I thank Valdemar Cardoso da Rocha Junior, responsible for introducing me to the wonderful world of scientific research in the 1990s. I also thank him for provoking me about the power of Information Theory, which undoubtedly contributed decisively to this work.

\looseness=-1
To Yoshua Bengio for the personal discussions we had regarding the out-of-distribution detection problem in deep learning during my stay at the Montreal Institute for Learning Algorithms (MILA) as Visiting Researcher\footnote{\url{https://mila.quebec/en/person/david-macedo}}. The first question he asked was: "Why does it work?"

\looseness=-1
To Adriano Oliveira, Cleber Zanchettin, and the alumni of the Deep Learning Course of the CIn/UFPE Computer Science Graduate Program for the extremely pleasant discussions that we have had and all interesting work that we have done together since 2016 when we started this rewarding journey\footnote{\url{https://deeplearning.cin.ufpe.br}}\textsuperscript{,}\footnote{\url{https://dlmacedo.com/courses/deeplearning}}.

To my family, especially my parents, José and Mary, my wife, Janaina, and my children, J\'essica and Daniel, for giving me love throughout my life.

%% file: default_content/epigraph.tex




\begin{epigraph}[]{David Macêdo}
``The most important power is the power of believing.''
\end{epigraph}

\begin{epigraph}[AI: Artificial Intelligence Movie]{David}
``Please make me a real boy so that my mommy can love me!\\Please, Blue Fairy!''
\end{epigraph}

%% file: default_content/abstract.tex
\abstract

Despite theoretical achievements and encouraging practical results, deep learning still presents limitations in many areas, such as reasoning, causal inference, interpretability, and explainability. From an application point of view, one of the most impactful restrictions is related to the robustness of these systems. Indeed, current deep learning solutions are well known for not informing whether they can reliably classify an example during inference. Modern neural networks are usually overconfident, even when they are wrong. Therefore, building robust deep learning applications is currently a cutting-edge research topic in computer vision, natural language processing, and many other areas. One of the most effective ways to build more reliable deep learning solutions is to improve their performance in the so-called out-of-distribution detection task, which essentially consists of ``know that you do not know'' or ``know the unknown''. In other words, out-of-distribution detection capable systems may reject performing a nonsense classification when submitted to instances of classes on which the neural network was not trained. This thesis tackles the defiant out-of-distribution detection task by proposing novel loss functions and detection scores. Uncertainty estimation is also a crucial auxiliary task in building more robust deep learning systems. Therefore, we also deal with this robustness-related task, which evaluates how realistic the probabilities presented by the deep neural network are. To demonstrate the effectiveness of our approach, in addition to a substantial set of experiments, which includes state-of-the-art results, we use arguments based on the principle of maximum entropy to establish the theoretical foundation of the proposed approaches. Unlike most current methods, our losses and scores are seamless and principled solutions that produce accurate predictions in addition to fast and efficient inference. Moreover, our approaches can be incorporated into current and future projects simply by replacing the loss used to train the deep neural network and computing a rapid score for detection.


%% file: default_content/resumo.tex
\resumo

Apesar das conquistas teóricas e resultados práticos encorajadores, o aprendizado profundo ainda apresenta limitações em muitas áreas, como raciocínio, inferência causal, interpretabilidade e explicabilidade. Do ponto de vista da aplicação, uma das restrições mais impactantes está relacionada à robustez desses sistemas. De fato, as soluções atuais de aprendizado profundo são bem conhecidas por não informar se podem classificar um exemplo de maneira confiável durante a inferência. As redes neurais modernas geralmente são superconfiantes, mesmo quando estão erradas. Portanto, construir aplicativos robustos de aprendizado profundo é atualmente um tópico de pesquisa de ponta em visão computacional, processamento de linguagem natural e muitas outras áreas. Uma das maneiras mais eficazes de construir soluções de aprendizado profundo mais confiáveis é melhorar seu desempenho na chamada tarefa de detecção fora de distribuição, que consiste essencialmente em ``saber que você não sabe'' ou ``conhecer o desconhecido''. Em outras palavras, sistemas com capacidade de detecção fora de distribuição podem rejeitar a realização de uma classificação sem sentido quando submetidos a instâncias de classes nas quais a rede neural não foi treinada. Esta tese aborda a desafiadora tarefa de detecção fora da distribuição, propondo novas funções de perda e pontuações de detecção. A estimativa de incerteza também é uma tarefa auxiliar crucial na construção de sistemas de aprendizado profundo mais robustos. Portanto, tratamos também dessa tarefa relacionada à robustez, que avalia quão realistas são as probabilidades apresentadas pela rede neural profunda. Para demonstrar a eficácia de nossa abordagem, além de um conjunto substancial de experimentos, que incluí resultados estado-da-arte, utilizamos argumentos baseados no princípio da máxima entropia para estabelecer a fundamentação teórica das abordagens propostas. Ao contrário da maioria dos métodos atuais, além de apresentarem inferência rápida e eficiente, nossas perdas e pontuações são soluções baseadas em princípios e não produzem efeitos colaterais indesejados. Além disso, nossas abordagens podem ser incorporadas em projetos atuais e futuros simplesmente substituindo a perda usada para treinar a rede neural profunda e computando uma pontuação rápida para detecção.


%% file: chapters/1.introduction.tex
\chapter{Introduction}\label{chap:intro}

\begin{quotation}[]{George Box}
``All models are wrong, but some are useful.''
\end{quotation}

\begin{quotation}[NeurIPS 2020]{Balaji Lakshminarayanan}
``All models are wrong, but some that know when they are wrong are useful.''
\end{quotation}

\begin{quotation}[]{Neil deGrasse Tyson}
``One of the greatest challenges in this world is knowing enough a subject to think you are right, but not enough to know you are wrong!''
\end{quotation}

\begin{quotation}[]{David Macêdo}
``Modern deep neural networks are just like humans.\\
They are usually overconfident, even when they are entirely wrong!''
\end{quotation}

In this introductory chapter, we explain the context, motivation, and objective of this work. Then we describe the concept of \gls{ood} detection, our specific problem of interest in the sub-area of deep learning robustness. Next, we define the objectives of this study. Among them is the design of high-performance solutions for OOD detection. Moreover, we present the peer review publications we authored in this and related research fields. Finally, we present an outline of the subject of the following chapters.\\

\newpage\section{Context: Why deep learning?}

AI is a field in computer science. Nowadays, it is hard to present an unique definition on which everybody could agree on. We understand AI as the study of systems capable of solving problems that are not handled well by using algorithm-based programming paradigms. It usually involves tasks for which a standard algorithmic solution is hard or impractical to design, implement, or even conceive. Otherwise, we could simply use a straightforward conventional so-called algorithmic-based programming approach to tackle the mentioned task.

This AI concept is strongly based on the phase ``field of study that gives computers the ability to learn without being explicitly programmed'' commonly attributed to Arthur Samuel. Although he probably never explicitly wrote or said the mentioned quotation, it works as a good abstract of thoughts he expressed in his seminal papers \citep{Samuel1959SomeSI,Samuel1967SomeSI}.

Machine learning is the field of AI in which the system learns from exploring the available data rather than relying on humans to algorithmize them. Machine learning established a significant advance by allowing AI systems to acquire their knowledge directly from data. 

\begin{figure}
\small
\centering
\caption[The Curse of Dimensionality]{The Curse of Dimensionality}
\includegraphics[width=0.75\textwidth]{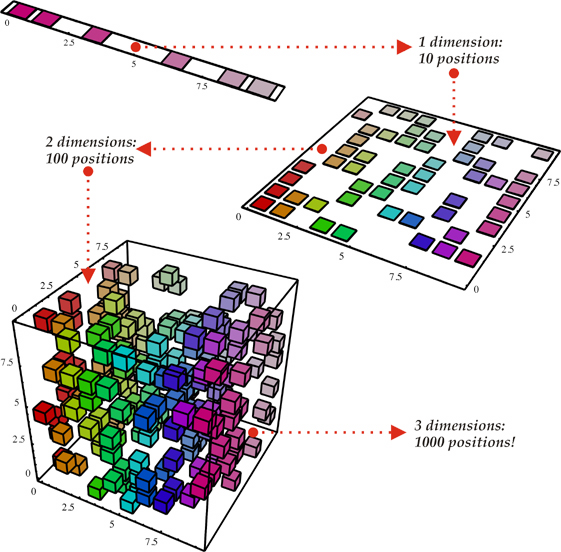}
\vskip 0.5cm
\begin{justify}
{Source:~\cite{Bengio2015DeepMotivations}. The curse of dimensionality: As the dimension grows, we need exponentially more training examples to avoid example density rarefaction in the feature space.}
\end{justify}
\label{fig:curse-of-dimensionality}
\end{figure}

Nevertheless, the conventional machine learning approach to AI has its limitations. Indeed, in this context, the curse of dimensionality \citep{Bellman1957DynamicProgramming,Bellman1961AdaptiveTour,Hughes1968OnRecognizers,Bengio2006TheMachines} may be understood as the empirical tendency to a dramatic training examples density rarefaction as the data dimensionality grows (\figref{fig:curse-of-dimensionality}). In such cases, at least apparently, it is reasonable to expect that the extremely low training example density could make generalization very difficult, as the test examples would be extremely distant from the examples used for training. Hence, at first sight, it appears that the number of training examples should grow exponentially faster than the data dimensionality to allow generalization. 

To mitigate the curse of dimensionality, machine learning methods generally rely on hand-designed high-level features, which allow such systems to sometimes produce high performance and scale appropriately \citep[pg.~3]{Goodfellow2016DeepLearning}. These engineered features are presented to the learning algorithm as low-dimensional vectors that, in some sense, bring the problem back to a treatable dimension, somehow avoiding extreme example density rarefaction.

\looseness=-1
Indeed, \tabref{tab:datasets-relative-density} presents datasets and their more relevant characteristics. We can see that the example density is reasonably high in datasets with few features (three first lines). Therefore, we may tackle these problems practically without relying on feature extraction or selection. However, for datasets with a high number of features (three last lines), we usually use feature engineering to tackle them to obtain high performance. In simple terms, the use of features brings the problem back to a dimensional that classical machine learning approaches can handle appropriately. 

\begin{figure*}
\small
\centering
\caption[The Fundamental Shallow Machine Learning Limitation]{The Fundamental Shallow Machine Learning Limitation}
\includegraphics[width=0.9\textwidth,trim={0 0.5cm 0 1cm},clip]{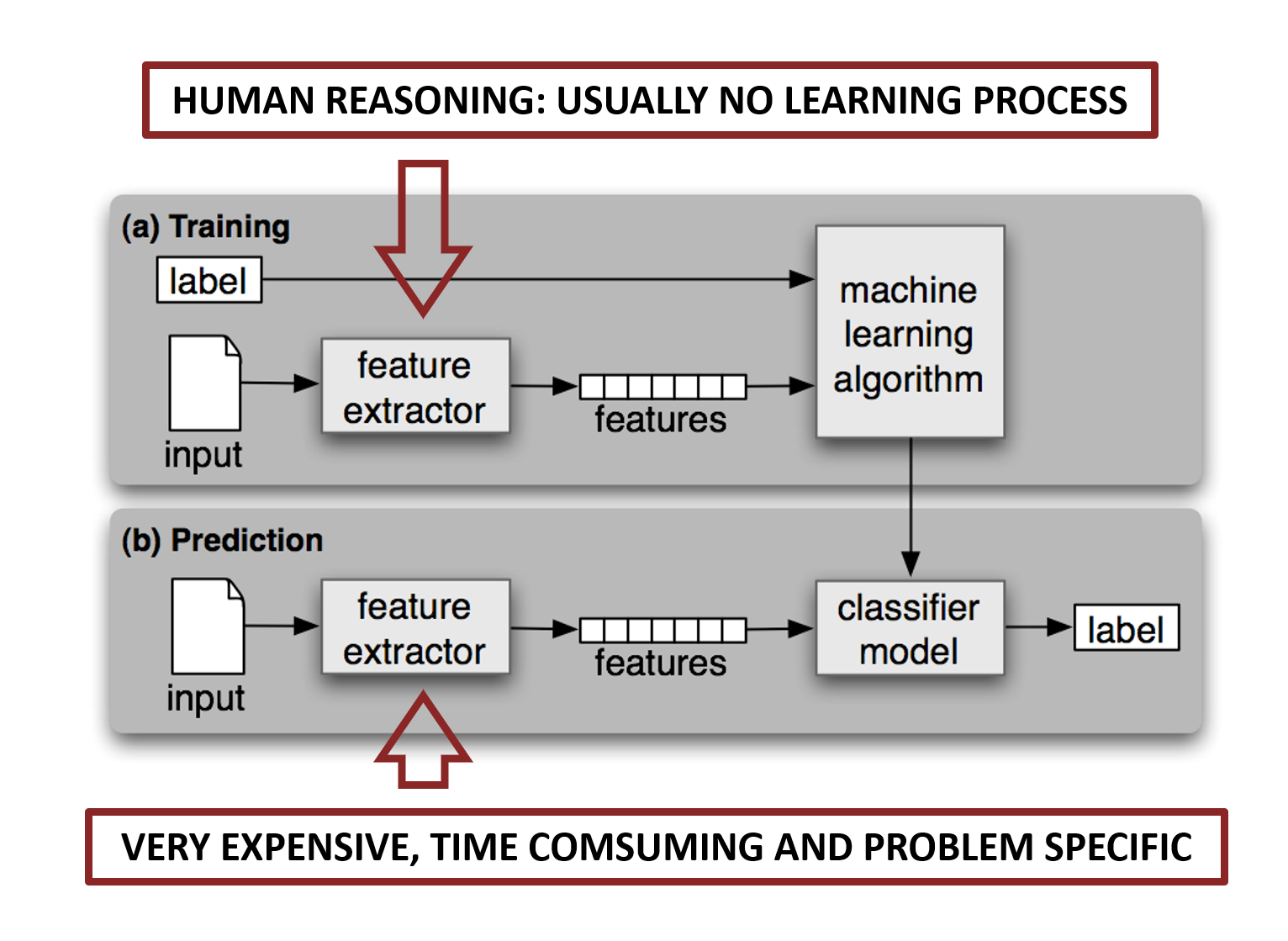}
\begin{justify}
{Source: Adapted from \url{https://www.nltk.org/book/ch06.html}. The shallow machine learning limitation: Numerous relevant real-world problems are high-dimensional. In such cases, the hand-designed feature engineering usually required by classical machine learning approaches presents drawbacks.}
\end{justify}
\label{fig:ml_problem}
\end{figure*}

\begin{figure}
\small
\centering
\caption[Self-Supervised Learning]{Self-Supervised Learning}
\includegraphics[width=0.9\textwidth]{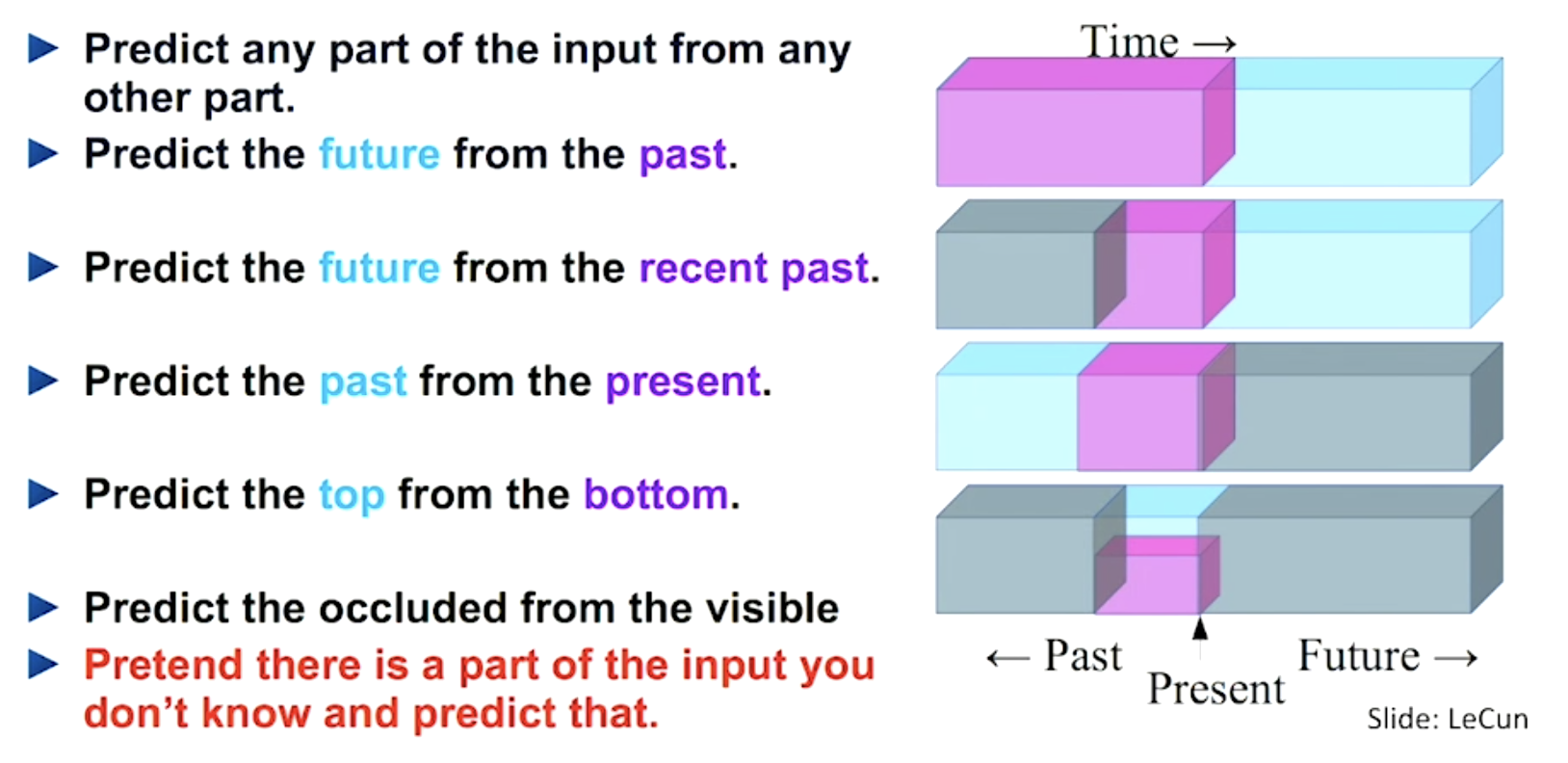}
\begin{justify}
{Source:~\url{https://www.youtube.com/watch?v=7I0Qt7GALVk}. Self-supervision allows training deep neural networks without labels using a pretext task. Given a portion of the data, the model is trained to predict what is missing.}
\end{justify}
\label{fig:self-supervision}
\end{figure}

However, relying on hand-designed high-level features brings significant drawbacks to systems based on traditional shallow machine learning. In fact, it is undesirable for many reasons. First, as a specific feature engineering is usually needed for each particular task, the overall solution becomes much less general-purpose than it could otherwise be. 

\looseness=-1
Second, feature engineering is commonly a costly and time-consuming process that requires human knowledge much more specialized than labeling data. Finally, for many real-world relevant tasks, it is challenging (if not impossible) to design high-quality features manually.

Recently, theoretical studies have shown that increasing the depth of AI models is an extremely effective way to deal with the curse of dimensionality \citep{Bengio2007ScalingAI, Delalleau2011ShallowNetworks, Pascanu2013OnActivations, Montufar2014OnNetworks}.  Indeed, some research results suggest that growing the depth of a learning model is exponentially more efficient than increasing the \hln{width} of traditional shallow machine learning solutions \citep{Bengio2013RepresentationPerspectives, Bengio2006TheMachines, Bengio2009LearningAI, Montufar2012WhenMixtures, Montufar2014OnNetworks}.

Therefore, the recognition that deep architectures are exponentially efficient in tackling the curse of dimensionality made deep neural networks extremely successful, as they virtually altogether avoid time-consuming, specific purpose, and expensive feature engineering (\figref{fig:ml_problem}). Moreover, modern self-supervised\footnote{\url{https://ai.facebook.com/blog/self-supervised-learning-the-dark-matter-of-intelligence}} techniques are making even labeling data unnecessary \citep{9226466,DBLP:journals/corr/abs-2011-00362,DBLP:journals/pami/JingT21,DBLP:journals/corr/abs-2006-08218} (\figref{fig:self-supervision}).

\tabref{tab:datasets-relative-density} presents datasets on which shallow machine learning (first three lines) and deep learning (last three lines) are usually applied \emph{without} relying on feature engineering. The difference in the density of examples in each case is staggering. The fact that deep learning allows satisfactory generalization in datasets with such small example density as ImageNet \citep{Deng2009ImageNetDatabase} without using feature engineering is extraordinary. Indeed, it was something unthinkable years ago.

\begin{table}
\small
\setlength{\tabcolsep}{0pt}
\renewcommand{\arraystretch}{1.5}
\centering
\caption[Example Density of Traditional Datasets]{Example Density of Traditional Datasets}
\label{tab:datasets-relative-density}
\begin{tabularx}{\columnwidth}{lYYYY}
\toprule
Dataset & Examples & Features & \makecell{Volumes\\(2\textsuperscript{Features})} & Example Density \mbox{(Examples/Volumes)}\\
\midrule
Iris & 150 & 4 & 16 & \textcolor{blue}{9.375}\\
Tic-Tac-Toe & 958 & 9 & 512 & \textcolor{blue}{1.871}\\
Adult & 48,842 & 14 & 16,384 & \textcolor{blue}{2.981}\\
\midrule
MNIST & 60,000 & 784 (28x28) & \num{1.017e236} & \textcolor{red}{\num{5.897e-232}}\\
SVHN & 600,000 & 3,072 (3x32x32) & \num{5.809e924} & \textcolor{red}{\num{1.032e-919}}\\
ImageNet & 1,200,000 & 196,608~(3x256x256) & \num{2.003e19728} & \textcolor{red}{\num{5.991e-19723}}\\
\bottomrule
\end{tabularx}
\vskip 0.25cm
\begin{justify}
{Source: The Author (2022). In this table, the first three lines present examples of datasets in which classical machine learning works properly without relying on feature engineering. The last three lines give examples of datasets in which deep learning works appropriately.}
\end{justify}
\end{table}

\newpage

\looseness=-1
After the breakthroughs in speech recognition \citep{Hinton2012DeepGroups} and computer vision \citep{Krizhevsky2012ImageNetNetworks}, other important advances followed in natural language processing \citep{Sutskever2014SequenceNetworks, Bahdanau2014NeuralTranslate}, speech processing \citep{ALAM2020302}, and even in structured data such as tabular data \citep{DBLP:journals/corr/abs-2110-01889} and time series \citep{DBLP:journals/corr/abs-2004-13408}. 

\begin{figure}
\small
\centering
\caption[Transformer Architecture]{Transformer Architecture}
\includegraphics[width=\textwidth]{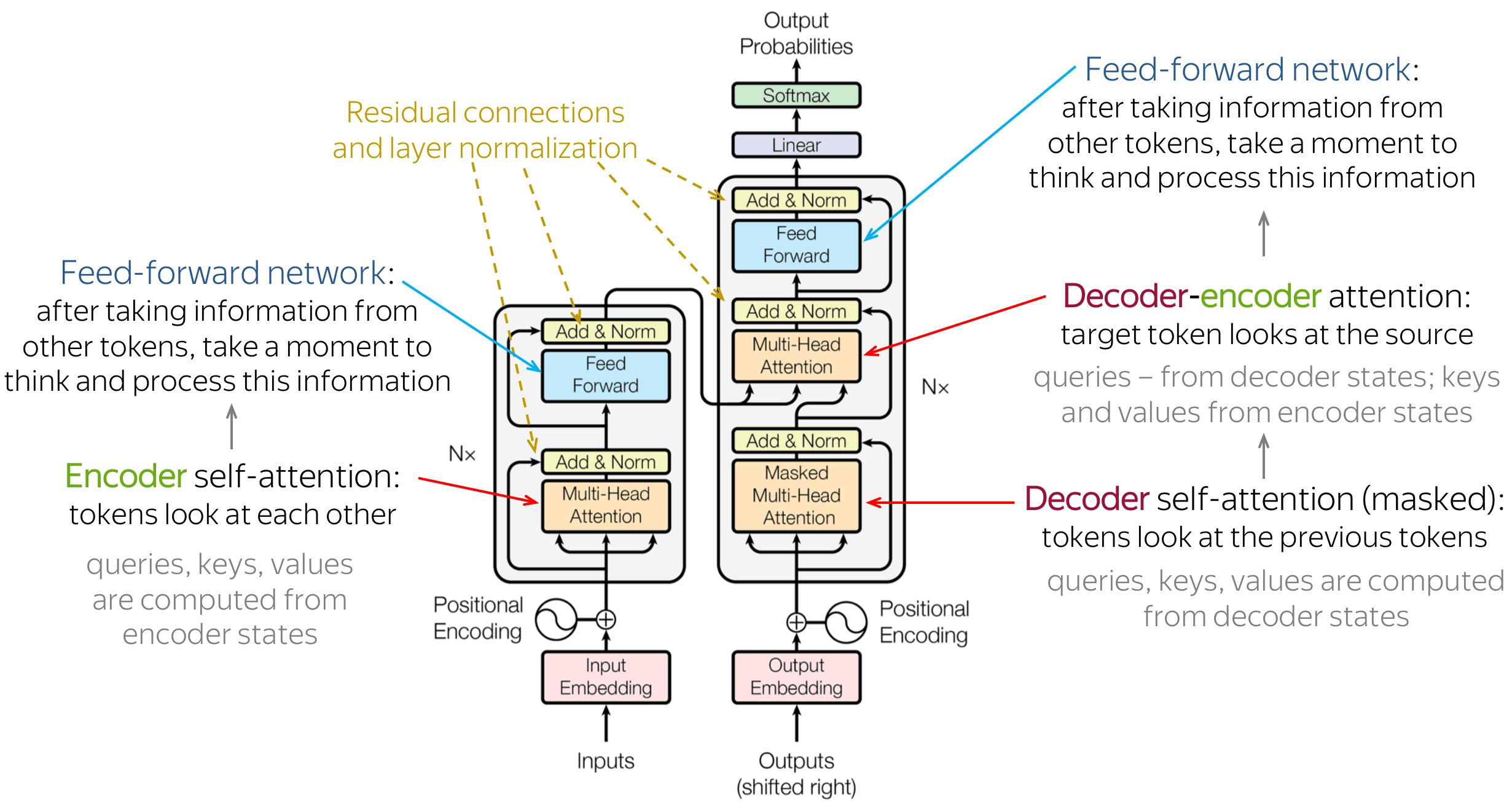}
\begin{justify}
{Source:~\url{http://www.ai2news.com/blog/42747/}. The Transformer understands everything as a mapping from an input set (or sequence) of tokens to an output set (or sequence) of tokens. All input need to be converted to tokens to be processed by the Transformer-based architectures \citep{DBLP:conf/nips/VaswaniSPUJGKP17}. It was highly inspired by the seminal Sequence-To-Sequence paper \citep{Sutskever2014SequenceNetworks}.}
\end{justify}
\label{fig:transformer}
\end{figure}

\begin{figure}
\small
\centering
\caption[Vision Transformer (ViT)]{Vision Transformer (ViT)}
\includegraphics[width=0.95\textwidth,trim={0 0.5cm 0 1cm},clip]{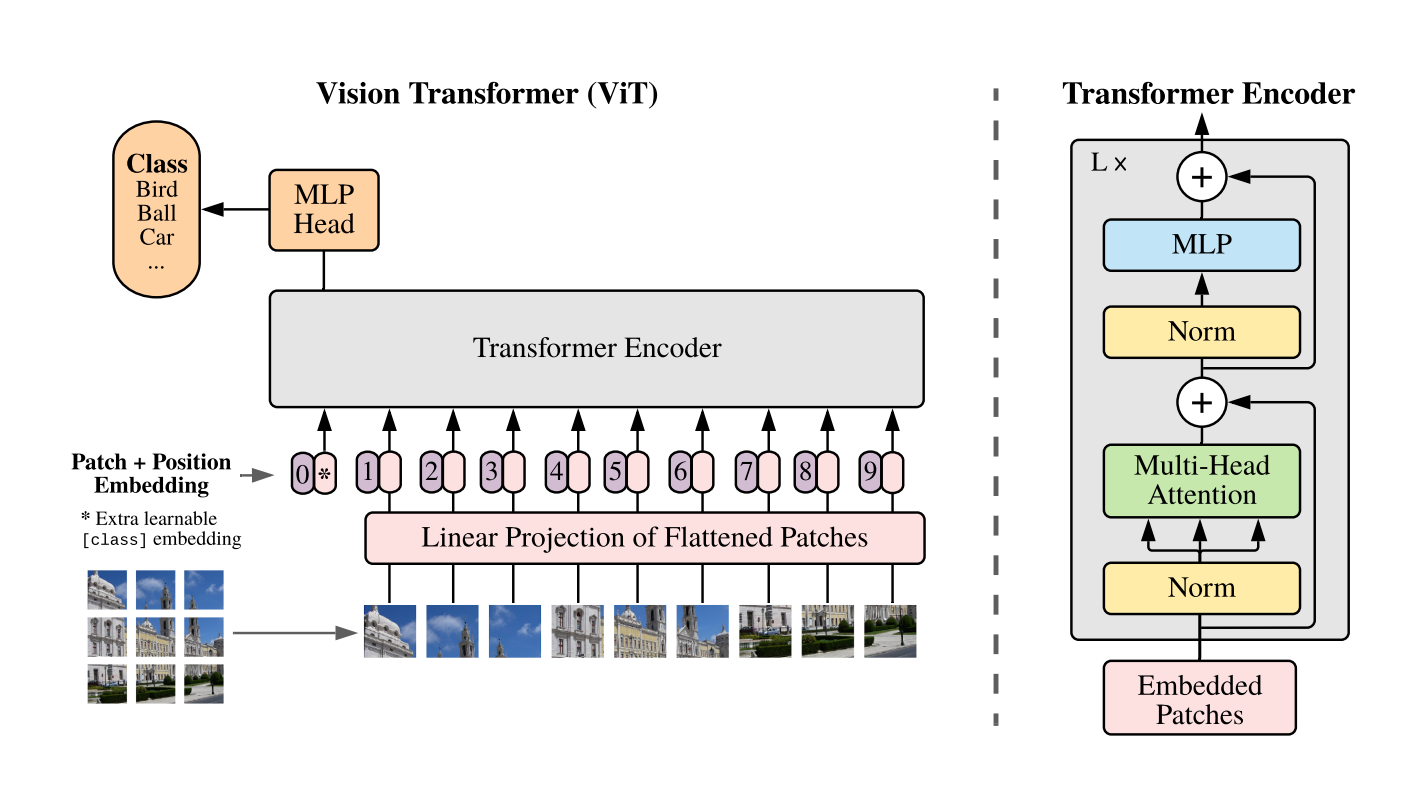}
\begin{justify}
{Source:~\cite{DBLP:conf/iclr/DosovitskiyB0WZ21}. The Vision Transformer (ViT) creates tokens by splitting the images into patches, which then feed the Transformer architecture as proposed in \cite{DBLP:conf/nips/VaswaniSPUJGKP17}. The classification token is inspired by the BERT paper \citep{DBLP:conf/naacl/DevlinCLT19}.}
\end{justify}
\label{fig:vit}
\end{figure}

Currently, deep learning is applied to solve tasks in a broad area of applications such as mathematics, physics, chemistry, engineering, biology, health care, finance, agriculture, and many others \citep[pg.~9]{Goodfellow2016DeepLearning} \citep{https://doi.org/10.48550/arxiv.2202.02306}. The advances in the use of deep learning for scientific discoveries\footnote{\url{https://www.youtube.com/watch?v=XtJVLOe4cfs}} \citep{https://doi.org/10.48550/arxiv.2110.13041} are so profound that some argue that we are witnessing a new scientific revolution\footnote{\url{https://thegradient.pub/ai-scientific-revolution/}}.

Transforms\footnote{\url{https://jalammar.github.io/illustrated-transformer/}}\textsuperscript{,}\footnote{\url{https://lena-voita.github.io/nlp_course/seq2seq_and_attention.html}} \citep{DBLP:conf/nips/VaswaniSPUJGKP17} allowed us to use a single universal architecture to construct image, text, video, audio, and multimodal models (Fig.~\ref{fig:transformer}). After BERT \citep{DBLP:conf/naacl/DevlinCLT19} and GPT \citep{DBLP:conf/nips/BrownMRSKDNSSAA20} successfully applied the groundbreaking Transformer architecture for text, the so-called Vision Transformers (ViT) \citep{DBLP:conf/iclr/DosovitskiyB0WZ21} (Fig.~\ref{fig:vit}) effectively applied this template architecture for images.

\begin{figure}
\small
\centering
\caption[Contrastive Learning]{Contrastive Learning}
\includegraphics[width=0.55\textwidth]{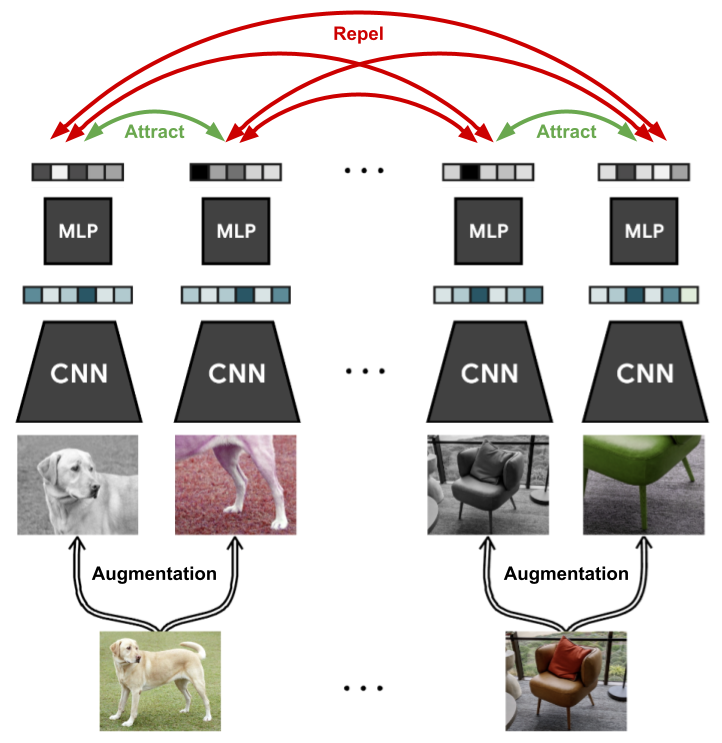}
\begin{justify}
{Source:~\cite{DBLP:conf/icml/ChenK0H20}. Contrastive learning allows supervised or unsupervised training of deep neural networks. Positive examples (i.e., examples we desire to approximate in the feature space) may be constructed using only data augmentation without any supervision.}
\end{justify}
\label{fig:constrastive}
\end{figure}

\begin{figure}
\small
\centering
\caption[CLIP Architecture]{CLIP Architecture}
\includegraphics[width=\textwidth]{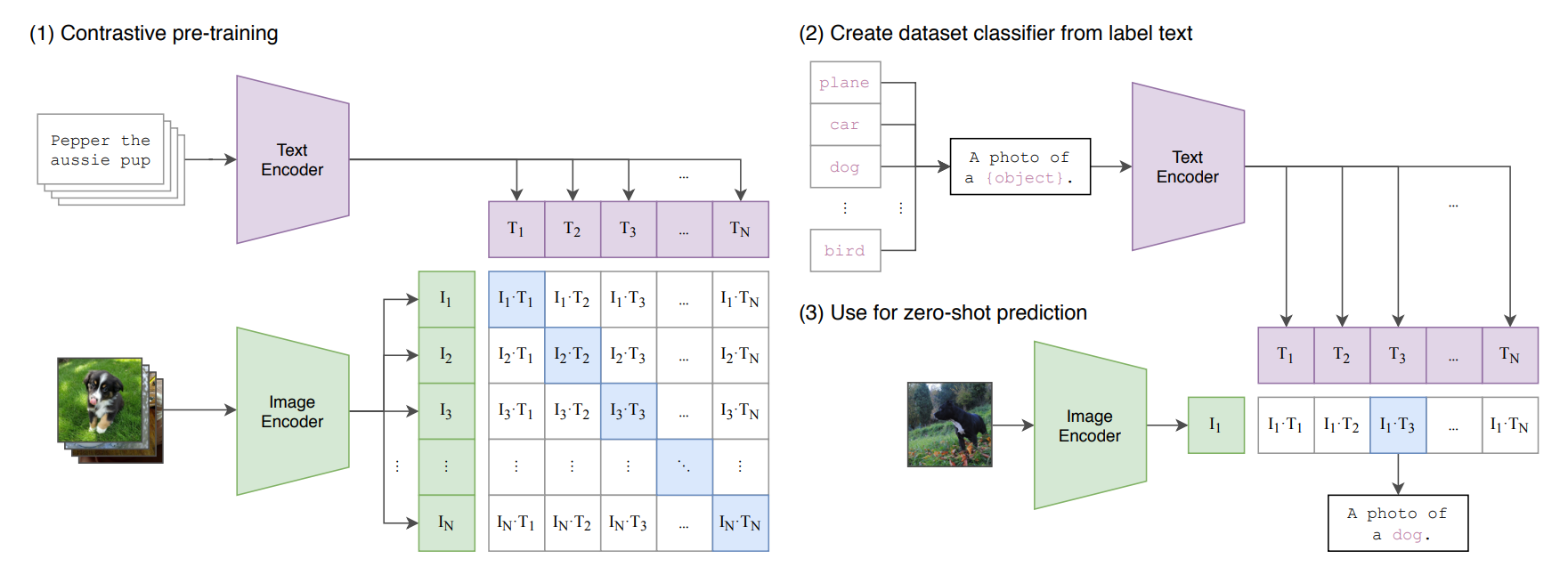}
\begin{justify}
{Source:~\cite{DBLP:conf/icml/RadfordKHRGASAM21}. OpenAI CLIP uses supervised contrastive learning to approximate embeddings of images and related texts in a joint embedding space. Zero-shot classification is allowed during inference.}
\end{justify}
\label{fig:clip}
\end{figure}

\begin{figure}
\small
\centering
\caption[Masked Autoencoder (MAE)]{Masked Autoencoder (MAE)}
\includegraphics[width=\textwidth]{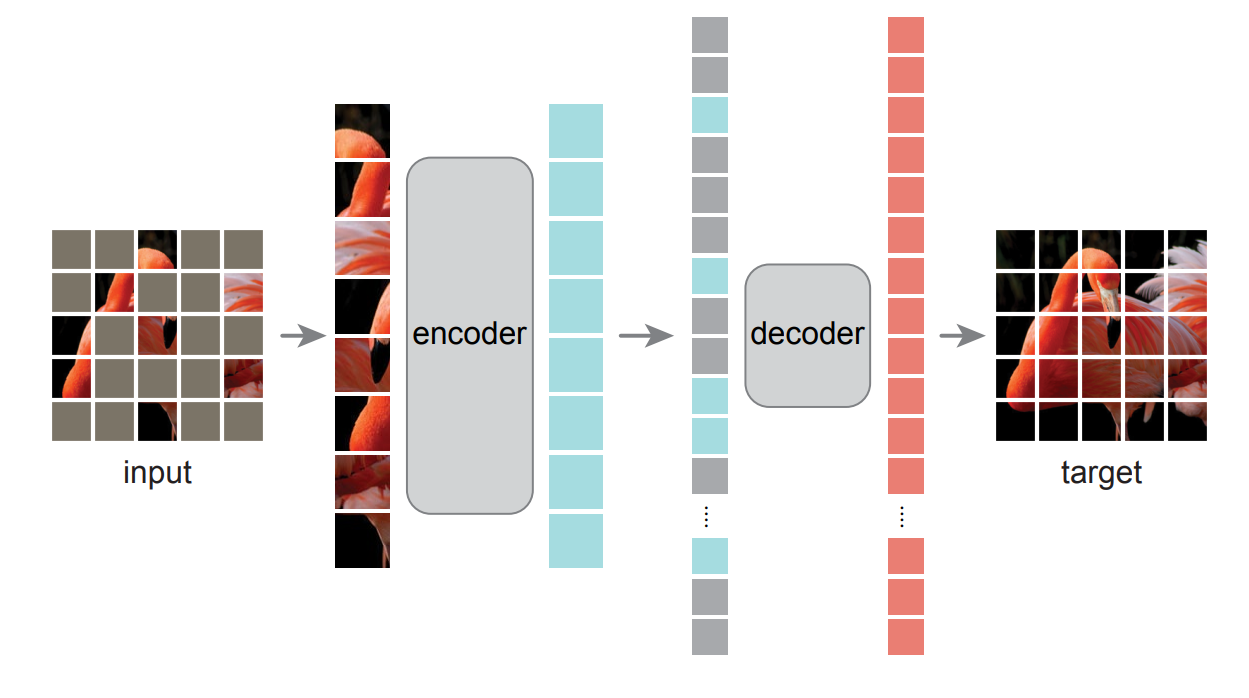}
\begin{justify}
{Source:~\cite{https://doi.org/10.48550/arxiv.2111.06377}. In the Masked Autoencoder (MAE), about 75\% of the image patches are hidden from the decoder during the self-supervised pretraining. The decoder is trained with patches produced by the encoder and mask tokens and needs to reconstruct the original image.}
\end{justify}
\label{fig:mae}
\end{figure}

\begin{figure}
\small
\centering
\caption[Generalist Agent (Gato)]{Generalist Agent (Gato)}
\includegraphics[width=0.9\textwidth]{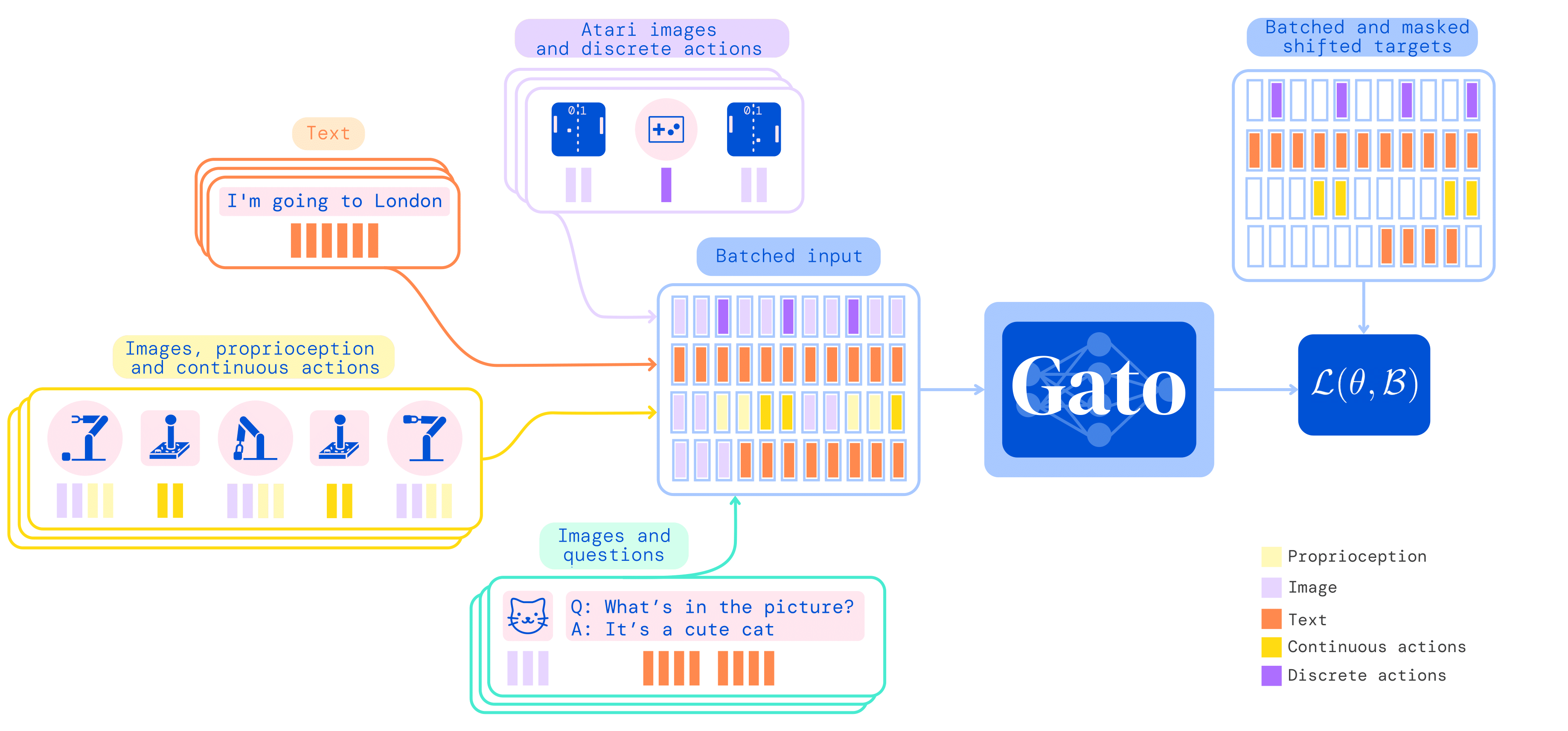}
\begin{justify}
{Source:~\cite{https://doi.org/10.48550/arxiv.2205.06175}. DeepMind Generalist Agent (Gato) can simultaneously handle many media and tasks using a unified Transformer-based architecture.}
\end{justify}
\label{fig:gato}
\end{figure}

\newpage

Contrastive learning\footnote{\url{https://ai.googleblog.com/2020/04/advancing-self-supervised-and-semi.html}} essential idea consists of maximizing similarity (i.e. minimizing the distance) between the (high-level) representation of input data that we desire or know to be similar, while doing the opposite when we desire the input data to be semantically different \citep{DBLP:conf/icml/ChenK0H20, DBLP:journals/entropy/Albelwi22} (Fig.~\ref{fig:constrastive}).

Notice that contrastive learning can be used in conjunction with data augmentation to create input data that we want to produce similar representations, which allows unsupervised training. Therefore, contrastive learning can be used in both supervised and unsupervised settings. Self-supervised learning may or not be used in combination with contrastive learning.

CLIP \citep{DBLP:conf/icml/RadfordKHRGASAM21} from OpenAI leverage supervised contrastive learning to build a vision-language model that constructs a \emph{joint embedding space} in which representations of images are near to representations of related texts. CLIP\footnote{\url{https://openai.com/blog/clip/}} allows zero-shot classification during inference and is a relevant example of supervised contrastive learning (Fig.~\ref{fig:clip}).

Self-supervision using masking and autoregressive techniques are making possible pretraining without labels large models that subsequently can be downloaded and fine-tuned with few examples of supervised data. Sometimes, they may even be used in zero-shot or few-shot settings. Currently, self-supervision is largely used regardless of media type (image, speech, text, etc.). For example, while BERT and MAE \citep{https://doi.org/10.48550/arxiv.2111.06377} (Fig.~\ref{fig:mae}) used a masking pretext, the GPT used an autoregressive pretext task. Currently, \hl{some researches} are exploring single models \citep{https://doi.org/10.48550/arxiv.2205.06175} that are both multimodal and multitask\footnote{\url{https://www.deepmind.com/publications/a-generalist-agent}} (Fig.~\ref{fig:gato}).

\begin{figure}
\small
\centering
\caption[Imagen: Photorealistic Examples]{Imagen: Photorealistic Examples}
\includegraphics[width=\textwidth]{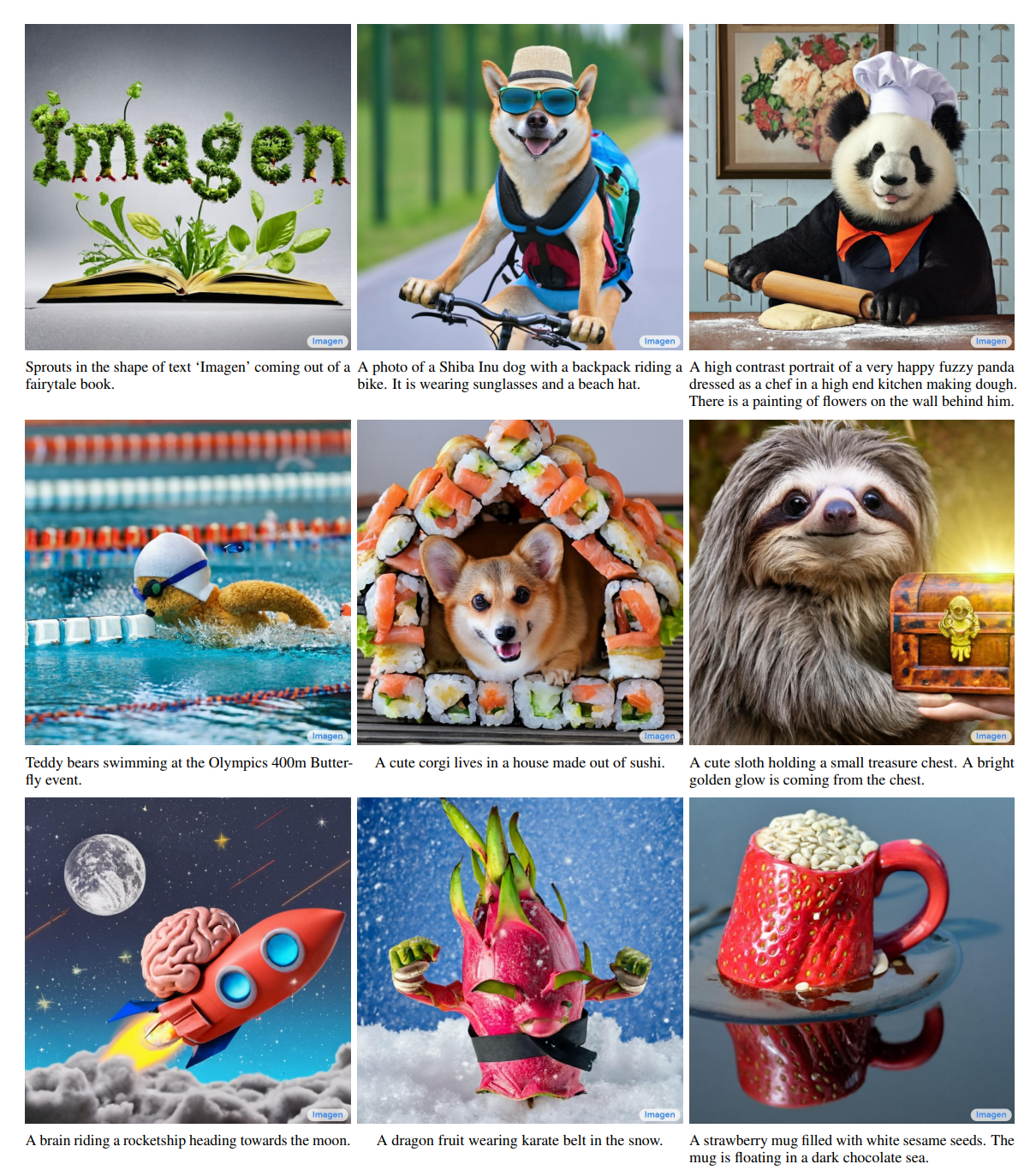}
\begin{justify}
{Source:~\cite{https://doi.org/10.48550/arxiv.2205.11487}. Google Imagen: Examples of photorealistic 1024x1024 high-resolution images produced conditioned on texts. It is also possible to generate artistic content.}
\end{justify}
\label{fig:imagen-examples}
\end{figure}

\begin{figure}
\small
\centering
\caption[DALL-E~2: Photorealistic Examples]{DALL-E~2: Photorealistic Examples}
\includegraphics[width=\textwidth]{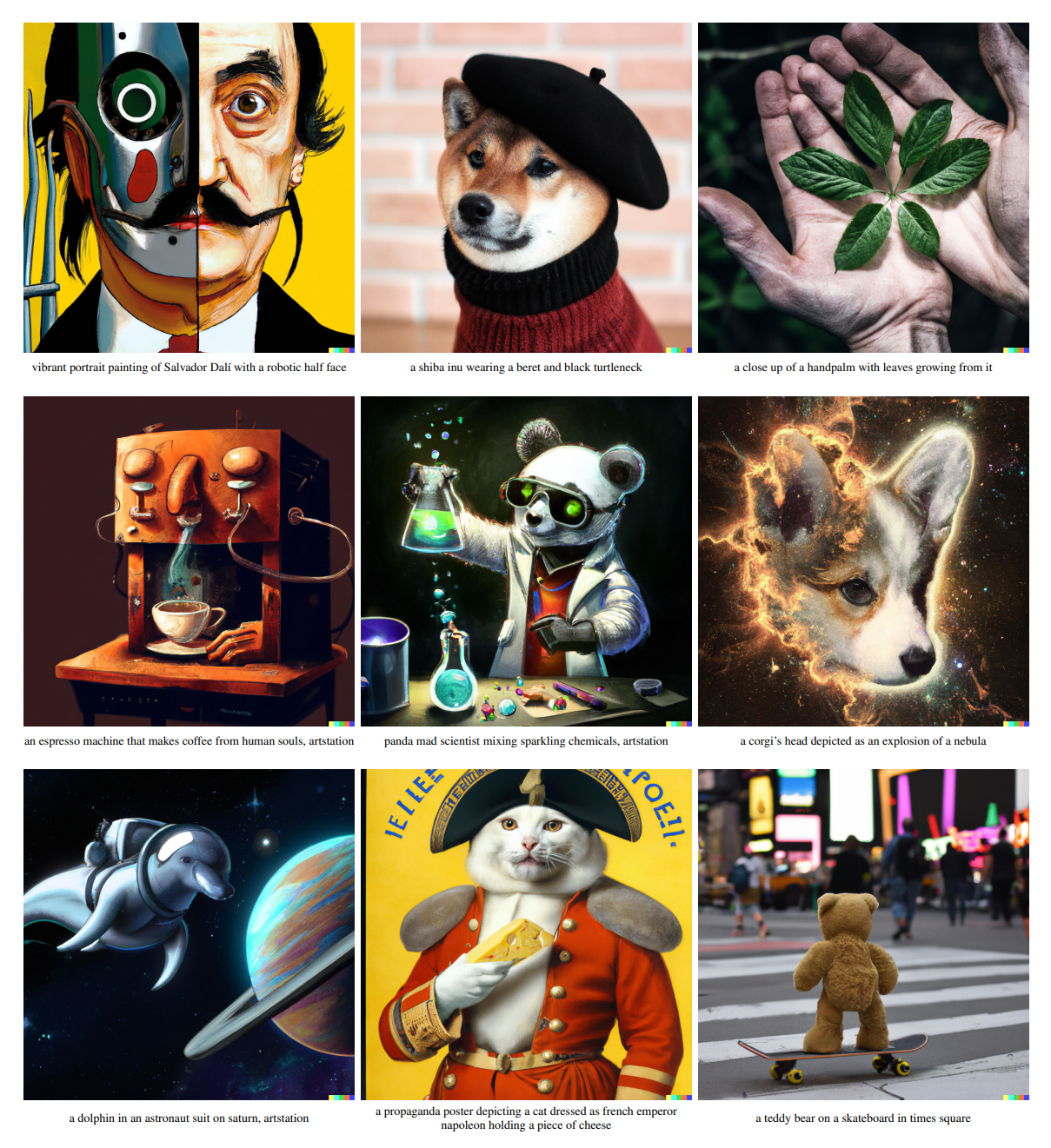}
\begin{justify}
{Source:~\cite{https://doi.org/10.48550/arxiv.2204.06125}. OpenAI DALL-E~2: Examples of 1024x1024 high-resolution images generated conditioned on the texts shown below each image.}
\end{justify}
\label{fig:dalle2-examples}
\end{figure}

\begin{figure}
\small
\centering
\caption[CLRS Algorithmic Reasoning Benchmark]{CLRS Algorithmic Reasoning Benchmark}
\includegraphics[width=\textwidth]{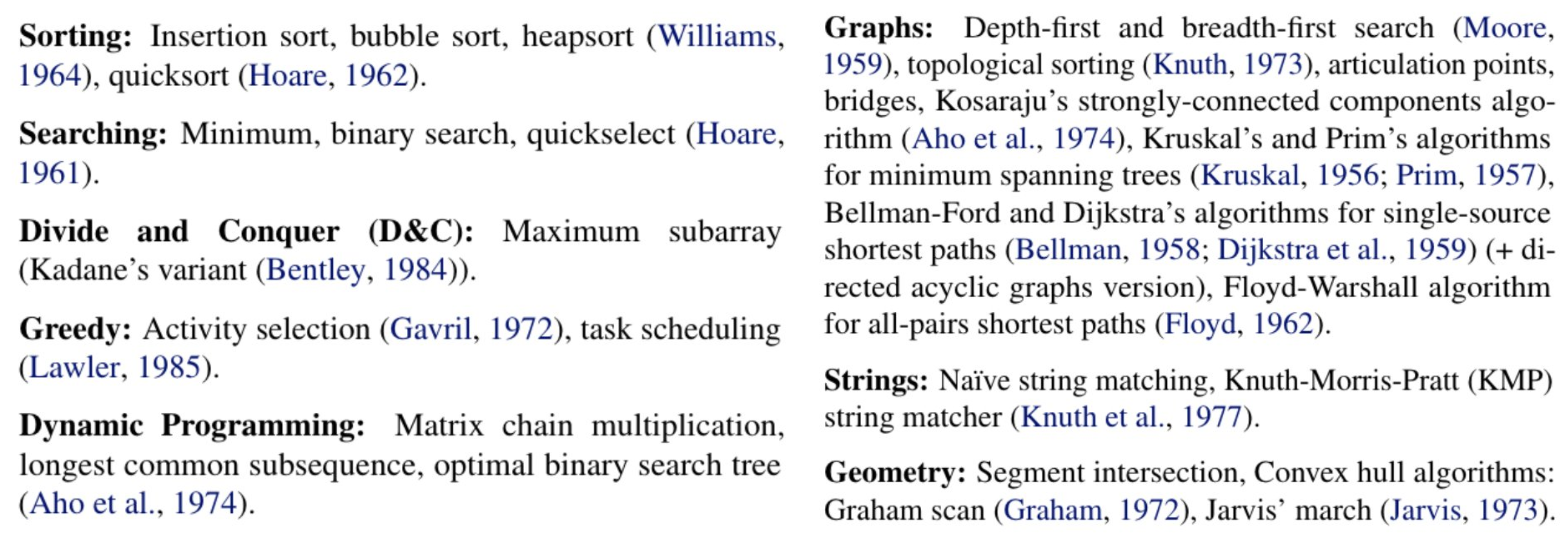}
\vskip 0.5cm
\begin{justify}
{Source:~\cite{deepmind2022clrs}. Examples of classical algorithms that neural networks are being shown to be able to learn. The \hln{CLRS Algorithmic Reasoning Benchmark} provides datasets for these tasks. CLRS stands for ``Cormen, Leiserson, Rivest, and Stein'' in homage to the authors of the ``Introduction to Algorithms'' classical textbook \citep{DBLP:books/daglib/0023376}.}
\end{justify}
\label{fig:algo}
\end{figure}

\begin{figure}
\small
\centering
\caption[Math Word Problems]{Math Word Problems}
\includegraphics[width=\textwidth]{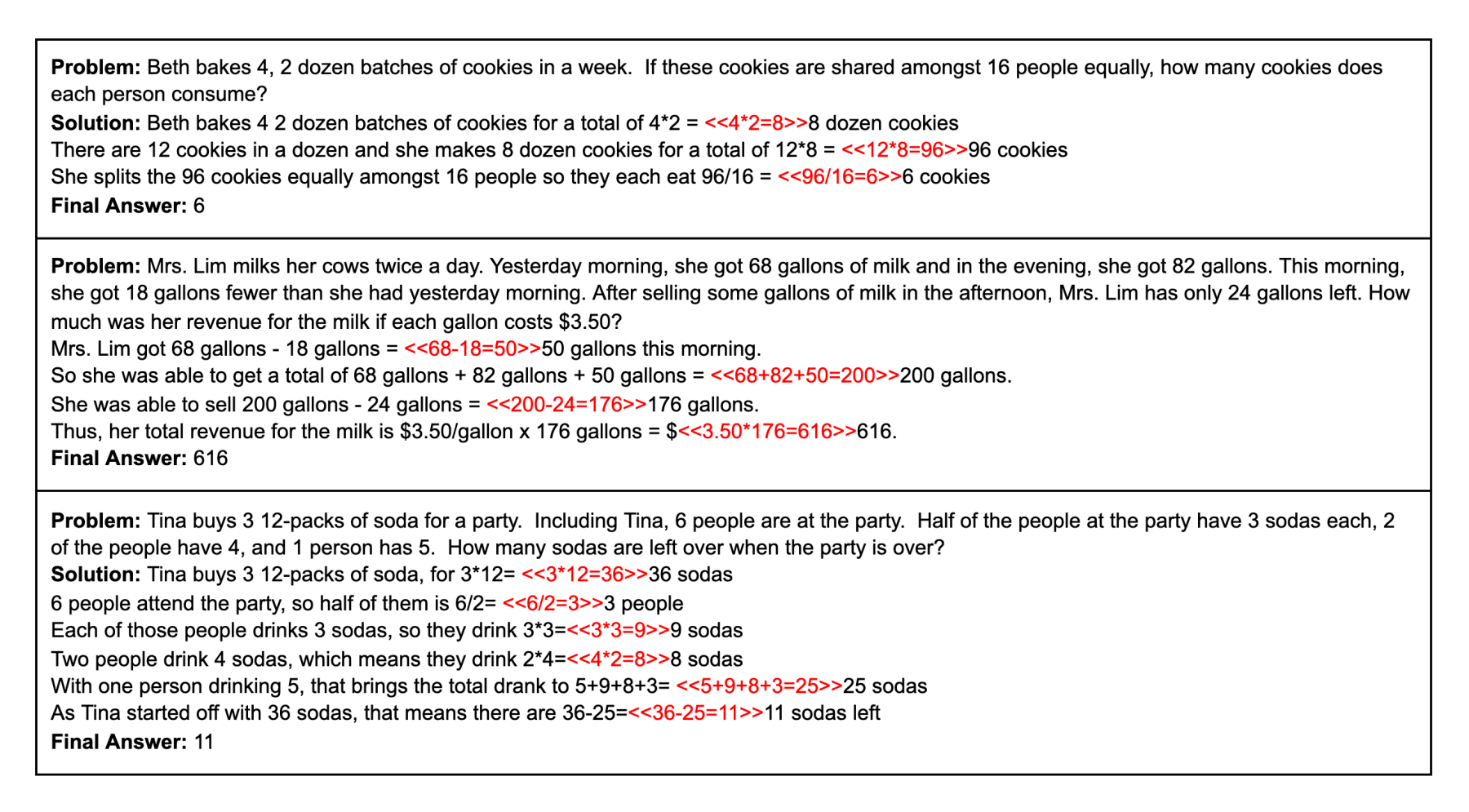}
\begin{justify}
{Source:~\cite{DBLP:journals/corr/abs-2110-14168}. Large language models are being adapted to tackle grade school math problems. Considering that humans think using natural language, it is reasonable to tackle reasoning-based tasks using improved language models.}
\end{justify}
\label{fig:math}
\end{figure}

\newpage

The advances are not restricted to discriminative tasks such as classification, object detection, and semantic segmentation. Instead, they are present in generative tasks as well. In fact, diffusion-based models such as Imagen\footnote{\url{https://imagen.research.google/}} from Google \citep{https://doi.org/10.48550/arxiv.2205.11487} (Fig.~\ref{fig:imagen-examples}) and DALL-E~2\footnote{\url{https://openai.com/dall-e-2/}} from OpenAI (Fig.~\ref{fig:dalle2-examples}) are producing promising results \citep{https://doi.org/10.48550/arxiv.2204.06125}.

We have recently seen advances in areas that we may have believed that neural networks could never reach. For example, some works show that it is possible to learn reasoning to perform classical algorithmic tasks such as sorting, searching, dynamic programming, graph algorithms, string algorithms, and geometric algorithms (Fig.~\ref{fig:algo}) \citep{deepmind2022clrs}. 

\begin{figure}
\small
\centering
\caption[Prompt Engineering: Chain of Thought (CoT)]{Prompt Engineering: Chain of Thought (CoT)}
\subfloat[]{\includegraphics[width=\textwidth]{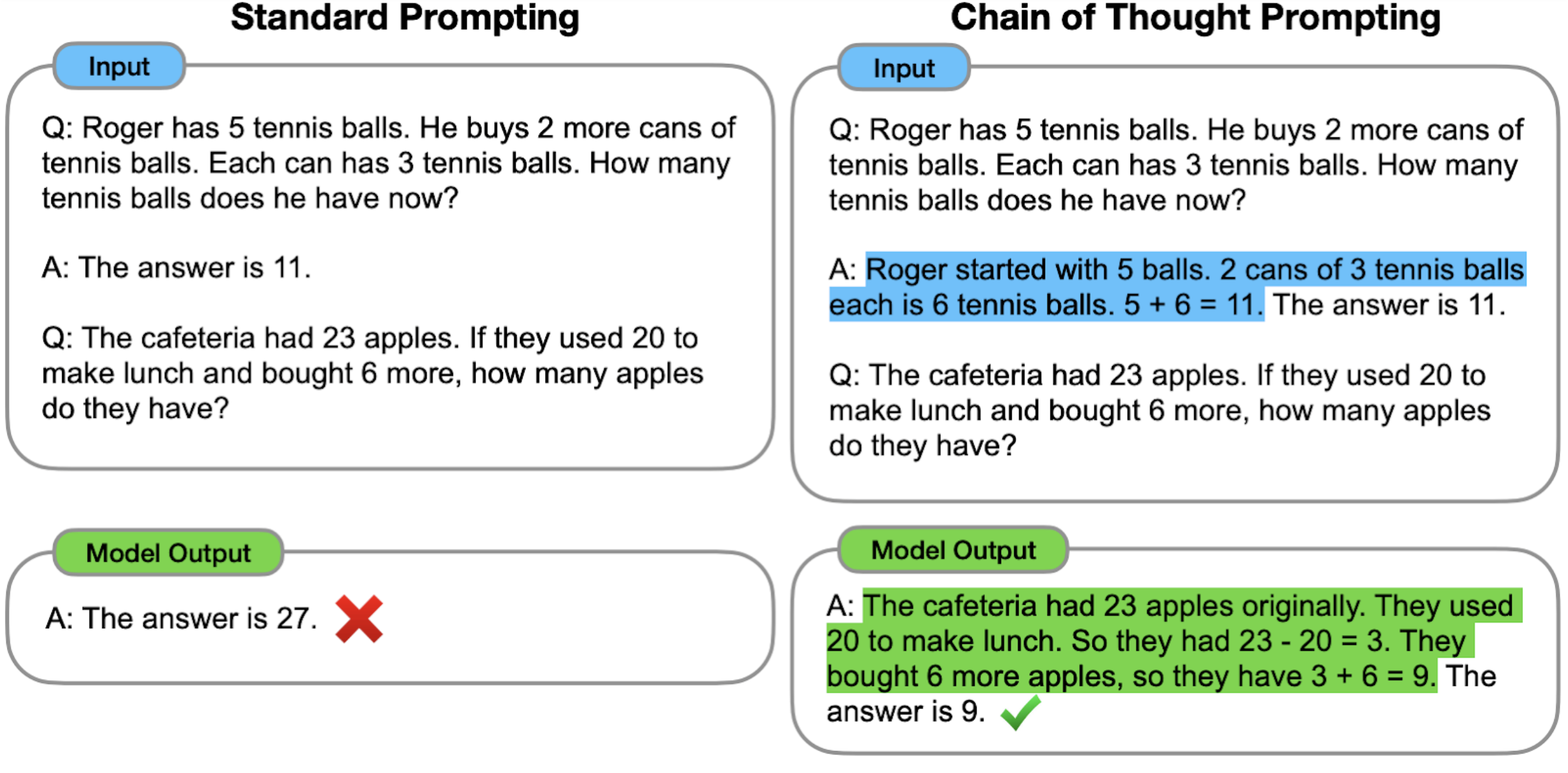}}
\\
\subfloat[]{\includegraphics[width=\textwidth]{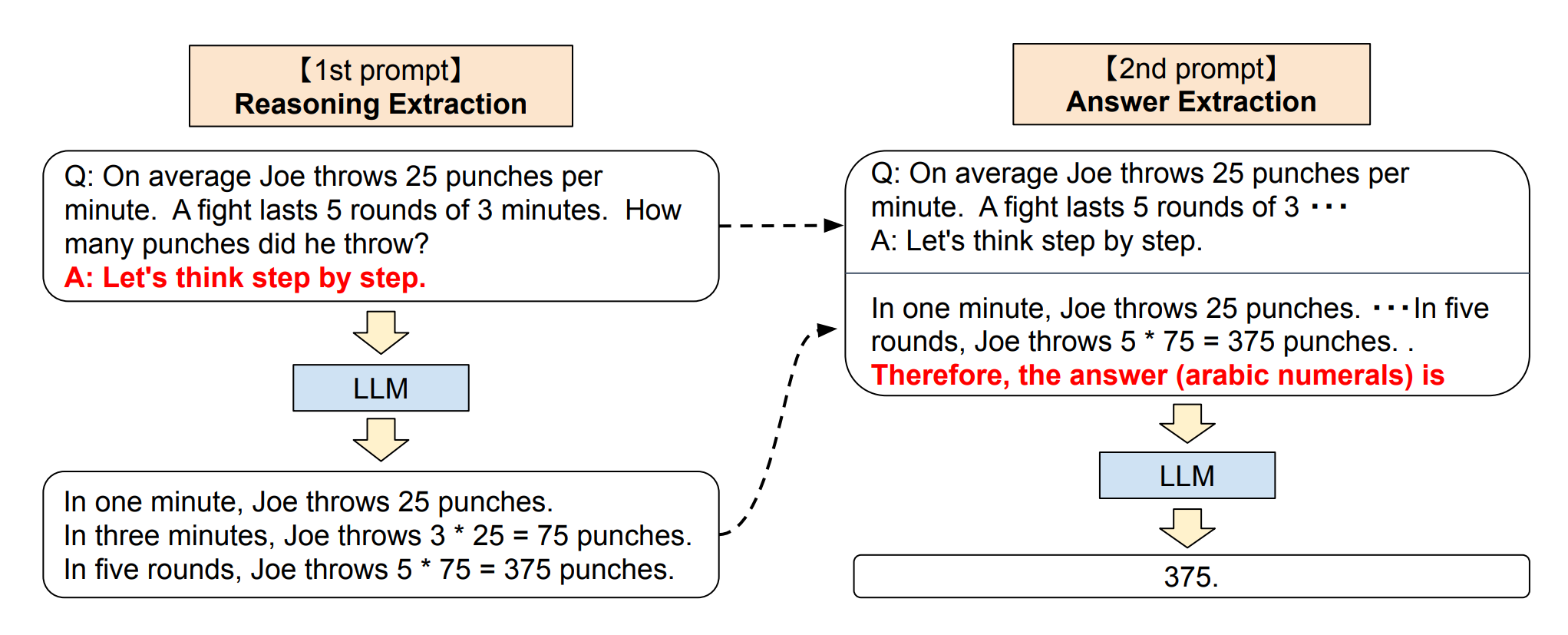}}
\begin{justify}
{Sources:~\cite{https://doi.org/10.48550/arxiv.2201.11903} and \cite{https://doi.org/10.48550/arxiv.2205.11916}. Chain of Thought (CoT) Prompt Engineering Approaches: a) A CoT approach using a single inference \citep{https://doi.org/10.48550/arxiv.2201.11903}. b) The Zero-shot-CoT pipeline requires two prompted inferences. The first inference prompts the model to ``think step by step'' to obtain a complete reasoning path from the large pretrained language model. We ask for a synthetic conclusion in the second inference that prompts the model with ``Therefore, the answer is:''. These simple prompt engineering procedures significantly increase the model performance in the analyzed reasoning task \citep{https://doi.org/10.48550/arxiv.2205.11916}.}
\end{justify}
\label{fig:cot}
\end{figure}

\looseness=-2
Related to approaches that use deep learning for reasoning-related tasks, we have seen that large language models may be adapted to solve math\footnote{\url{https://openai.com/blog/formal-math/}}\textsuperscript{,}\footnote{\url{https://openai.com/blog/grade-school-math/}} \citep{DBLP:journals/corr/abs-2110-14168} (Fig.~\ref{fig:math}), arithmetic, logical, and symbolic reason-based problems expressed in natural language. One possible idea is to ask the model to tackle the problem by prompting the model. Similarly to humans, ``thinking'' step by step helps to solve logical statements\footnote{\url{https://ai.googleblog.com/2022/05/language-models-perform-reasoning-via.html}} \citep{https://doi.org/10.48550/arxiv.2201.11903,https://doi.org/10.48550/arxiv.2205.11916} (Fig.~\ref{fig:cot}). Some recent works are using language models for robotics\footnote{\url{https://say-can.github.io/}}. 

\looseness=-1
Nowadays, we have much more resources efficient models \citep{DBLP:conf/icml/TanL19,https://doi.org/10.48550/arxiv.2110.02178} and also techniques such as quantization \citep{https://doi.org/10.48550/arxiv.2103.13630}, pruning \citep{https://doi.org/10.48550/arxiv.2101.09671}, and distillation \citep{DBLP:journals/ijcv/GouYMT21} that allow us to deploy deep neural networks into constrained embedded systems and devices. Finally, we have expectations\footnote{\url{https://spectrum.ieee.org/analog-ai}} that these models will consume much less energy in the near future \citep{https://doi.org/10.48550/arxiv.2204.05149, deepphysical}.

\newpage

\clearpage

\section{Problem: Deep Learning Robustness}

Despite recent advances in self-supervision \citep{DBLP:journals/corr/abs-2002-05709,DBLP:journals/corr/abs-2006-07733}, it is indisputable that the vast majority of success cases of deep learning practical applications are based on pure supervised learning or fine-tuning of pretrained models. In supervised deep learning, the classification task is the cornerstone of building more specialized tasks in computer vision (object detection \citep{DBLP:conf/cvpr/RedmonDGF16}, semantic segmentation \citep{DBLP:journals/corr/ChenPK0Y16}, instance segmentation \citep{DBLP:conf/nips/RenHGS15,DBLP:conf/iccv/HeGDG17}, etc.) and natural language processing (text classification \citep{10.1145/3439726}, named entity recognition \citep{9039685}, etc.).

\begin{figure*}
\small
\centering
\caption[The Supervised Learning Paradigm and Unknown Distributions]{The Supervised Learning Paradigm and Unknown Distributions}
\includegraphics[width=0.9\textwidth,trim={0 0 0 0},clip]{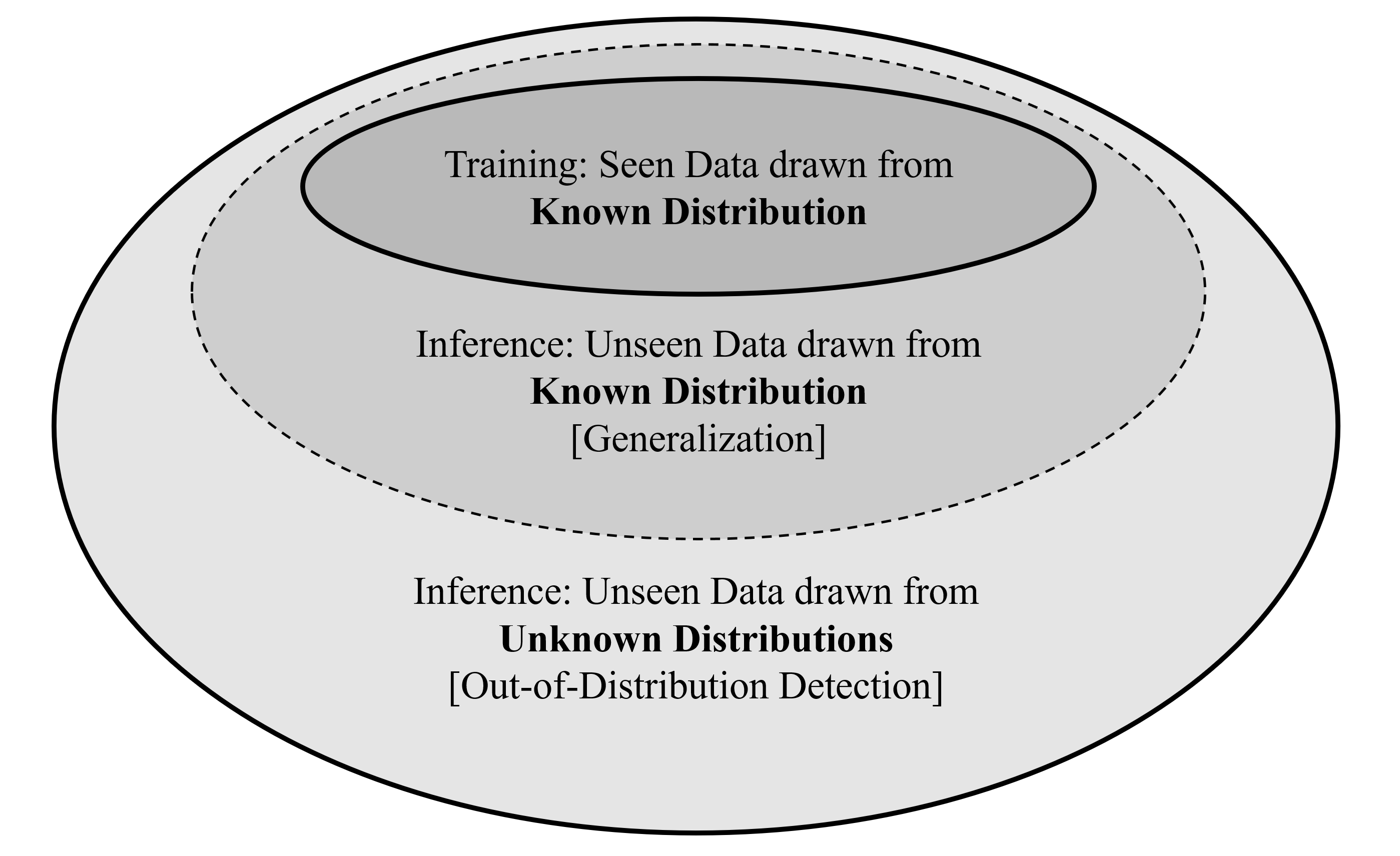}
\begin{justify}
{Source:~The Author (2022). The Supervised Learning Paradigm and the Unknown Distributions: The machine learning paradigm is based on using data to understand the world and generalize to novel situations. However, to acquire this capability, the model is presented with data that necessarily represent a limited portion of the real-world complexity. In classification problems, during training, the model is trained with examples of known classes. During inference, the system usually satisfactorily generalizes whether unseen examples of these known classes have been presented. However, if a sample that does not represent an instance of a known class is shown to the system, \emph{how may it know that it does not know?} Therefore, one of the significant challenges to current deep learning (and, in general, machine learning) systems is realizing when it cannot reliably perform the task it was trained to and behave appropriately.}
\end{justify}
\label{fig:problem}
\end{figure*}

However, deep learning classifiers present a significant drawback in their current form. \hl{We usually expect to present that an instance of a known class is presented to the neural network for inference.} If this holds, neural networks commonly show satisfactory performance. However, in real-world applications, which are becoming even more common after the deep learning advent, this assumption may not be fulfilled.

\newpage

Indeed, it is unrealistic to expect training examples to thoroughly consider the complexity to which the system will be submitted when in the field. We emphasize that this is not a problem exclusively for deep learning, but rather for machine learning classification systems in general. The possibility of being presented during test an instance that does not belong to any training class is a limitation of the learning paradigm itself in its current form (\figref{fig:problem}). The relevance of this problem \hln{incentivizes} researchers to propose alternatives to the principle of Empirical Risk Minimization (ERM) \citep{DBLP:conf/nips/Vapnik91}. For example, \cite{DBLP:journals/corr/abs-2003-00688} propose the principle of Risk Extrapolation (REx) that incorporates the unknown distribution aspect of the problem into the learning theory.

\begin{figure*}
\small
\centering
\caption[Unknown Distributions Examples]{Unknown Distributions Examples}
\includegraphics[width=\textwidth,trim={0 0 0 0},clip]{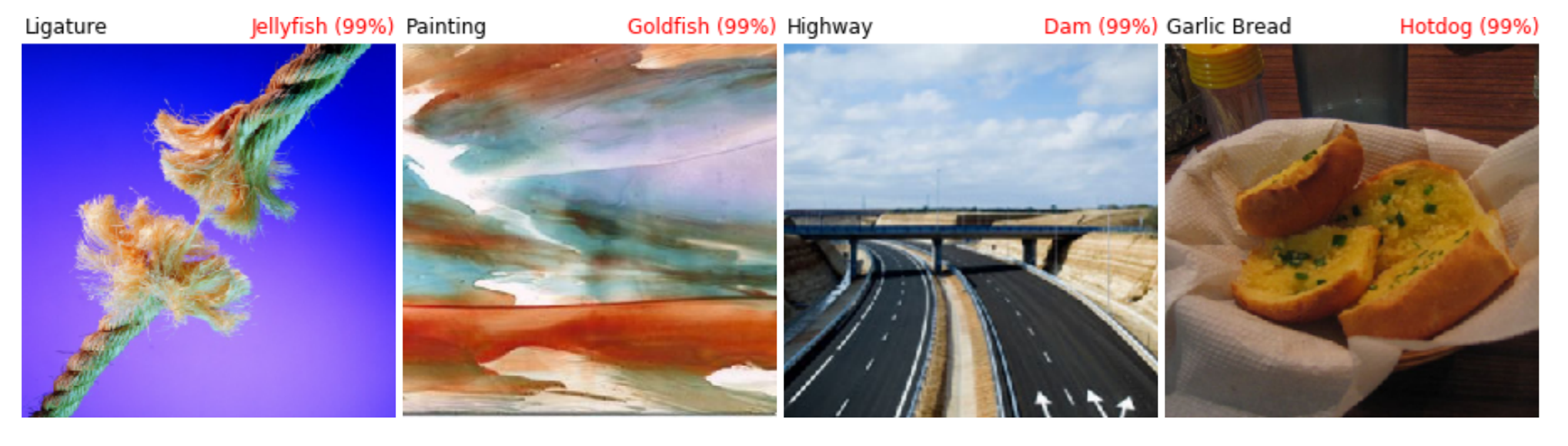}
\begin{justify}
{Source:~\cite{DBLP:journals/corr/abs-1907-07174}. Images from ImageNet-O dataset, which was created for the OOD detection task. These examples do not belong to ImageNet classes. For each image, the black text represents the correct ImageNet-O class not presented in ImageNet. The red text is the confidence of the ResNet-50 prediction that the image belongs to an ImageNet class. Images of unknown distributions are wrongly assigned highly confident predictions by the model trained on ImageNet.}
\end{justify}
\label{fig:natural}
\end{figure*}

The ability to detect whether an input applied to a neural network does not represent an example of trained classes is essential to building robust applications in medicine, finance, agriculture, engineering, fraud detection, and many others. In such situations, it is better to have a system that can recognize that the sample should not be classified. Instead, current systems classify unknown class instances and usually present very high confidence for them (\figref{fig:natural}).

Indeed, the rapid adoption of neural networks in modern real-world applications makes the development of systems that can detect when dealing with examples that belong to unknown distributions a primary necessity from a practical point of view. Besides interpretability, casual inference, reasoning, common sense, privacy, fairness, security, and other robustness aspects, this constitutes one of the primary challenges to construct more reliable deep learning systems.

\begin{figure*}
\small
\centering
\caption[Out-of-Distribution Detection Problem Statement]{Out-of-Distribution Detection Problem Statement}
\includegraphics[width=\textwidth,trim={0 0 0 0},clip]{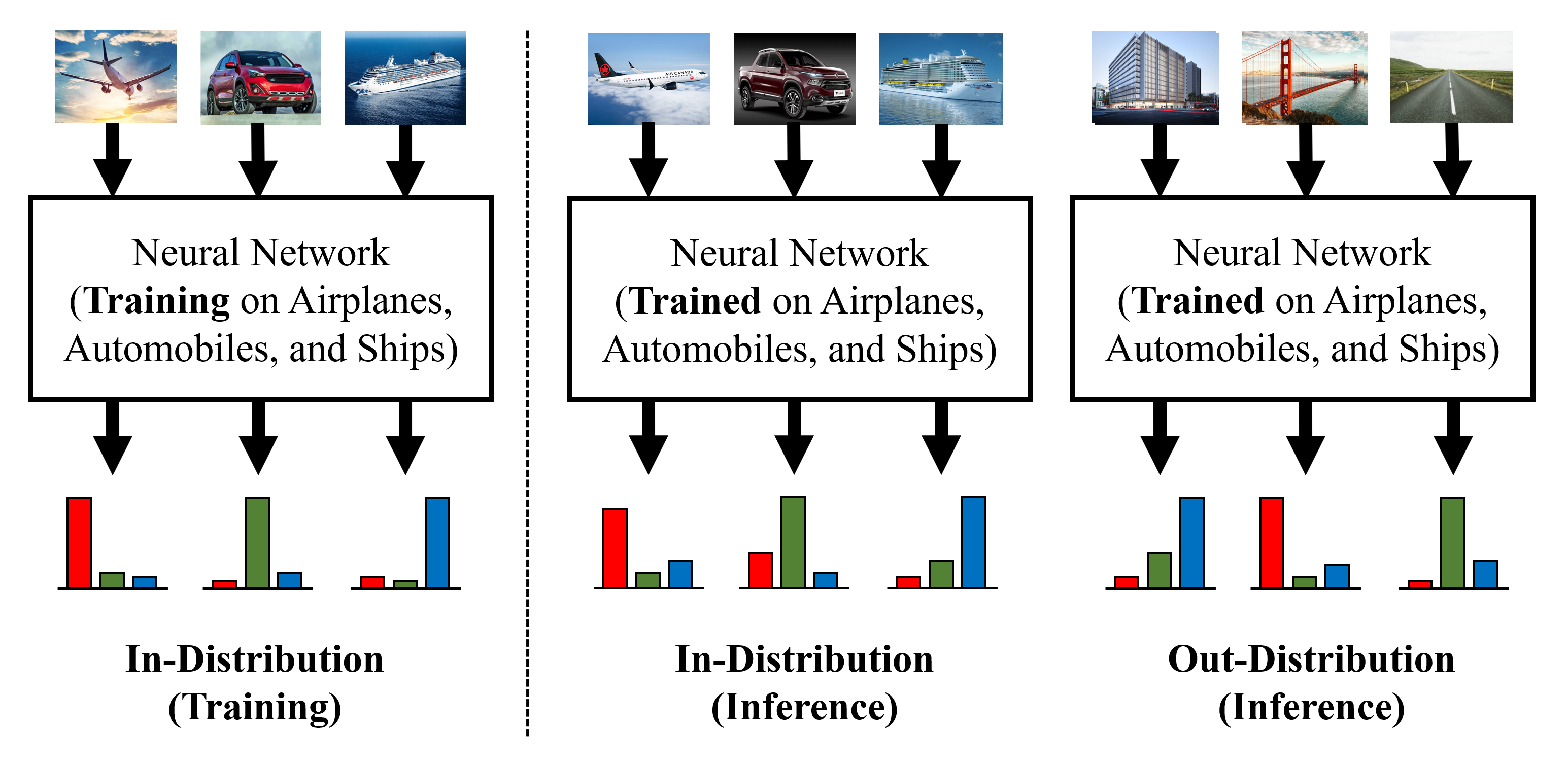}
\begin{justify}
{Source:~The Author (2022). Out-of-distribution detection: Deep models are trained on a limited set of a priori known classes, learning a combined in-distribution (left). During the inference, if an instance of a trained class (an example sampled from the in-distribution) is presented to the considered system, the performance is usually satisfactory (middle). However, in real-world applications, the system is commonly subjected to examples of unknown classes (instances sampled from unknown distributions, collectively called out-distributions). In such situations, current models usually provide high-confidence predictions. It is a problem that affects any classification machine learning system, rather than only deep learning models. Out-of-distribution detection is the ability to detect such situations and consequently avoid nonsense classification.}
\end{justify}
\label{fig:ood_problem_statement}
\end{figure*}

\newpage

\looseness=-1
This problem has been studied under many similar or related nomenclatures, such as spurious patterns \citep{DBLP:journals/prl/VasconcelosFB95}, open set recognition \citep{Scheirer_2013_TPAMI, Scheirer_2014_TPAMIb} and open-world recognition \citep{7298799,Rudd_2018_TPAMI}. Recently, to quantify the advance in our ability to construct a more reliable deep learning, \cite{hendrycks2017baseline} defined out-of-distribution (OOD) detection as the task of evaluating whether a sample does not come from the \gls{id} on which a neural network was trained (\figref{fig:ood_problem_statement}).

\looseness=-1
OOD detection is closely related to anomaly and novelty detection \citep{10.1145/3439950}. However, in OOD detection, we have multiple (usually many more than two) classes. From an anomaly detection perspective, examples that belong to any of these classes are considered ``normal''. Additionally, we have labels that individually identify examples from each of these ``normal'' classes, which collectively represent what we call the in-distribution. There are no training examples of the ``abnormal'' class, which are called out-of-distribution examples in the context of OOD detection. We have to decide whether we have an in-distribution (``normal'') example or an out-of-distribution (``abnormal'') example during inference. In the first case, we also have to predict the correct class.


The relevance of building deep learning systems that support out-of-distribution detection can not be overestimated. Indeed, in his Turing Award Lecture\footnote{\url{https://www.youtube.com/watch?v=llGG62fNN64}} and in the invited talk at NeurIPS 2019\footnote{\url{https://slideslive.com/38922304/from-system-1-deep-learning-to-system-2-deep-learning}}, Yoshua Bengio stated that out-of-distribution detection is currently one of the major challenges to move AI forward.

There are many real-world cases in which \gls{ood} detection support is important for the solution to work properly. For example, Andrej Karpathy, the AI Tesla Director, explains how a self-driving car system trained on a set of standard stop signs usually has a hard time identifying ``weird'' stop signs{\small\footnote{\url{https://www.youtube.com/watch?v=hx7BXih7zx8}}}. In bacteria detection approaches, it is crucial to know when a new bacterium may exist\footnote{\url{https://ai.googleblog.com/2019/12/improving-out-of-distribution-detection.html}}. 

\hln{Out-of-distribution capabilities are particularly important in many medical applications} \citep{DBLP:journals/corr/abs-2109-14885}. \hln{For example, it is critical when dealing with applications dealing with Electronic Health Records. Food classifiers should be able to reject other kind of images} \citep{DBLP:journals/corr/abs-2110-11334}. Therefore, in \gls{ood} detection capable approaches, the system presents an auxiliary task that evaluates whether it is indeed able to perform the primary classification task reliably.  

\cite{hendrycks2017baseline} introduced benchmarks for \gls{ood} detection. They established the baseline performance by proposing an \gls{ood} detection approach that uses the maximum predicted probability as the score for detection. Despite being a fundamental task to build reliable AI systems, most current \gls{ood} detection approaches are based on ad hoc techniques that produce unwanted side effects and add troublesome requirements to the solution.

\begin{figure*}
\small
\centering
\caption[Reliability Diagrams and Expected Calibration Errors]{Reliability Diagrams and Expected Calibration Errors}
\includegraphics[width=\textwidth,trim={0 0 0 0},clip]{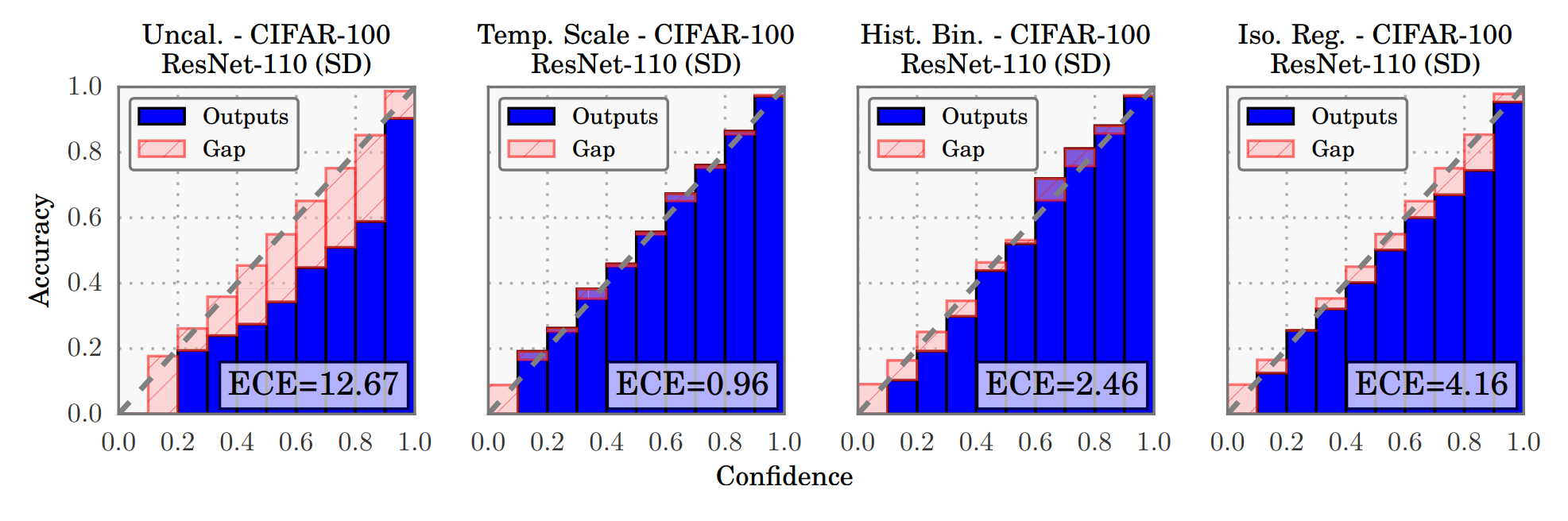}
\begin{justify}
{Source: \cite{Guo2017OnCO}. Reliability diagrams and Expected Calibration Error (ECE) for CIFAR-100: Before calibration (far left), the ECE is higher than after using calibration techniques (middle left, middle right, far right).}
\end{justify}
\label{fig:reliability_diagrams}
\end{figure*}


In addition to OOD detection, uncertain estimation is relevant to producing more robust deep learning systems. Indeed, \cite{Guo2017OnCO} showed that neural networks are usually uncalibrated because their predicting probabilities are not representative of the actual correctness likelihood, which may be a significant drawback in many applications. It is reasonable to want the predicted probability to represent somehow the chance of the classification to be correct.

Besides showing that deep networks are usually uncalibrated (Fig.~\ref{fig:reliability_diagrams}), \cite{Guo2017OnCO} used the Expected Calibration Error (ECE) to measure this. Finally, they also proposed many approaches to calibrate them.

\hln{On the one hand, OOD detection deals with rejecting examples that should not even be classified because they are \emph{not} instances of trained classes. On the other hand, uncertainty estimation tackles the problem of assigning realistic probabilities to \emph{in-distribution} samples. We have observed in the literature that this two tasks are usually study simultaneously}\footnote{\url{https://sites.google.com/view/udlworkshop2021/home}}.

\newpage

\section{Motivation}\label{sec:motivation}

The explanations above show the potential of deep learning to be used even more broadly in practice. It certainly has the potential to improve the quality of our lives. Since the groundbreaking \hln{results obtained} in 2012 with the ImageNet competition, we have noticed that deep learning has been expanding fast in the last ten years.

Indeed, from a technology used mainly in computer vision a decade ago, deep learning has currently been delivering remarkable state-of-the-art results also in natural language processing, speech, and audio processing. The adoption of deep learning approaches is not restricted to unstructured data, as methods for tabular data and time series have been proposed. We have also witnessed promising approaches to tackle robotics using pretrained language models.

We have recently seen deep learning approaches being used to promote relevant scientific discoveries and even to tackle reasoning-like tasks using large pretrained language models. It is somewhat surprising that neural networks, historically associated with perception-like tasks, are being used successfully for reasoning tasks.

Novel technologies are making it possible to train and use deep learning models with a significantly reduced amount of \hlf{labeled} data (few-shot and one-shot approaches) or even no \hlf{labeled} data at all (self-supervision and contrastive learning). Recent techniques are allowing deploying deep models into resource-constrained embedded devices. Novel promising technologies show that it is possible to dramatically reduce the carbon footprint and energy consumption to train and perform inferences with such systems. 

However, for some applications, it is crucial to reject no-sense predictions or mismatch something known with something unknown (e.g., self-driving cars). In other cases, it is critical to produce predictions with probabilities that reflect the real chance that the system is correct in its predictions. For example, for cancer diagnoses, the system must inform a probability that reflects accurate chances that it is indeed correct.

Therefore, contributing to the incorporation of those capabilities into AI-based systems is the primary motivation of this work. We believe constructing more robust deep learning systems is essential to improving people's lives. 

\begin{figure*}
\small
\centering
\caption[Current Approaches vs. Desired Solutions]{Current Approaches vs. Desired Solutions}
\subfloat[]{\includegraphics[width=\textwidth,trim={0 0 0 0},clip]{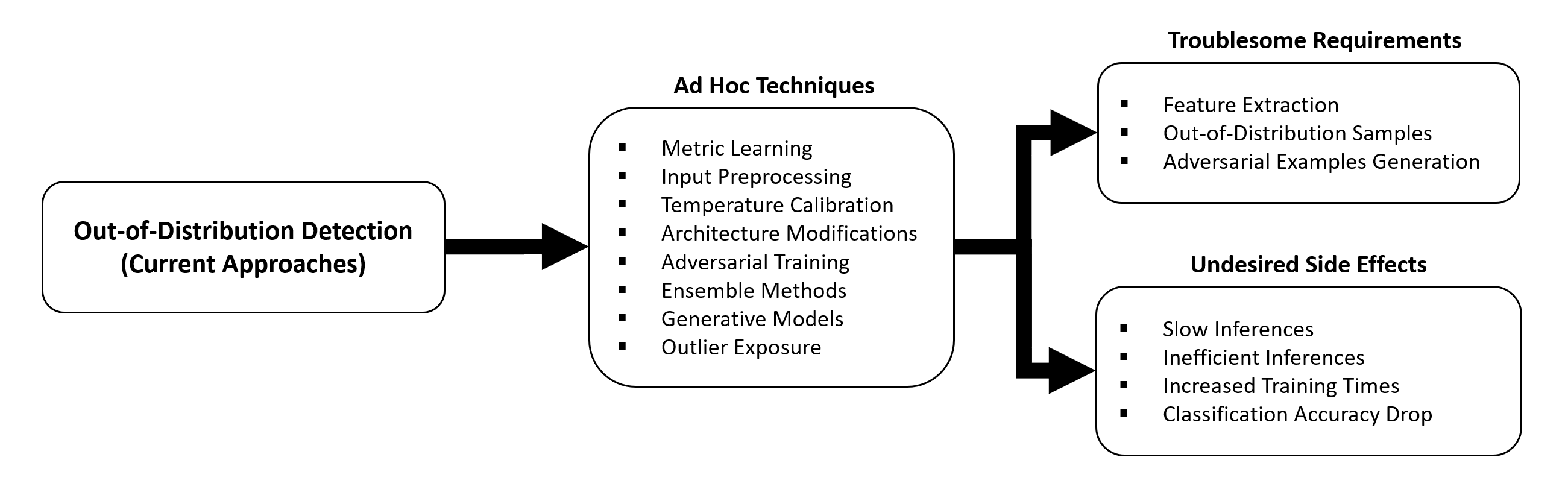}\label{fig:overview_current}}
\\
\subfloat[]{\includegraphics[width=\textwidth,trim={0 0 0 0},clip]{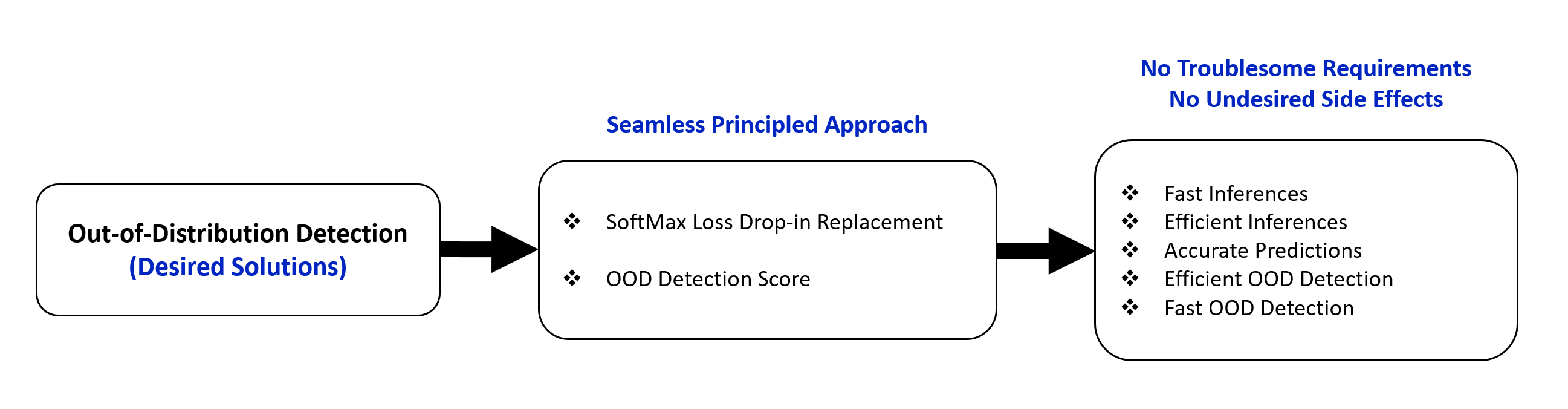}\label{fig:overview_goal}}
\begin{justify}
{Source: The Author (2022). (a) Current approaches typically use a combination of conventional ad hoc techniques, such as input preprocessing, temperature calibration, adversarial training, and metric learning. However, these techniques add troublesome requirements and undesired side effects to the overall solution. Unrealistic access to OOD examples in design time is a common undesired requirement. Slow and inefficient inferences compared to those performed by neural networks trained using standard procedures) and Classification Accuracy (ACC) drop are usually unwanted side effects. (b) Unlike current approaches, in this work, our goal is to design solutions that completely avoid conventional ad hoc techniques and associated side effects and requirements. Consequently, models trained using our loss should produce accurate predictions (no classification accuracy drop) and fast and efficient inferences. Unlike many current OOD detection methods, feature extraction and metric learning after neural network training should also be avoided. The solutions have to be based on theoretical motivations. In this figure, desired solutions represent the characteristics of the solutions we aim to achieve.}
\end{justify}
\label{fig:overview}
\end{figure*}

\newpage\section{Objective}\label{sec:goal}

The main objective of this work is to design an \gls{ood} detection approach that, unlike current ones, allows out-of-distribution detection without relying on ad hoc techniques that add extra requirements or unwanted side effects to the overall solution (\figref{fig:overview_current}).

Throughout this work, we argue that the unsatisfactory \gls{ood} detection performance of modern neural networks is mainly due to the drawbacks of the currently used loss rather than architecture limitations. Indeed, the SoftMax loss\footnote{We follow the ``SoftMax loss'' expression as defined in \cite{liu2016large}.} (i.e., the combination of the output linear layer, the SoftMax activation, and the cross-entropy loss) anisotropy \hln{caused by the linear layer} does not induce the concentration of high-level representations in the feature space \citep{DBLP:conf/eccv/WenZL016,Hein2018WhyRN}, which makes \gls{ood} detection difficult \citep{Hein2018WhyRN}. Furthermore, the SoftMax loss generates extremely low entropy (high confidence) posterior probability distributions \citep{Guo2017OnCO} in conflict with the fundamental principle of maximum~entropy.

Therefore, we aim to contemplate novel losses that are able to work as drop-in replacements to the one currently used to train neural networks, therefore avoiding the need to change the model. No changes to the training procedure should be required. \hln{Hence, to enforce isotropy, we aim to design an exclusively distance-based loss able to train deep neural networks end-to-end in a straightforward way.} Furthermore, to perform inference time \gls{ood} detection, we intend to design novel scores\footnote{\hln{A score is a scalar that measures the likelihood of an instance belonging to the in or the out-distribution.}} that need to be fast and computationally efficient\footnote{In this work, we consider that an approach does not present classification accuracy drop if it never presents a classification accuracy more than one percent lower than the correspondent SoftMax loss trained model.} (\figref{fig:overview_goal}).

To construct the desired solutions, our scores need to be defined \emph{a priori} rather than be trainable. In addition, we aim to base the proposed approaches on solid theoretical foundations, most likely based on the fundamentals of information theory. For all these reasons, we call the overall solutions to be designed seamless and principled \gls{ood} detection methods.

To achieve these objectives, we investigate the following fundamental research questions:

\begin{enumerate}[label=\textbf{RQ{\arabic*}}.]
\item Is it possible to perform OOD by changing only the losses? How should we calculate the score in such cases?
\item Is there any theoretic motivation (e.g., a fundamental principle) that sustains such \gls{ood} detection methods?
\item Which characteristics would such solutions present?
\end{enumerate}

\newpage
In summary, we intend to study whether it is possible to design \gls{ood} detection approaches competitive to (or that outperform) current practices, avoiding their troublesome requirements and unwanted side effects. To answer \textbf{RQ1}, \textbf{RQ2}, and \textbf{RQ3}, we intend to investigate current approaches limitations and propose novel \gls{ood} detection solutions by focusing exclusively on neural networks and information theory foundations.

\newpage\section{Publications}

Here, we present the list of our \emph{deep learning} publications:\footnote{For an updated version, please visit: \url{https://scholar.google.com/citations?user=hypWII4AAAAJ&hl=en}}:

\subsection*{Books}

\subsubsection*{Foundations}

\begin{itemize}
\item
David Macêdo.
\textbf{Enhancing Deep Learning Performance Using Displaced Rectifier Linear Unit.}
\textit{Editora Dialética}, ISBN Physical book: 978-65-252-3056-6, ISBN E-book: 978-65-252-3075-7, 2022.
\end{itemize}

\subsection*{Related Papers}

\subsubsection*{Foundations}

\begin{itemize}
\item
D Macêdo, TI Ren, C Zanchettin, ALI Oliveira, T Ludermir.
\textbf{Entropic Out-of-Distribution Detection.}
\textit{2021 International Joint Conference on Neural Networks (IJCNN).}
\end{itemize}

\begin{itemize}
\item
D Macêdo, TI Ren, C Zanchettin, ALI Oliveira, T Ludermir.
\textbf{Entropic Out-of-Distribution Detection: Seamless Detection of Unknown Examples.}
\textit{IEEE Transactions on Neural Networks and Learning Systems, 2022.}
\end{itemize}

\begin{itemize}
\item
D Macêdo, T Ludermir.
\textbf{Enhanced Isotropy Maximization Loss: Seamless and High-Performance Out-of-Distribution Detection Simply Replacing the SoftMax Loss.}
\textit{arXiv preprint arXiv:2105.14399 (paper under review).}
\end{itemize}

\begin{itemize}
\item
D Macêdo, C Zanchettin, T Ludermir.
\textbf{Distinction Maximization Loss: Efficiently Improving Out-of-Distribution Detection and Uncertainty Estimation by Replacing the Loss and Calibrating.}
\textit{arXiv preprint arXiv:2205.05874 (paper under review).}
\end{itemize}

\newpage

\subsection*{Additional Papers}

\subsubsection*{Foundations}

\begin{itemize}
\item
D Macêdo, C Zanchettin, ALI Oliveira, T Ludermir.
\textbf{Enhancing Batch Normalized Convolutional Networks using Displaced Rectifier Linear Units: A Systematic Comparative Study.}
\textit{Expert Systems with Applications, 2019.}
\end{itemize}

\begin{itemize}
\item
D Macêdo, P Dreyer, T Ludermir, C Zanchettin.
\textbf{Training Aware Sigmoidal Optimizer.}
\textit{arXiv preprint arXiv:2102.08716.}
\end{itemize}

\subsubsection*{Computer Vision}

\begin{itemize}
\item
MEN Gomes, D Macêdo, C Zanchettin, PSG de~Mattos~Neto, A Oliveira
\textbf{Multi-human Fall Detection and Localization in Videos.}
\textit{Computer Vision and Image Understanding, 2022.}
\end{itemize}

\begin{itemize}
\item
A Ayala, B Fernandes, F Cruz, D Macêdo, C Zanchettin.
\textbf{Convolution Optimization in Fire Classification.}
\textit{IEEE Access, 2022.}
\end{itemize}

\begin{itemize}
\item
W Costa, D Macêdo, C Zanchettin, LS Figueiredo, V Teichrieb.
\textbf{Multi-Cue Adaptive Emotion Recognition Network.}
\textit{arXiv preprint arXiv:2111.02273.}
\end{itemize}

\begin{itemize}
\item
A Ayala, B Fernandes, F Cruz, D Macêdo, ALI Oliveira, C Zanchettin.
\textbf{Kutralnet: A portable deep learning model for fire recognition.}
\textit{2020 International Joint Conference on Neural Networks (IJCNN).}
\end{itemize}

\begin{itemize}
\item
RB das Neves, LF Verçosa, D Macêdo, BLD Bezerra, C Zanchettin.
\textbf{A Fast Fully Octave Convolutional Neural Network for Document Image Segmentation.}
\textit{2020 International Joint Conference on Neural Networks (IJCNN).}
\end{itemize}

\begin{itemize}
\item
H Felix, WM Rodrigues, D Macêdo, F Simões, ALI Oliveira, V Teichrieb, C Zanchettin.
\textbf{Squeezed Deep 6DoF Object Detection Using Knowledge Distillation.}
\textit{2020 International Joint Conference on Neural Networks (IJCNN).}
\end{itemize}

\begin{itemize}
\item
JLP Lima, D Macêdo, C Zanchettin.
\textbf{Heartbeat Anomaly Detection using Adversarial Oversampling.}
\textit{2019 International Joint Conference on Neural Networks (IJCNN).}
\end{itemize}

\begin{itemize}
\item
D Castro, D Pereira, C Zanchettin, D Macêdo, BLD Bezerra.
\textbf{Towards Optimizing Convolutional Neural Networks for Robotic Surgery Skill Evaluation.}
\textit{2019 International Joint Conference on Neural Networks (IJCNN).}
\end{itemize}

\begin{itemize}
\item
CC de Amorim, D Macêdo, C Zanchettin.
\textbf{Spatial-Temporal Graph Convolutional Networks for Sign Language Recognition.}
\textit{ICANN 2019: Artificial Neural Networks and Machine Learning.}
\end{itemize}

\begin{itemize}
\item
AG Santos, CO de Souza, C Zanchettin, D Macêdo, ALI Oliveira, T Ludermir.
\textbf{Reducing SqueezeNet Storage Size with Depthwise Separable Convolutions.}
\textit{2018 International Joint Conference on Neural Networks (IJCNN).}
\end{itemize}

\begin{itemize}
\item
LA de Oliveira Junior, HR Medeiros, D Macêdo, C Zanchettin, ALI Oliveira, T Ludermir.
\textbf{SegNetRes-CRF: A Deep Convolutional Encoder-Decoder Architecture for Semantic Image Segmentation.}
\textit{2018 International Joint Conference on Neural Networks (IJCNN).}
\end{itemize}

\subsubsection*{Natural Language Processing}

\begin{itemize}
\item
J Pereira, D Macêdo, C Zanchettin, A Oliveira, R Fidalgo.
\textbf{PictoBERT: Transformers for Next Pictogram Prediction.}
\textit{Expert Systems With Applications, 2022.}
\end{itemize}

\begin{itemize}
\item
J Moreira, C Oliveira, D Macêdo, C Zanchettin, L Barbosa.
\textbf{Distantly-Supervised Neural Relation Extraction with Side Information using BERT.}
\textit{2020 International Joint Conference on Neural Networks (IJCNN).}
\end{itemize}

\begin{itemize}
\item
J Abreu, L Fred, D Macêdo, C Zanchettin.
\textbf{Hierarchical Attentional Hybrid Neural Networks for Document Classification.}
\textit{ICANN 2019: Artificial Neural Networks and Machine Learning.}
\end{itemize}

\begin{itemize}
\item
AB Duque, LLJ Santos, D Macêdo, C Zanchettin.
\textbf{Squeezed Very Deep Convolutional Neural Networks for Text Classification.}
\textit{ICANN 2019: Artificial Neural Networks and Machine Learning.}
\end{itemize}

\begin{itemize}
\item
FAO Santos, KL Ponce-Guevara, D Macêdo, C Zanchettin.
\textbf{Improving Universal Language Model Fine-Tuning using Attention Mechanism.}
\textit{2019 International Joint Conference on Neural Networks (IJCNN).}
\end{itemize}

\subsubsection*{Speech Processing}

\begin{itemize}
\item
JAC Nunes, D Macêdo, C Zanchettin.
\textbf{AM-MobileNet1D: A Portable Model for Speaker Recognition.}
\textit{2020 International Joint Conference on Neural Networks (IJCNN).}
\end{itemize}

\begin{itemize}
\item
JAC Nunes, D Macêdo, C Zanchettin.
\textbf{Additive Margin SincNet for Speaker Recognition.}
\textit{2019 International Joint Conference on Neural Networks (IJCNN).}
\end{itemize}

\subsubsection*{Tabular Data}

\begin{itemize}
\item
AA Silva, AS Xavier, D Macêdo, C Zanchettin, ALI Oliveira
\textbf{An Adapted GRASP Approach for Hyperparameter Search on Deep Networks Applied to Tabular Data.}
\textit{2022 International Joint Conference on Neural Networks (IJCNN).}
\end{itemize}

\subsubsection*{Time Series}

\begin{itemize}
\item
PM Vasconcelos, D Macêdo, LM Almeida, RGL Neto, CA Benevides, Cleber Zanchettin, Adriano LI Oliveira.
\textbf{Identification of Microorganism Colony Odor Signature using InceptionTime.}
\textit{2021 IEEE International Conference on Systems, Man, and Cybernetics (SMC).}
\end{itemize}

\subsubsection*{Information Security}

\begin{itemize}
\item
PF de Araujo-Filho, G Kaddoum, DR Campelo, AG Santos, D Macêdo, C Zanchettin.
\textbf{Intrusion Detection for Cyber–Physical Systems using Generative Adversarial Networks in Fog Environment.}
\textit{IEEE Internet of Things Journal, 2020.}
\end{itemize}

\subsubsection*{Financial Services}

\begin{itemize}
\item
JCS Silva, D Macêdo, C Zanchettin, ALI Oliveira, AT de Almeida Filho.
\textbf{Multi-Class Mobile Money Service Financial Fraud Detection by Integrating Supervised Learning with Adversarial Autoencoders.}
\textit{2021 International Joint Conference on Neural Networks (IJCNN).}
\end{itemize}

\subsubsection*{Power Efficiency}

\begin{itemize}
\item
PFC Barbosa, BA da Silva, D Macêdo, C Zanchettin, RM de Moraes.
\textbf{Otimização do Consumo de Energia em Redes Ad Hoc Aloha Empregando Deep Learning.}
\textit{2019 Workshop em Desempenho de Sistemas Computacionais e de Comunicação (WPERFORMANCE).}
\end{itemize}

%% file: chapters/2.background.tex
\chapter{Background}\label{chap:background}

\begin{quotation}[]{Francis Bacon}
``Truth is the daughter of time, not of authority.''
\end{quotation}

\begin{quotation}[]{Isaac Newton}
``If I have seen further, it is by standing on the shoulders of giants.''
\end{quotation}

\begin{quotation}[Sworn testimony]{Witness}
``I swear to tell the truth, the whole truth, and nothing but the truth.''
\end{quotation}

\begin{quotation}[Braveheart]{William Wallace}
``If I swear to him, then all that I am is dead already.''
\end{quotation}

\begin{quotation}[Braveheart]{Last Scene}
Magistrate: ``The prisoner wishes to say a word.''\\William Wallace: ``Freeeedommm!''
\end{quotation}

In this chapter, considering one of our aims is to build isotropic (exclusively distance-based) losses designed to be applied to \gls{ood} detection, we discuss both current \gls{ood} detection approaches and distance-based losses. Moreover, we also introduce the maximum entropy principle that will have a fundamental role in our solutions.\\

\newpage\section{Current Approaches}

\begin{figure*}
\small
\centering
\caption[Out-of-Distribution Detection Relevance]{Out-of-Distribution Detection Relevance}
\includegraphics[width=\textwidth,trim={0 0 0 0},clip]{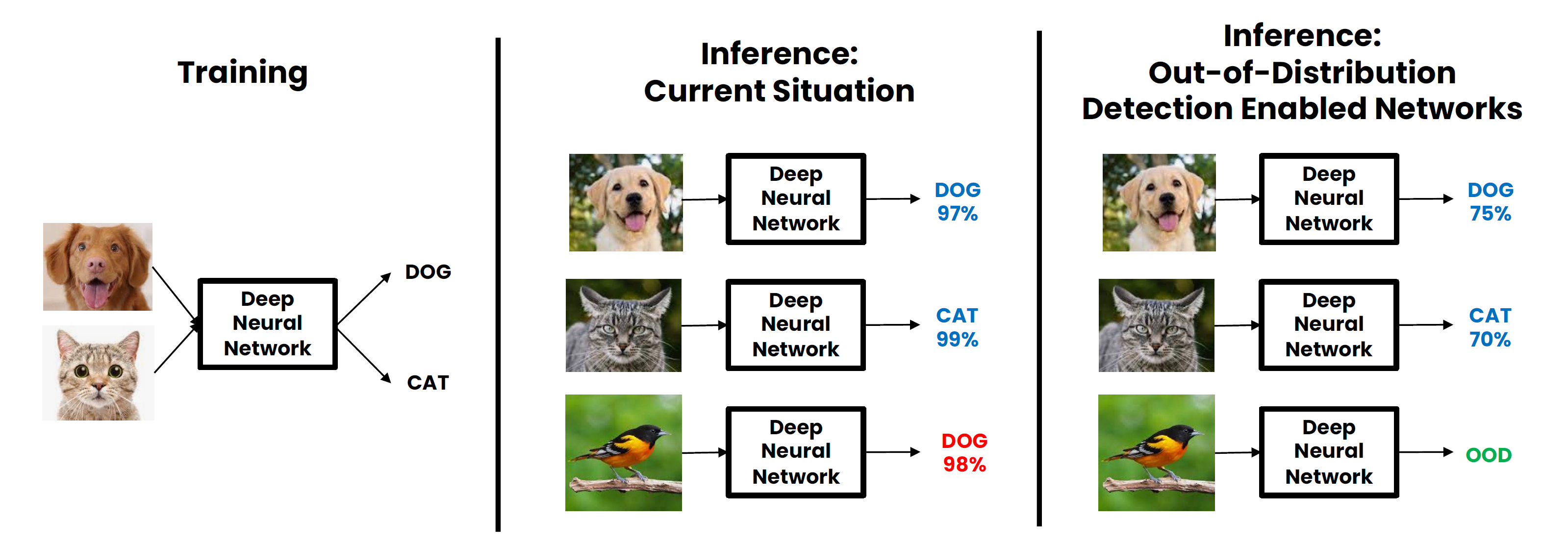}
\begin{justify}
{Source:~The Author (2022). Out-of-Distribution Detection Relevance: In the real world, the deep neural network is subject to out-of-distribution (OOD) examples. Ideally, the system should be able to recognize such situations.}
\end{justify}
\label{fig:ood_task_statement}
\end{figure*}

The relevance of the out-of-distribution detection auxiliary task is demonstrated in Fig.~\ref{fig:ood_task_statement} while Table~\ref{tab:overview} presents the limitations of major current approaches. In the following paragraphs, we explain in more detail their drawbacks. 

\gls{odin} was proposed in \cite{liang2018enhancing} by combining \emph{input preprocessing} with \emph{temperature calibration}. Although it significantly outperforms the baseline, the input preprocessing introduced in ODIN considerably increases the inference time by requiring an initial forward pass, a backpropagation, and finally a second forward pass to perform an inference that can be used for \gls{ood} detection.

Considering that backpropagation is typically slower than a forward pass, input preprocessing makes ODIN inferences at least three times slower than normal. Additionally, input preprocessing multiplies the inference power consumption and computational cost by at least three. Those are severe limitations from an economic and environmental perspective \citep{Schwartz2019GreenA}. Several subsequent \gls{ood} detection proposals incorporated input preprocessing and its drawbacks \citep{liang2018enhancing, lee2018simple, Hsu2020GeneralizedOD, DeVries2018LearningNetworks}.

Temperature calibration consists of changing the scale of the logits of a \emph{pretrained model}. Both input preprocessing and temperature calibration require hyperparameter tuning. \hlf{When proposing novel approaches, we should prefer the ones with the smallest possible number of hyperparameters to make the solution easier to use. It is essential in deep learning because we may require increased computational resources to perform extensive hyperparameter validation.}

Moreover, ODIN requires unrealistic access to \gls{ood} samples to validate hyperparameters. Even if the supposed \gls{ood} samples are available during design-time, using these examples to tune hyperparameters makes the solution overfit to detect this particular type of out-of-distribution.

In real-world applications, the system will likely operate under different/novel/unknown out-of-distributions, and the estimated \gls{ood} detection performance could degrade significantly. Therefore, using \gls{ood} samples to validate hyperparameters may produce over-optimistic \gls{ood} detection performance~estimations \citep{shafaei2018biased}.

\begin{table*}
\footnotesize
\caption{OOD Detection Approaches: Special Requirements and Side Effects}
\label{tab:overview}
\centering
\begin{tabularx}{\textwidth}{l|YY|YY}
\toprule
\multirow{3}{*}{Approach}
& \multicolumn{2}{c|}{Special Requirement}
& \multicolumn{2}{c}{Side Effect}\\
\cmidrule{2-5}
& Hyperparameter & Additional & Inefficient Inference & Classification\\
& Tuning & Data & or OOD Detection & Accuracy Drop\\
\midrule
ODIN\textsuperscript{1} & \color{red}Required & \color{blue}Not Required & \color{red}Present & \color{blue}Not Present\\
\midrule
Mahalanobis\textsuperscript{2} & \color{red}Required & \color{blue}Not Required & \color{red}Present & \color{blue}Not Present\\
\midrule
ACET\textsuperscript{3} & \color{red}Required & \color{blue}Not Required & \color{blue}Not Present & \color{red}Present\\
\midrule
Outlier Exposure\textsuperscript{4} & \color{blue}Not Required & \color{red}Required & \color{blue}Not Present & \color{blue}Not Present\\
\midrule
GODIN\textsuperscript{5} & \color{red}Required & \color{blue}Not Required & \color{red}Present & \color{red}Present\\
\midrule
Gram Matrices\textsuperscript{6} & \color{blue}Not Required & \color{blue}Not Required & \color{red}Present & \color{blue}Not Present\\
\midrule
Scaled Cosine\textsuperscript{7} & \color{blue}Not Required & \color{blue}Not Required & \color{blue}Not Present & \color{red}Present\\
\midrule
Energy-based\textsuperscript{8} & \color{red}Required & \color{red}Required & \color{blue}Not Present & \color{blue}Not Present\\
\midrule
Deep Ensemble\textsuperscript{9} & \color{blue}Not Required & \color{blue}Not Required & \color{red}Present & \color{blue}Not Present\\
\midrule
DUQ\textsuperscript{10} & \color{red}Required & \color{blue}Not Required & \color{red}Present & \color{blue}Not Present\\
\midrule
SNGP\textsuperscript{11} & \color{red}Required & \color{blue}Not Required & \color{red}Present & \color{blue}Not Present\\
\bottomrule
\end{tabularx}
\small
\begin{justify}
{Source: The Author (2022).
\textsuperscript{1}\cite{liang2018enhancing},
\textsuperscript{2}\cite{lee2018simple},
\textsuperscript{3}\cite{Hein2018WhyRN},
\textsuperscript{4}\cite{hendrycks2018deep},
\textsuperscript{5}\cite{Hsu2020GeneralizedOD},
\textsuperscript{6}\cite{Sastry2019DetectingOE},
\textsuperscript{7}\cite{techapanurak2019hyperparameterfree},
\textsuperscript{8}\cite{DBLP:journals/corr/abs-2010-03759},
\textsuperscript{9}\cite{lakshminarayanan2017simple},
\textsuperscript{10}\cite{Amersfoort2020SimpleAS},
\textsuperscript{11}\cite{DBLP:conf/nips/LiuLPTBL20}.}
\end{justify}
\end{table*}


The Mahalanobis distance-based method\footnote{For the rest of this work, the expression ``the Mahalanobis distance-based method'' is replaced by ``the Mahalanobis method''.} \citep{lee2018simple} overcomes the need for access to \gls{ood} samples by validating hyperparameters using adversarial examples, producing more realistic \gls{ood} detection performance estimates. However, the use of adversarial examples has the disadvantage of adding a cumbersome procedure to the solution. Even worse, the generation of adversarial samples itself requires hyperparameters, such as the maximum perturbations. Although adequate hyperparameters may be known for research datasets, they may be difficult to find for novel real-world~data. 

Moreover, as the Mahalanobis approach also requires input preprocessing, the previously mentioned drawbacks associated with this technique are still present in the Mahalanobis solution. Hence, both \gls{odin} \citep{liang2018enhancing} and the Mahalanobis distance-based method \citep{lee2018simple} use \emph{input preprocessing}, which produces remarkably slow and energy-inefficient inferences, which is undesired because it is important to make deep learning more computationally efficient \citep{Schwartz2019GreenA}. 

Furthermore, \emph{feature ensemble} introduced in the Mahalanobis approach also presents limitations. In fact, feature ensembles require training of ad hoc classification and regression models on features extracted from many network layers. Finally, the Mahalanobis method involves feature extraction and metric learning. Similar to the Mahalanobis distance-based approach, \cite{Scheirer_2013_TPAMI,Scheirer_2014_TPAMIb,7298799,Rudd_2018_TPAMI} require metric learning on features extracted from pretrained~models.


\looseness=-1
Hyperparameter tuning is also a drawback to methods based on adversarial training, such as \gls{acet} \citep{Hein2018WhyRN}, as we need to define the appropriated adversarial perturbations. Adversarial training is known for increasing training time \citep{DBLP:conf/iclr/WongRK20}, reducing classification accuracy \citep{Raghunathan2019AdversarialTC}, and presenting limited scalability when dealing with large-size images \citep{DBLP:conf/nips/ShafahiNG0DSDTG19}. Furthermore, adversarial training can cause a drop in classification accuracy \citep{Raghunathan2019AdversarialTC}.

In some cases, OOD detection proposals require architecture modifications \citep{yu2019unsupervised} or ensemble methods \citep{vyas2018out, lakshminarayanan2017simple}. Despite significantly improving the OOD detection performance, loss enhancement (regularization) techniques, such as outlier exposure \citep{hendrycks2018deep, papadopoulos2019outlier}, background methods \citep{NIPS2018_8129}, and the energy-based fine-tuning \citep{DBLP:journals/corr/abs-2010-03759} require the addition of carefully chosen extra/outlier/background data and expand memory usage. Moreover, they usually add hyperparameters to the solution. 

Solutions based on uncertainty (or confidence) estimation (or calibration) \citep{kendall2017uncertainties,Leibig2017LeveragingUI,malinin2018predictive,kuleshov2018accurate,subramanya2017confidence} usually present additional complexity, slow and energy-inefficient inferences \citep{Schwartz2019GreenA}, and OOD detection performance typically worse than ODIN \citep{shafaei2018biased,Hsu2020GeneralizedOD}.

\looseness=-1
The Entropic Open-Set loss and the Objectosphere loss were proposed in \cite{NIPS2018_8129}. These two losses used background samples to improve the performance of detecting unknown inputs. The Entropic Open-Set loss works like the usual SoftMax loss in the in-distribution training data, producing a low entropy for these samples. However, it forces maximum entropy in the background samples. The Objectosphere loss is the Entropic Open-Set loss with an added regularization factor that forces the feature magnitude of in-distribution samples to be near a predefined value $\xi$ while minimizing the feature magnitude of background samples.

Methods that use data-augmentation have been proposed. \cite{DBLP:conf/nips/TackMJS20} improve OOD detection in a self-supervised setting. \cite{Sastry2019DetectingOE} analyzed statistics of the activations of the pretrained model on training and validation data to detect OOD examples.


In 2019, on the one hand, IsoMax \citep{https://doi.org/10.48550/arxiv.1908.05569} proposed a \emph{non-squared Euclidean distance last layer to address out-of-distribution detection} in an \emph{end-to-end trainable way} (i.e., no feature extraction). On the other hand, \gls{sc} \citep{techapanurak2019hyperparameterfree} proposed using a \emph{cosine distance}. Although the scale factor in IsoMax is a \emph{constant} scalar called the entropy scale, Scaled Cosine requires the addition of a \emph{block of layers} to learn the scale factor. This is made up of an exponential function, batch normalization, and a linear layer that has the feature layer as input. Moreover, to present high performance, it is necessary to avoid applying weight decay to this \emph{extra learning block}. We believe that this additional learning block, which adds an ad hoc linear layer in the final of the neural network, may make the solution prone to \emph{overfitting} and explain the classification accuracy drop mentioned by the authors.

In 2020, \gls{godin} \citep{Hsu2020GeneralizedOD} cited and was heavily inspired by Scaled Cosine. GODIN kept the \emph{extra learning block} to learn the scale factor and also avoided applying weight decay to it. In addition to the usual affine transformation and cosine distance from Scaled Cosine, it presents a variant that uses a Euclidean distance-based last layer, similar to IsoMax. The major contribution of GODIN was to allow using the input preprocessing introduced in ODIN without the need for out-of-distribution data. However, input preprocessing increases the inference latency (i.e., reduces the inference efficiency) approximately four times \citep{DBLP:journals/corr/abs-2006.04005}.

\looseness=-1
We emphasize that both \cite{techapanurak2019hyperparameterfree, Hsu2020GeneralizedOD} presents \emph{classification accuracy drop} \hln{of significantly more than one percent} in some situations, which is a harmful side effect, as classification is commonly the primary aim of the system \citep{carlini2019evaluating}.

\looseness=-1
Moreover, \gls{sngp} \citep{DBLP:conf/nips/LiuLPTBL20} cited, followed, and improved the idea introduced by IsoMax in 2019: A \emph{distance-based output layer} for OOD detection. In a similar direction, \gls{duq} \citep{Amersfoort2020SimpleAS} also proposed a modified \emph{distance-based loss to address OOD detection}. However, unlike IsoMax variants (e.g., IsoMax, IsoMax+, and DisMax), SNGP and DUQ produce inferences not as efficient as those produced by a deterministic neural network \citep{DBLP:conf/nips/LiuLPTBL20}. In 2021, the IsoMax+ loss \citep{macedo2021enhanced} introduced the \emph{isometric} distance. Considering that OOD and uncertainty estimation are related and relevant auxiliary tasks, modern approaches such as SNGP simultaneously tackle both.

\newpage\section{Distance-based Losses}

Recently, neural network distance-based losses have been proposed in the context of face recognition. For example, the contrastive \citep{DBLP:conf/nips/SunCWT14} and triplet \citep{DBLP:conf/cvpr/SchroffKP15} losses use the high-level feature (embeddings) \emph{pairwise} distances. In both cases, the SoftMax \emph{function}\footnote{We follow the ``SoftMax \emph{function}'' expression as defined in \cite{liu2016large}.} is not present, and the \emph{squared} Euclidean distance is used. One of the main drawbacks is the need for using Siamese neural networks, which adds complexity to the solution and expands memory requirements during training \citep{DBLP:conf/eccv/WenZL016}. 

\looseness=-1
Additionally, the \emph{triplet sampling} and \emph{pairwise training}, which implicate the recombination of the training samples with dramatic data expansion, slow convergence and produce instability \citep{DBLP:conf/eccv/WenZL016}. Finally, no prototypes are learned during training. The challenge of training networks using \emph{purely} distance-based losses while avoiding \emph{triplet sampling} and \emph{pairwise training} was discussed in \cite{DBLP:conf/eccv/WenZL016}, which proposed a \emph{squared} Euclidean distance-based \emph{regularization} procedure.

\looseness=-1
The center loss \citep{DBLP:conf/eccv/WenZL016} has two parameters $\alpha$ and $\lambda$. We call a loss isotropic when its dependency on the high-level features (embeddings) is performed \emph{exclusively} through distances, which are usually calculated to the class prototypes. Any metric may be used for such purpose. Any \emph{valid}\footnote{\url{https://en.wikipedia.org/wiki/Metric_(mathematics)}} distance may be used to construct an isometric loss. In this sense, the center loss is \emph{not} isotropic, as it presents affine transformation in its SoftMax classification term. Thus, the center loss inherits the drawbacks of the SoftMax loss affine transformation (See Chapter 3).

In \cite{Snell2017PrototypicalNF}, the authors proposed a solution based on \emph{squared} Euclidean distance to address few-shot learning. However, this approach does not work as a SoftMax loss drop-in replacement, as it does not simultaneously learn high-level features (embeddings) \emph{and prototypes} using \emph{exclusively} stochastic gradient descent (SGD) and \emph{end-to-end} backpropagation. Indeed, despite learning embeddings using regular SGD and backpropagation, \emph{additional offline procedures are required to calculate the class prototypes}.

\newpage\section{Maximum Entropy Principle}

The principle of maximum entropy, formulated by E. T. Jaynes to unify the statistical mechanics and information theory entropy concepts \citep{PhysRev.106.620, PhysRev.108.171}, states that when estimating probability distributions, we should choose the one that produces the maximum entropy consistent with the given constraints \citep{10.5555/1146355}. Following this principle, we avoid introducing additional assumptions or bias\footnote{\url{https://mtlsites.mit.edu/Courses/6.050/2003/notes}} not presented in the data. 

From a set of trial probability distributions that satisfactorily describe the available prior knowledge, the distribution that presents the maximal information entropy, which is the least informative option, represents the best possible choice. In other words, we must produce posterior probability distributions as under-confident as possible as long as they match the correct predictions for accurate classification.

\begin{figure*}
\small
\centering
\caption[The Principle of Maximum Entropy]{The Principle of Maximum Entropy}
\includegraphics[width=\textwidth,trim={0 0 0 0},clip]{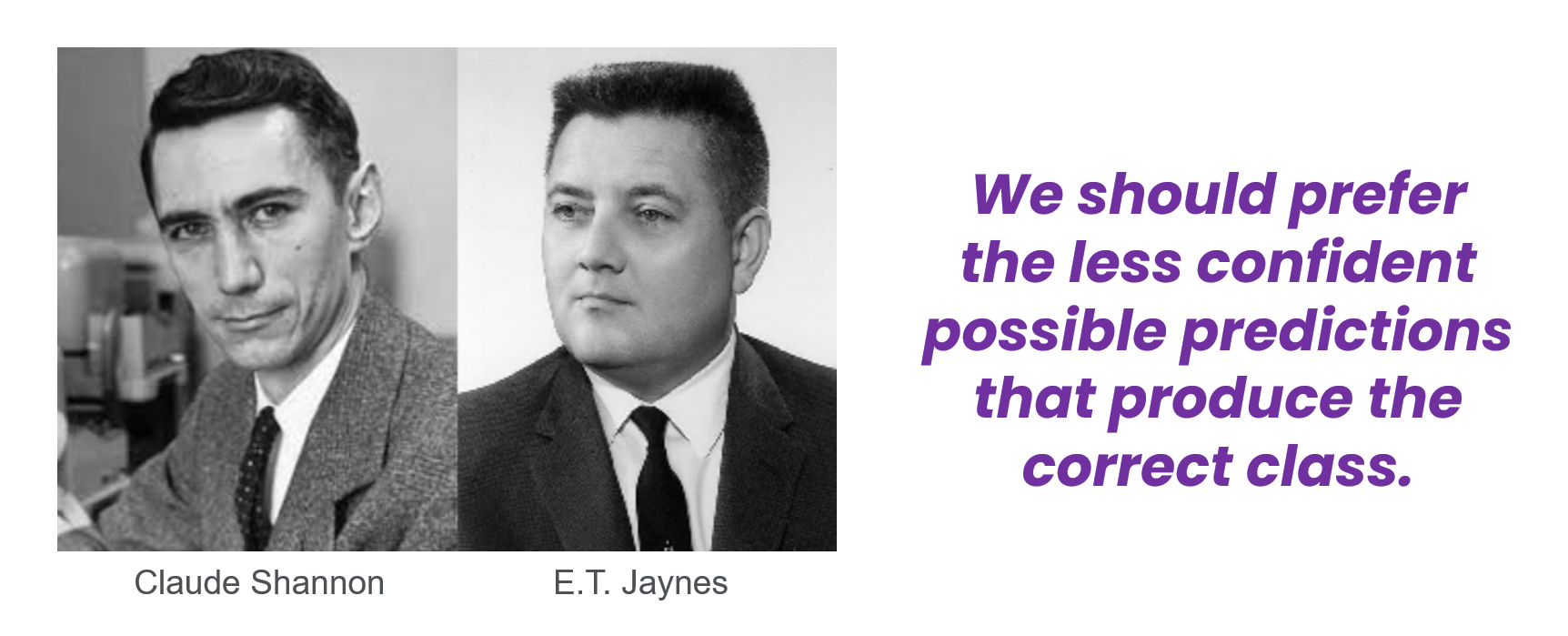}
\begin{justify}
{Source: The Author (2022). The Principle of Maximum Entropy: From a set of probability distributions that satisfactorily describe the prior knowledge available, the distribution that presents the maximal information entropy (i.e., the least informative option) represents the best possible choice. The maximum entropy principle produces the least biased possible probability distribution (no extra assumptions not presented in data).}
\end{justify}
\label{fig:maximum_entropy_principle}
\end{figure*}

\clearpage\section{\hl{Data Augmentation}}

\begin{figure*}
\small
\centering
\caption[\hl{Data Augmentation Approaches}]{\hl{Data Augmentation Approaches}}
\includegraphics[width=0.95\textwidth,trim={0 0 0 0},clip]{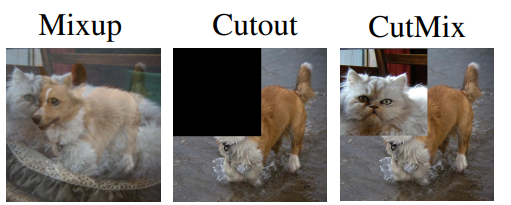} 
\begin{justify}
{Source:~\cite{DBLP:conf/iccv/YunHCOYC19}. \hl{Some recent data augmentation approaches are capable of increasing the OOD detection performance in some cases.}}
\end{justify}
\label{fig:data_aug}
\end{figure*}

\hl{Some recent studies have shown that data augmentation may enhance OOD detection. For example,} \cite{DBLP:conf/iccv/YunHCOYC19} \hl{showed that composing two images improves out-of-distribution detection performance} (Fig.~\ref{fig:data_aug}). \hl{In Mixup} \citep{zhang2018mixup}\hl{, two images are added pixel by pixel using some weighted proportion. Cutout} \citep{DBLP:journals/corr/abs-1708-04552} \hl{removed some portion of an image. CutMix replaces a piece of an image with a frame taken from another.}

\hl{Building novel data augmentation methods to improve robustness is an interesting proposal because they do not impact the inference delay. However, we need to be careful to avoid increasing the training time. It is also important to avoid adding too many hyperparameters to the solution when designing novel data augmentation strategies.}

%% file: chapters/3.proposal.tex
\chapter{Entropic Losses}\label{chap:seamless_approach}

\begin{quotation}[]{William of Ockham}
``More things should not be used than are necessary.''
\end{quotation}

\begin{quotation}[]{Albert Einstein}
``Things should be made as simple as possible, but not simpler.''
\end{quotation}


\begin{quotation}[]{Richard Feynman}
``The prize is in the pleasure of finding the thing out, the kick in the discovery, the observation that other people use it [my work] -- those are the real things, the honors are unreal to me.''
\end{quotation}


\looseness=-1
In this chapter, we present our solutions to the out-of-distribution and uncertainty estimation problems. We initially propose to use our isotropic distance-based loss IsoMax combined with the entropic score. Then, we evolve IsoMax to IsoMax+ by changing the initialization and performing what we call the isometrization of the distances used in IsoMax. We also propose the minimum distance score for out-of-distribution detection. Moreover, starting from IsoMax+, we add some modifications to present the state-of-the-art DisMax loss. Even more, we propose a novel score called the max-mean logit entropy score. Finally, we also built a fast way to achieve state-of-the-art uncertainty estimation by calibrating the temperature of DisMax trained models. We collectively call all proposed losses \emph{entropic losses} because all of them are based on the principle of maximum entropy.\\

\newpage\section{Isotropy Maximization Loss}\label{sec:isotropy_maximization_loss}

\begin{figure*}
\small
\centering
\caption[Seamless Approach: Loss and Score]{Seamless Approach: Loss and Score}
\includegraphics[width=0.8\textwidth,trim={0 0 0 0},clip]{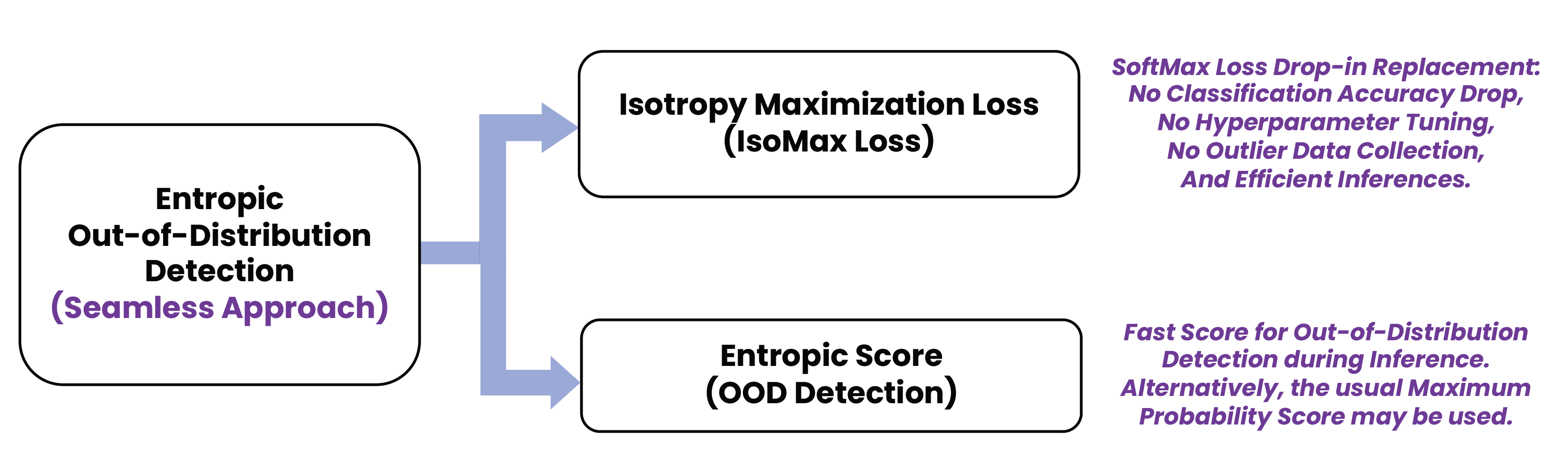}
\begin{justify}
{Source:~The Author (2022). The OOD detection solution is composed of two parts: the IsoMax loss and the Entropic Score.}
\end{justify}
\label{fig:loss_score}
\end{figure*}

The first component of our seamless and principled approach is the \gls{isomax} loss. The second is the rapid entropy score used for \gls{ood} detection. The entropic score is defined as the negative entropy of the neural network output probabilities. We chose the entropic score also based on the maximum entropy principle (\figref{fig:loss_score}). 

\hln{As will be explained in detail in the next section,} the IsoMax loss uses \emph{exclusively distance-based logits to fix the SoftMax loss anisotropy caused by its affine transformation}. Moreover, we introduce the entropic scale, a \emph{constant scalar multiplicative factor} applied to the logits \emph{throughout training} that is, nevertheless, \emph{removed before inference} to achieve high entropy posterior probability distributions in agreement with the maximum entropy principle.

The entropic scale is equivalent to the $\beta$ of the SoftMax \emph{function}. However, training with a \emph{predefined constant} entropic scale and then \emph{removing it before inference} is completely different from \emph{temperature calibration}. On the one hand, in ODIN and similar methods based on \emph{temperature calibration}, the temperature of a \emph{pretrained} model is validated \emph{after training}, which nevertheless was performed with a temperature equal to one. This validation usually requires unrealistic access to \gls{ood} or adversarial examples. Furthermore, over-optimistic performance estimation is commonly produced \citep{shafaei2018biased}. On the other hand, our approach requires neither hyperparameter validation nor access to \gls{ood} or adversarial data.

The intuitions that associate the unsatisfactory OOD detection performance of current neural networks with the SoftMax loss anisotropy and disagreement with the maximum entropy principle. The IsoMax loss trained models present accurate predictions and fast inferences that are energy- and computation-efficient. It does not require additional data. When using the entropy, rather than just one output, all network outputs are considered. In applications where the requirements and side effects of current OOD techniques are not a concern, future work may combine \hl{current approaches} with our loss to achieve even higher OOD detection performance.

\newpage

\begin{figure*}
\small
\centering
\caption[Structural Blocks: SoftMax Loss vs. IsoMax Loss]{Structural Blocks: SoftMax Loss vs. IsoMax Loss}
\subfloat[]{\includegraphics[width=0.95\textwidth,trim={0 0 0 0},clip]{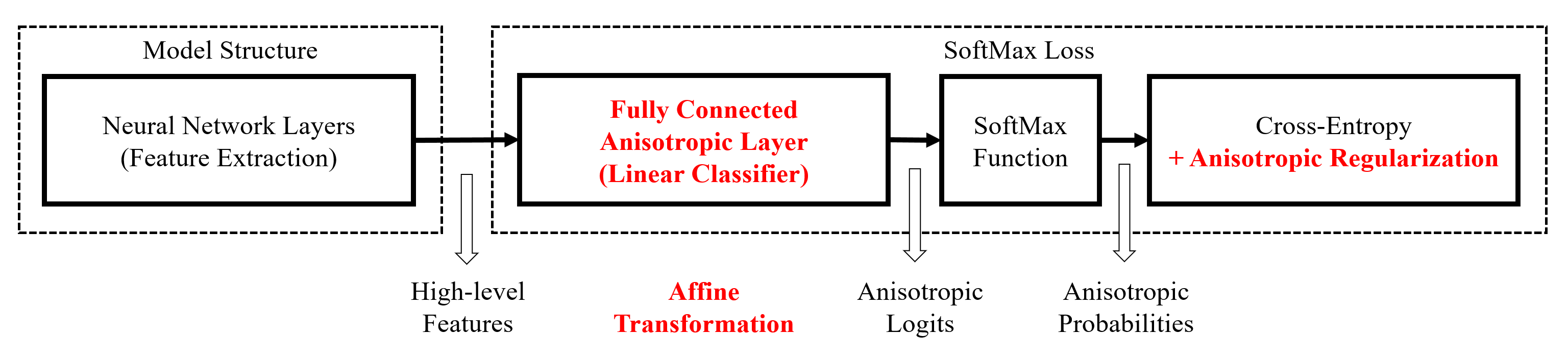}\label{fig:softmax_loss}} 
\\
\vskip -0.05cm
\subfloat[]{\includegraphics[width=0.95\textwidth,trim={0 0 0 0},clip]{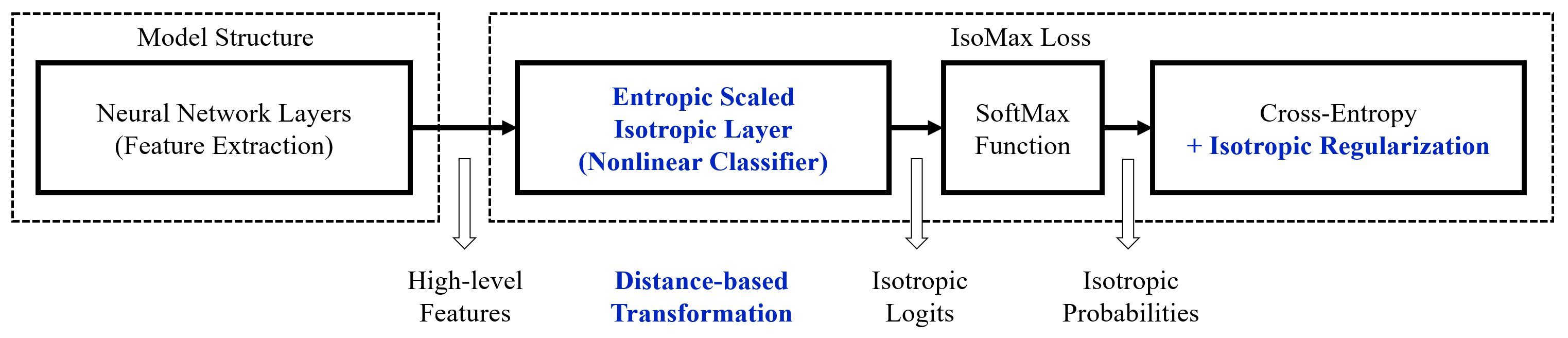}\label{fig:isomax_loss}}
\vskip -0.1cm
\begin{justify}
{Source: The Author (2022). Loss structural blocks: (a) SoftMax loss. Adapted from \cite{liu2016large}. (b) IsoMax loss (ours). In contrast to the SoftMax loss affine transformation, the IsoMax loss nonlinear isotropic layer incentivizes the concentration of high-level representations around learnable prototypes, facilitating \gls{ood} detection while avoiding the need for feature extraction and metric learning post-processing phases. Moreover, the exclusively distance-based nonlinearity increases the neural network representation power. During training, the isotropic layer is multiplied by a constant value called the entropic scale. For inference, the entropic scale is removed to produce high entropy posterior probability distributions, as recommended by the maximum entropy principle. Finally, the \gls{ood} detection uses the entropic score, which is defined as the negative entropy of neural network output probabilities.}
\end{justify}
\label{fig:softmax_isomax}
\end{figure*}

The swap of the SoftMax loss with the IsoMax loss requires changes in neither the model's architecture\footnote{\hln{Following a modern encoder-decoder and self-supervision terminology, we do not consider the last layer part of the architecture.}} nor training procedures or parameters. \figref{fig:softmax_isomax} synthesizes the differences between the SoftMax loss and the IsoMax loss.

\subsection{Initial Considerations}

\begin{figure*}
\small
\centering
\caption[Separable Features vs. Discriminative Features]{Separable Features vs. Discriminative Features}
\includegraphics[width=0.9\textwidth,trim={0 0 0 0},clip]{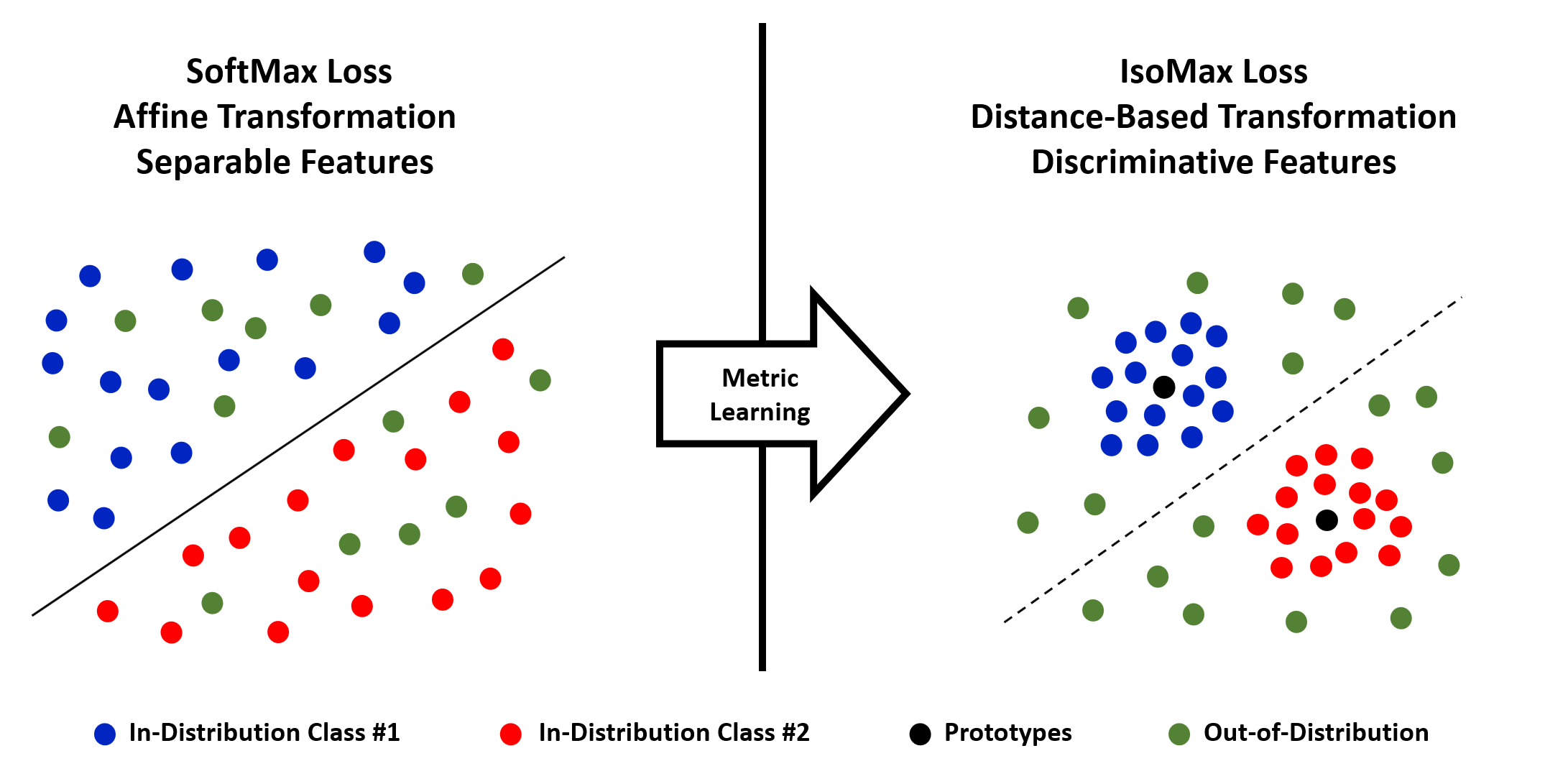}\label{fig:separable-discriminative}
\begin{justify}
{Source: Adapted from \cite{DBLP:conf/eccv/WenZL016}. SoftMax loss produces separable features \citep{DBLP:conf/eccv/WenZL016}. Post-processing metric learning on features extracted from SoftMax loss trained networks may convert from the situation on the left to the situation on the right \citep{lee2018simple, Mensink2013DistanceBasedIC, Scheirer_2013_TPAMI, Scheirer_2014_TPAMIb, 7298799,Rudd_2018_TPAMI}. The IsoMax loss, which is an exclusively distance-based (isotropic) loss, tends to generate more discriminative features \citep{DBLP:conf/eccv/WenZL016}. No feature extraction and subsequent metric learning are required when the IsoMax loss is used for training.}
\end{justify}
\label{fig:features}
\end{figure*}

Let $\bm{x}$ represent the input applied to a neural network and $\bm{f}_{\bm{\theta}}(\bm{x})$ represent the high-level feature vector produced by it. For this work, the underlying structure of the network does not matter.
Considering $k$ to be the correct class for a particular training example $\bm{x}$,
we can write the SoftMax loss associated with this specific training sample as:

\begin{align}
\label{eq:loss_softmax}
\mathcal{L}_{S}(\hat{y}^{(k)}|\bm{x})&=-\log\left(\frac{\exp(\bm{w}_k^\top\bm{f}_{\bm{\theta}}(\bm{x})\!+\!b_k)}{\sum\limits_j\exp(\bm{w}_j^\top\bm{f}_{\bm{\theta}}(\bm{x})\!+\!b_j)}\right)
\end{align}

\newpage
In Equation~\eqref{eq:loss_softmax}, $\bm{w}_j$ and $b_j$ represent the weights and biases associated with class $j$, respectively. 
From a geometric perspective, the term $\bm{w}_j^\top\bm{f}_{\bm{\theta}}(\bm{x})\!+\!b_j$ represents a hyperplane in the high-level feature space. It divides the feature space into two subspaces called positive and negative subspaces. The deeper inside the positive subspace the feature vector $\bm{f}_{\bm{\theta}}(\bm{x})$ of an example is located, the more likely the example is believed to belong to the considered class.

Therefore, training neural networks using SoftMax loss does not lead to agglomeration of representations of examples associated with a particular class into a limited region of the hyperspace, as it produces \emph{separable features} rather than \emph{discriminative features} \citep{DBLP:conf/eccv/WenZL016} (\figref{fig:features}, left).

The immediate consequence is the propensity of neural networks trained with SoftMax loss to make high confidence predictions on examples that stay in regions far away from the training examples, which explains their unsatisfactory \gls{ood} detection~performance \citep{Hein2018WhyRN}. Indeed, the SoftMax loss is based on affine transformations, which are essentially internal products. Consequently, the last layer representations of such networks tend to align in the direction of the weight vector, producing locally preferential directions in space and subsequently~anisotropy.

The SoftMax loss anisotropy is usually corrected by using metric learning on neural network pretrained features \citep{lee2018simple, Mensink2013DistanceBasedIC, Scheirer_2013_TPAMI, Scheirer_2014_TPAMIb, 7298799,Rudd_2018_TPAMI}. For example, the high \gls{ood} detection performance of the Mahalanobis approach \citep{lee2018simple} indicates that deploying locally isotropic spaces around class prototypes improves the \gls{ood} detection performance. In such solutions, a mapping from the extracted features to a novel embedding space is constructed and class prototypes are produced. The distance may be predefined (e.g., Euclidean distance) or learned (e.g., Mahalanobis distance).


However, approaches based on feature extraction and metric learning present drawbacks \citep{musgrave2020metric}. First, ad hoc procedures are required after neural network training. Additionally, they usually present hyperparameters to tune, usually requiring unrealistic access to design-time \gls{ood} or adversarial samples.

\looseness=-1
Therefore, a possible option to build a seamless approach to the \gls{ood} detection is to design an isotropic (\emph{exclusively} distance-based) loss that works as a SoftMax loss drop-in replacement. Hence, we obtain prototypes based classification while avoiding metric learning post-processing (\figref{fig:features}, right).

\subsection{Seamless Isotropic Loss}\label{sec:seamless_isotropic_loss}

Let $\bm{f}_{\bm{\theta}}(\bm{x})$ represent the high-level feature (embedding) associated with $\bm{x}$, $\bm{p}_{\bm{\phi}}^j$ represent the prototype associated with the class $j$, and $d()$ represent a distance function. To construct an isotropic loss, we need to avoid \emph{direct} dependency on $\bm{f}_{\bm{\theta}}(\bm{x})$ or $\bm{p}_{\bm{\phi}}^j$. Therefore, the loss has to be a function that \emph{exclusively} depends on the embedding-prototype distances given by $d({\bm{f}_{\bm{\theta}}(\bm{x}),\bm{p}_{\bm{\phi}}^j})$. Therefore, we can write:

\begin{align}\label{eq:isotropic_loss1}
\begin{split}
\mathcal{L}_{I}\!=\!g(d({\bm{f}_{\bm{\theta}}(\bm{x}),\bm{p}_{\bm{\phi}}^j}))
\end{split}
\end{align}

In the previous equation, $g()$ represents a scalar function. The expression $d({\bm{f}_{\bm{\theta}}(\bm{x}),\bm{p}_{\bm{\phi}}^j})$ represents the isotropic layer, where its weights are given by the learnable prototypes $\bm{p}_{\bm{\phi}}^j$. We decided to normalize the embedding-prototype distances using the SoftMax \emph{function} to allow interpretation in terms of probabilities. Therefore, the embedding-prototype distances represent the logits of the SoftMax function and correspond to the output of the isotropic layer. We also decided to use the cross-entropy for efficient optimization. Therefore, we can write the following:

\begin{align}\label{eq:isotropic_loss2}
\begin{split}
\mathcal{L}_{I}(\hat{y}^{(k)}|\bm{x})
=-\log\left(\frac{\exp(-d(\bm{f}_{\bm{\theta}}(\bm{x}),\bm{p}_{\bm{\phi}}^k))}{\sum\limits_j\exp(-d(\bm{f}_{\bm{\theta}}(\bm{x}),\bm{p}_{\bm{\phi}}^j))}\right)
\end{split}
\end{align}

In the above equation, $k$ represents the correct class, while the negative logarithm represents the cross-entropy. The negative terms before the distances are necessary to indicate the negative correlation between distances and probabilities. The expression between the outermost parentheses applied to the term $-d({\bm{f}_{\bm{\theta}}(\bm{x}),\bm{p}_{\bm{\phi}}^j})$ represents the SoftMax function.

\begin{figure*}
\small
\centering
\caption[SoftMax Loss Drawbacks and IsoMax Loss Advantages]{SoftMax Loss Drawbacks and IsoMax Loss Advantages}
\subfloat[]{\includegraphics[width=0.7\textwidth,trim={0 0 0 0},clip]{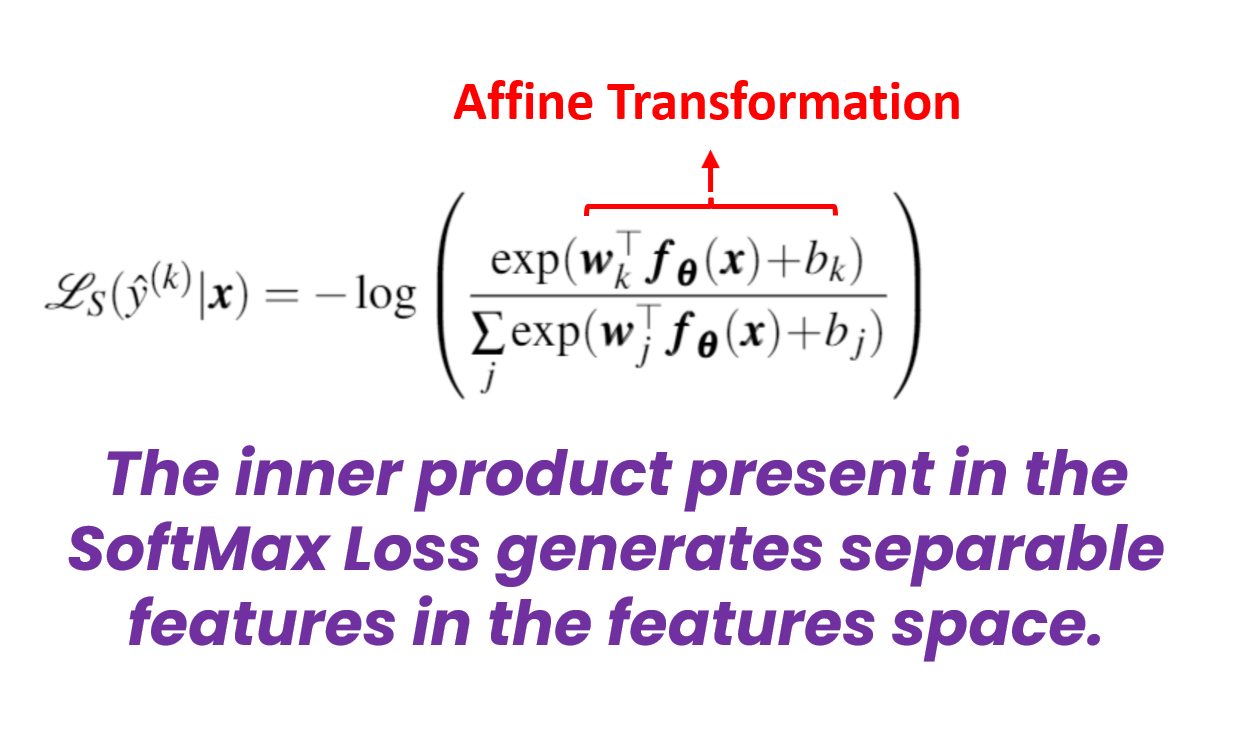}}
\\
\subfloat[]{\includegraphics[width=0.7\textwidth,trim={0 0 0 0},clip]{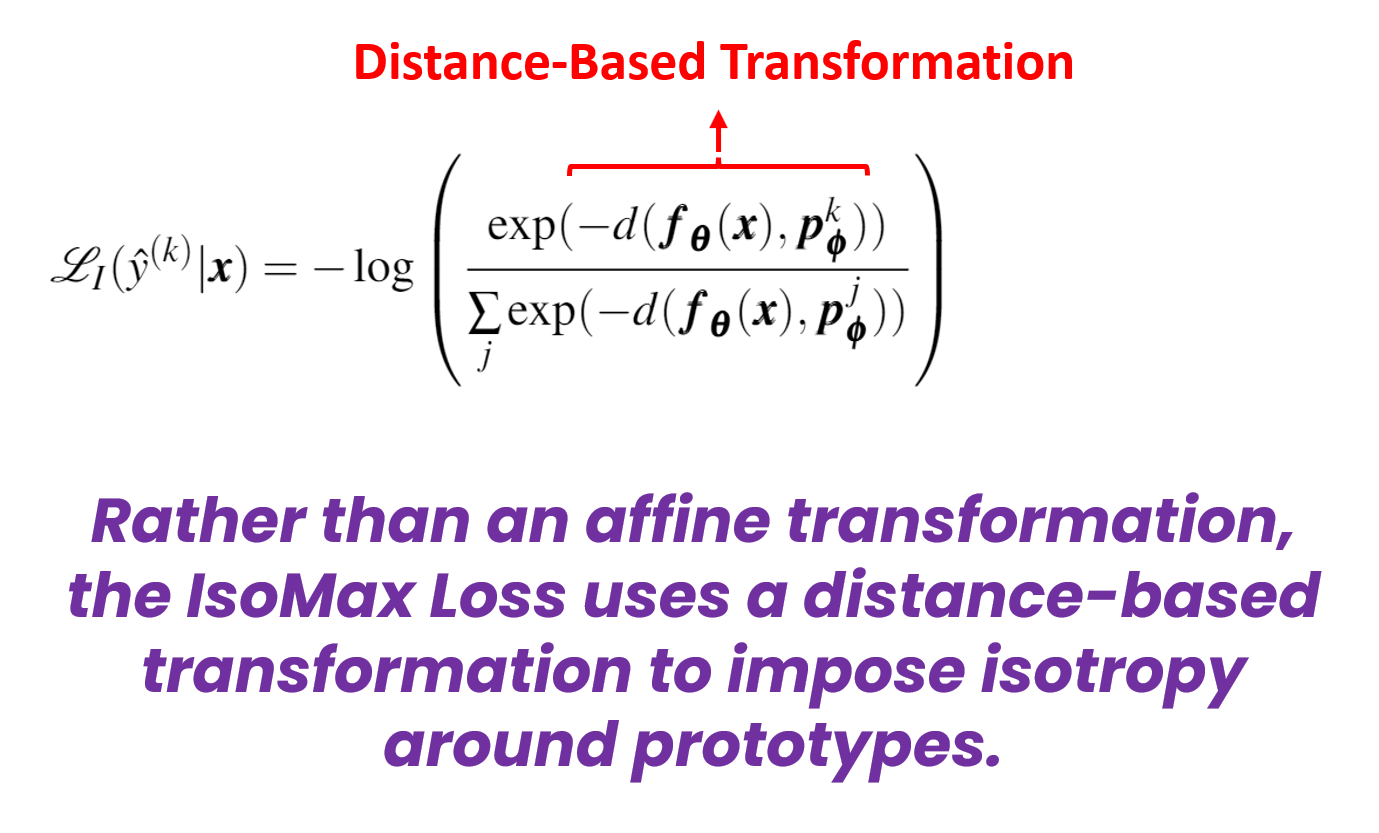}}
\begin{justify}
{Source: The Author (2022). SoftMax loss produces separable features \hln{(above)}, while distance-based losses tend to generate discriminative features \hln{(below)}. Out-of-distribution examples are more discernible when in-distribution examples are concentrated around prototypes.}
\end{justify}
\label{fig:softmax_vs_isomax}
\end{figure*}

We need to choose a distance that allows IsoMax to work as a SoftMax drop-in replacement. Hence, the loss needs to learn \emph{both} high-level features \emph{and prototypes} using \emph{exclusively} SGD and \emph{end-to-end} backpropagation, as, based on the goal we defined for this work, \emph{no additional offline procedures are allowed}. \hl{We emphasize that the prototypes are learned using backpropagation just like other weights.} We also require the training using IsoMax loss to be as consistent and stable as the typical SoftMax loss neural network~training.

The covariance matrix makes it hard to use the Mahalanobis distance to train a neural network directly, as it contains components that are not differentiable. Therefore, we decide to use \emph{Euclidean} distance. We have reasons to prefer the \emph{nonsquared} Euclidean distance rather than the \emph{squared} Euclidean distance. First, the nonsquared Euclidean distance is a real metric that obeys the Cauchy–Schwarz inequality while the squared Euclidean distance is not. Using a metric that follows the Cauchy–Schwarz inequality is essential because of our previous geometric considerations such as features separability and isotropy. Additionally, using the squared Euclidean distance is actually equivalent to using a linear model with a particular parameterization \citep{Snell2017PrototypicalNF, Mensink2013DistanceBasedIC}, which we prefer to avoid increasing the representative power of the proposed loss.

Moreover, we experimentally observed that training neural networks using exclusively SGD and end-to-end backpropagation is more stable and consistent when using nonsquared Euclidean distance rather than squared Euclidean distance. Indeed, squared Euclidean distance-based logits are harder to \emph{seamlessly} train than nonsquared Euclidean distance-based logits because numeric calculus instabilities are much more likely to occur when performing calculations and derivations with values of the order of $\mathcal{O}(e^{-d^{2}})$ than $\mathcal{O}(e^{-d})$.

Finally, even in the cases in which we were eventually able to seamlessly train neural networks using the squared Euclidean distance, we observed lower \gls{ood} detection performance than using the nonsquared Euclidean distance-based alternative. Therefore, we choose the nonsquared Euclidean distance.

\looseness=-1
Unlike metric learning-based \gls{ood} detection approaches, rather than learning a metric from a preexisting feature space (metric learning on features extracted from a pretrained model), when using the IsoMax loss, we learn a feature space that is, from the start, consistent with the chosen metric. Indeed, the minimization of Equation~\eqref{eq:isotropic_loss2} is achieved by making the expression inside the outer parentheses go to one. It is only possible by reducing the distances between the high-level features (embeddings) and the associated class prototypes, while simultaneously keeping high distances among class prototypes. Hence, the main aim of metric learning, which is to reduce intraclass distances while increasing interclass distances, is performed naturally during the neural network training, avoiding the need for feature extraction and metric learning post-processing phases (\figref{fig:softmax_vs_isomax}).

\newpage
\subsection{Principle of Maximum Entropy and Entropic Scale}\label{sec:principle_of_maximum_entropy}

\begin{figure*}
\small
\centering
\caption[The Entropic Maximization Trick]{The Entropic Maximization Trick}
\subfloat[]{\includegraphics[width=\textwidth,trim={0 0 0 0},clip]{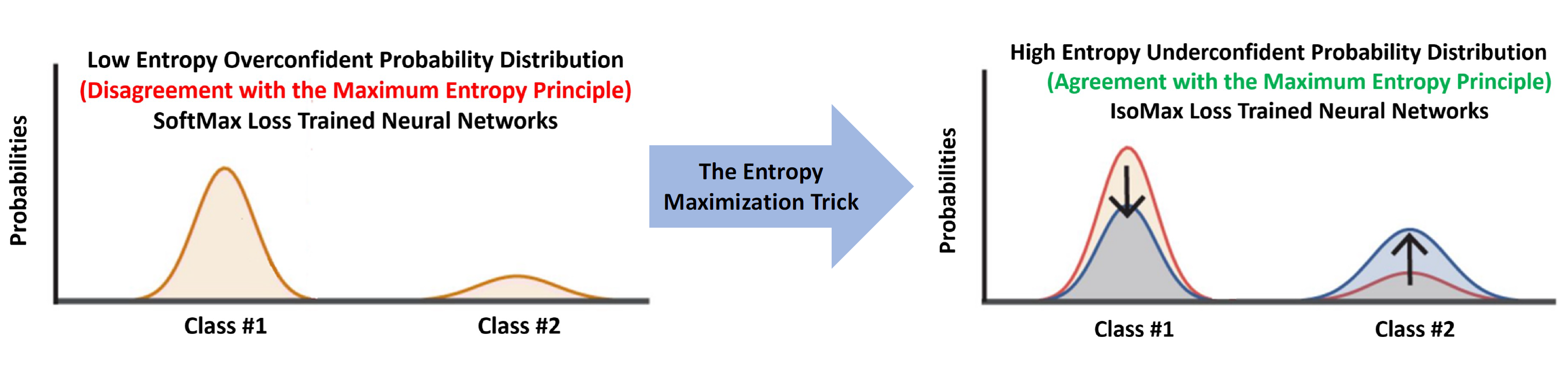}}
\\
\subfloat[]{\includegraphics[width=\textwidth,trim={0 0 0 0},clip]{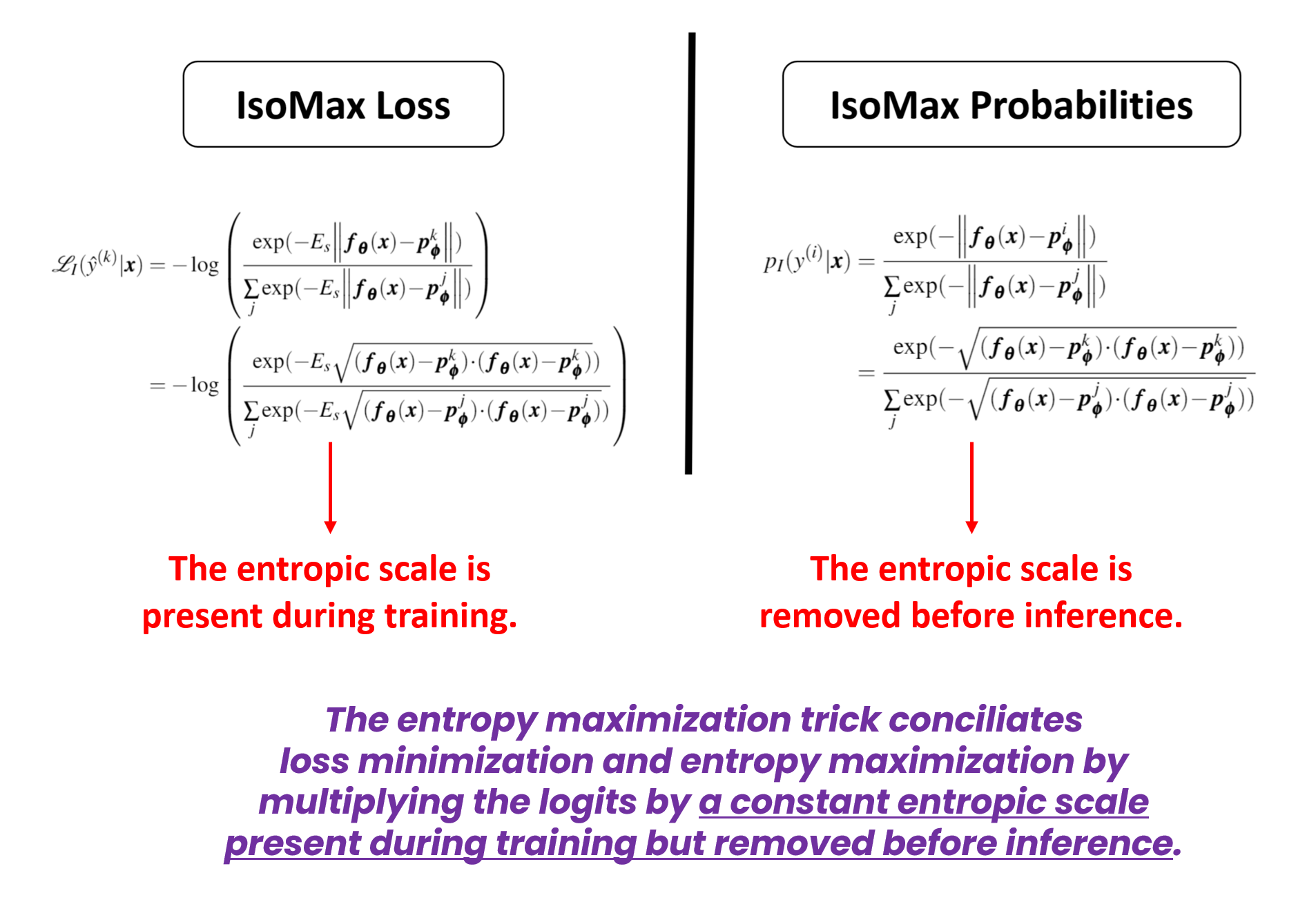}}
\begin{justify}
{Source: The Author (2022). High entropy distributions (IsoMax Loss) may produce the same predictions and, consequently, classification accuracies of low entropy distributions (SoftMax Loss). However, the former provide a much higher OOD detection performance than the latter, regardless of using maximum probability or entropy as the score. SoftMax loss trained neural networks produce overconfident low-entropy probability distributions in disagreement with the maximum entropy principle. Our \emph{entropy maximization trick}, which consists in training using logits multiplied by a constant factor called the entropic scale \emph{that is nevertheless removed before inference}, enables IsoMax to generate underconfident high-entropy (almost maximum entropy) probability distributions in agreement with the principle of maximum entropy.}
\end{justify}
\label{fig:entropy_maximization_trick}
\end{figure*}

Isotropy increases \gls{ood} detection performance \hln{(See Chapter} \ref{chap:experiments}). However, for further gains, we need to circumvent the cross-entropy extreme propensity to produce significantly low entropy posterior probability distributions. Additionally, as established in our goal, we need to achieve this without losing IsoMax seamlessness.


The principle of maximum entropy has been studied as a \emph{regularization factor} \citep{DBLP:conf/nips/DubeyGRN18, pereyra2017regularizing}. In some cases, it has also been used as a direct optimization procedure without connection to cross-entropy minimization or backpropagation. For example, in \cite{514879,berger-etal-1996-maximum,pmlr-v5-shawe-taylor09a}, the maximization of the entropy subject to a constraint on the expected classification error was shown to be equivalent to solving an unconstrained Lagrangian. Although theoretically well grounded \citep{10.5555/534975, Williamson2005ObjectiveBN, Pages493-533, DBLP:journals/mima/Hosni13}, \emph{direct} entropy maximization presents high computational complexity \citep{10.5555/534975,Pages493-533}.

Alternatively, modern neural networks are trained using computationally efficient cross-entropy. However, this procedure does not prioritize high entropy (low confident) posterior probability distributions. \emph{Actually, the opposite is true}. Indeed, the minimization of cross-entropy has the undesired side effect of producing low entropy (overconfident) posterior probability distributions \citep{Guo2017OnCO}. Unlike the previously mentioned works, we use the principle of maximum entropy neither to motivate the construction of regularization mechanisms, such as label smoothing or the confidence penalty \citep{DBLP:conf/nips/DubeyGRN18,pereyra2017regularizing}, nor to perform \emph{direct} maximum entropy optimization \citep{514879,berger-etal-1996-maximum,pmlr-v5-shawe-taylor09a}. Indeed, the entropy is not even calculated during training.


In the opposite direction, we use the principle of maximum entropy as \emph{motivation} to construct high-entropy (low-confident) posterior probabilities, still relying on computationally efficient cross-entropy minimization. 
Since our approach does \emph{not directly} maximize the entropy, we cannot state that IsoMax produces the maximum entropy posterior probability distribution. However, the entropies of the probability distributions produced by IsoMax are high enough to improve the \gls{ood} detection performance significantly.

\begin{align}
\mathcal{L}_{\textsf{SoftMax}}\!=\!-\log\left(\frac{\exp(L_k)}{\sum\limits_j\exp(L_j)}\right)\to 0\label{eq:cross_entropy_softmax_minimization1}\\
\implies
\mathcal{P}(y|\bm{x})\to 1\label{eq:cross_entropy_softmax_minimization2}\\
\implies \mathcal{H}_{\textsf{SoftMax}}\to 0\label{eq:cross_entropy_softmax_minimization3}
\end{align}

\newpage
Equation \eqref{eq:cross_entropy_softmax_minimization1} describes the behavior of cross-entropy and entropy for the SoftMax loss. $L_j$ represents the logits associated with class $j$, and $L_k$ represents the logits associated with the correct class $k$. When minimizing the loss (Equation \eqref{eq:cross_entropy_softmax_minimization1}), high probabilities are generated (Equation \eqref{eq:cross_entropy_softmax_minimization2}). Consequently, significantly low entropy posterior probability distributions are produced (Equation \eqref{eq:cross_entropy_softmax_minimization3}). Hence, the usual cross-entropy minimization tends to generate unrealistic overconfident (low entropy) probability distributions. Therefore, we have an opposition between cross-entropy minimization and the principle of maximum~entropy.

\begin{align}
\mathcal{L}_{\textsf{IsoMax}}\!=\!-\log\left(\frac{\exp(-E_s\!\times\!D_k)}{\sum\limits_j\exp(-E_s\!\times\!D_j)}\right)\to 0\label{eq:cross_entropy_isomax_minimization1}\\
\centernot\implies \mathcal{P}(y|\bm{x})\to 1\label{eq:cross_entropy_isomax_minimization2}\\
\centernot\implies \mathcal{H}_{\textsf{IsoMax}} \to 0\label{eq:cross_entropy_isomax_minimization3}
\end{align}

The IsoMax loss conciliates these contradictory objectives (loss minimization and entropy maximization) by multiplying the logits by a constant positive scalar $E_s$, which is presented during training but removed before inference. Equation \eqref{eq:cross_entropy_isomax_minimization1} demonstrates how the entropic scale (presented at training time but removed at inference time) allows the production of high entropy posterior distributions despite using cross-entropy minimization.  $D_j$ represents the distances associated with class $j$, and $D_k$ represents the distances associated with the correct class $k$. The $E_s$ present during training allows the term $-E_s\!\times\!D_k$ to become high enough (less negative compared to $-E_s\!\times\!D_j$) to produce a low loss (Equation \eqref{eq:cross_entropy_isomax_minimization1}) \emph{without} producing high probabilities for the correct classes, as they are calculated with the $E_s$ removed (Equation \eqref{eq:cross_entropy_isomax_minimization2}). Thus, it is possible to build posterior probability distributions with high entropy (Equation \eqref{eq:cross_entropy_isomax_minimization3}) in agreement with the fundamental principle of maximum entropy despite using cross-entropy to minimize the loss (\figref{fig:entropy_maximization_trick}).


Therefore, returning to Equation \eqref{eq:isotropic_loss2}, multiplying the embedding-prototype distances by $E_s$, and making $d()$ equal to the nonsquared Euclidean distance, we write the IsoMax loss as:

\begin{align}\label{eq:loss_isomax}
\begin{split}
\mathcal{L}_{I}(\hat{y}^{(k)}|\bm{x})
&=-\log^\dagger\left(\frac{\exp(-E_s\norm{\bm{f}_{\bm{\theta}}(\bm{x})\!-\!\bm{p}_{\bm{\phi}}^k})}{\sum\limits_j\exp(-E_s\norm{\bm{f}_{\bm{\theta}}(\bm{x})\!-\!\bm{p}_{\bm{\phi}}^j})}\right)
\\
&=-\log^\dagger\left(\frac{\exp(-E_s\sqrt{(\bm{f}_{\bm{\theta}}(\bm{x})\!-\!\bm{p}_{\bm{\phi}}^k)\!\cdot\!(\bm{f}_{\bm{\theta}}(\bm{x})\!-\!\bm{p}_{\bm{\phi}}^k)})}{\sum\limits_j\exp(-E_s\sqrt{(\bm{f}_{\bm{\theta}}(\bm{x})\!-\!\bm{p}_{\bm{\phi}}^j)\!\cdot\!(\bm{f}_{\bm{\theta}}(\bm{x})\!-\!\bm{p}_{\bm{\phi}}^j)})}\right)
\end{split}
\end{align}
\blfootnote{\textsuperscript{$\dagger$}The probability (i.e., the expression between the outermost parentheses) and logarithm operations are computed sequentially and separately for higher OOD detection performance (see the source code).}

\newpage

We emphasize that removing the entropic scale after the training does not affect the ability of the solution to represent the prior knowledge available, as it does not change the predictions. Therefore, the expression for the probabilities with the entropic scale removed is preferable, as it remarkably increases the entropy of the posterior distributions in agreement with the principle of maximum entropy. 
Hence, inference probabilities for the IsoMax loss are defined as follows:

\begin{align}\label{eq:probability_isomax}
\begin{split}
p_{I}(y^{(i)}|\bm{x})
&=\frac{\exp(-\norm{\bm{f}_{\bm{\theta}}(\bm{x})\!-\!\bm{p}_{\bm{\phi}}^i})}{\sum\limits_j\exp(-\norm{\bm{f}_{\bm{\theta}}(\bm{x})\!-\!\bm{p}_{\bm{\phi}}^j})}\\
&=\frac{\exp(-\sqrt{(\bm{f}_{\bm{\theta}}(\bm{x})\!-\!\bm{p}_{\bm{\phi}}^k)\!\cdot\!(\bm{f}_{\bm{\theta}}(\bm{x})\!-\!\bm{p}_{\bm{\phi}}^k)})}{\sum\limits_j\exp(-\sqrt{(\bm{f}_{\bm{\theta}}(\bm{x})\!-\!\bm{p}_{\bm{\phi}}^j)\!\cdot\!(\bm{f}_{\bm{\theta}}(\bm{x})\!-\!\bm{p}_{\bm{\phi}}^j)})}
\end{split}
\end{align}

\subsection{Initialization and Implementation Details}

We experimentally observed that using the common Xavier \citep{glorot2010understanding} and Kaiming \citep{He2016DelvingClassification} initializations for prototypes makes the \gls{ood} detection performance oscillate. Sometimes it improves, sometimes it decreases. Hence, we decided to always initialize all prototypes to zero vector. It indeed makes sense, as this is the most natural value for untrained embeddings. Weight decay is applied to prototypes, as they are regular trainable parameters in~our~solution.

To calculate losses based on cross-entropy, deep learning libraries usually combine the logarithm and probability calculations into a single computation. However, we experimentally observed that sequentially computing these calculations as stand-alone operations improves IsoMax performance \hln{because it is more effective in keeping the initial output entropy high.}

The class prototypes have the same dimension as the neural network last layer representations. Naturally, the number of prototypes is equal to the number of classes. Therefore, the IsoMax loss has fewer parameters than the SoftMax loss, as it has no bias to be learned. \hl{We remember that the prototypes are updated during the regular backpropagation procedure, just like any other parameters.}

Finally, we verified classification accuracy drop and low or oscillating \gls{ood} detection performance when trying to integrate $E_s$ with cosine similarity \citep{liu2016large, DBLP:journals/spl/WangCLL18, DBLP:conf/cvpr/DengGXZ19} or the affine transformations used in SoftMax loss. \hl{The above trick (i.e., to initialize prototypes to zero vector) cannot be performed in cosine similarity and SoftMax loss cases.} Therefore, our only option is to confirm that we need to use distance. Moreover, the \emph{nonsquared} Euclidean distance as the best option to build the IsoMax~loss. 

\newpage\subsection{Entropic Score}

Out-of-distribution detection approaches typically define a score to be used after inference to evaluate whether an example should be considered \gls{ood}. In a seminal work, \cite{Shannon1948ACommunication,DBLP:journals/bstj/Shannon48a} demonstrated that entropy represents the optimum measure of the randomness of a source of symbols. More broadly, we currently understand entropy as a measure of the uncertainty. Therefore, considering that the uncertainty in classifying a specific sample should be an optimum metric to evaluate whether a particular example is \gls{ood}, we define our score to perform \gls{ood} detection, called the entropic score, as the \emph{negative entropy} of the output probabilities:

\begin{align}\label{eq:entropic_score}
\mathcal{ES}\!=\!-\sum_{i=1}^{N}{p(y^{(i)}|\bm{x})}\log p(y^{(i)}|\bm{x})
\end{align}


Using the negative entropy as a score to evaluate whether a particular sample is \gls{ood}, we consider the information provided by \emph{all} available network outputs rather than just one. For instance, ODIN and ACET only use the maximum probability, while the Mahalanobis method only uses the distance to the nearest prototype. During IsoMax training, the embedding-prototype distances are affected. On the one hand, the distances from embeddings to the correct class prototype are reduced to increase classification accuracy. On the other hand, the distances from embeddings to the wrong class prototypes are increased. Consequently, based on the Equation~\eqref{eq:probability_isomax}, the probabilities of in-distribution examples increase. Therefore, it is reasonable to expect that samples with lower entropy more likely belong to the in-distribution. Additionally, using this predefined mathematical expression as score avoids the need to train an ad hoc additional regression model to detect \gls{ood} samples that is otherwise required, for example, in the Mahalanobis approach.

Even more important, since no regression model needs to be trained, there is no need for unrealistic access to \gls{ood} or adversarial samples for hyperparameter validation\footnote{We call hyperparameter validation (or tuning) the procedure of selecting the model for test evaluation (i.e., model section) based on the performance achieved on the validation partition of a set of models trained with different hyperparameters.}. Since the entropic score is a predefined mathematical expression rather than a trainable model, it is available as soon as the neural network training finishes, avoiding additional training of ad hoc models in a post-processing phase. \hl{In practice, we may set the threshold in two ways. First, we may collect random out-of-distribution samples and also use the train or validation set to define the threshold. This way, we may have an estimation of the performance we may expect in the field.} \hl{The second option is only using the train or validation set without access to a random out-of-distribution. In this case, we may set a threshold by making, for example, 95\% of your in-distribution data to be considered in-distribution using the chosen threshold. In this case, however, we will not have an estimation of the expected performance. In this work, we followed the literature and often used threshold independent detection metrics to ensure the robustness of the score used.} 

\newpage\section{Enhanced Isotropy Maximization Loss}\label{sec:enhanced_isotropy_maximization_loss}

In this section, we present the enhanced IsoMax for training and the minimum distance score for inference. By combining those methods, we develop a seamless, scalable, and high-performance out-of-distribution detection approach.

We started from the IsoMax loss. First, we normalize both the prototypes and the features. Second, we change the initialization of the prototypes. Third, we add the \emph{distance scale}, which is a \emph{learnable scalar parameter} that multiplies the \emph{feature-prototype distances}. We call the combination of these three modifications the \emph{isometrization} of the \emph{feature-prototype distances}. We call the proposed modified version of IsoMax the \gls{isomax+}. Finally, we use the \emph{minimum feature-prototype distance} as a score to perform OOD detection. Considering that the minimum feature-prototype distance is calculated to perform the classification, \emph{the OOD detection task presents essentially zero computational cost} because we simply reuse this value as a score to perform OOD detection.

IsoMax+ keeps the solution seamless (i.e., avoids the previously mentioned special requirements and side effects) while significantly increasing the OOD detection performance. Similar to IsoMax loss, IsoMax+ works as a \emph{SoftMax loss drop-in replacement} because no procedures other than regular neural network training are required. 


\subsection{Isometric Distance}

\begin{figure}
\centering
\small
\caption[\hl{Isometric Distances}]{\hl{Isometric Distances}}
\subfloat[]{
\begin{tikzpicture}[tdplot_main_coords, scale = 1.2]
\node[align=center] at (0,0,3.6) {\color{black} \textit{IsoMax n-dimensional Euclidean space}};
\coordinate (P1) at ({1.75/sqrt(2)},{-1/sqrt(2)},{1.5/sqrt(2)});
\coordinate (P2) at ({(-1/sqrt(2))/2},{(1/sqrt(2))/2},{2/sqrt(2)});
\coordinate (P3) at ({2.5/sqrt(2)},{1.75/sqrt(2)},{0});
\coordinate (F1) at ({1.3/sqrt(2)},{2.5/sqrt(2)},{2/sqrt(2)});
%
%
%
%
%
\draw[dotted, gray] (0,0,0) -- (-2,0,0);
\draw[dotted, gray] (0,0,0) -- (0,-2,0);
\draw[dotted, gray] (0,0,0) -- (0,0,-2);
\draw[-stealth] (0,0,0) -- (3.60,0,0) node[below left] {$x$};
\draw[-stealth] (0,0,0) -- (0,2.50,0) node[below right] {$y$};
\draw[-stealth] (0,0,0) -- (0,0,2.70) node[above] {$z$};
\draw[thick, -stealth] (0,0,0) -- (P1) node[above] {$P_1$};
\draw[thick, -stealth] (0,0,0) -- (P2) node[above] {$P_2$};
\draw[thick, -stealth] (0,0,0) -- (P3) node[below] {$P_3$};
\draw[thick, -stealth, blue] (0,0,0) -- (F1) node[above right] {$F$};
%
%
\end{tikzpicture}
}
\hskip 0.5 cm
\subfloat[]{
\begin{tikzpicture}[tdplot_main_coords, scale = 0.8]
\node[align=center] at (0,0,3.6) {\color{black} \textit{(n-1)-sphere of radius one in the}\\\textit{IsoMax+ n-dimensional Euclidean space}\\\\};
\coordinate (P1) at ({1.75/sqrt(2)},{-1/sqrt(2)},{1.5/sqrt(2)});
\coordinate (P2) at ({(-1/sqrt(2))/2},{(1/sqrt(2))/2},{2/sqrt(2)});
\coordinate (P3) at ({2.5/sqrt(2)},{1.75/sqrt(2)},{0});
\coordinate (F1) at ({1.3/sqrt(2)},{2.5/sqrt(2)},{2/sqrt(2)});
\shade[ball color = lightgray, opacity = 0.5] (0,0,0) circle (2.0cm);
\tdplotsetrotatedcoords{0}{0}{0};
\draw[dotted, tdplot_rotated_coords, gray] (0,0,0) circle (2);
\tdplotsetrotatedcoords{90}{90}{90};
\draw[dotted, tdplot_rotated_coords, gray] (2,0,0) arc (0:180:2);
\tdplotsetrotatedcoords{0}{90}{90};
\draw[dotted, tdplot_rotated_coords, gray] (2,0,0) arc (0:180:2);
\draw[dotted, gray] (0,0,0) -- (-2,0,0);
\draw[dotted, gray] (0,0,0) -- (0,-2,0);
\draw[dotted, gray] (0,0,0) -- (0,0,-2);
\draw[-stealth] (0,0,0) -- (3.60,0,0) node[below left] {$x$};
\draw[-stealth] (0,0,0) -- (0,2.50,0) node[below right] {$y$};
\draw[-stealth] (0,0,0) -- (0,0,2.70) node[above] {$z$};
\draw[thick, -stealth] (0,0,0) -- (P1) node[above] {$P_1$};
\draw[thick, -stealth] (0,0,0) -- (P2) node[above] {$P_2$};
\draw[thick, -stealth] (0,0,0) -- (P3) node[below] {$P_3$};
\draw[thick, -stealth, blue] (0,0,0) -- (F1) node[above right] {$F$};
%
%
\end{tikzpicture}
}
\begin{justify}
{Source: The Authors (2022). \hl{The illustration presents the advent of the isometric distances in IsoMax+. $P_1$, $P_2$, and $P_3$ represent prototypes of classes $1$, $2$, and $3$, respectively. $F$ denotes the feature associated with a given image. IsoMax does not restrict prototypes and features to the (n-1)-sphere. In contrast, IsoMax+ does precisely this by rescaling prototypes and features.}}
\end{justify}
\label{fig:isometric_distances}
\end{figure}
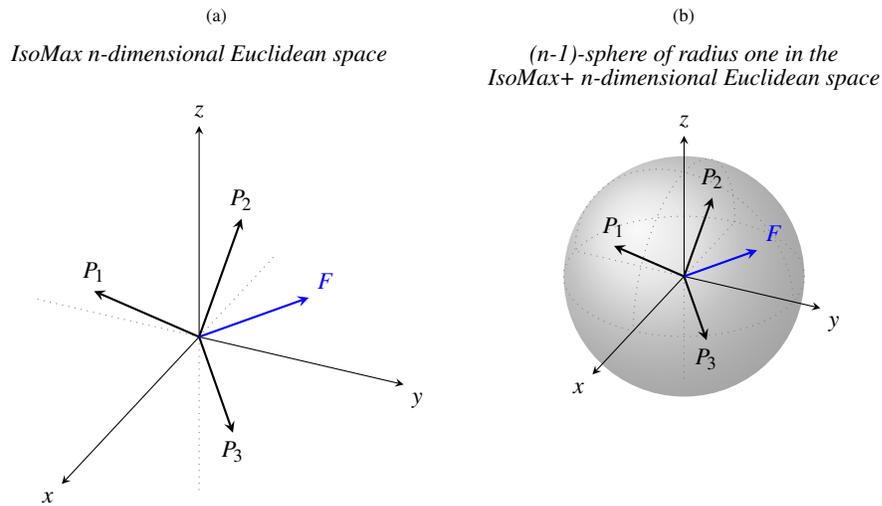

We consider an input $\bm{x}$ applied to a neural network that performs a parametrized transformation $\bm{f}_{\bm{\theta}}(\bm{x})$. We also consider $\bm{p}_{\bm{\phi}}^j$ to be the \emph{learnable prototype} associated with class $j$. Additionally, let the expression $\norm{\bm{f}_{\bm{\theta}}(\bm{x})\!-\!\bm{p}_{\bm{\phi}}^j}$ represent the \emph{nonsquared Euclidean distance} between $\bm{f}_{\bm{\theta}}(\bm{x})$ and $\bm{p}_{\bm{\phi}}^j$. Finally, we consider $\bm{p}_{\bm{\phi}}^k$ as the learnable prototype associated with the correct class for the input $\bm{x}$. \hl{Thus, we can rewrite for convenience the IsoMax loss} (equation \eqref{eq:loss_isomax}) using the equation below:

\begin{align}
\mathcal{L}_{\textsf{IsoMax}}&=-\log^\dagger\left(\frac{\exp(-E_s\norm{\bm{f}_{\bm{\theta}}(\bm{x})\!-\!\bm{p}_{\bm{\phi}}^k})}{\sum\limits_{j}\exp(-E_s\norm{\bm{f}_{\bm{\theta}}(\bm{x})\!-\!\bm{p}_{\bm{\phi}}^j})}\right)\label{eq:isomax_loss_full}
\end{align}
\blfootnote{\textsuperscript{$\dagger$}The probability (i.e., the expression between the outermost parentheses) and logarithm operations are computed sequentially and separately for higher OOD detection performance (see the source code).}

In the above equation, $E_s$ represents the entropic scale. From Equation~\eqref{eq:isomax_loss_full}, we observe that the distances from IsoMax loss are given by the expression $\mathcal{D}\!=\!\norm{\bm{f}_{\bm{\theta}}(\bm{x})\!-\!\bm{p}_{\bm{\phi}}^j}$. During inference, probabilities calculated based on these distances are used to produce the negative entropy, which serves as a score to perform OOD detection. However, because the features $\bm{f}_{\bm{\theta}}(\bm{x})$ are not normalized, examples with low norms are unjustifiably favored to be considered OOD examples because they tend to produce high entropy. Additionally, because the weights $\bm{p}_{\bm{\phi}}^j$ are not normalized, examples from classes that present prototypes with low norms are unjustifiably favored to be considered OOD examples for the same reason. Thus, we propose replacing $\bm{f}_{\bm{\theta}}(\bm{x})$ with its normalized version given by $\widehat{\bm{f}_{\bm{\theta}}(\bm{x})}\!=\!\bm{f}_{\bm{\theta}}(\bm{x})/\norm{\bm{f}_{\bm{\theta}}(\bm{x})}$. Additionally, we propose replacing $\bm{p}_{\bm{\phi}}^j$ with its normalized version given by $\widehat{\bm{p}_{\bm{\phi}}^j}\!=\!\bm{p}_{\bm{\phi}}^j/\norm{\bm{p}_{\bm{\phi}}^j}$. The expression $\norm{\bm{v}}$ represents the 2-norm of a given vector $\bm{v}$.

However, while the distances in the original IsoMax loss may vary from zero to infinity, the distance between two normalized vectors is always equal to or lower than two. To avoid this unjustifiable and unreasonable restriction, we introduce the \emph{distance scale} $d_s$, which is a \emph{scalar learnable parameter}. Naturally, we require the distance scale to always be positive by taking its absolute value $\abs{d_s}$.

Feature normalization makes the solution isometric regardless of the norm of the features produced by the examples. The distance scale is class independent because it is a \emph{single} scalar value learnable during training. The weight normalization and the class independence of the distance scale make the solution isometric regarding all classes. The final distance is isometric because it produces an isometric treatment of all features, prototypes, and classes.
Therefore, we can write the \emph{isometric distances} used by the IsoMax+ loss as $\mathcal{D_I}=\abs{d_s}\:\norm{\widehat{\bm{f}_{\bm{\theta}}(\bm{x})}\!-\!\widehat{\bm{p}_{\bm{\phi}}^j}}$. Returning to Equation~\eqref{eq:isomax_loss_full}, we can finally write the expression for the IsoMax+ loss as follows (see Algorithm 1 \hl{and Fig.~}\ref{fig:isometric_distances}):

\begin{equation}
\begin{aligned}
\mathcal{L}_{\textsf{IsoMax+}}=
&-\log^{\dagger\dagger}\left(\frac{\exp(-E_s\:\abs{d_s}\:\norm{\widehat{\bm{f}_{\bm{\theta}}(\bm{x})}\!-\!\widehat{\bm{p}_{\bm{\phi}}^k}})}{\sum\limits_{j}\exp(-E_s\:\abs{d_s}\:\norm{\widehat{\bm{f}_{\bm{\theta}}(\bm{x})}\!-\!\widehat{\bm{p}_{\bm{\phi}}^j}})}\right)\label{eq:isomax2_loss_full}
\end{aligned}
\end{equation}
\blfootnote{\textsuperscript{$\dagger$}\textsuperscript{$\dagger$}The probability (i.e., the expression between the outermost parentheses) and logarithm operations are computed sequentially and separately for higher OOD detection performance (see Algorithm 1 and the source code).}

\begin{algorithm*}[!t]
\label{alg:pseudo-code1}
\captionsetup{font=small}
\caption{PyTorch for the Enhanced IsoMax Loss.}
\definecolor{codeblue}{rgb}{0.25,0.5,0.5}
\lstset{
basicstyle=\fontsize{8.5 pt}{8.5 pt}\ttfamily\bfseries,
commentstyle=\fontsize{8.5 pt}{8.5 pt}\color{codeblue},
keywordstyle=\fontsize{8.5 pt}{8.5 pt}\color{magenta},
}
\begin{lstlisting}[language=python]
class IsoMaxPlusLossFirstPart(nn.Module):
    """This part replaces the model classifier output layer nn.Linear()"""
    def __init__(self, num_features, num_classes):
        super(IsoMaxPlusLossFirstPart, self).__init__()
        self.num_features = num_features
        self.num_classes = num_classes
        self.prototypes = nn.Parameter(torch.Tensor(num_classes, num_features))
        nn.init.normal_(self.prototypes, mean=0.0, std=1.0)
        self.distance_scale = nn.Parameter(torch.Tensor(1)) 
        nn.init.constant_(self.distance_scale, 1.0)

    def forward(self, features):
        distances = torch.abs(self.distance_scale) * torch.cdist(
            F.normalize(features), F.normalize(self.prototypes),
            p=2.0, compute_mode="donot_use_mm_for_euclid_dist")
        logits = -distances
        return logits

class IsoMaxPlusLossSecondPart(nn.Module):
    """This part replaces the nn.CrossEntropyLoss()"""
    def __init__(self, entropic_scale = 10.0):
        super(IsoMaxPlusLossSecondPart, self).__init__()
        self.entropic_scale = entropic_scale

    def forward(self, logits, targets):
        """Probabilities and logarithms are calculated
        separately and sequentially"""
        """Therefore, nn.CrossEntropyLoss() must not be
        used to calculate the loss"""
        distances = -logits
        probabilities_for_training = nn.Softmax(dim=1)
        (-self.entropic_scale * distances)
        probabilities_at_targets = 
        probabilities_for_training[range(distances.size(0)), targets]
        loss = -torch.log(probabilities_at_targets).mean()
        return loss
\end{lstlisting}
\end{algorithm*}

Applying the entropy maximization trick (i.e., the removal of the entropic scale $E_s$ for inference) \citep{macdo2019isotropic}, we can write the expression for the IsoMax+ loss probabilities used during inference for performing OOD detection when using the maximum probability score or the entropic score \citep{macdo2019isotropic}:

\begin{align}\label{eq:probability_class_isomax2}
\mathcal{P}_{\textsf{IsoMax+}}(y^{(i)}|\bm{x})&=\frac{\exp(-\:\abs{d_s}\:\norm{\widehat{\bm{f}_{\bm{\theta}}(\bm{x})}\!-\!\widehat{\bm{p}_{\bm{\phi}}^i}})}{\sum\limits_{j}\exp(-\:\abs{d_s}\:\norm{\widehat{\bm{f}_{\bm{\theta}}(\bm{x})}\!-\!\widehat{\bm{p}_{\bm{\phi}}^j}})}
\end{align}

\newpage
Different from IsoMax loss, where the prototypes are initialized to a zero vector, we initialized all prototypes using a normal distribution with a mean of zero and a standard deviation of one. This approach is necessary because we normalize the prototypes when using IsoMax+ loss. The distance scale is initialized to one, and we added no hyperparameters to the solution.

\subsection{Minimum Distance Score}

Motivated by the desired characteristics of the isometric distances used in IsoMax+, we use the minimum distance as score for performing OOD detection. Naturally, the \gls{mds} for IsoMax+ is given by:

\begin{align}
\text{MDS}\!=\!\min_j \left(\norm{\widehat{\bm{f}_{\bm{\theta}}(\bm{x})}\!-\!\widehat{\bm{p}_{\bm{\phi}}^j}}\right)
\label{eq:min_distance_score}
\end{align}

In this equation, $\abs{d_s}$ was removed because, after training, it is a scale factor that does not affect the detection decision. Considering that the minimum distance is computed to perform the classification because the predicted class is the one that presents \emph{the lowest feature-prototype distance}, when using the minimum distance score, \emph{the OOD detection exhibits essentially zero latency and computational cost} because we simply reuse the minimum distance that was already calculated for classification purpose.

\begin{figure*}
\small
\centering
\caption[IsoMax+ Loss PyTorch Code Usage]{IsoMax+ Loss PyTorch Code Usage}
\subfloat[]{\includegraphics[width=\textwidth,trim={0 0 0 0},clip]{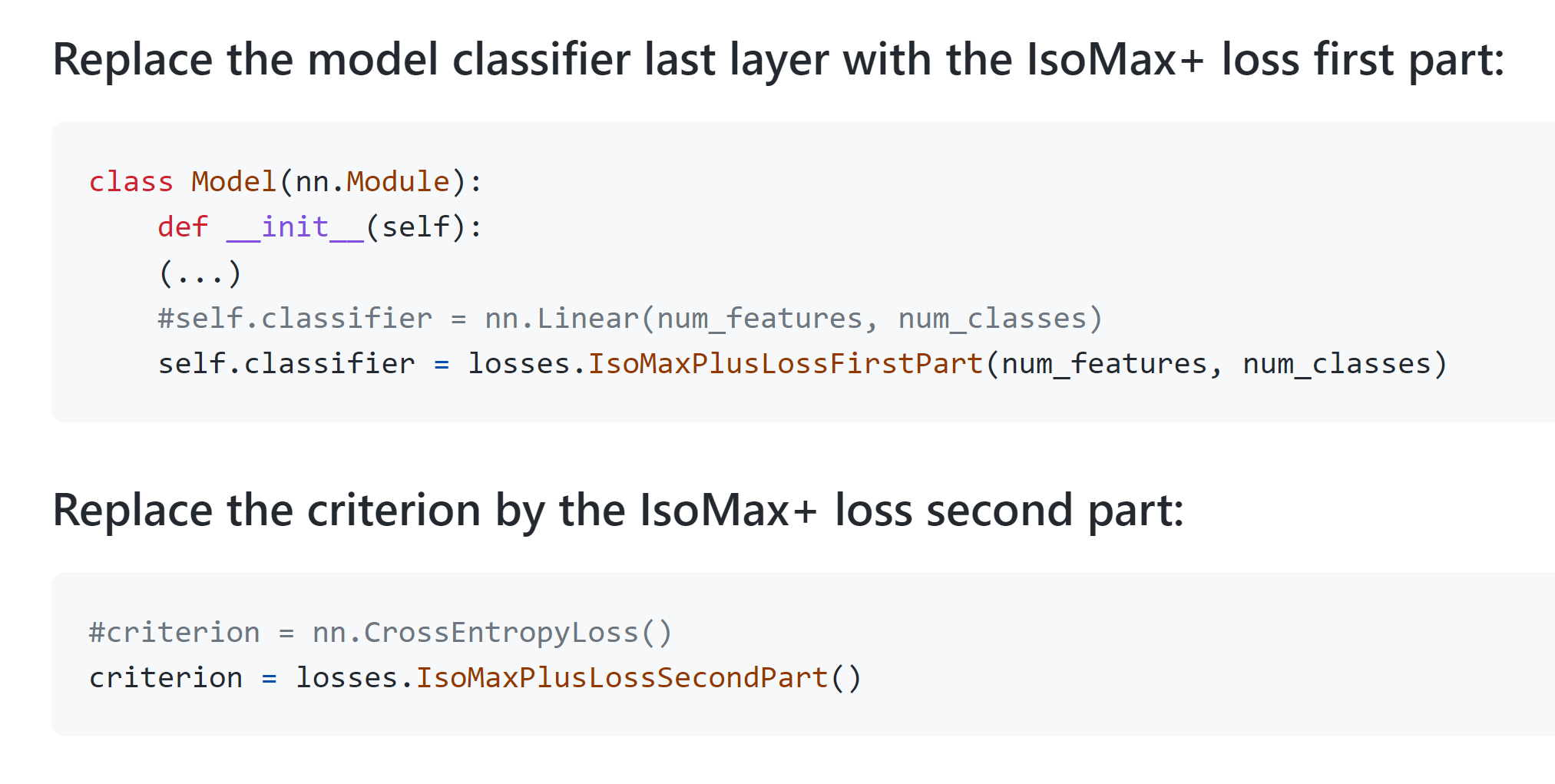}}
\\
\subfloat[]{\includegraphics[width=\textwidth,trim={0 0 0 0},clip]{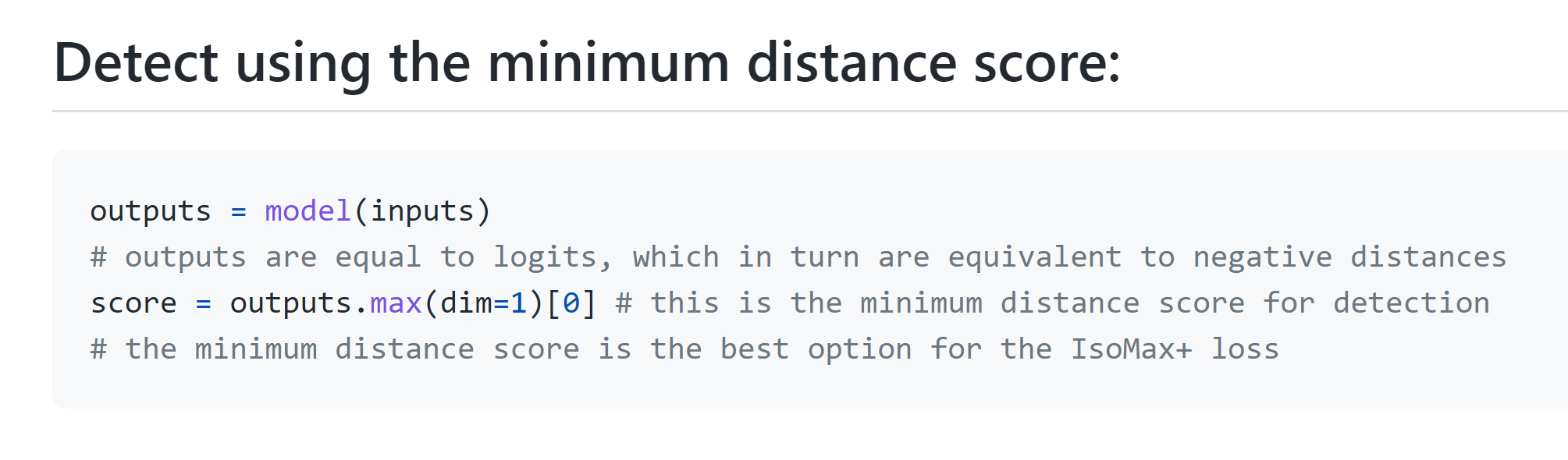}}
\begin{justify}
{Source: The Author (2022). IsoMax+ Loss PyTorch Code Usage: a) Replace the last linear layer by the IsoMax+ loss code first part. Replace the cross-entropy loss with the IsoMax+ loss code second part. b) Detect using the minimum distance score.}
\end{justify}
\label{fig:code_train_detect}
\end{figure*}

\newpage\section{Distinction Maximization Loss}\label{sec:distiction_maximization_loss}

We started from IsoMax+ loss \citep{macedo2021enhanced} to construct the \gls{dismax}. First, we create the \emph{enhanced} logits (logits+) by using \emph{all} feature-prototype distances, rather than just the feature-prototype distance to the correct class. Second, we introduce the \gls{fpr} by minimizing the Kullback–Leibler (KL) divergence between the output probability distribution associated with a \emph{compound} image and a target probability distribution containing \emph{fractional} probabilities. \hl{The motivation of training on fractional probabilities is to force the neural network present outputs with higher entropies.} 

We call DisMax dagger (DisMax\textsuperscript{$\dagger$}) the variant using FPR. Otherwise, we simply call it DisMax. Third, we construct a \emph{composite} score for OOD detection that combines three components: the maximum logit+, the \emph{mean} logit+, and the entropy of the network output. Fourth, we present a simple and fast temperature-scaling procedure that \hl{allows DisMax trained models to produce a high-performance uncertainty estimation.} Like IsoMax+, DisMax works as a drop-in replacement for SoftMax loss. The trained models keep deterministic neural network inference efficiency.

\begin{figure}
\centering
\small
\caption[All-Distances-Aware Logits, Enhanced Logits, or Logits+]{All-Distances-Aware Logits, Enhanced Logits, or Logits+}
\subfloat[]{
\begin{tikzpicture}[tdplot_main_coords, scale = 0.8]
\node[align=center] at (0,0,3.6) {\color{black} \textit{(n-1)-sphere of radius one in the}\\\textit{IsoMax+ n-dimensional Euclidean space}\\\\};
\coordinate (P1) at ({1.75/sqrt(2)},{-1/sqrt(2)},{1.5/sqrt(2)});
\coordinate (P2) at ({(-1/sqrt(2))/2},{(1/sqrt(2))/2},{2/sqrt(2)});
\coordinate (P3) at ({2.5/sqrt(2)},{1.75/sqrt(2)},{0});
\coordinate (F1) at ({1.3/sqrt(2)},{2.5/sqrt(2)},{2/sqrt(2)});
\shade[ball color = lightgray, opacity = 0.5] (0,0,0) circle (2.0cm);
\tdplotsetrotatedcoords{0}{0}{0};
\draw[dotted, tdplot_rotated_coords, gray] (0,0,0) circle (2);
\tdplotsetrotatedcoords{90}{90}{90};
\draw[dotted, tdplot_rotated_coords, gray] (2,0,0) arc (0:180:2);
\tdplotsetrotatedcoords{0}{90}{90};
\draw[dotted, tdplot_rotated_coords, gray] (2,0,0) arc (0:180:2);
\draw[dotted, gray] (0,0,0) -- (-2,0,0);
\draw[dotted, gray] (0,0,0) -- (0,-2,0);
\draw[dotted, gray] (0,0,0) -- (0,0,-2);
\draw[-stealth] (0,0,0) -- (3.60,0,0) node[below left] {$x$};
\draw[-stealth] (0,0,0) -- (0,2.50,0) node[below right] {$y$};
\draw[-stealth] (0,0,0) -- (0,0,2.70) node[above] {$z$};
\draw[thick, -stealth] (0,0,0) -- (P1) node[above] {$P_1$};
\draw[thick, -stealth] (0,0,0) -- (P2) node[above] {$P_2$};
\draw[thick, -stealth] (0,0,0) -- (P3) node[below] {$P_3$};
\draw[thick, -stealth, blue] (0,0,0) -- (F1) node[above right] {$F$};
\draw[ultra thick, dashed, olive] (F1) -- (P1);
\end{tikzpicture}
}
\hskip 0.5 cm
\subfloat[]{
\begin{tikzpicture}[tdplot_main_coords, scale = 0.8]
\node[align=center] at (0,0,3.6) {\color{black} \textit{(n-1)-sphere of radius one in the}\\\textit{DisMax n-dimensional Euclidean space}\\\\};
\coordinate (P1) at ({1.75/sqrt(2)},{-1/sqrt(2)},{1.5/sqrt(2)});
\coordinate (P2) at ({(-1/sqrt(2))/2},{(1/sqrt(2))/2},{2/sqrt(2)});
\coordinate (P3) at ({2.5/sqrt(2)},{1.75/sqrt(2)},{0});
\coordinate (F1) at ({1.3/sqrt(2)},{2.5/sqrt(2)},{2/sqrt(2)});
\shade[ball color = lightgray, opacity = 0.5] (0,0,0) circle (2.0cm);
\tdplotsetrotatedcoords{0}{0}{0};
\draw[dotted, tdplot_rotated_coords, gray] (0,0,0) circle (2);
\tdplotsetrotatedcoords{90}{90}{90};
\draw[dotted, tdplot_rotated_coords, gray] (2,0,0) arc (0:180:2);
\tdplotsetrotatedcoords{0}{90}{90};
\draw[dotted, tdplot_rotated_coords, gray] (2,0,0) arc (0:180:2);
\draw[dotted, gray] (0,0,0) -- (-2,0,0);
\draw[dotted, gray] (0,0,0) -- (0,-2,0);
\draw[dotted, gray] (0,0,0) -- (0,0,-2);
\draw[-stealth] (0,0,0) -- (3.60,0,0) node[below left] {$x$};
\draw[-stealth] (0,0,0) -- (0,2.50,0) node[below right] {$y$};
\draw[-stealth] (0,0,0) -- (0,0,2.70) node[above] {$z$};
\draw[thick, -stealth] (0,0,0) -- (P1) node[above] {$P_1$};
\draw[thick, -stealth] (0,0,0) -- (P2) node[above] {$P_2$};
\draw[thick, -stealth] (0,0,0) -- (P3) node[below] {$P_3$};
\draw[thick, -stealth, blue] (0,0,0) -- (F1) node[above right] {$F$};
\draw[ultra thick, dashed, purple] (F1) -- (P1);
\draw[ultra thick, dashed, purple] (F1) -- (P2);
\draw[ultra thick, dashed, purple] (F1) -- (P3);
\end{tikzpicture}
\label{fig:softmax}
}
\begin{justify}
{Source: The Author (2022). The illustration presents the difference between IsoMax+ \citep{macedo2021enhanced} and DisMax with respect to \emph{logit formation}. $P_1$, $P_2$, and $P_3$ represent prototypes of classes $1$, $2$, and $3$, respectively. $F$ denotes the feature associated with a given image. Like all current losses, IsoMax+ constructs \emph{each} logit associated with $F$ considering its distance from a \emph{single} prototype (olive dashed line). In contrast, DisMax loss builds \emph{each} logit associated with $F$ considering its distance from \emph{all} prototypes (purple dashed lines). We use the terms \emph{all-distances-aware} logits, \emph{enhanced} logits, or logits+ indistinctly.}
\end{justify}
\label{fig:logits+}
\end{figure}
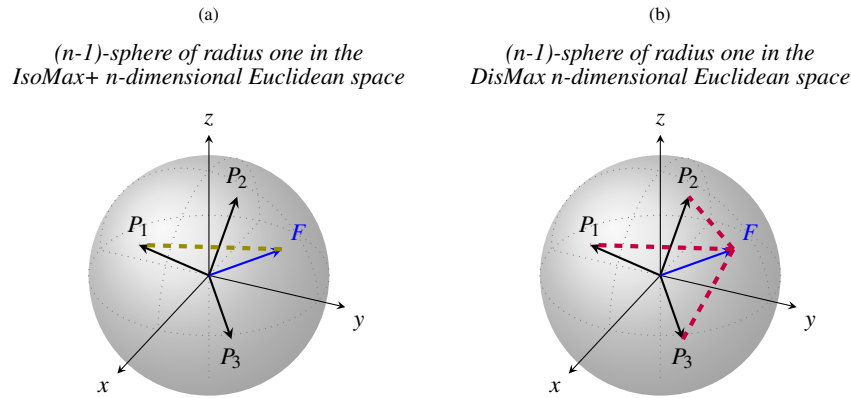

\subsection{All-Distances-Aware Logits}

In IsoMax loss variants (e.g., IsoMax and IsoMax+), logits are formed from distances and are commonly used to calculate the score to perform OOD detection. Hence, it is essential to build logits that contain semantic information relevant to separating in-distribution (ID) from OOD during inference. IsoMax+ uses \emph{isometric} distances \citep{macedo2021enhanced}.

\looseness=-1
In IsoMax+, the logits are simply the negatives of the isometric distances. We have two motivations to add the \emph{mean} isometric distance considering \emph{all} prototypes to the isometric distance associated with \emph{each} class to construct what we call \emph{all-distances-aware} logits, \emph{enhanced} logits, or logits+.

First, considering that IsoMax+ is an isotropic loss, the pairwise distances between the prototypes and ID examples are forced to become increasingly smaller. Therefore, after training, it is reasonable to believe that ID feature-prototype distances are, on average, smaller than the distances from the prototypes to OOD samples, which were not forced to be closer to the prototypes. Hence, adding the mean distance to the logits used in IsoMax+ can help distinguish between ID and OOD more effectively. Second, taking \emph{all} feature-prototype distances to compose the logits makes them a more stable information for OOD detection (Fig.~\ref{fig:logits+}).

\begin{center}
\begin{minipage}{\textwidth}
\vskip 2 cm
\small
\begin{equation}
\centering
\vspace{\baselineskip}
\label{eq:logits+}
\tikzmarknode{x}{\highlight{olive}{$L^j_{+}$}}=-\left(\tikzmarknode{y}{\highlight{red}{$D_I^j$}}\:+\tikzmarknode{z}{\highlight{blue}{$\frac{1}{N}\sum\limits_{n=1}^N D_I^n$}}\right)
\vspace{\baselineskip}
\end{equation}
\begin{tikzpicture}[overlay,remember picture,>=stealth,nodes={align=left,inner ysep=1pt},<-]
\path (x.north) ++ (0,0.5em) node[anchor=south east,color=olive!67, align=center] (scalep){\textbf{all-distances-aware logit}\\\textbf{for the j-th class}};
\draw [color=olive!67](x.north) |- ([xshift=0ex,color=olive]scalep.south west);
\path (y.south) ++ (0,-1.0em) node[anchor=north west,color=red!67, align=center] (mean){\textbf{isometric distance to}\\\textbf{the j-th class prototype}};
\draw [color=red!67](y.south) |- ([xshift=0ex,color=red]mean.south east);
\path (z.north) ++ (0em,0.5em) node[anchor=south west,color=blue!67, align=center] (mean){\textbf{mean isometric distance}\\\textbf{to all prototypes}};
\draw [color=blue!67](z.north) |- ([xshift=0ex,color=blue]mean.south east);
\end{tikzpicture}
\end{minipage}
\end{center}

\begin{center}
\begin{minipage}{\textwidth}
\vskip 2 cm
\small
\begin{equation}
\vspace{\baselineskip}
\label{eq:dismax_prob}
\tikzmarknode{x}{\highlight{olive}{$\mathcal{P}_{\textsf{DisMax}}(y^{(i)}|\bm{x})$}}=\frac{\exp(E_s\tikzmarknode{z}{\highlight{olive}{$L^i_{+}$}}/T)}{\sum\limits_{j=1}^N\exp(E_s\tikzmarknode{y}{\highlight{olive}{$L^j_{+}$}}/T)}
\vspace{\baselineskip}
\end{equation}
\begin{tikzpicture}[overlay,remember picture,>=stealth,nodes={align=left,inner ysep=1pt},<-]
\path (x.north) ++ (0,0.5em) node[anchor=south east,color=olive!67, align=center] (scalep){\textbf{predicted probability}\\\textbf{distribution}};
\draw [color=olive!67](x.north) |- ([xshift=0ex,color=olive]scalep.south west);
\path (y.south) ++ (0,-0.5em) node[anchor=north west,color=olive!67,align=center] (mean){\textbf{all-distances-aware logit}\\\textbf{for the j-th class}};
\draw [color=olive!67](y.south) |- ([xshift=0ex,color=olive]mean.south east);
\path (z.north) ++ (0em,0.5em) node[anchor=south west,color=olive!67,align=center] (mean){\textbf{all-distances-aware logit}\\\textbf{for the i-th class}};
\draw [color=olive!67](z.north) |- ([xshift=0ex,color=olive]mean.south east);
\end{tikzpicture}
\end{minipage}
\end{center}

\looseness=-1
Therefore, we consider an input $\bm{x}$ and a network that performs a transformation $\bm{f}_{\bm{\theta}}(\bm{x})$. We also consider $\bm{p}_{\bm{\phi}}^j$ to be the learnable prototype associated with class $j$. Moreover, considering that $\norm{\bm{v}}$ represents the 2-norm of a vector $\bm{v}$, and $\widehat{\bm{v}}$ represents the 2-norm normalization of $\bm{v}$, we can write the \emph{isometric distance} relative to class $j$ as $D_I^j=\abs{d_s}\:\norm{\widehat{\bm{f}_{\bm{\theta}}(\bm{x})}\!-\!\widehat{\bm{p}_{\bm{\phi}}^j}}$, where $\abs{d_s}$ represents the absolute value of the \emph{learnable} scalar called distance scale \citep{macedo2021enhanced}. Finally, we can write the proposed \emph{enhanced} logit for class $j$ using the equation \eqref{eq:logits+}. $N$ is the number of classes. Probabilities are given by equation \eqref{eq:dismax_prob}, where $T$ is the temperature. $E_s$
is the entropic scale, \emph{which is removed after training} \citep{macdo2019isotropic, DBLP:journals/corr/abs-2006.04005, macedo2021enhanced}, \emph{but before calibration}. For the rest of this section, distance means \emph{isometric} distance.

\newpage\subsection{Fractional Probability Regularization}

We often train neural networks using \emph{unitary} probabilities. Indeed, the usual cross-entropy loss forces a \emph{probability equal to one} on a given training example. Consequently, we commonly train neural networks by providing a tiny proportion of points in the learning manifold. Hence, we propose what we call \emph{fractional} probability regularization (FPR). The idea is to force the network to learn more diverse points in the learning manifold. Consequently, we confront target and predicted probability distributions also on \emph{fractional} probability values rather than only \emph{unitary} probability manifold points.


\begin{center}
\begin{minipage}{\textwidth}
\vskip 2 cm
\small
\begin{align}
\vspace{\baselineskip}
\label{eq:fpr}
\tikzmarknode{x}{\highlight{blue}{$\mathcal{Q}_{\textsf{Target}}(y^{(i)}|\bm{\tilde{x}})$}}=\tikzmarknode{y}{\highlight{blue}{$\frac{1}{4}{\sum\limits_{m=1}^4\delta[y^{(i)}-y^{(j^m)}]}$}}
\vspace{\baselineskip}
\end{align}
\begin{tikzpicture}[overlay,remember picture,>=stealth,nodes={align=left,inner ysep=1pt},<-]
\path (x.north) ++ (0,0.5em) node[anchor=south east,color=blue!67,align=center] (scalep){\textbf{desired probability distribution}\\\textbf{for the compound image}};
\draw [color=blue!67](x.north) |- ([xshift=0ex,color=blue]scalep.south west);
\path (y.south) ++ (0,-0.5em) node[anchor=north west,color=blue!67,align=center] (mean){\textbf{sum one quarter to the}\\\textbf{probability of each class}\\\textbf{in the compound image}};
\draw [color=blue!67](y.south) |- ([xshift=0ex,color=blue]mean.south east);
\end{tikzpicture}
\end{minipage}
\end{center}

\vskip 1cm

\begin{center}
\begin{minipage}{\textwidth}
\vskip 2 cm
\small
\begin{equation}
\label{eq:dismax_loss}
\vspace{\baselineskip}
\tikzmarknode{w}{\highlight{olive}{$\mathcal{L}_{\textsf{DisMax}}$}}=-\log^*\left(\frac{\exp(E_s \tikzmarknode{x}{\highlight{olive}{$L^k_{+}$}})}{\sum\limits_j\exp(E_s \tikzmarknode{y}{\highlight{olive}{$L^j_{+}$}})}\right) + 
\tikzmarknode{a}{\alpha}\cdot
\tikzmarknode{z}{\highlight{blue}{\kld{\mathcal{P}_{\textsf{DisMax}}(y|\bm{\tilde{x}})}{\mathcal{Q}_{\textsf{Target}}(y|\bm{\tilde{x}})}}}
\vspace{\baselineskip}
\end{equation}
\begin{tikzpicture}[overlay,remember picture,>=stealth,nodes={align=left,inner ysep=1pt},<-]
\path (x.north) ++ (0,1.5em) node[anchor=south west,color=olive!67,align=center] (scalep){\textbf{enhanced logit}\\\textbf{for the correct class k}};
\draw [color=olive!67](x.north) |- ([xshift=0ex,color=olive]scalep.south east);
\path (y.south) ++ (0,-1.0em) node[anchor=north east,color=olive!67,align=center] (mean){\textbf{enhanced logit}\\\textbf{for the j-th class}};
\draw [color=olive!67](y.south) |- ([xshift=0ex,color=olive]mean.south west);
\path (z.south) ++ (0em,-0.5em) node[anchor=north west,color=blue!67, align=center] (mean){\textbf{fractional probability}\\\textbf{regularization}};
\draw [color=blue!67](z.south) |- ([xshift=0ex,color=blue]mean.south east);
\path (w.north) ++ (0,0.5em) node[anchor=south east,color=olive!67,align=center] (scalep){\textbf{loss}\\\textbf{function}};
\draw [color=olive!67](w.north) |- ([xshift=0ex,color=olive]scalep.south west);
\end{tikzpicture}
\end{minipage}
\end{center}
\blfootnote{\textcolor{black}{\textsuperscript{*}The probability (i.e., the expression between the outermost parentheses) and logarithm operations are computed sequentially and separately for optimal OOD detection performance.}}

\begin{figure}
\small
\centering
\caption[Fractional Probability Regularization]{Fractional Probability Regularization}
\includegraphics[width=\textwidth]{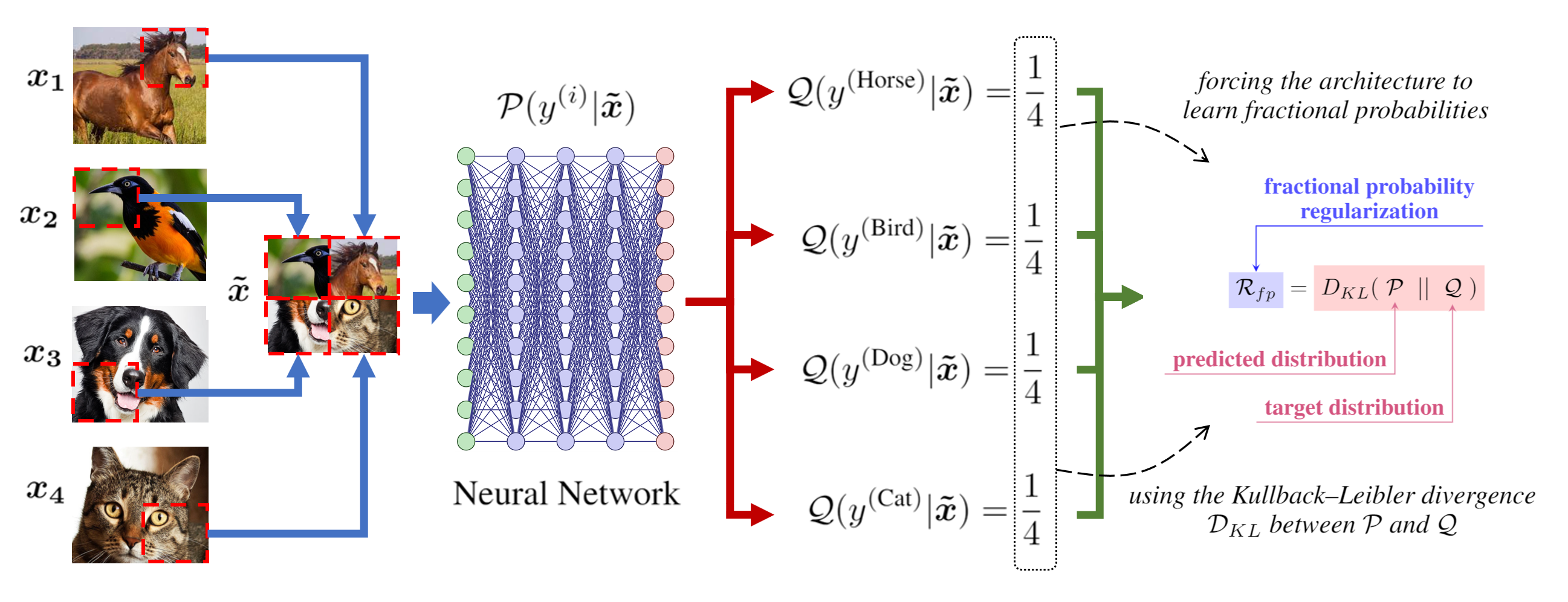}
\begin{justify}
{Source: The Author (2022). We use images composed of patches of four randomly selected training examples. The KL divergence regularization term forces the network to predict \emph{fractional} probabilities on compound images.}
\end{justify}
\label{fig:fpr}
\end{figure}

Therefore, our batch is divided into two halves. In the first half, we use the regular \emph{unitary} probability training. For the second batch, we construct images specifically composed of patches of four others (Fig. \ref{fig:fpr}). We construct our target probability distribution $Q$ for those images by adding a quarter probability for each class corresponding to a patch of the compound image. Finally, we minimize the KL divergence regularization between our predicted and target probability distributions in the second half. These procedures do not increase training memory size requirements. Considering $\delta$ the Kronecker delta function\footnote{\url{https://en.wikipedia.org/wiki/Kronecker_delta}}, equation \eqref{eq:fpr} represents the FPR in math terms. By combining the \emph{enhanced} logits and the FPR, equation \eqref{eq:dismax_loss} presents the mathematical expression of the DisMax loss. Considering OOD is related to the uncertainty estimation, we validate $\alpha$ without requiring access to OOD data by choosing the pretrained model that produces the lower expected calibration error (ECE) after the temperature calibration.

We recognize a similarity between CutMix \citep{DBLP:conf/iccv/YunHCOYC19} and FPR: both are based on the combination of images to create compound data. However, we identify many differences. CutMix combines two images, while FPR combines four images. Moreover, the combination procedure is entirely different. In CutMix, a portion of an image is replaced by a patch of \emph{variable} size, format, and position that comes from another image. In FPR, patches of the \emph{same} size, format, and \emph{predefined} positions from four different images are combined into a single one. This simplification introduced by FPR allowed us to combine four images instead of only two. Indeed, trying to extend CutMix by replacing portions of an image with patches from three others while allowing \emph{random} sizes, shapes, and positions produces \emph{patch superpositions}, making \emph{the calculation of the pairwise ratio of the areas of the superposed patches extremely hard}. Therefore, this simplification made it possible to simultaneously combine four rather than only two images, in addition to avoiding the beta distribution and the $\alpha$ hyperparameter. In FPR, \emph{fractional} probabilities, rather than losses, are proportional to areas.

While CutMix is applied randomly to some batches with probability~$p$, FPR is applied to half of each batch, avoiding loss or gradient oscillations. CutMix neither creates a target distribution containing \emph{fractional} probabilities nor forces the predicted probabilities to follow it by minimizing the KL divergence between them. Indeed, CutMix does \emph{not} use the KL divergence at all. CutMix calculates the regular cross-entropy loss of the compound image considering the labels of the original images and takes a linear interpolation between the resulting loss values weighted by the ratio of the areas of the patch and the remaining image. While CutMix operates on losses, FPR operates directly on probabilities \emph{before} calculating loss values. The concept of \emph{fractional} probabilities is not even present in CutMix. Unlike CutMix, the mentioned procedure can be easily expanded to combine even more than four images. Finally, CutMix \emph{increases the training time} and \emph{presents hyperparameters} \citep{DBLP:conf/iccv/YunHCOYC19}.


Even if we consider adapting the mosaic augmentation from the objection detection task \citep{DBLP:journals/corr/abs-2004-10934} to classification and add to it a CutMix-like regularization, most of the previously mentioned differences between FPR and CutMix still holds to differentiate FPR from this CutMix-like classification-ported mosaic augmentation.

\subsection{Max-Mean Logit Entropy Score}

For OOD detection, we propose a score composed of three parts. The first part is the maximum logit+. The second part is the \emph{mean} logit+. Incorporating the mean value of the logits into the score is an \emph{independent procedure relative to the logit formation}. It can be applied regardless of the type of logit (e.g., usual or enhanced) used during training. Finally, we subtract the entropy calculated considering the probabilities of the neural network output. We call this \emph{composite} score \gls{mmles}. It is given by equation \eqref{eq:mmles}. We call MMLES a \emph{composite} score because it is formed by the sum of many other scores.

\vskip 2cm

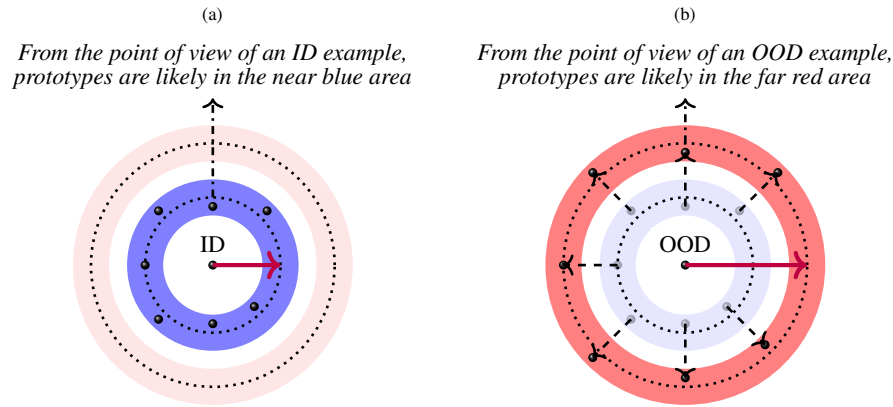
\begin{figure}
\centering
\small
\caption[Max-Mean Logit Entropy Score]{Max-Mean Logit Entropy Score}
\subfloat[]{
\begin{tikzpicture}[line width=1pt, scale=0.6]
\fill[blue!50,even odd rule] (0,0) circle (1.1) (0,0) circle (1.9);
\fill[red!10,even odd rule] (0,0) circle (2.3) (0,0) circle (3.1);
\node[above=0pt of {(0,0)}, outer sep=2pt,fill=white] {ID};
\shade[ball color = black, opacity = 1.0] (0,0) circle (0.1cm);
\shade[ball color = black, opacity = 1.0] (45:1.7) circle (0.1cm);
\shade[ball color = black, opacity = 1.0] (90:1.3) circle (0.1cm);
\shade[ball color = black, opacity = 1.0] (135:1.7) circle (0.1cm);
\shade[ball color = black, opacity = 1.0] (180:1.5) circle (0.1cm);
\shade[ball color = black, opacity = 1.0] (225:1.7) circle (0.1cm);
\shade[ball color = black, opacity = 1.0] (270:1.3) circle (0.1cm);
\shade[ball color = black, opacity = 1.0] (315:1.3) circle (0.1cm);
\foreach \x in {1.5} {\draw [dotted] (0,0) circle (\x cm);}
\foreach \x in {2.7} {\draw [dotted] (0,0) circle (\x cm);}
\draw[dash dot,black,->] (90:1.5) -- (90:3.7) node[above,align=center] {\color{black} \textit{From the point of view of an ID example,}\\\textit{prototypes are likely in the near blue area}};
\draw[->, ultra thick, purple] (0,0) -- (0:1.5);
\end{tikzpicture}
}
\hskip 0.5 cm
\subfloat[]{
\begin{tikzpicture}[line width=1pt, scale=0.6]
\fill[blue!10,even odd rule] (0,0) circle (1.1) (0,0) circle (1.9);
\fill[red!50,even odd rule] (0,0) circle (2.3) (0,0) circle (3.1);
\node[above=0pt of {(0,0)}, outer sep=2pt,fill=white] {OOD};
\shade[ball color = black, opacity = 1.0] (0,0) circle (0.1cm);
\shade[ball color = black, opacity = 0.3] (45:1.7) circle (0.1cm);
\shade[ball color = black, opacity = 0.3] (90:1.3) circle (0.1cm);
\shade[ball color = black, opacity = 0.3] (135:1.7) circle (0.1cm);
\shade[ball color = black, opacity = 0.3] (180:1.5) circle (0.1cm);
\shade[ball color = black, opacity = 0.3] (225:1.7) circle (0.1cm);
\shade[ball color = black, opacity = 0.3] (270:1.3) circle (0.1cm);
\shade[ball color = black, opacity = 0.3] (315:1.3) circle (0.1cm);
\shade[ball color = black, opacity = 1.0] (45:2.9) circle (0.1cm);
\shade[ball color = black, opacity = 1.0] (90:2.5) circle (0.1cm);
\shade[ball color = black, opacity = 1.0] (135:2.9) circle (0.1cm);
\shade[ball color = black, opacity = 1.0] (180:2.7) circle (0.1cm);
\shade[ball color = black, opacity = 1.0] (225:2.9) circle (0.1cm);
\shade[ball color = black, opacity = 1.0] (270:2.5) circle (0.1cm);
\shade[ball color = black, opacity = 1.0] (315:2.5) circle (0.1cm);
\draw [->, dashed] (45:1.7) -- (45:2.9);
\draw [->, dashed] (90:1.3) -- (90:2.5);
\draw [->, dashed] (135:1.7) -- (135:2.9);
\draw [->, dashed] (180:1.5) -- (180:2.7);
\draw [->, dashed] (225:1.7) -- (225:2.9);
\draw [->, dashed] (270:1.3) -- (270:2.5);
\draw [->, dashed] (315:1.3) -- (315:2.5);
\foreach \x in {1.5} {\draw [dotted] (0,0) circle (\x cm);}
\foreach \x in {2.7} {\draw [dotted] (0,0) circle (\x cm);}
\draw[dash dot,black,->] (90:2.7) -- (90:3.7) node[above,align=center] {\color{black} \textit{From the point of view of an OOD example,}\\\textit{prototypes are likely in the far red area}};
\draw[->, ultra thick, purple] (0,0) -- (0:2.7);
\end{tikzpicture}
}
\begin{justify}
{Source: The Author (2022). In addition to the maximum logit and the negative entropy, the MMLES incorporates the \emph{mean} logit+. We observed that the prototypes are generally closer to ID samples than OOD samples, which is true \emph{regardless of whether the ID sample belongs to the class of the considered prototype}. Hence, incorporating this \emph{all-distances-aware} information increases the OOD detection performance (see Chapter \ref{chap:experiments}).}
\end{justify}
\label{fig:mmles_fig}
\end{figure}

\begin{center}
\begin{minipage}{\textwidth}
\vskip 1 cm
\small
\begin{equation}
\vspace{\baselineskip}
\label{eq:mmles}
\tikzmarknode{x}{\highlight{blue}{$\mathcal{S}_{\textsf{MMLES}}$}}= \tikzmarknode{w}{\highlight{olive}{$\max_j (L^j_{+})$}} + \tikzmarknode{y}{\highlight{blue}{$\frac{1}{N}\sum\limits_{n=1}^N L^n_{+}$}} - \tikzmarknode{z}{\highlight{blue}{$\mathcal{H(\mathcal{P}_{\textsf{DisMax}})}$}}
\vspace{\baselineskip}
\end{equation}
\begin{tikzpicture}[overlay,remember picture,>=stealth,nodes={align=left,inner ysep=1pt},<-]
\path (x.north) ++ (0,1em) node[anchor=south east,color=blue!67, align=center] (scalep){\textbf{max-mean logit}\\\textbf{entropy score}};
\draw [color=blue!67](x.north) |- ([xshift=0ex,color=blue]scalep.south west);
\path (y.south) ++ (0,-1em) node[anchor=north west,color=blue!67] (mean){\textbf{mean logit+}};
\draw [color=blue!67](y.south) |- ([xshift=0ex,color=blue]mean.south east);
\path (z.north) ++ (0em,1em) node[anchor=south east,color=blue!67] (mean){\textbf{entropy}};
\draw [color=blue!67](z.north) |- ([xshift=0ex,color=blue]mean.south west);
\path (w.south) ++ (0,-1em) node[anchor=north east,color=olive!67] (mean){\textbf{maximum logit+}};
\draw [color=olive!67](w.south) |- ([xshift=0ex,color=olive]mean.south west);
\end{tikzpicture}
\end{minipage}
\end{center}


\hl{The motivation for the MMLES score is presented in} Fig.~\ref{fig:mmles_fig}. \hl{The experiments presented in Chapter} \ref{chap:experiments} \hl{show that, as explained in the mentioned figure, prototypes are usually closer to the in-distrbutions than to out-of-distributions.}

\subsection{Temperature Calibration}

Unlike the usual SoftMax loss, the IsoMax loss variants produce \emph{underestimated} probabilities \emph{to obey the Maximum Entropy Principle} \citep{macdo2019isotropic,https://doi.org/10.48550/arxiv.1908.05569,DBLP:journals/corr/abs-2006.04005,macedo2021enhanced}. Therefore, \emph{we need to perform a temperature calibration after training to improve the uncertainty estimation}. To find an optimal temperature, which was kept equal to one during training, we used the L-BFGS-B algorithm with approximate gradients and bounds equal to 0.001 and 100 \citep{doi:10.1137/0916069, 10.1145/279232.279236, 10.1145/2049662.2049669} for ECE minimization on the validation set. This calibration takes only a few seconds \hl{in our experiments}.

\hl{Therefore, we introduced the temperature and calibrated it to improve our proposed solutions' uncertainty estimation. We choose to tackle the uncertainty estimation problem using temperature calibration because} \cite{Guo2017OnCO} \hl{showed that this is a simple and effective approach. Our experiments showed that mixing our modified technique for temperature calibration with our entropic losses produced state-of-the-art results.}

%% file: chapters/4.experiments.tex
\chapter{Experiments}\label{chap:experiments}



\begin{quotation}[]{Richard Feynman}
``It doesn't matter how beautiful your theory is, it doesn't matter how smart you are. If it doesn't agree with experiment, it's wrong.''
\end{quotation}

\begin{quotation}[Cinema Paradiso]{Alfredo}
``Whatever you do, love it like you loved that
projection booth of the Paradiso when you were little...''
\end{quotation}

\begin{quotation}[]{David Macêdo}
``Similar to humans, neural networks suffer from the Dunning–Kruger effect.''
\end{quotation}

\begin{quotation}[Barbarians from Netflix]{Arminius}
``Everything I know about fighting, I know from you. What do we fight for?\\I ask myself, what were you fighting for? And was it worth it?''
\end{quotation}

In this chapter, we compare our entropic losses with the main state-of-the-art approaches. The experiments are presented in an evolutionary way. We start with IsoMax experiments. Then, we present IsoMax+ experiments. Finally, we present the DisMax experiments. The code to reproduce all experiments of this work is available online\footnote{\url{https://github.com/dlmacedo/entropic-out-of-distribution-detection}}\textsuperscript{,}\footnote{\url{https://github.com/dlmacedo/distinction-maximization-loss}}.\\

\newpage\section{Isotropy Maximization Loss}


\looseness=-1
We trained from scratch 100-layer DenseNetBCs (growth rate $k\!=\!12$) \citep{Huang2017DenselyNetworks}, 34-layer ResNets \citep{He_2016}, and 28-layer WideResNets (widening factor $k\!=\!10$) \citep{DBLP:conf/bmvc/ZagoruykoK16}. We trained on CIFAR10 \citep{Krizhevsky2009LearningImages}, CIFAR100 \citep{Krizhevsky2009LearningImages}, SVHN \citep{Netzer2011ReadingLearning}, and TinyImageNet \citep{Deng2009ImageNetDatabase} using SoftMax and IsoMax losses.

We used SGD with the Nesterov moment equal to 0.9 with a \hl{batch size of 64 for CIFAR10, CIFAR100, and TinyImageNet.} We used an initial learning rate of 0.1. The weight decay was 0.0001, and we did not use dropout. We trained during 300 epochs for CIFAR10 and CIFAR100; and during 90 epochs for TinyImageNet. We used a learning rate decay rate equal to ten applied in epoch numbers 150, 200, and 250 for CIFAR10 and CIFAR100; and 30 and 60 for TinyImageNet. These values are commonly used in papers in general that train these models in these datasets. They were used in OOD detection papers \citep{liang2018enhancing,lee2018simple}.

We used resized images from the TinyImageNet \citep{Deng2009ImageNetDatabase}\footnote{\label{odin_code}\url{https://github.com/facebookresearch/odin}} and the Large-scale Scene UNderstanding (LSUN) \citep{Yu2015LSUNLoop}\textsuperscript{\ref{odin_code}} as OOD data. Finally, we separately added these OOD data to the ID validation sets to construct the respective OOD detection test sets. For example, the OOD detection test set for ID CIFAR10 and OOD TinyImageNet is composed by combing the CIFAR validation set and the TinyImageNet OOD data. In some cases, CIFAR10 and SVHN work as OOD and the respective validation set is used to construct the OOD detection test set. We emphasize that our solution does not require validation sets for training or validation. \hlf{In the following subsection, we provide extensive justifications for why validating the entropic scale does not produce significantly different performance.} Thus, validation sets are used exclusively for building OOD detection test sets.


For text data experiments, we followed the experimental setting presented in \cite{hendrycks2018deep}. Therefore, we used the 20 Newsgroups as in-distribution data. The 20 Newsgroups is a text classification dataset of newsgroup documents. It has 20 classes and approximately 20,000 examples that are split evenly among the classes. We used the standard 60/40 train/test split. We used the IMDB, Multi30K, and Yelp Reviews datasets as out-distribution. IMDB is a dataset of movie review sentiment classification. Multi30K is a dataset of English-German image descriptions, of which we use the English descriptions. Yelp Reviews is a dataset of restaurant reviews. We trained 2-layer GRUs \citep{DBLP:conf/emnlp/ChoMGBBSB14} using SoftMax and IsoMax losses.

\newpage

We evaluated the performance using three detection metrics. First, we calculated the \gls{tnr} using the adequate threshold. In addition, we evaluated the \gls{auroc} and the \gls{dtacc}, which corresponds to the maximum classification probability over all possible thresholds $\delta$:
\begin{align}
\begin{split}
1- \min_{\delta} \big\{ P_{\texttt{in}} \left( o \left( \mathbf{x} \right) \leq \delta \right) P \left(\mathbf{x}\text{ is from }P_{\texttt{in}}\right)
\\+ P_{\texttt{out}} \left( o \left( \mathbf{x} \right) >\delta \right) P \left(\mathbf{x}\text{ is from }P_{\texttt{out}}\right)\big\},
\end{split}
\end{align}
where $o(\mathbf{x})$ is a given OOD detection score. It is assumed that both positive and negative samples have equal probability. \hl{This is the way this metric is defined and used in the literature.} AUROC and DTACC are threshold independent.

\subsection{Entropic Scale, High Entropy, and Out-of-Distribution Detection}\label{entropic_scale}

\begin{figure*}
\small
\centering
\caption[IsoMax Loss Effects]{IsoMax Loss Effects}
\subfloat[]{\includegraphics[width=0.2375\textwidth]{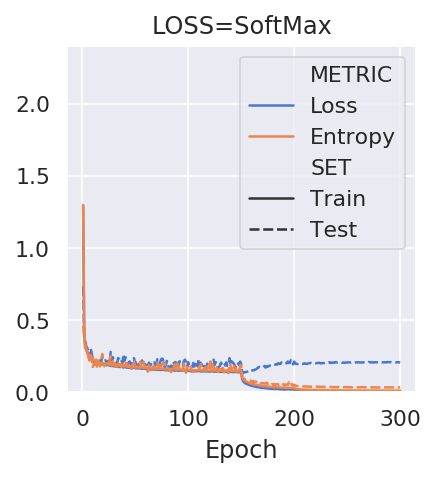}\label{fig:train_loss_entropies_softmax_new}}
\subfloat[]{\includegraphics[width=0.2375\textwidth]{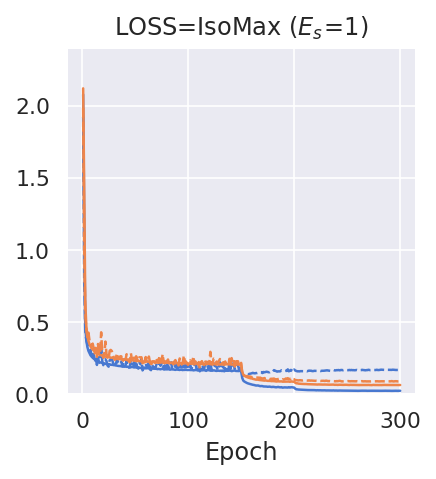}\label{fig:train_loss_entropies_isomax1_new}}
\subfloat[]{\includegraphics[width=0.2375\textwidth]{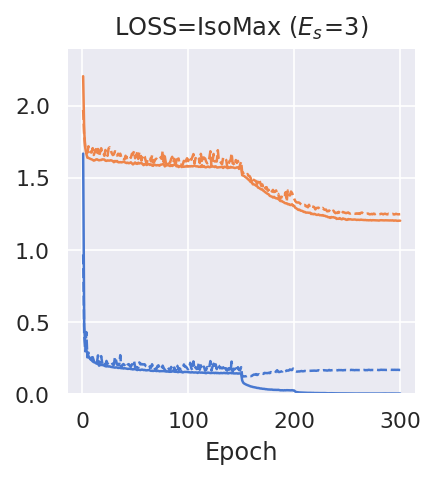}\label{fig:train_loss_entropies_isomax3_new}}
\subfloat[]{\includegraphics[width=0.2375\textwidth]{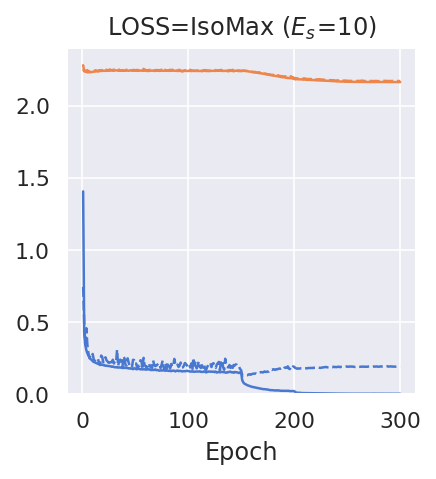}\label{fig:train_loss_entropies_isomax10_new}}
\\
\subfloat[]{\includegraphics[width=0.925\textwidth]{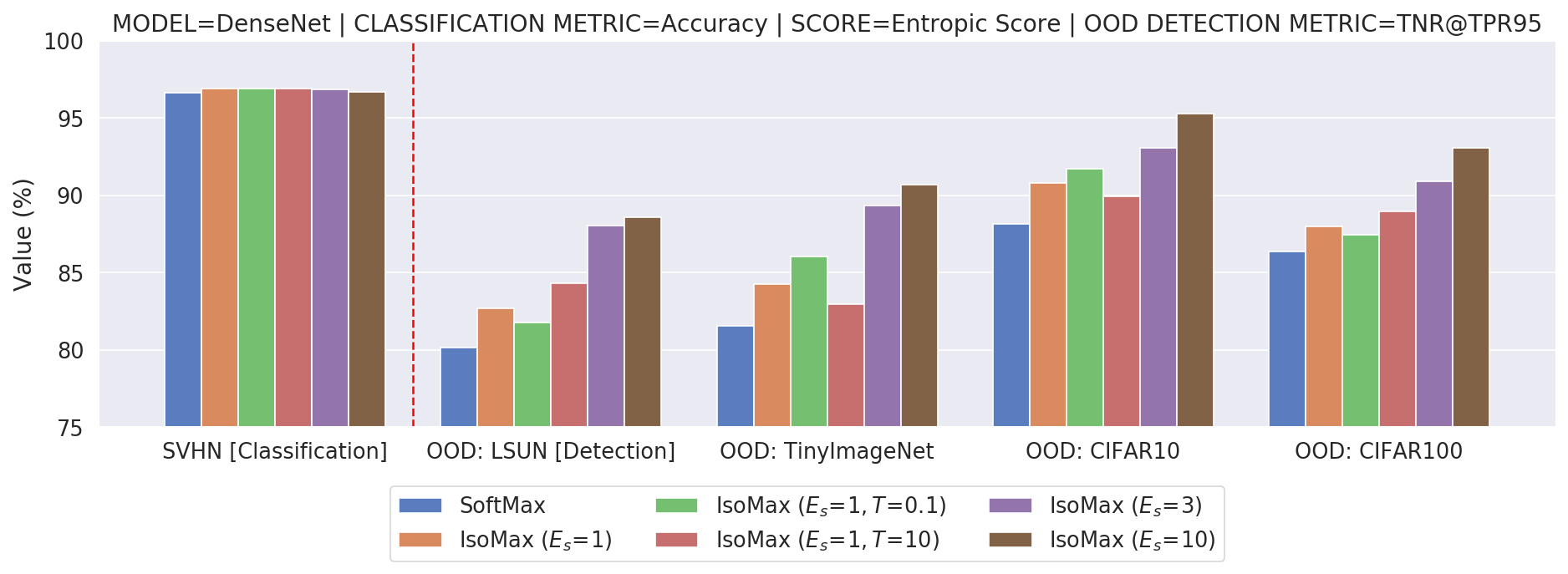}\label{fig:entropic_scale_parametrization_new}}
\begin{justify}
{Source: The Author (2022). (a) SoftMax loss simultaneously minimizes both the cross-entropy and the entropy of the posterior probabilities. (b)~IsoMax loss produces low entropy posterior probabilities for a low entropic scale \mbox{($E_s\!\!=\!1$)}. (c) IsoMax loss produces medium mean entropy for an intermediate entropic scale \mbox{($E_s\!\!=\!3$)}. (d) IsoMax loss minimizes the cross-entropy while producing high mean entropies for the high entropic scale \mbox{($E_s\!\!=\!10$)}. \emph{``The entropy maximization trick''} is the fundamental mechanism that allows us to migrate from low entropy posterior probability distributions (a,b) to high entropy posterior probability distributions (d). Higher entropic scale values correlate to higher mean entropies as recommended by the principle of maximum entropy. Regardless of training with a high entropic scale, if we do not remove it for inference (``the entropy maximization trick''), the IsoMax loss always produces posterior probability distributions with entropies as low as those generated by the SoftMax loss. An entropic scale equal to ten is sufficient to virtually produce the maximum possible entropy posterior probability distribution, as the highest possible value of the entropy is $\log(N)$, where $N$ is the number of classes. (e) The left side of the dashed vertical red line presents classification accuracies. The dashed vertical red line right side shows OOD detection performance using the entropic score and the TNR@TPR95 metric. Higher mean entropies produce increased OOD detection performance regardless of the out-distribution. We emphasize that OOD examples were never used during training and no validation was required to tune hyperparameters. Additionally, isotropy enables IsoMax loss to produce higher OOD performance than SoftMax loss, even for the unitary value of the entropic scale. Training using \mbox{$E_s\!\!=\!1$} and then making the temperature \mbox{$T\!\!=\!0.1$} or \mbox{$T\!\!=\!10$} during inference produces lower OOD detection performance than training using \mbox{$E_s\!\!=\!10$} and removing it for inference, which consists in our proposal. Finally, IsoMax loss presents similar classification accuracy compared with SoftMax regardless of the entropic scale. The entropic score, which is the negative entropy of the network output probabilities, was used as the score.}
\end{justify}
\label{fig:train_losses_entropies_and_entropic_scale_parametrization}
\end{figure*}

\begin{figure*}
\small
\centering
\caption[Accuracy and Out-of-Distribution Detection Study]{Accuracy and Out-of-Distribution Detection Study}
\includegraphics[width=0.925\textwidth]{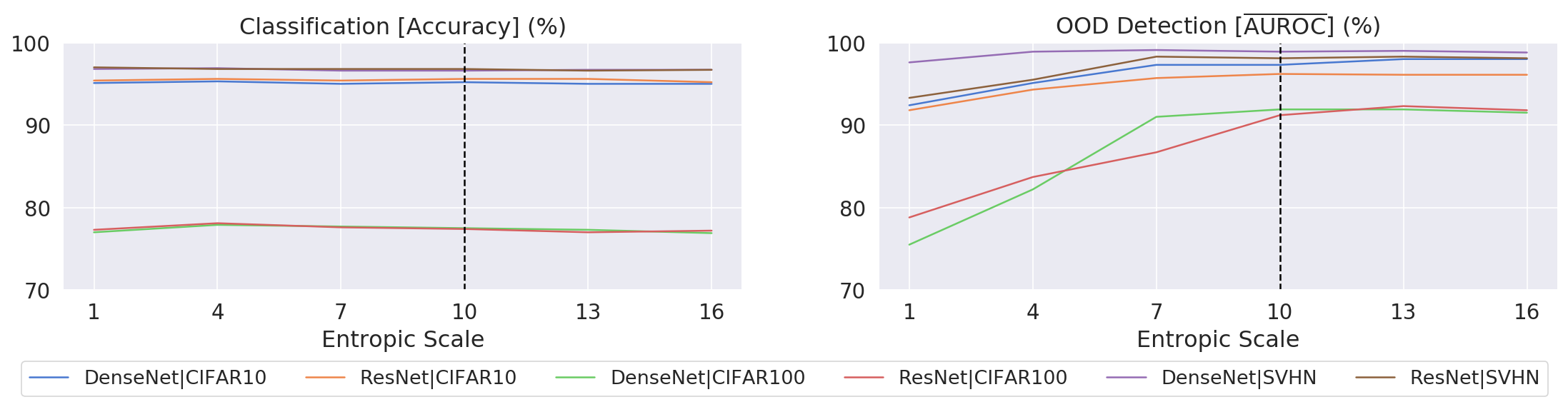}\label{fig:entropic_score_study}
\begin{justify}
{Source: The Author (2022). Study of classification accuracy and OOD detection performance dependence on the entropic scale. $\overline{\mathrm{AUROC}}$ represents the mean AUROC considering all out-of-distribution data. The classification accuracy and the mean OOD detection performance are approximately stable for \mbox{$E_s\!\!=\!10$} or higher regardless of the dataset and model. It also does not depend on the number of training classes N, which is probably explained by the fact that the entropic scale is inside an exponential function while the entropy increases only with the logarithm of the number of classes. Making $E_s$ learnable did not considerably improve or decrease the OOD detection results.}
\end{justify}
\label{fig:train_losses_entropies_and_entropic_scale_parametrization_new2}
\end{figure*}

To experimentally show that higher entropic scales lead to higher entropy posterior probability distributions and improved OOD detection performance, we trained DenseNets on SVHN using the SoftMax loss and IsoMax loss with distinct entropic scale values. We used the entropic score and the TNR@TPR95 to evaluate OOD detection performance (Fig.~\ref{fig:train_losses_entropies_and_entropic_scale_parametrization}).

As expected, Fig.~\ref{fig:train_losses_entropies_and_entropic_scale_parametrization}a shows that the SoftMax loss generates posterior distributions with extremely low entropy. Fig.~\ref{fig:train_losses_entropies_and_entropic_scale_parametrization}b illustrates that the unitary entropic scale \mbox{($E_s\!\!=\!1$)} does not increase the posterior distribution mean entropy. In other words, isotropy alone is not enough to circumvent the cross-entropy propensity to produce low entropy posterior probability distributions and the entropic scale is indeed necessary. Nevertheless, the replacement of anisotropic affine-based logits by isotropic distance-based logits is enough to produce initial OOD detection performance gains regardless of the out-distribution (Fig.~\ref{fig:train_losses_entropies_and_entropic_scale_parametrization}e).

\looseness=-1
Fig.~\ref{fig:train_losses_entropies_and_entropic_scale_parametrization}c shows that an intermediate entropic scale \mbox{($E_s\!\!=\!3$)} provides medium entropy probability distributions with corresponding additional OOD detection performance gains for all out-distributions (Fig.~\ref{fig:train_losses_entropies_and_entropic_scale_parametrization}e). Fig.~\ref{fig:train_losses_entropies_and_entropic_scale_parametrization}d illustrates that a high entropic scale \mbox{($E_s\!\!=\!10$)} produces even higher entropy probability distributions and the highest OOD detection performance regardless of the out-distribution considered (Fig.~\ref{fig:train_losses_entropies_and_entropic_scale_parametrization}e). Despite the entropy scale value used, the cross-entropy is minimized and the classification accuracies produced by SoftMax and IsoMax losses are extremely similar.

Hence, for a high entropic scale, the IsoMax loss can indeed minimize the cross-entropy while producing high entropy posterior probability distributions as recommended by the principle of maximum entropy. More importantly, higher entropy posterior probability distributions directly correlate with increased OOD detection performances despite the OOD data. An entropic scale \mbox{$E_s\!\!=\!10$} is enough to produce significantly high entropy probability distributions. Additionally, we experimentally observed no further gains for entropic scales higher than ten. 

\looseness=-1
Fig.~\ref{fig:train_losses_entropies_and_entropic_scale_parametrization_new2} shows that \emph{regardless of the combination of dataset and model}, the classification accuracy and the mean OOD detection performance are stable for \mbox{$E_s\!\!=\!10$} or higher, as the entropic scale is already high enough to ensure near-maximal entropy. We speculate that \mbox{$E_s\!\!=\!10$} worked satisfactorily regardless of the number of training classes because it is used inside an exponential function, while the entropy increases only with the logarithm of the number of training classes. Hence, an eventual validation of $E_s$ would produce an insignificant performance increase. Actually, this is not possible because we consider access to OOD or outlier samples forbidden. Moreover, making $E_s$ learnable did not significantly affect the OOD detection performance.


\looseness=-1
Considering that the entropic scale $E_{s}$ is present in Equation \eqref{eq:loss_isomax}, we are initially led to think that it needs to be tuned to achieve the highest possible OOD detection performance. However, we emphasize that the experiments showed that the dependence of the OOD detection performance on the entropic scale is remarkably well-behaved. Essentially, the OOD detection performance monotonically increases with the entropic scale until it reaches a saturation point near $E_{s}\!\!=\!10$ regardless of the dataset or the number of training classes (Fig.~\ref{fig:train_losses_entropies_and_entropic_scale_parametrization_new2}). It may be explained by the fact that the entropic scale is inside an exponential function and that we experimentally observed that $\exp{(-10\!\times\!d)}$ is enough to produce almost maximum entropy regardless of the dataset, model or number of training classes under consideration.

Finally, even if we consider that validating for entropic scales higher than ten would allow some minor OOD detection performance improvement, we usually cannot do this because we do not often have access to out-distribution data. Hence, the well-behaved dependence of the OOD detection performance on the entropic scale allowed us to define the $E_{s}$ as a \emph{constant scalar} equals to 10 rather than a hyperparameter that needs to be tuned for each novel dataset and model. It is the reason our approach may be used without requiring access to OOD/outlier data. Therefore, we kept the entropic scale as a constant (global value) equal to ten for all subsequent experiments.Hence, no validation of the entropic score was need for different models, datasets, or number of classes.


The entropic scale is constant regardless of the data and model considered. Hence, our approach does not require hyperparameter tuning or extra validation data. We have experimental evidence and theoretical insights to explain why this generalizes well. This constant value is used regardless of the model, in-distribution, and out-of-distribution.

The entropic scale is inside an exponential function, so the value of ten is enough to obtain almost the maximum entropy possible. Moreover, increasing this value further does not help, as the maximum possible entropy is already achieved. Therefore, we observe that gains in OOD detection performance increase fast in the beginning and saturate after some point. For example, consider the $d$ as a distance. Hence, $e^{-1d}$ and $e^{-3d}$ are not as small as $e^{-10d}$. However, $e^{-10d}$ and $e^{-20d}$ are too small numbers to make any difference. This behavior is monotonic in relation to the entropic scale because we deal with a monotonic function: the exponential. A substantial amount of experiments was performed to confirm this experimentally. This behavior is the same regardless of the dataset and model.

In machine learning and deep learning, we have many examples of constant global values (i.e., they do not need to be tuned) regardless of the data and model used. For example, in the ADAM \citep{DBLP:journals/corr/KingmaB14} optimizer, we all use $\beta_1$ equals 0.9, and $\beta_2$ equals 0.99. When using SGD with momentum, we commonly use the moment equals 0.9. These are experimentally recognized as values that work well across essentially all circumstances. Like those values, many experiments have demonstrated that the entropic scale equals ten works well.

Consequently, we do not need extra validation data for tuning it, as the same constant value is used regardless of the data and models used for training. Besides the experimental evidence, this value is inside an exponential function gives theoretical insights to justify this fact.

\newpage
\subsection{Implementation Detail: Separating Logarithms from Probabilities}

\hln{Table} \ref{tab:implementation_detail} \hln{shows that computing logarithms separated from probabilities is much more efficient in preserving the maximum normalized entropy present at the beginning of the training. Therefore, the implementations of all entropic losses calculate logarithms and probabilities as two separated and sequential operations.}

\begin{table*}
\renewcommand{\arraystretch}{1.5}
\footnotesize
\centering
\caption[\hln{Implementation Detail: Separating Logarithms from Probabilities}]{\hln{Implementation Detail: Separating Logarithms from Probabilities.}}
\label{tab:implementation_detail}
\begin{tabularx}{\textwidth}{lYYYY}
\toprule
Data & Loss Value & Mean Maximum Probability & Mean Normalized Entropy & Classification Accuracy (\%)\\
Partition & LogProb / Log+Prob & LogProb / Log+Prob & LogProb / Log+Prob & LogProb / Log+Prob\\
\midrule
Train & 1.697978 / 1.448191 & 0.308275 / 0.144888 & 0.796333 / 0.987172 & 36.25 / 48.34\\
Test & 1.379901 / 1.487199 & 0.479694 / 0.167632 & 0.628158 / 0.979525 & 49.06 / 53.30\\
\bottomrule
\end{tabularx}
\small
\begin{justify}
{Source: The Author (2022). \hln{IsoMax results for DenseNet trained one epoch on CIFAR10. Normalized entropy means the entropy divided by the maximum entropy. \mbox{LogProb} means using logarithms and probabilities combined into a single operation. \mbox{Log+Prob} means using logarithms and probabilities as separated and sequential operations, which is the option adopted in all entropic losses. We made the code deterministic for these experiments; therefore, many executions of the same experiment produce the same results. We kept everything else the same in both \mbox{LogProb} and \mbox{Log+Prob} cases.}}
\end{justify}
\end{table*}

\subsection{Classification Accuracy}

\begin{figure*}
\small
\centering
\caption[Classification Accuracy: SoftMax vs. IsoMax]{Classification Accuracy: SoftMax vs. IsoMax}
\subfloat[]{\includegraphics[width=\textwidth]{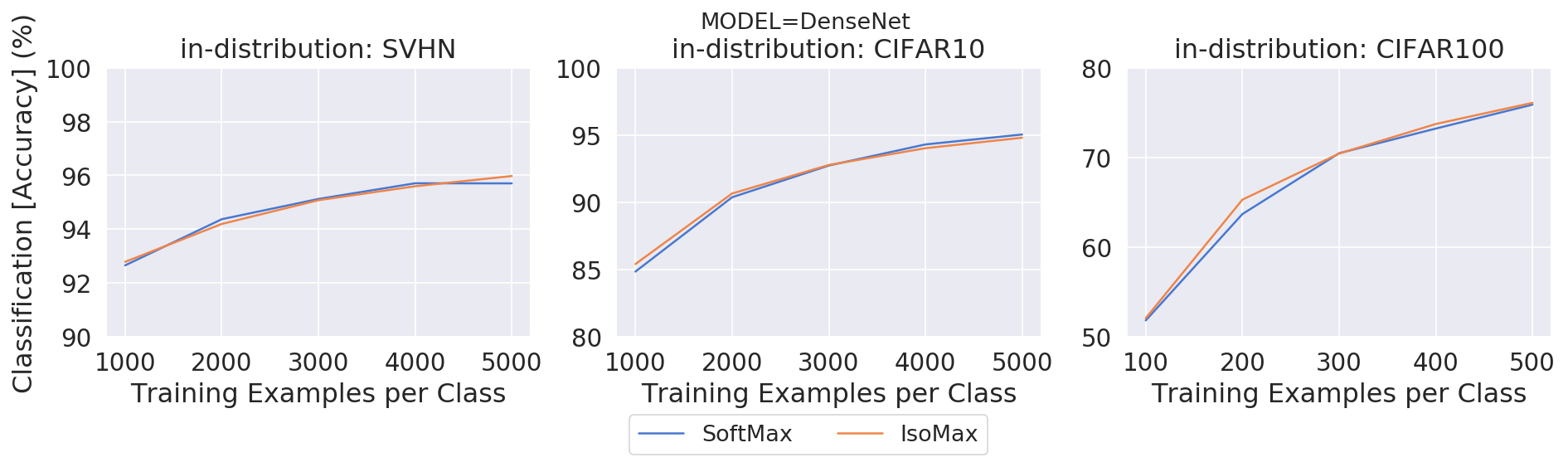}} 
\\
\subfloat[]{\includegraphics[width=\textwidth]{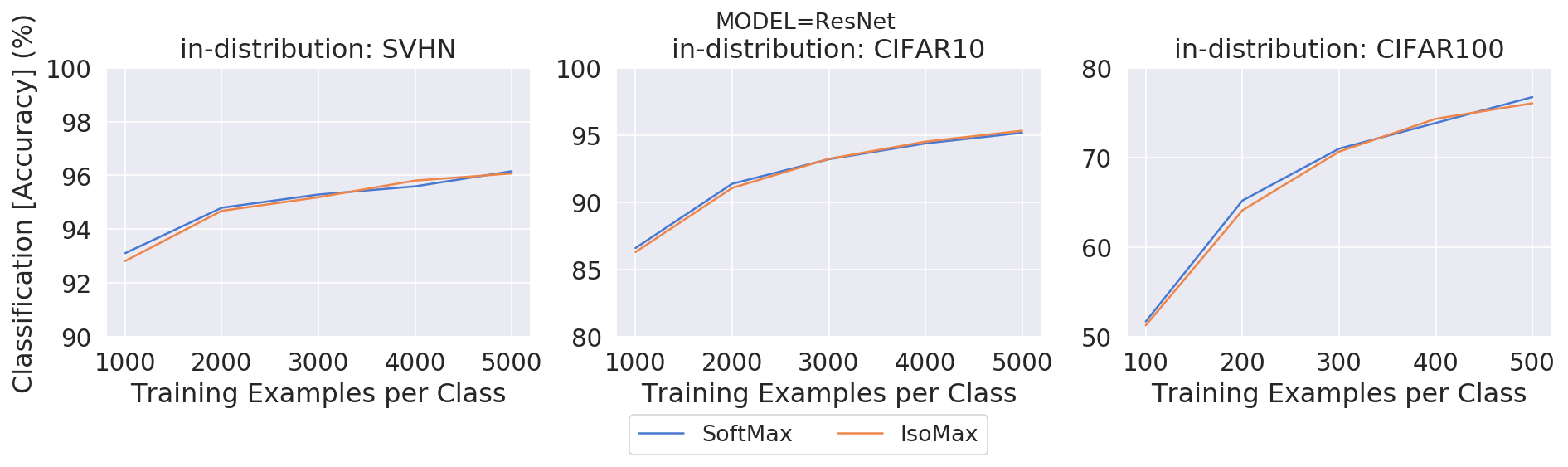}}
\begin{justify}
{Source: The Author (2022). IsoMax loss presents test accuracy similar to SoftMax loss (no classification accuracy drop) for different numbers of training examples per class on several datasets and models. Simultaneously, IsoMax usually produces much higher OOD detection performance (Fig.~\ref{fig:detection_robstness}).}
\end{justify}
\label{fig:classification_robstness}
\end{figure*}

Table \ref{tbl:addon_odd} shows that IsoMax loss consistently produces classification accuracy similar to SoftMax loss, regardless of being used as a baseline approach or combined with additional techniques to improve OOD detection performance. Fig.~\ref{fig:classification_robstness} shows that this is also true even for a varying number of training examples per class.

\clearpage\subsection{Out-of-Distribution Detection}

The IsoMax loss enhanced versions almost always outperform the OOD detection performance of the corresponding SoftMax loss enhanced version when both are using the same add-on technique. The major drawback of adding label smoothing or center loss regularization is the need to tune the hyperparameters presented by these add-on techniques. Additionally, different hyperparameters values need to be validated for each pair of in-distribution and out-distribution. As mentioned before, it is highly optimistic to assume access to OOD data, as we usually do not know what OOD data the solution we will face in the field.

\looseness=-1
Even considering this best-case scenario, SoftMax loss combined with label smoothing or center loss regularization always presented significantly lower OOD detection performance than IsoMax loss without using them and, consequently, avoiding unrealistically optimist access to OOD data and the mentioned validations. Enhancing SoftMax loss or IsoMax loss with ODIN presents the same problems from a practical perspective. In CIFAR100, SoftMax loss with outlier exposure produces lower performance than IsoMax loss without it for both DenseNet and ResNet.

\begin{table*}
\setlength{\tabcolsep}{4pt}
\footnotesize
\centering
\caption[IsoMax: Classification Accuracy and Enhanced Versions]{IsoMax: Classification Accuracy and Enhanced Versions}
\label{tbl:addon_ood}
\begin{tabularx}{\textwidth}{lll|YYY}
\toprule
\multirow{4}{*}{\begin{tabular}[c]{@{}c@{}}\\Model\end{tabular}} & \multirow{4}{*}{\begin{tabular}[c]{@{}c@{}}\\In-Data\\(training)\end{tabular}} & \multirow{4}{*}{\begin{tabular}[c]{@{}c@{}}\\OOD Detection\\Approach\end{tabular}} & 
\multicolumn{3}{c}{Enhanced Versions of SoftMax loss and IsoMax loss.}\\
&&& \multicolumn{3}{c}{Add-on Techniques produce different Side Effects and Requirements.}\\
\cmidrule{4-6}
&&& Class. Accuracy (\%) [$\uparrow$] & TNR@TPR95 (\%) [$\uparrow$] & AUROC (\%) [$\uparrow$]\\
&&& SoftMax / IsoMax & SoftMax / IsoMax & SoftMax / IsoMax\\
\midrule
\multirow{15}{*}{\begin{tabular}[c]{@{}c@{}}DenseNet\end{tabular}}
& \multirow{5}{*}{\begin{tabular}[c]{@{}c@{}}CIFAR10\end{tabular}} 
& Baseline & \bf{95.4$\pm$0.3} / \bf{95.2$\pm$0.3} & 53.3$\pm$0.4 / \bf{84.1$\pm$0.3} & 91.9$\pm$0.4 / \bf{97.3$\pm$0.3}\\
&& + Label Smoothing & \bf{95.2$\pm$0.4} / \bf{95.0$\pm$0.3} & \textbf{70.7$\pm$0.4} / 60.3$\pm$0.3 & \textbf{94.9$\pm$0.3} / 80.8$\pm$0.3\\
&& + Center Loss Regularization & \bf{95.3$\pm$0.3} / \bf{95.1$\pm$0.4} & 54.7$\pm$0.6 / \bf{87.1$\pm$0.3} & 92.8$\pm$0.3 / \bf{97.6$\pm$0.3}\\
&& + ODIN & \bf{95.4$\pm$0.3} / \bf{95.2$\pm$0.3} & 91.9$\pm$0.3 / {\color{blue}\bf{95.3$\pm$0.4}} & \bf{98.2$\pm$0.3} / \bf{98.4$\pm$0.4}\\
&& + Outlier Exposure & \bf{95.3$\pm$0.4} / \bf{95.6$\pm$0.4} & {\color{blue}93.8$\pm$0.3} / \bf{94.7$\pm$0.3} & {\color{blue}\bf{98.5$\pm$0.3}} / {\color{blue}\bf{98.8$\pm$0.4}}\\
\cmidrule{2-6} 
& \multirow{5}{*}{\begin{tabular}[c]{@{}c@{}}CIFAR100\end{tabular}} 
& Baseline & \bf{77.5$\pm$0.6} / \bf{77.5$\pm$0.4} & 22.3$\pm$0.7 / \bf{45.1$\pm$0.6} & 77.4$\pm$0.8 / \bf{91.9$\pm$0.6}\\
&& + Label Smoothing & \bf{77.0$\pm$0.4} / \bf{77.2$\pm$0.6} & 31.5$\pm$0.6 / \bf{33.0$\pm$0.6} & 82.0$\pm$0.6 / \bf{88.3$\pm$0.7}\\
&& + Center Loss Regularization & \bf{77.2$\pm$0.7} / \bf{77.0$\pm$0.6} & 30.3$\pm$0.7 / \bf{43.7$\pm$0.6} & 79.5$\pm$0.8 / \bf{92.2$\pm$0.6}\\
&& + ODIN & \bf{77.5$\pm$0.6} / \bf{77.5$\pm$0.4} & {\color{blue}64.4$\pm$0.7} / {\color{blue}\bf{83.1$\pm$0.8}} & {\color{blue}92.5$\pm$0.6} / {\color{blue}\bf{96.9$\pm$0.7}}\\
&& + Outlier Exposure & \bf{77.8$\pm$0.6} / \bf{77.5$\pm$0.7} & 23.0$\pm$0.7 / \bf{36.4$\pm$0.8} & 80.5$\pm$0.8 / \bf{89.6$\pm$0.6}\\
\cmidrule{2-6} 
& \multirow{5}{*}{\begin{tabular}[c]{@{}c@{}}SVHN\end{tabular}} 
& Baseline & \bf{96.6$\pm$0.2} / \bf{96.6$\pm$0.2} & 90.1$\pm$0.2 / \bf{95.9$\pm$0.1} & 98.2$\pm$0.2 / \bf{98.9$\pm$0.2}\\
&& + Label Smoothing & \bf{96.6$\pm$0.2} / \bf{96.7$\pm$0.3} & 87.5$\pm$0.2 / \bf{93.5$\pm$0.2} & 97.0$\pm$0.2 / \bf{97.8$\pm$0.3}\\
&& + Center Loss Regularization & \bf{96.7$\pm$0.3} / \bf{96.6$\pm$0.2} & 88.0$\pm$0.3 / \bf{95.9$\pm$0.2} & 97.9$\pm$0.2 / \bf{98.9$\pm$0.1}\\
&& + ODIN & \bf{96.6$\pm$0.2} / \bf{96.6$\pm$0.2} & 95.5$\pm$0.2 / \bf{96.7$\pm$0.2} & 98.8$\pm$0.1 / \bf{99.1$\pm$0.1}\\
&& + Outlier Exposure & \bf{96.6$\pm$0.3} / \bf{96.7$\pm$0.3} & {\color{blue}\bf{99.9$\pm$0.1}} / {\color{blue}\bf{99.9$\pm$0.1}} & {\color{blue}\bf{99.9$\pm$0.1}} / {\color{blue}\bf{99.9$\pm$0.1}}\\
\midrule
\multirow{15}{*}{\begin{tabular}[c]{@{}c@{}}ResNet\end{tabular}}
& \multirow{5}{*}{\begin{tabular}[c]{@{}c@{}}CIFAR10\end{tabular}} 
& Baseline & \bf{95.4$\pm$0.3} / \bf{95.6$\pm$0.4} & 50.8$\pm$0.4 / \bf{78.6$\pm$0.3} & 91.2$\pm$0.3 / \bf{96.1$\pm$0.4}\\
&& + Label Smoothing & \bf{95.5$\pm$0.4} / \bf{95.4$\pm$0.3} & 54.0$\pm$0.4 / \bf{63.5$\pm$0.3} & 78.9$\pm$0.4 / \bf{84.5$\pm$0.3}\\
&& + Center Loss Regularization & \bf{95.6$\pm$0.3} / \bf{95.4$\pm$0.4} & 53.5$\pm$0.4 / \bf{80.6$\pm$0.3} & 90.5$\pm$0.3 / \bf{96.6$\pm$0.2}\\
&& + ODIN & \bf{95.4$\pm$0.3} / \bf{95.6$\pm$0.4} & 73.7$\pm$0.3 / \bf{85.6$\pm$0.3} & 93.4$\pm$0.2 / \bf{97.2$\pm$0.3}\\
&& + Outlier Exposure & \bf{95.5$\pm$0.3} / \bf{95.6$\pm$0.3} & {\color{blue}91.1$\pm$0.3} / {\color{blue}\bf{94.2$\pm$0.4}} & {\color{blue}97.7$\pm$0.2} / {\color{blue}\bf{98.6$\pm$0.3}}\\
\cmidrule{2-6} 
& \multirow{5}{*}{\begin{tabular}[c]{@{}c@{}}CIFAR100\end{tabular}} 
& Baseline & \bf{77.3$\pm$0.6} / \bf{77.4$\pm$0.7} & 22.4$\pm$0.6 / \bf{41.8$\pm$0.7} & 80.5$\pm$0.6 / \bf{90.1$\pm$0.7}\\
&& + Label Smoothing & \bf{77.7$\pm$0.5} / \bf{77.3$\pm$0.5} & 21.0$\pm$0.7 / \bf{33.4$\pm$0.6} & 81.6$\pm$0.6 / \bf{85.9$\pm$0.6}\\
&& + Center Loss Regularization & \bf{77.9$\pm$0.5} / \bf{77.4$\pm$0.5} & 29.5$\pm$0.9 / \bf{48.2$\pm$0.7} & 80.3$\pm$0.6 / \bf{90.6$\pm$0.6}\\
&& + ODIN & \bf{77.3$\pm$0.6} / \bf{77.4$\pm$0.7} & {\color{blue}64.0$\pm$0.6} / {\color{blue}\bf{77.9$\pm$0.6}} & {\color{blue}92.7$\pm$0.7} / {\color{blue}\bf{95.8$\pm$0.6}}\\
&& + Outlier Exposure & \bf{77.3$\pm$0.5} / \bf{77.0$\pm$0.5} & 37.8$\pm$0.9 / \bf{41.7$\pm$0.8} & 86.6$\pm$0.6 / \bf{88.6$\pm$0.7}\\
\cmidrule{2-6} 
& \multirow{5}{*}{\begin{tabular}[c]{@{}c@{}}SVHN\end{tabular}} 
& Baseline & \bf{96.8$\pm$0.2} / \bf{96.7$\pm$0.2} & 71.7$\pm$0.3 / \bf{92.4$\pm$0.2} & 94.7$\pm$0.3 / \bf{98.0$\pm$0.2}\\
&& + Label Smoothing & \bf{96.9$\pm$0.2} / \bf{96.9$\pm$0.3} & \bf{86.3$\pm$0.2} / \bf{86.4$\pm$0.2} & \textbf{97.2$\pm$0.3} / 94.9$\pm$0.2\\
&& + Center Loss Regularization & \bf{96.9$\pm$0.2} / \bf{96.7$\pm$0.2} & 77.2$\pm$0.2 / \bf{82.2$\pm$0.2} & 94.5$\pm$0.2 / \bf{96.1$\pm$0.3}\\
&& + ODIN & \bf{96.8$\pm$0.2} / \bf{96.7$\pm$0.2} & 79.9$\pm$0.3 / \bf{93.5$\pm$0.3} & 95.1$\pm$0.2 / \bf{98.2$\pm$0.3}\\
&& + Outlier Exposure & \bf{96.9$\pm$0.3} / \bf{96.8$\pm$0.2} & {\color{blue}\bf{99.9$\pm$0.1}} / {\color{blue}\bf{99.9$\pm$0.1}} & {\color{blue}\bf{99.9$\pm$0.1}} / {\color{blue}\bf{99.9$\pm$0.1}}\\
\bottomrule
\end{tabularx}
\small
\begin{justify}
{Source: The Author (2022). Comparison of Enhanced Versions of SoftMax Loss and IsoMax Loss: Adding label smoothing, center loss regularization, ODIN, and outlier exposure to SoftMax loss and IsoMax loss. The side effects and requirements added to the solution depend on the add-on technique. Regardless of using SoftMax loss or IsoMax loss, adding label smoothing (LS) requires validation of the hyperparameter $\epsilon$ \citep{DBLP:conf/cvpr/SzegedyVISW16}. The values searched for $\epsilon$ were 0.1 \citep{DBLP:conf/cvpr/SzegedyVISW16, DBLP:journals/corr/abs-2007-03212} and 0.01 \citep{DBLP:journals/corr/abs-2007-03212}. Adding center loss regularization (CLR) to SoftMax loss or IsoMax loss requires validation of the hyperparameter $\lambda$ \citep{DBLP:conf/eccv/WenZL016}. The values searched for $\lambda$ were 0.01 and 0.003 \citep{DBLP:conf/eccv/WenZL016}. We emphasize that the center loss is not used as a stand-alone loss, but rather combined as a regularization term with a preexisting baseline loss. In the case of the original paper, the baseline loss was the SoftMax loss. In this paper, we added the mentioned regularization term \citep[Equation (5)]{DBLP:conf/eccv/WenZL016} also to IsoMax loss to construct the center loss enhanced version of our loss. Regardless of using SoftMax loss or IsoMax loss, adding ODIN \citep{liang2018enhancing} requires validation of the hyperparameters $\epsilon$ and $T$. The values searched for these hyperparameters were the same used in the original paper \citep{liang2018enhancing}. Adding ODIN implies using input preprocessing, which makes inferences much slower, and energy- and cost-inefficient. We used OOD data to validate the LS, CLR, and ODIN hyperparameters. Adding outlier exposure (OE) \citep{hendrycks2018deep} to SoftMax loss or IsoMax loss requires collecting outlier data. We used the same outlier data used in \cite{hendrycks2018deep}. The add-on techniques were not applied to the SoftMax and IsoMax losses combined, but rather individually. The values of the performance metrics TNR@TPR95 and AUROC were averaged over all out-of-distribution data. All results used the entropic score, as it always overcame the maximum probability score. The results represent the mean and standard deviation of five executions. The best values are bold when they overcome the competing approach value outside the margin of error given by the standard deviations. For OOD detection, the variants of SoftMax and IsoMax losses that presented the best performance for each combination of architecture and dataset are blue.}
\end{justify}
\label{tbl:addon_odd}
\end{table*}

\begin{table*}
\setlength{\tabcolsep}{2pt}
\renewcommand{\arraystretch}{0.5}
\footnotesize
\centering
\caption[IsoMax: OOD Detection Performance]{IsoMax: OOD Detection Performance}
\vspace{-0.05in}
\begin{tabularx}{\textwidth}{lll|YYY}
\toprule
\multirow{4}{*}{\begin{tabular}[c]{@{}c@{}}\\Model\end{tabular}} & \multirow{4}{*}{\begin{tabular}[c]{@{}c@{}}\\In-Data\\(training)\end{tabular}} & \multirow{4}{*}{\begin{tabular}[c]{@{}c@{}}\\Out-Data\\(unseen)\end{tabular}} & 
\multicolumn{3}{c}{Baseline Out-of-Distribution Detection Approaches Comparison.}\\
&&& \multicolumn{3}{c}{Fast and Energy-Efficient Inferences. No outlier data used.}\\
\cmidrule{4-6}
&&& TNR@TPR95 (\%) [$\uparrow$] & AUROC (\%) [$\uparrow$] & DTACC (\%) [$\uparrow$]\\
&&& \multicolumn{3}{c}{SoftMax+MPS / SoftMax+ES / IsoMax+ES (ours)}\\
\midrule
\multirow{9}{*}{\begin{tabular}[c]{@{}c@{}}DenseNet~~\end{tabular}}
& \multirow{3}{*}{\begin{tabular}[c]{@{}c@{}}CIFAR10~~\end{tabular}} 
& SVHN & 32.1$\pm$0.3 / 33.1$\pm$0.4 / \bf77.1$\pm$0.3 & 86.5$\pm$0.4 / 86.8$\pm$0.3 / \bf96.7$\pm$0.4 & 79.8$\pm$0.3 / 79.8$\pm$0.3 / \bf91.8$\pm$0.3\\
&& TinyImageNet~~& 55.7$\pm$0.3 / 59.7$\pm$0.4 / \bf88.1$\pm$0.4 & 93.5$\pm$0.4 / 94.1$\pm$0.4 / \bf97.9$\pm$0.3 & 87.5$\pm$0.2 / 87.9$\pm$0.3 / \bf93.3$\pm$0.3\\
&& LSUN & 64.8$\pm$0.4 / 69.6$\pm$0.3 / \bf94.6$\pm$0.2 & 95.1$\pm$0.3 / 95.8$\pm$0.2 / \bf98.9$\pm$0.3 & 89.8$\pm$0.3 / 90.1$\pm$0.3 / \bf95.0$\pm$0.4\\
\cmidrule{2-6} 
& \multirow{3}{*}{\begin{tabular}[c]{@{}c@{}}CIFAR100\end{tabular}} 
& SVHN & 20.5$\pm$0.6 / {\bf24.8$\pm$0.8} / {\bf23.6$\pm$0.9} & 80.1$\pm$0.7 / 81.8$\pm$0.7 / \bf88.8$\pm$0.6 & 73.8$\pm$0.6 / 74.4$\pm$0.7 / \bf83.9$\pm$0.6\\
&& TinyImageNet & 19.3$\pm$0.9 / 23.8$\pm$0.8 / \bf49.0$\pm$0.6 & 77.1$\pm$0.6 / 78.7$\pm$0.7 / \bf92.8$\pm$0.6 & 70.5$\pm$0.6 / 71.3$\pm$0.7 / \bf86.5$\pm$0.6\\
&& LSUN & 18.7$\pm$0.6 / 24.3$\pm$0.8 / \bf63.1$\pm$0.6 & 75.8$\pm$0.7 / 77.8$\pm$0.6 / \bf94.8$\pm$0.6 & 69.4$\pm$0.8 / 70.3$\pm$0.9 / \bf89.2$\pm$0.6\\
\cmidrule{2-6} 
& \multirow{3}{*}{\begin{tabular}[c]{@{}c@{}}SVHN\end{tabular}} 
& CIFAR10 & 81.5$\pm$0.2 / 83.7$\pm$0.3 / \bf94.1$\pm$0.2 & 96.5$\pm$0.3 / 96.9$\pm$0.2 / \bf98.5$\pm$0.2 & 91.9$\pm$0.3 / 92.1$\pm$0.2 / \bf95.0$\pm$0.2\\
&& TinyImageNet & 88.3$\pm$0.2 / 90.1$\pm$0.2 / \bf97.2$\pm$0.3 & 97.7$\pm$0.3 / 98.1$\pm$0.2 / \bf99.1$\pm$0.2 & 93.4$\pm$0.4 / 93.8$\pm$0.2 / \bf96.3$\pm$0.3\\
&& LSUN & 86.4$\pm$0.2 / 88.4$\pm$0.4 / \bf96.8$\pm$0.2 & 97.3$\pm$0.2 / 97.8$\pm$0.3 / \bf99.1$\pm$0.2 & 92.8$\pm$0.2 / 93.0$\pm$0.4 / \bf96.0$\pm$0.2\\
\midrule
\multirow{9}{*}{\begin{tabular}[c]{@{}c@{}}ResNet\end{tabular}}
& \multirow{3}{*}{\begin{tabular}[c]{@{}c@{}}CIFAR10\end{tabular}} 
& SVHN & 43.2$\pm$0.4 / 44.5$\pm$0.3 / \bf83.6$\pm$0.4 & 91.6$\pm$0.3 / 92.0$\pm$0.3 / \bf97.1$\pm$0.4 & 86.5$\pm$0.2 / 86.4$\pm$0.4 / \bf91.9$\pm$0.3\\
&& TinyImageNet & 46.4$\pm$0.4 / 48.0$\pm$0.3 / \bf70.3$\pm$0.3 & 89.8$\pm$0.4 / 90.1$\pm$0.2 / \bf94.6$\pm$0.4 & 84.0$\pm$0.4 / 84.2$\pm$0.3 / \bf88.3$\pm$0.4\\
&& LSUN & 51.2$\pm$0.4 / 53.3$\pm$0.2 / \bf82.3$\pm$0.3 & 92.2$\pm$0.4 / 92.6$\pm$0.4 / \bf96.9$\pm$0.3 & 86.5$\pm$0.2 / 86.6$\pm$0.4 / \bf91.5$\pm$0.3\\
\cmidrule{2-6} 
& \multirow{3}{*}{\begin{tabular}[c]{@{}c@{}}CIFAR100\end{tabular}} 
& SVHN & 15.9$\pm$0.8 / 18.0$\pm$0.7 / \bf20.2$\pm$0.6 & 71.3$\pm$0.6 / 72.7$\pm$0.7 / \bf85.3$\pm$0.6 & 66.1$\pm$0.6 / 66.3$\pm$0.7 / \bf79.7$\pm$0.6\\
&& TinyImageNet & 18.5$\pm$0.8 / 22.4$\pm$0.6 / \bf50.6$\pm$0.7 & 74.7$\pm$0.6 / 76.3$\pm$0.7 / \bf92.0$\pm$0.7 & 68.8$\pm$0.6 / 69.1$\pm$0.6 / \bf85.6$\pm$0.7\\
&& LSUN & 18.3$\pm$0.8 / 22.4$\pm$0.5 / \bf54.9$\pm$0.6 & 74.7$\pm$0.6 / 76.5$\pm$0.7 / \bf93.3$\pm$0.6 & 69.1$\pm$0.5 / 69.4$\pm$0.7 / \bf87.6$\pm$0.8\\
\cmidrule{2-6} 
& \multirow{3}{*}{\begin{tabular}[c]{@{}c@{}}SVHN\end{tabular}} 
& CIFAR10 & 67.3$\pm$0.2 / 67.7$\pm$0.3 / \bf92.3$\pm$0.2 & 89.8$\pm$0.2 / 89.7$\pm$0.3 / \bf98.0$\pm$0.2 & 87.0$\pm$0.3 / 86.9$\pm$0.3 / \bf94.1$\pm$0.2\\
&& TinyImageNet & 66.8$\pm$0.3 / 67.3$\pm$0.2 / \bf94.6$\pm$0.2 & 89.0$\pm$0.3 / 89.0$\pm$0.2 / \bf98.3$\pm$0.2 & 86.8$\pm$0.2 / 86.6$\pm$0.4 / \bf94.8$\pm$0.4\\
&& LSUN & 62.1$\pm$0.2 / 62.5$\pm$0.3 / \bf90.9$\pm$0.4 & 86.0$\pm$0.2 / 85.8$\pm$0.2 / \bf97.8$\pm$0.2 & 84.2$\pm$0.2 / 84.1$\pm$0.3 / \bf93.6$\pm$0.4\\
\bottomrule
\end{tabularx}
\vspace{-0.05in}
\small
\begin{justify}
{Source: The Author (2022). OOD Detection: Fast and Energy-Efficient Inferences. No extra/outlier/background data used. The approaches do not require outlier/background/extra data. No additional procedures other than typical straightforward neural network training is required, and no classification accuracy drop is observed. All approaches present fast and energy-efficient inferences. Neither adversarial training, input preprocessing, temperature calibration, feature ensemble, nor metric learning is used. Since there is no need to tune hyperparameters, no access to OOD or adversarial examples is required. SoftMax+MPS means training with SoftMax loss and performing OOD detection using the maximum probability score, which is the approach defined in \cite{hendrycks2017baseline}. SoftMax+ES means training with SoftMax loss and performing OOD detection using the entropic score. IsoMax+ES means training with IsoMax loss and performing OOD detection using the entropic score. The results represent the mean and standard deviation of five executions. The best values are bold when they overcome the competing approach value outside the margin of error given by the standard deviations.}
\end{justify}
\label{tbl:expanded_fair_odd_new}
\end{table*}

\begin{table*}
\footnotesize
\caption[IsoMax: OOD Detection on TinyImageNet]{IsoMax: OOD Detection on TinyImageNet}\label{tbl:tinyimagenet}
\centering
\begin{tabularx}{\textwidth}{YYl|YY}
\toprule
\multirow{4}{*}{\begin{tabular}[c]{@{}c@{}}\\Model\end{tabular}} & \multirow{4}{*}{\begin{tabular}[c]{@{}c@{}}\\Accuracy (\%) [$\uparrow$]\\SoftMax / IsoMax\end{tabular}} & \multirow{4}{*}{\begin{tabular}[c]{@{}c@{}}\\Out-Data\\(unseen)\end{tabular}} & 
\multicolumn{2}{c}{Baseline Out-of-Distribution Detection Approaches Comparison.}\\
&&& \multicolumn{2}{c}{Fast and Energy-Efficient Inferences. No outlier data used.}\\
\cmidrule{4-5}
&&& TNR@TPR95 (\%) [$\uparrow$] & AUROC (\%) [$\uparrow$]\\
&&& \multicolumn{2}{c}{SoftMax+MPS / IsoMax+ES (ours)}\\
\midrule
\multirow{3}{*}{\begin{tabular}[c]{@{}c@{}}\shortstack{DenseNetBC100\\(small size)}\end{tabular}}
& \multirow{3}{*}{\begin{tabular}[c]{@{}c@{}}61.1$\pm$0.2 / \bf61.6$\pm$0.3\end{tabular}} 
& CIFAR10 & 24.3$\pm$3.3 / \bf38.9$\pm$3.7 & 81.1$\pm$1.3 / \bf88.5$\pm$2.2\\
&& CIFAR100 & 23.8$\pm$2.4 / \bf36.3$\pm$3.1 & 79.6$\pm$0.9 / \bf84.2$\pm$1.2\\
&& SVHN & 40.3$\pm$3.5 / \bf64.4$\pm$2.7 & 84.9$\pm$2.0 / \bf94.3$\pm$1.4\\
\midrule
\multirow{3}{*}{\begin{tabular}[c]{@{}c@{}}\shortstack{ResNet34\\(medium size)}\end{tabular}}
& \multirow{3}{*}{\begin{tabular}[c]{@{}c@{}}\bf65.6$\pm$0.3 / \bf65.4$\pm$0.3\end{tabular}} 
& CIFAR10 & 18.2$\pm$2.6 / \bf51.9$\pm$4.9 & 78.3$\pm$1.2 / \bf90.9$\pm$1.0\\
&& CIFAR100 & 16.7$\pm$2.4 / \bf50.5$\pm$4.7 & 76.3$\pm$1.4 / \bf88.2$\pm$1.2\\
&& SVHN & 19.3$\pm$5.2 / \bf89.1$\pm$5.4 & 73.8$\pm$1.7 / \bf98.0$\pm$0.8\\
\midrule
\multirow{3}{*}{\begin{tabular}[c]{@{}c@{}}\shortstack{WideResNet2810\\(big size)}\end{tabular}}
& \multirow{3}{*}{\begin{tabular}[c]{@{}c@{}}67.0$\pm$0.2 / \bf67.5$\pm$0.2\end{tabular}} 
& CIFAR10 & 30.9$\pm$5.8 / \bf56.0$\pm$4.3 & 84.1$\pm$2.4 / \bf91.4$\pm$1.9\\
&& CIFAR100 & 31.5$\pm$1.8 / \bf54.1$\pm$3.8 & 83.3$\pm$1.3 / \bf88.2$\pm$0.9\\
&& SVHN & 59.7$\pm$2.3 / \bf89.0$\pm$4.4 & 90.0$\pm$2.0 / \bf98.0$\pm$1.4\\
\bottomrule
\end{tabularx}
\small
\begin{justify} 
{The Author (2022). SoftMax+MPS means training with SoftMax loss and performing OOD detection using the maximum probability score \citep{hendrycks2017baseline}. IsoMax+ES means training with IsoMax loss and performing OOD detection using the entropic score. The results represent the mean and standard deviation of five executions. The best values are bold when they overcome the competing approach value outside the margin of error given by the standard deviations.}
\end{justify}
\end{table*}

\begin{table*}
\footnotesize
\caption[IsoMax: OOD Detection on Text Data]{IsoMax: OOD Detection on Text Data}\label{tbl:text_odd}
\centering
\begin{tabularx}{\textwidth}{lll|YY}
\toprule
\multirow{4}{*}{\begin{tabular}[c]{@{}c@{}}\\Model\end{tabular}} & \multirow{4}{*}{\begin{tabular}[c]{@{}c@{}}\\In-Data\\(training)\end{tabular}} & \multirow{4}{*}{\begin{tabular}[c]{@{}c@{}}\\Out-Data\\(unseen)\end{tabular}} & 
\multicolumn{2}{c}{Baseline Out-of-Distribution Detection Approaches Comparison.}\\
&&& \multicolumn{2}{c}{Fast and Energy-Efficient Inferences. No outlier data used.}\\
\cmidrule{4-5}
&&& TNR@TPR95 (\%) [$\uparrow$] & AUROC (\%) [$\uparrow$]\\
&&& \multicolumn{2}{c}{SoftMax+MPS / IsoMax+ES (ours)}\\
\midrule
\multirow{6}{*}{\begin{tabular}[c]{@{}c@{}}GRU2L~~~~\end{tabular}}
& \multirow{3}{*}{\begin{tabular}[c]{@{}c@{}}20 Newsgroups~~~~\end{tabular}} 
& IMDB & 31.9$\pm$6.9 / \bf58.0$\pm$12.6 & 86.8$\pm$2.0 / \bf92.2$\pm$2.0\\
&& Multi30K & 29.7$\pm$3.1 / \bf47.9$\pm$13.3 & 82.4$\pm$1.8 / \bf85.7$\pm$2.8\\
&& Yelp Reviews & 26.7$\pm$3.8 / \bf46.6$\pm$5.7 & 84.0$\pm$1.5 / \bf88.7$\pm$0.9\\
\cmidrule{2-5} 
& \multirow{3}{*}{\begin{tabular}[c]{@{}c@{}}TREC~~~~\end{tabular}} 
& IMDB & 35.8$\pm$8.4 / \bf54.3$\pm$5.3 & 88.9$\pm$1.8 / \bf93.2$\pm$1.2\\
&& Multi30K & 15.4$\pm$5.4 / \bf37.4$\pm$4.1 & 74.8$\pm$3.8 / \bf82.0$\pm$2.3\\
&& Yelp Reviews & 34.4$\pm$5.3 / \bf52.3$\pm$6.2 & 82.5$\pm$0.4 / \bf90.2$\pm$1.2\\
\bottomrule
\end{tabularx}
\small
\begin{justify} 
Source: The Author (2022). SoftMax+MPS means training with SoftMax loss and performing OOD detection using the maximum probability score \citep{hendrycks2017baseline}. IsoMax+ES means training with IsoMax loss and performing OOD detection using the entropic score. The results represent the mean and standard deviation of five executions. The best values are bold when they overcome the competing approach value outside the margin of error given by the standard deviations.
\end{justify}
\end{table*}

\begin{table*}
\setlength{\tabcolsep}{2pt}
\renewcommand{\arraystretch}{0.5}
\caption[IsoMax: Comparison with No Seamless Solutions]{IsoMax: Comparison with No Seamless Solutions}
\footnotesize
\centering
\begin{tabularx}{\textwidth}{lll|YY}
\toprule
\multirow{4}{*}{\begin{tabular}[c]{@{}c@{}}\\Model\end{tabular}} & \multirow{4}{*}{\begin{tabular}[c]{@{}c@{}}\\In-Data\\(training)\end{tabular}} & \multirow{4}{*}{\begin{tabular}[c]{@{}c@{}}\\Out-Data\\(unseen)\end{tabular}} & 
\multicolumn{2}{c}{ODIN, ACET, and Mahalanobis present troublesome requirements.}\\
&&& \multicolumn{2}{c}{ODIN, ACET, and Mahalanobis produce undesired side effects.}\\
\cmidrule{4-5}
&&& AUROC (\%) [$\uparrow$] & DTACC (\%) [$\uparrow$]\\
&&& \multicolumn{2}{c}{ODIN / ACET / IsoMax+ES (ours) / Mahalanobis}\\
\midrule
\multirow{9}{*}{\begin{tabular}[c]{@{}c@{}}DenseNet~~~~\end{tabular}}
& \multirow{3}{*}{\begin{tabular}[c]{@{}c@{}}CIFAR10~~~~\end{tabular}} 
& SVHN & 92.7$\pm$0.4 / NA / {\bf96.7$\pm$0.4} / \bf97.5$\pm$0.6 & 86.4$\pm$0.4 / NA / {\bf91.8$\pm$0.4} / \bf92.4$\pm$0.5\\
&& TinyImageNet~~~~& 97.3$\pm$0.4 / NA / {\bf97.9$\pm$0.4} / \bf98.5$\pm$0.6 & 92.1$\pm$0.4 / NA / {\bf93.5$\pm$0.4} / \bf94.3$\pm$0.6\\
&& LSUN & 98.4$\pm$0.4 / NA / {\bf98.9$\pm$0.4} / \bf99.1$\pm$0.6 & 94.3$\pm$0.3 / NA / {\bf95.1$\pm$0.4} / \bf95.9$\pm$0.4\\
\cmidrule{2-5} 
& \multirow{3}{*}{\begin{tabular}[c]{@{}c@{}}CIFAR100\end{tabular}} 
& SVHN & 88.1$\pm$0.6 / NA / 88.7$\pm$0.7 / \bf91.7$\pm$0.6 & 80.8$\pm$0.6 / NA / {\bf83.9$\pm$0.6} / \bf84.3$\pm$0.7\\
&& TinyImageNet & 85.2$\pm$0.6 / NA / 92.8$\pm$0.5 / \bf96.9$\pm$0.7 & 77.1$\pm$0.6 / NA / 86.7$\pm$0.5 / \bf91.7$\pm$0.6\\
&& LSUN & 85.8$\pm$0.6 / NA / 94.6$\pm$0.4 / \bf97.7$\pm$0.6 & 77.3$\pm$0.6 / NA / 89.2$\pm$0.7 / \bf93.5$\pm$0.5\\
\cmidrule{2-5} 
& \multirow{3}{*}{\begin{tabular}[c]{@{}c@{}}SVHN\end{tabular}} 
& CIFAR10 & 91.8$\pm$0.2 / NA / {\bf98.6$\pm$0.2} / \bf98.7$\pm$0.3 & 86.7$\pm$0.2 / NA / {\bf95.7$\pm$0.3} / \bf96.1$\pm$0.3\\
&& TinyImageNet & 94.8$\pm$0.2 / NA / {\bf99.3$\pm$0.2} / \bf99.7$\pm$0.3 & 90.2$\pm$0.2 / NA / 96.2$\pm$0.2 / \bf98.8$\pm$0.3\\
&& LSUN & 94.0$\pm$0.2 / NA / {\bf99.5$\pm$0.2} / \bf99.8$\pm$0.3 & 89.1$\pm$0.2 / NA / 96.1$\pm$0.2 / \bf99.0$\pm$0.1\\
\midrule
\multirow{9}{*}{\begin{tabular}[c]{@{}c@{}}ResNet\end{tabular}}
& \multirow{3}{*}{\begin{tabular}[c]{@{}c@{}}CIFAR10\end{tabular}} 
& SVHN & 86.4$\pm$0.4 / {\bf97.7$\pm$0.4} / {\bf97.4$\pm$0.4} / 95.5$\pm$0.6 & 77.7$\pm$0.4 / NA / {\bf91.8$\pm$0.4} / 89.0$\pm$0.5\\
&& TinyImageNet & 93.9$\pm$0.3 / 85.7$\pm$0.4 / 94.7$\pm$0.3 / \bf99.1$\pm$0.5 & 86.1$\pm$0.3 / NA / 88.5$\pm$0.3 / \bf95.4$\pm$0.5\\
&& LSUN & 93.4$\pm$0.4 / 85.9$\pm$0.4 / 96.8$\pm$0.3 / \bf99.5$\pm$0.3 & 85.7$\pm$0.4 / NA / 91.3$\pm$0.3 / \bf97.3$\pm$0.4\\
\cmidrule{2-5} 
& \multirow{3}{*}{\begin{tabular}[c]{@{}c@{}}CIFAR100\end{tabular}} 
& SVHN & 72.1$\pm$0.6 / {\bf91.1$\pm$0.6} / 85.2$\pm$0.6 / 84.3$\pm$0.5 & 67.8$\pm$0.4 / NA / {\bf79.9$\pm$0.6} / 76.4$\pm$0.7\\
&& TinyImageNet & 83.7$\pm$0.6 / 75.3$\pm$0.5 / {\bf92.4$\pm$0.5} / 87.7$\pm$0.6 & 75.7$\pm$0.5 / NA / {\bf85.8$\pm$0.5} / 84.3$\pm$0.5\\
&& LSUN & 81.8$\pm$0.6 / 69.7$\pm$0.5 / {\bf93.3$\pm$0.6} / 82.2$\pm$0.7 & 74.7$\pm$0.6 / NA / {\bf87.6$\pm$0.5} / 79.6$\pm$0.6\\
\cmidrule{2-5} 
& \multirow{3}{*}{\begin{tabular}[c]{@{}c@{}}SVHN\end{tabular}} 
& CIFAR10 & 92.1$\pm$0.2 / 97.3$\pm$0.3 / {\bf98.2$\pm$0.2} / 97.3$\pm$0.2 & 89.3$\pm$0.3 / NA / {\bf94.3$\pm$0.2} / \bf94.5$\pm$0.3\\
&& TinyImageNet & 92.8$\pm$0.2 / 97.6$\pm$0.2 / {\bf98.8$\pm$0.1} / \bf99.0$\pm$0.3 & 90.0$\pm$0.2 / NA / 94.6$\pm$0.3 / \bf98.7$\pm$0.2\\
&& LSUN & 90.6$\pm$0.2 / {\bf99.7$\pm$0.3} / 97.9$\pm$0.2 / \bf99.8$\pm$0.2 & 88.3$\pm$0.2 / NA / 93.7$\pm$0.3 / \bf99.4$\pm$0.2\\
\bottomrule
\end{tabularx}
\small
\begin{justify}
Source: The Author (2022). OOD Detection: Unfair comparison of approaches with different requirements and side effects. ODIN \citep{liang2018enhancing} uses input preprocessing, temperature calibration, and adversarial validation (hyperparameter tuning using adversarial examples). The Mahalanobis \citep{lee2018simple} solution uses input preprocessing, adversarial validation, feature extraction, feature ensemble, and metric learning. Input preprocessing makes the inferences of ODIN and the Mahalanobis method at least three times slower and at least three times less energy/computationally efficient than SoftMax or IsoMax inferences. Feature ensembles may limit the Mahalanobis method scalability to deal with large-size images used in real-world applications. ACET \citep{Hein2018WhyRN} uses adversarial training, which results in slower training, possibly reduced scalability for large-size images, and eventually, classification accuracy drop. The ODIN, the Mahalanobis approach, and ACET present hyperparameters that need to be validated for each combination of datasets and models presented in the table. Furthermore, considering that adversarial hyperparameters (e.g., the adversarial perturbation) used in ODIN/Mahalanobis/ACET were validated using the SVHN/CIFAR10/CIFAR100 validation sets and that these sets were reused as OOD detection test set, we conclude that the OOD detection performance reported by those papers, which we are reproducing in this table, may be overestimated. ODIN/Mahalanobis/ACET results were obtained using SoftMax loss rather than IsoMax loss as baseline. No outlier/extra/background data is used. IsoMax+ES means training with IsoMax loss and performing OOD detection using the entropic score. IsoMax+ES does not use these techniques and does not have such requirements, side effects, and hyperparameters to tune (as IsoMax+ES works as a baseline OOD detection approach, those techniques may be incorporated in future research). 
The results represent the mean and standard deviation of five executions. The best values are bold when they overcome the competing approach value outside the margin of error given by the standard deviations.
\end{justify}
\label{tbl:unfair_odd_new}
\end{table*}

\subsubsection{Fair Scenario}

Table~\ref{tbl:expanded_fair_odd_new} shows that models trained with the SoftMax loss using the maximum probability as the score \mbox{(SoftMax+MPS)} almost always present the worst OOD detection performance. In the case of models trained with SoftMax loss, replacing the maximum probability score by the entropic score (SoftMax+ES) produces small OOD detection performance gains. However, the combination of models trained using IsoMax loss with the entropic score (IsoMax+ES), which is the proposed solution, significantly improves, usually by several percentage points, the OOD detection performance across almost all datasets, models, out-distributions, and~metrics. We emphasize that the entropic score only produces high OOD detection performance when the probability distributions present high entropy (IsoMax+ES). For low entropy probability distributions (SoftMax+ES), the performance increase is minimal.

The IsoMax loss is well-positioned to replace SoftMax loss as a novel baseline OOD detection approach for neural networks OOD detection, as the former does not present an accuracy drop in relation to the latter and simultaneously improves the OOD detection performance. Additional techniques (e.g., input preprocessing, adversarial training, and outlier exposure) may be added to improve OOD detection performance gains further.

Table~\ref{tbl:tinyimagenet} shows that the results generalize to higher resolution images. Table~\ref{tbl:text_odd} presents results for text data. As expected, the results show that our approach is \emph{domain-agnostic} and therefore may be applied to data other than images.

\subsubsection{Unfair Scenario}

Table~\ref{tbl:unfair_odd_new} presents a perspective on how our proposed baseline OOD detection approach compares with no seamless solutions. Hence, we need to analyze the mentioned table considering that it shows an \emph{unfair} comparison of approaches that present different requirements and side effects. ODIN and the Mahalanobis solution use input preprocessing\footnote{To allow OOD detection, each inference requires a first neural network forward pass, a backpropagation, and a second forward pass.}. Consequently, they present solutions with much slower and less energy-efficient inferences than models trained with IsoMax loss, which are as fast and computation-efficient as the models trained with common SoftMax loss. Moreover, they also require validation using adversarial samples.

Additionally, ODIN requires temperature calibration, while the Mahalanobis approach uses feature ensemble and metric learning, which may be implicated in limited scalability for large-size images. ACET requires adversarial training, which may also prevent its use in large-size images. ODIN, the Mahalanobis approach, and ACET have hyperparameters tuned for each combination of in-data and models in the table. ODIN, the Mahalanobis method, and ACET used SoftMax loss rather than IsoMax loss. IsoMax+ES exhibits neither the mentioned unfortunate requirements nor undesired side effects. Additionally, taking into consideration that ODIN, Mahalanobis, and ACET used adversarial hyperparameters (e.g., the adversarial perturbation) that were validated using the validation sets of SVHN/CIFAR10/CIFAR100 and noticing that these sets composed the OOD detection test sets, we conclude that some overestimation may be present in the OOD detection performance reported by these papers.

Regardless of the previous considerations, Table~\ref{tbl:unfair_odd_new} shows that IsoMax+ES considerably outperforms ODIN in all evaluated scenarios. Therefore, in addition to avoiding hyperparameter tuning and access to OOD or adversarial samples, the results show that removing the entropic scale is much more effective in increasing the OOD detection performance than performing temperature calibration. Furthermore, IsoMax+ES usually outperforms ACET by a large margin. Moreover, in most cases, even operating under much more favorable conditions, the Mahalanobis method surpasses IsoMax+ES by less than 2\%. In some scenarios, the latter overcomes the former despite avoiding hyperparameter validation, being seamless and producing much faster and more computation-efficient inferences, as no input preprocessing technique is~required.

\subsection{Training Examples per Class Variation Study}

\begin{figure*}
\small
\centering
\caption[Training Examples per Class Variation Study: SoftMax vs. IsoMax]{Training Examples per Class Variation Study: SoftMax vs. IsoMax}
\subfloat[]{\includegraphics[width=0.8\textwidth]{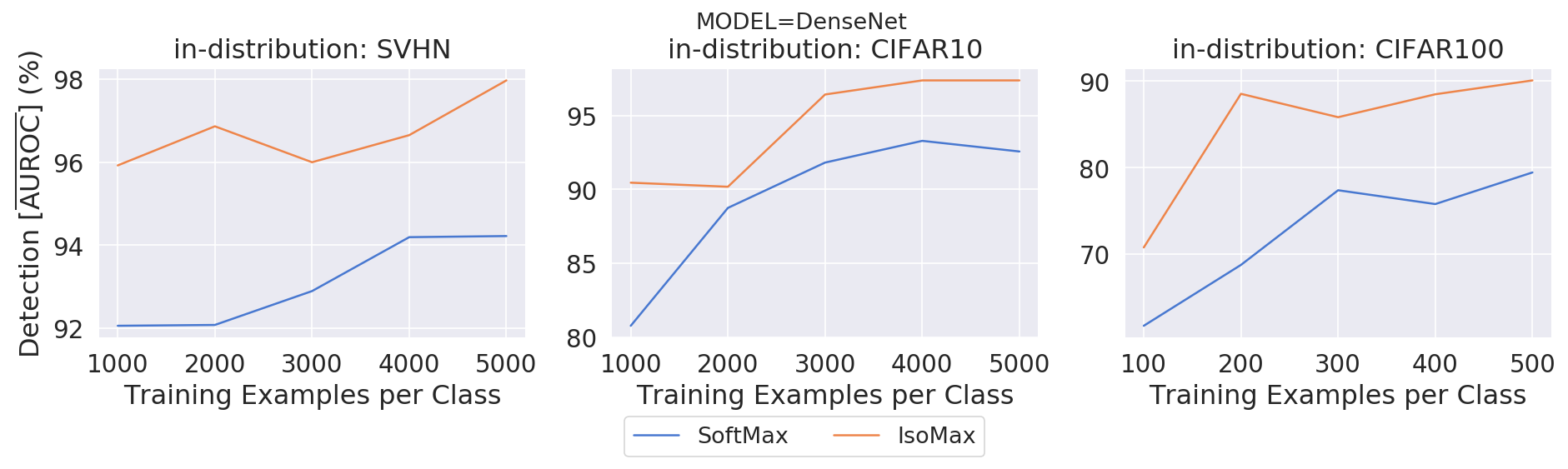}} 
\\
\vskip -0.05cm
\subfloat[]{\includegraphics[width=0.8\textwidth]{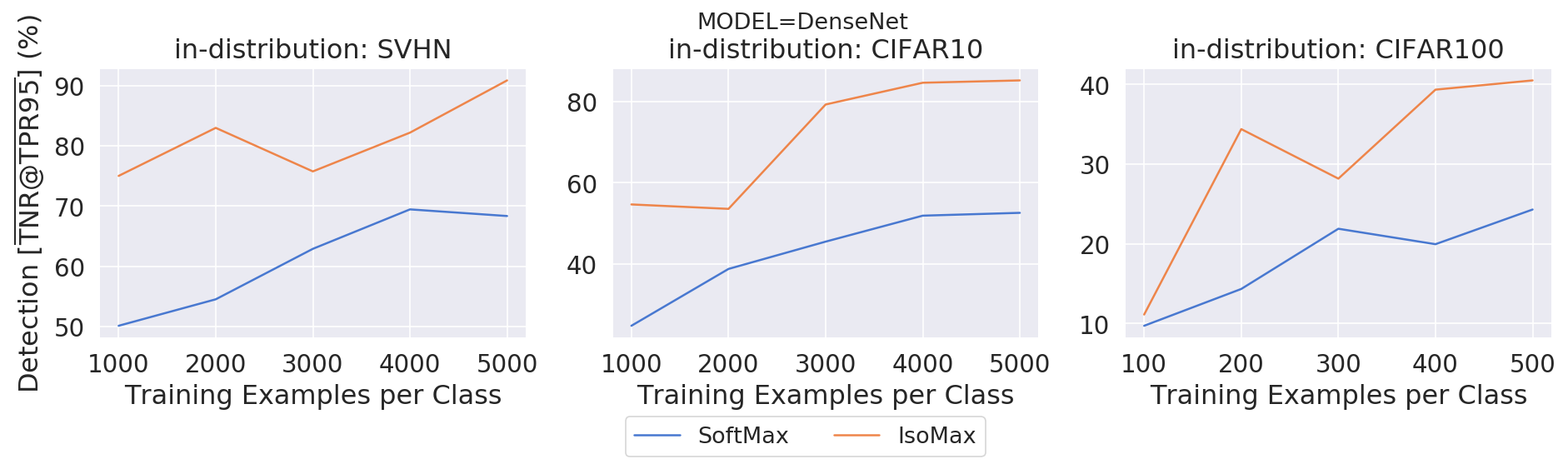}}
\\
\vskip -0.05cm
\subfloat[]{\includegraphics[width=0.8\textwidth]{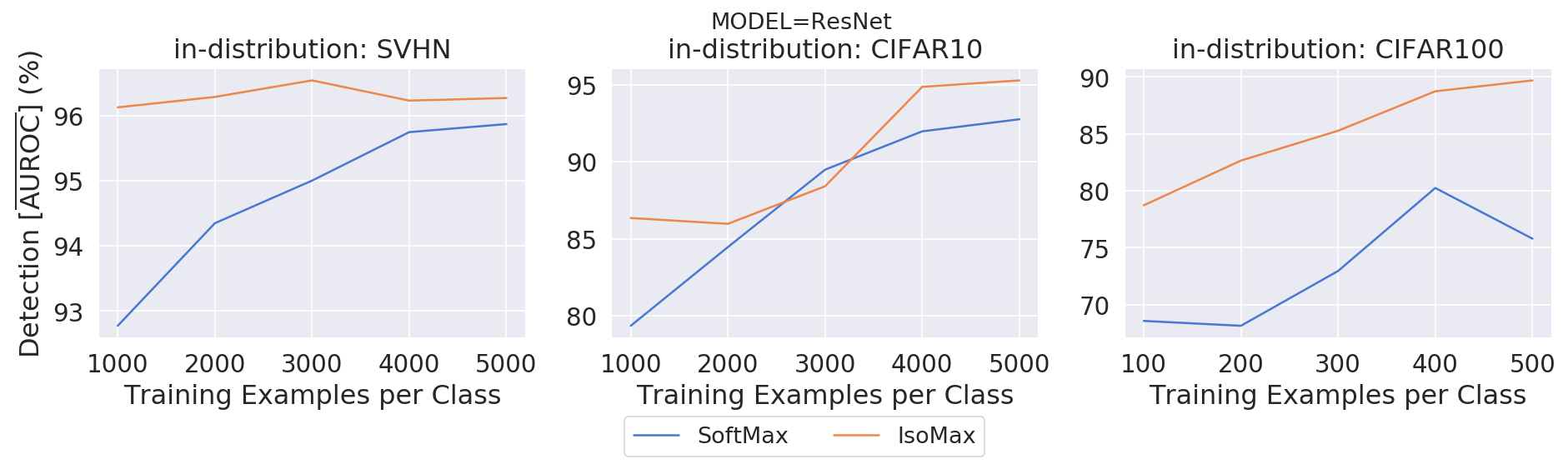}} 
\\
\vskip -0.05cm
\subfloat[]{\includegraphics[width=0.8\textwidth]{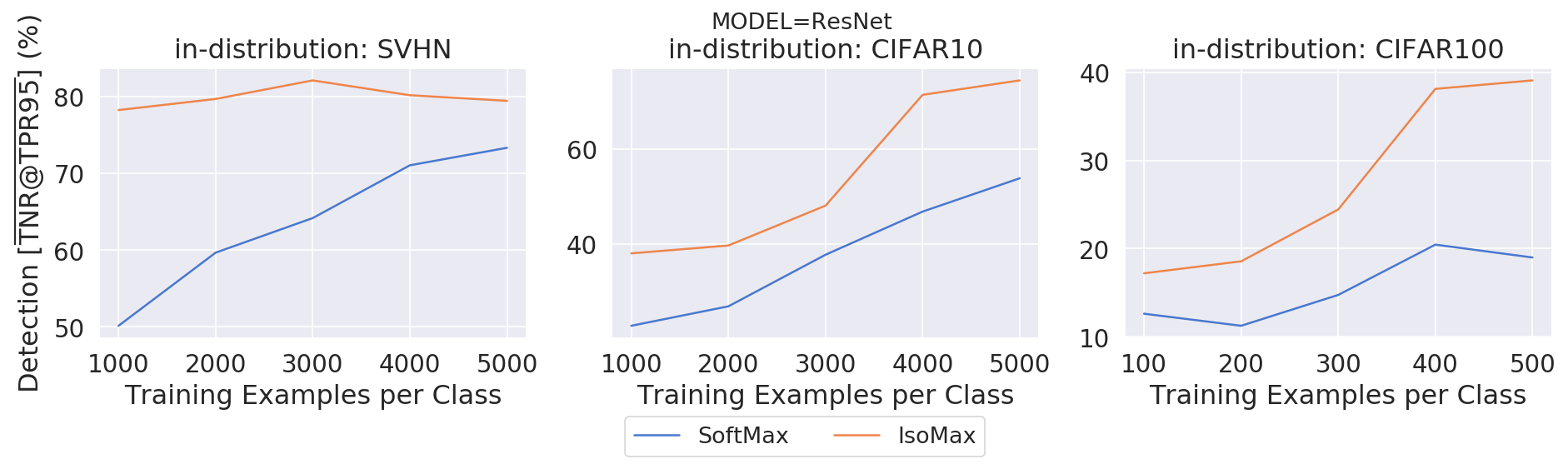}}
\vskip -0.05cm
\begin{justify}
{Source: The Author (2022). Training Examples per Class Variation Study: IsoMax loss consistently presents higher OOD detection performance than does SoftMax loss for different numbers of training examples per class on several datasets, models, and metrics. The entropic score was used for both SoftMax and IsoMax losses. \hl{For each in-distribution, we calculated the mean AUROC and TNR@TPR95 considering all possible OOD detection test sets. The vertical scales are automatically adjusted to better focus on the intervals the curves indeed variate.}}
\end{justify}
\label{fig:detection_robstness}
\end{figure*}

Fig.~\ref{fig:detection_robstness} presents the OOD detection performance of SoftMax and IsoMax losses in many models (DenseNet and ResNet), and datasets (SVHN, CIFAR10, and CIFAR100) for a range of training samples per class. For each in-distribution, the AUROC and TNR@TPR95 results were averaged over all possible combinations of OOD detection test set, except noise and fooling ones. The entropic score was used in all cases. It shows that IsoMax loss consistently and notably overcomes the OOD detection performance of SoftMax loss for virtually all combinations of evaluated datasets, training examples per class, metrics, and~models.

\subsection{Logits, Probabilities, Entropies, and Training}

\begin{figure*}
\small
\centering
\caption[Training Metrics: SoftMax vs. IsoMax]{Training Metrics: SoftMax vs. IsoMax}
\subfloat[]{\includegraphics[width=0.8\textwidth]{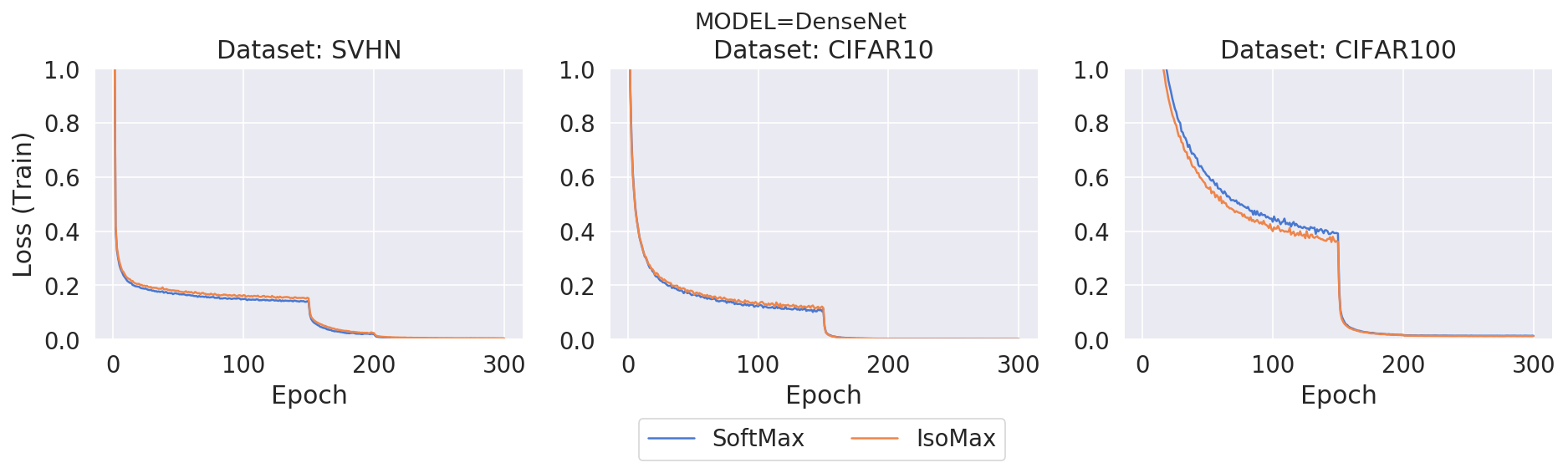}}
\\
\vskip -0.025cm
\subfloat[]{\includegraphics[width=0.8\textwidth]{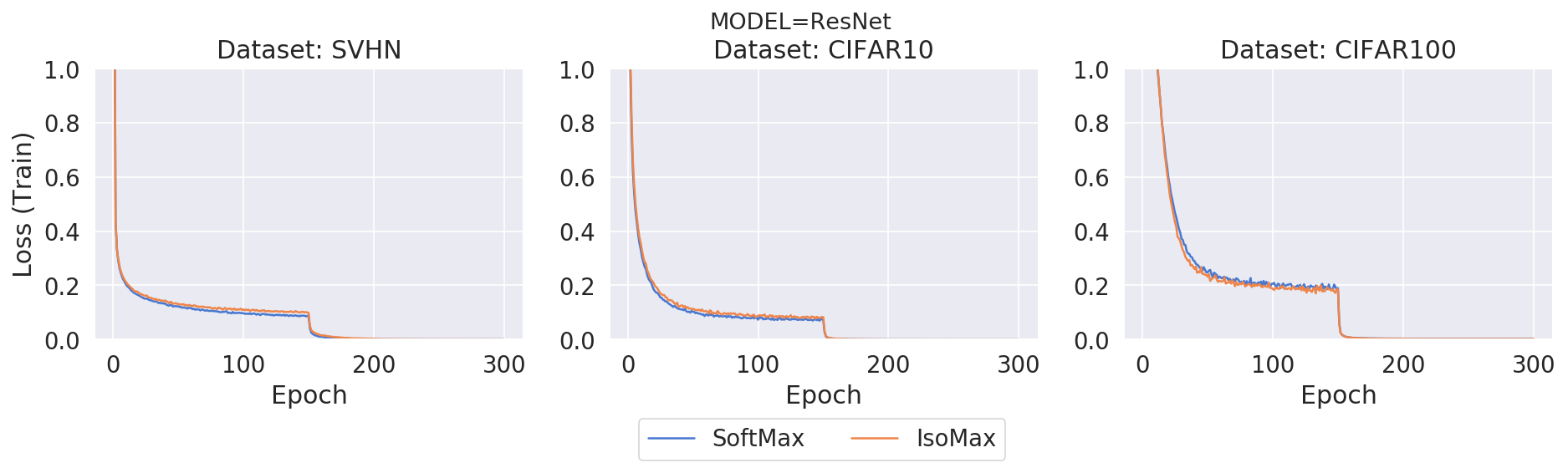}}
\\
\vskip -0.025cm
\subfloat[]{\includegraphics[width=0.8\textwidth]{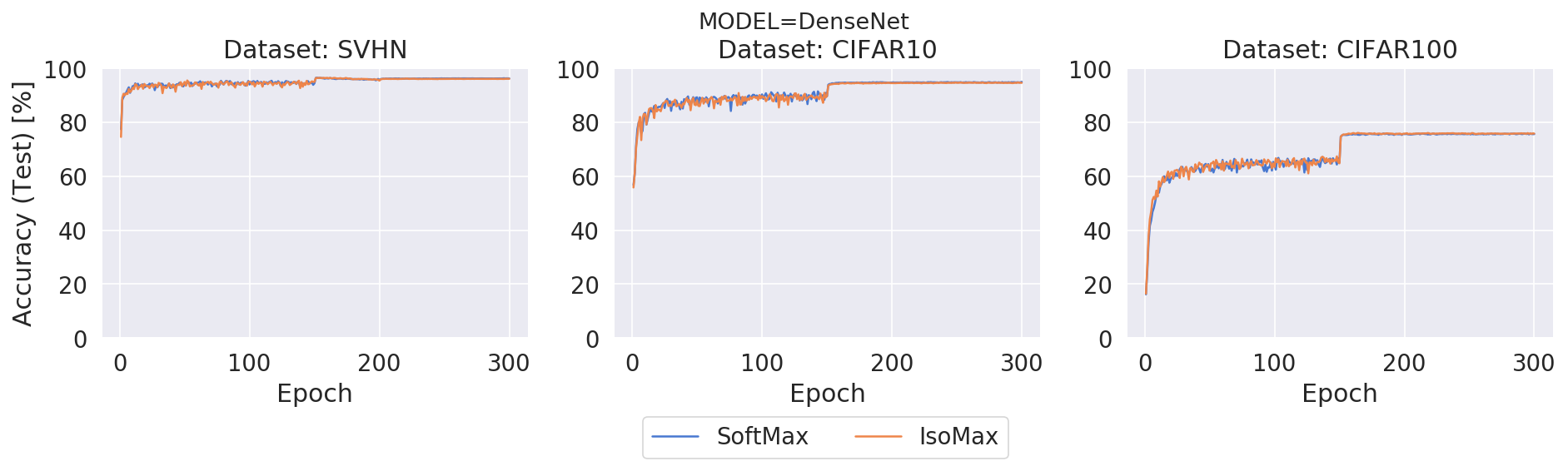}}
\\
\vskip -0.025cm
\subfloat[]{\includegraphics[width=0.8\textwidth]{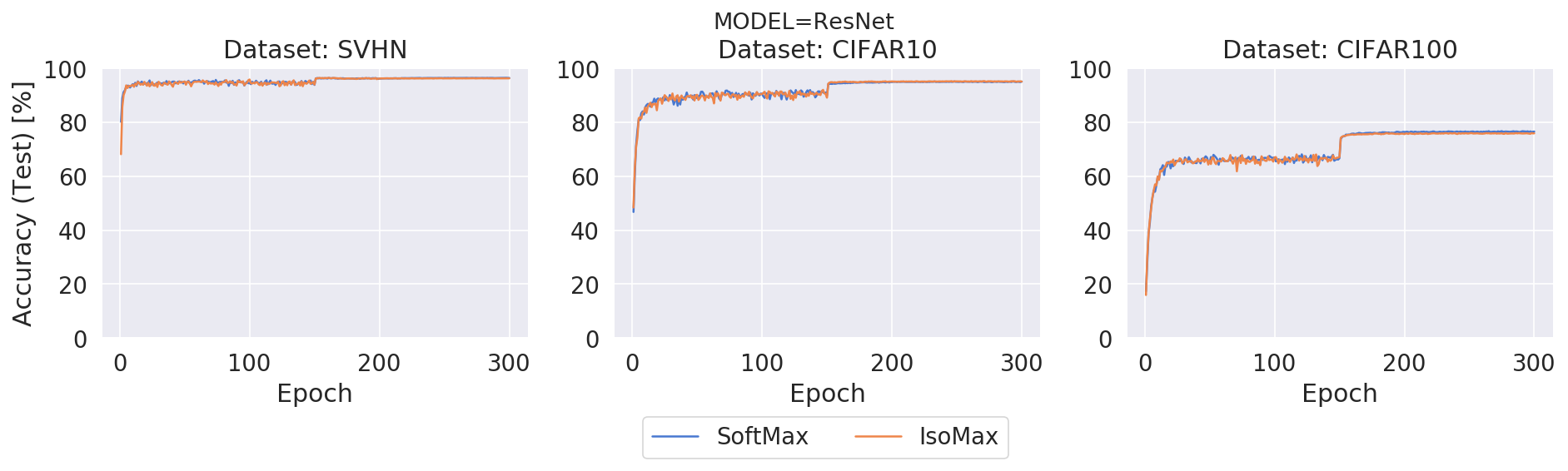}}
\\
\vskip -0.15cm
\begin{justify}
{Source: The Authors (2022). Training metrics: (a) Training loss values and (b) test accuracies for a range of models, datasets, and losses. SoftMax loss and IsoMax loss present remarkably similar metrics throughout training, which confirms that IsoMax loss training is consistent and stable.}
\end{justify}
\label{fig:training_metrics}
\end{figure*}

\begin{figure*}
\small
\centering
\caption[Logits: SoftMax vs. IsoMax]{Logits: SoftMax vs. IsoMax}
\subfloat[]{\includegraphics[width=0.8\textwidth]{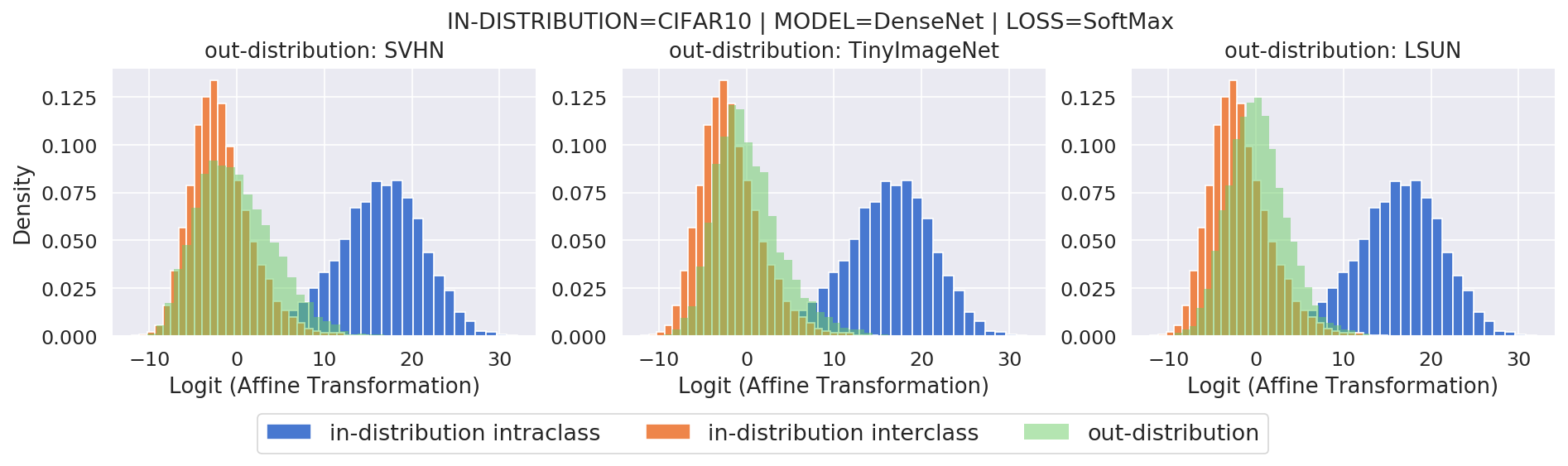}\label{fig:densent_softmax_logits_histograms}}
\\
\subfloat[]{\includegraphics[width=0.8\textwidth]{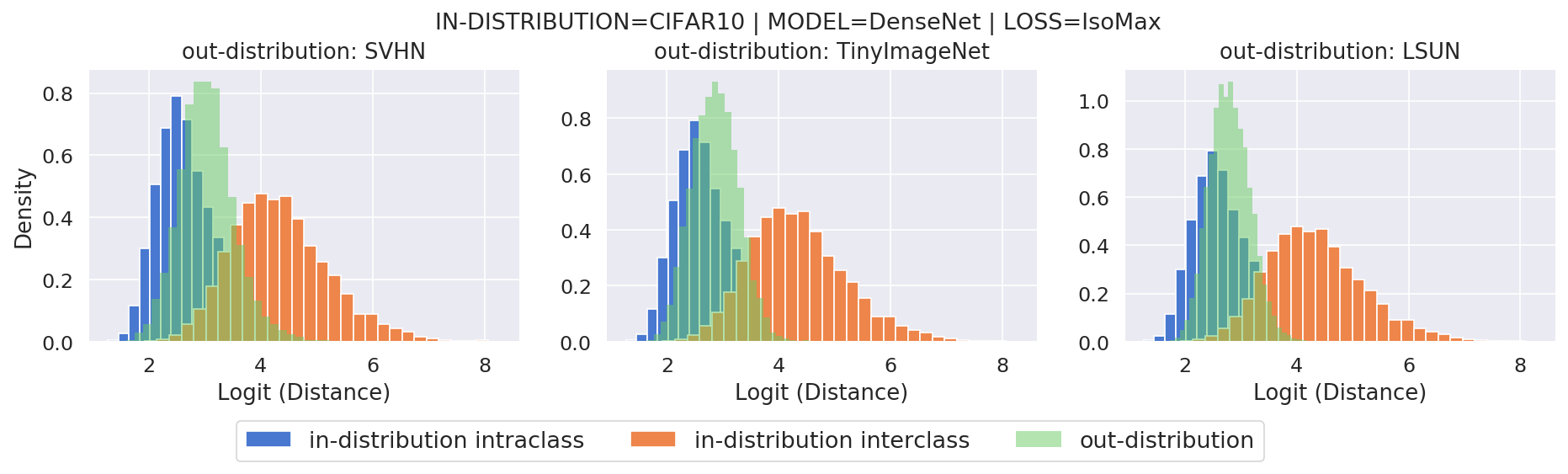}\label{fig:densenet_isomax_logits_histograms}} 
\vskip -0.05cm
\subfloat[]{\includegraphics[width=0.8\textwidth]{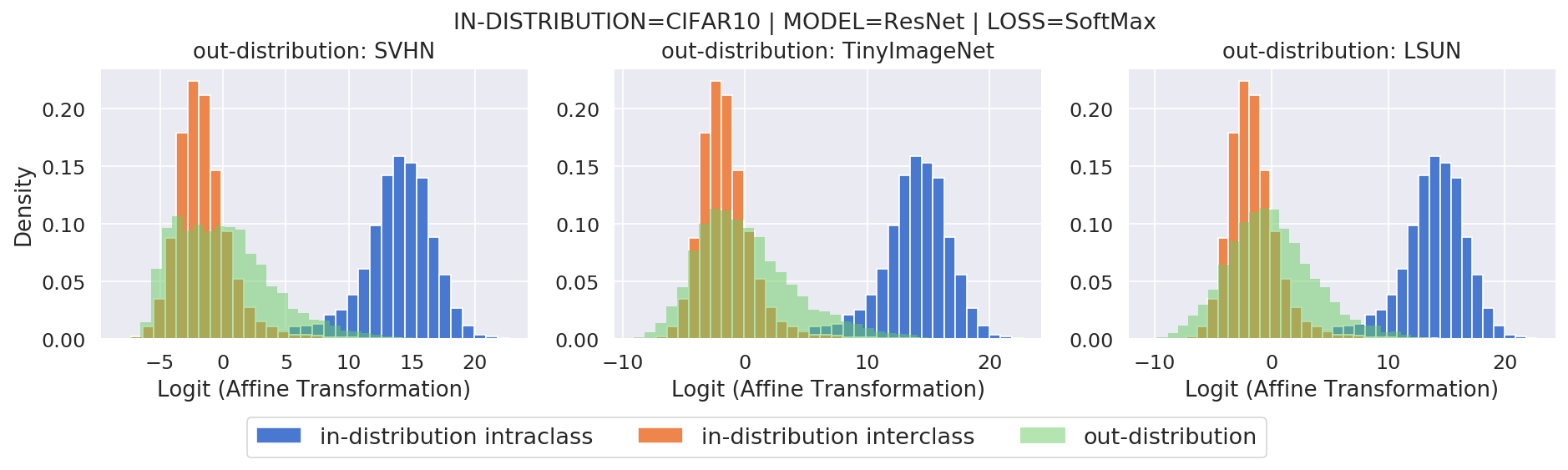}\label{fig:resnet_softmax_logits_histograms}} 
\\
\subfloat[]{\includegraphics[width=0.8\textwidth]{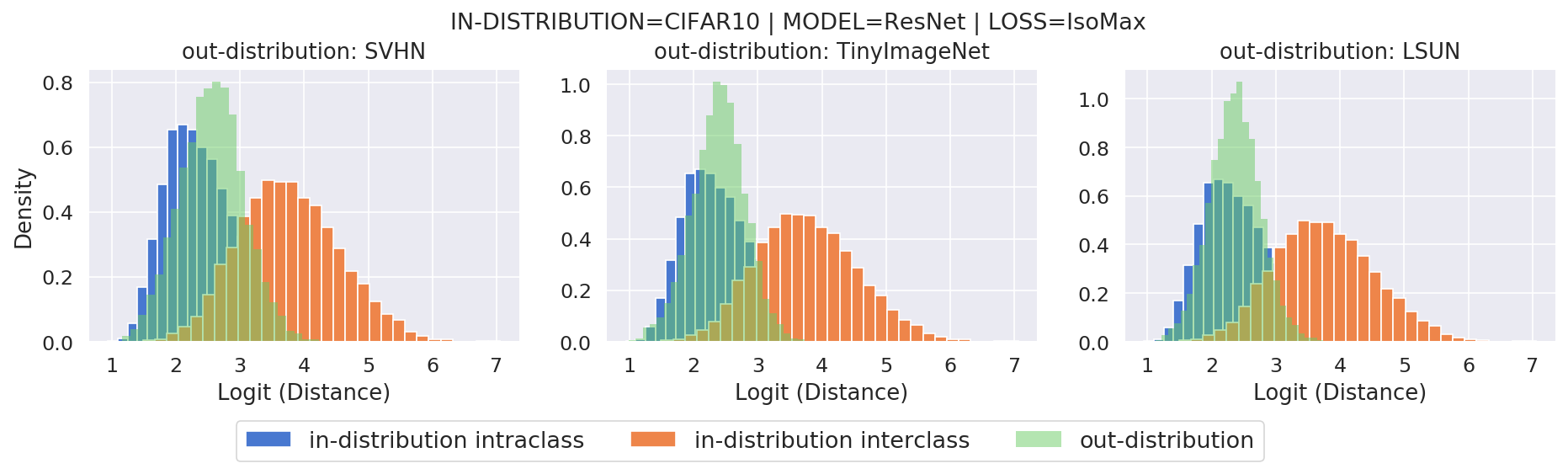}\label{fig:resnet_isomax_logits_histograms}} 
\begin{justify}
{Source: The Authors (2022). Logits: (a,c) In SoftMax loss, out-distribution logits mimic in-distribution interclass logits. (b,d) In IsoMax loss, out-distribution logits mimic in-distribution intraclass logits, which facilitates OOD detection, as there are many more interclass logits than intraclass logits, and the entropic score takes into consideration the information provided by all network outputs rather than just one. Distances are calculated from class prototypes.}
\end{justify}
\label{fig:logits_histograms}
\end{figure*}

Fig.~\ref{fig:training_metrics} illustrates that training metrics are remarkably similar for SoftMax and IsoMax losses. Fig.~\ref{fig:logits_histograms} shows that the in-distribution \emph{interclass} logits are more distinguishable from out-distribution logits when using IsoMax loss, which explains its increased OOD detection performance compared with the SoftMax loss because there are far more \emph{interclass} logits than \emph{intraclass} logits. Distances are calculated from class prototypes.

\begin{figure*}
\small
\centering
\caption[Probabilities and Entropies: SoftMax vs. IsoMax]{Probabilities and Entropies: SoftMax vs. IsoMax}
\subfloat[]{\includegraphics[width=0.8\textwidth]{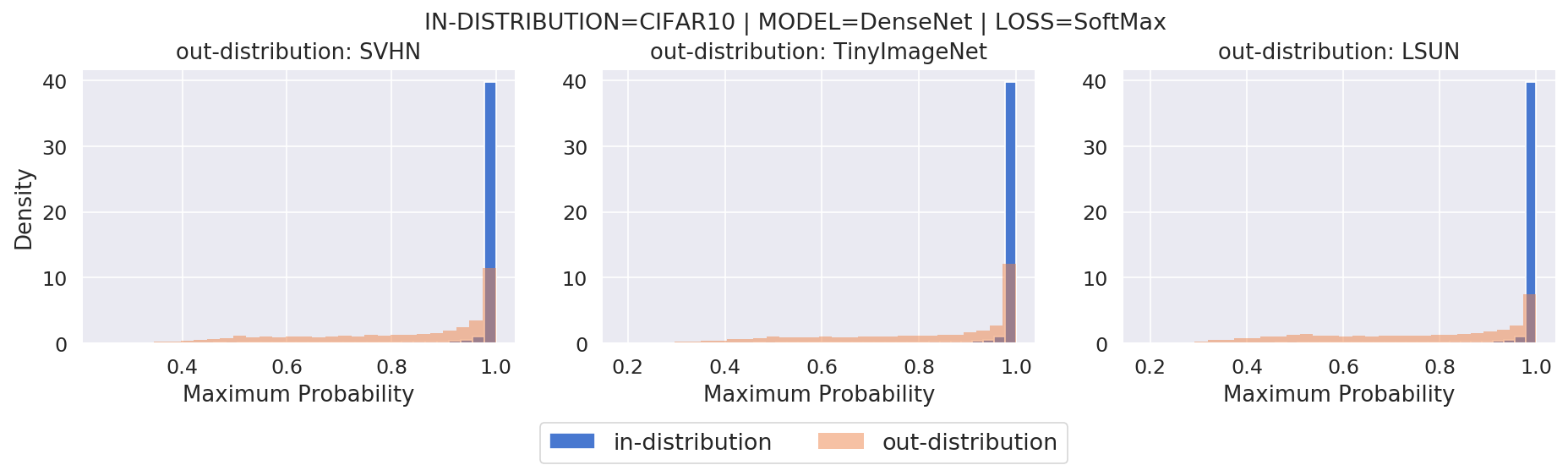}\label{fig:softmax_maxprobs_histograms}}
\\
\vskip -0.05cm
\subfloat[]{\includegraphics[width=0.8\textwidth]{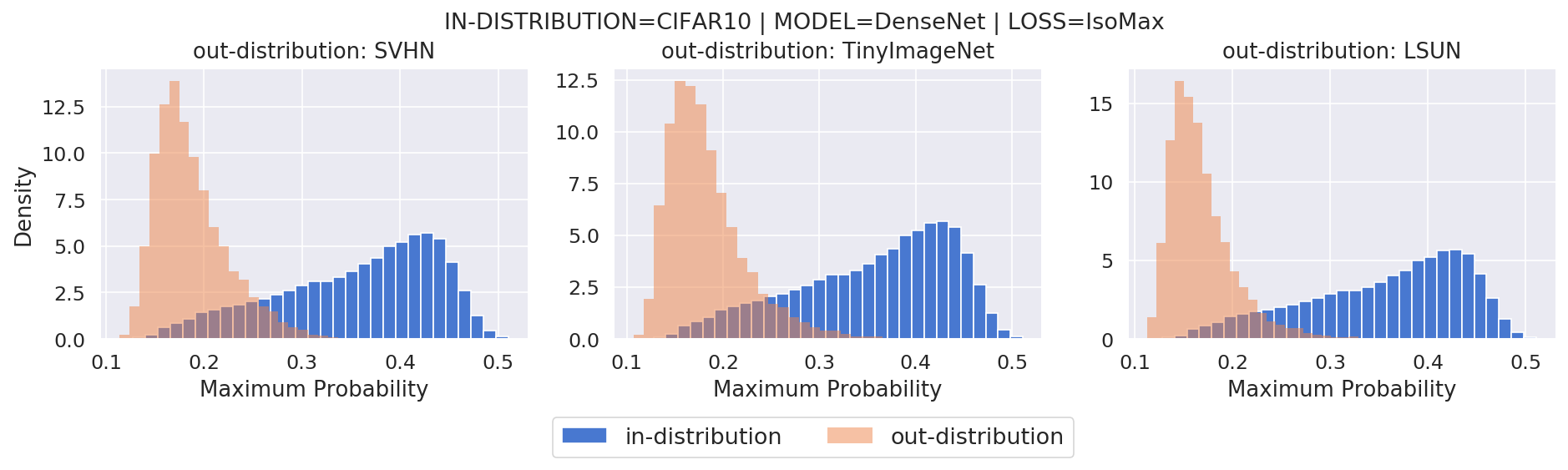}\label{fig:isomax_maxprobs_histograms}} 
\\
\vskip -0.05cm
\subfloat[]{\includegraphics[width=0.8\textwidth]{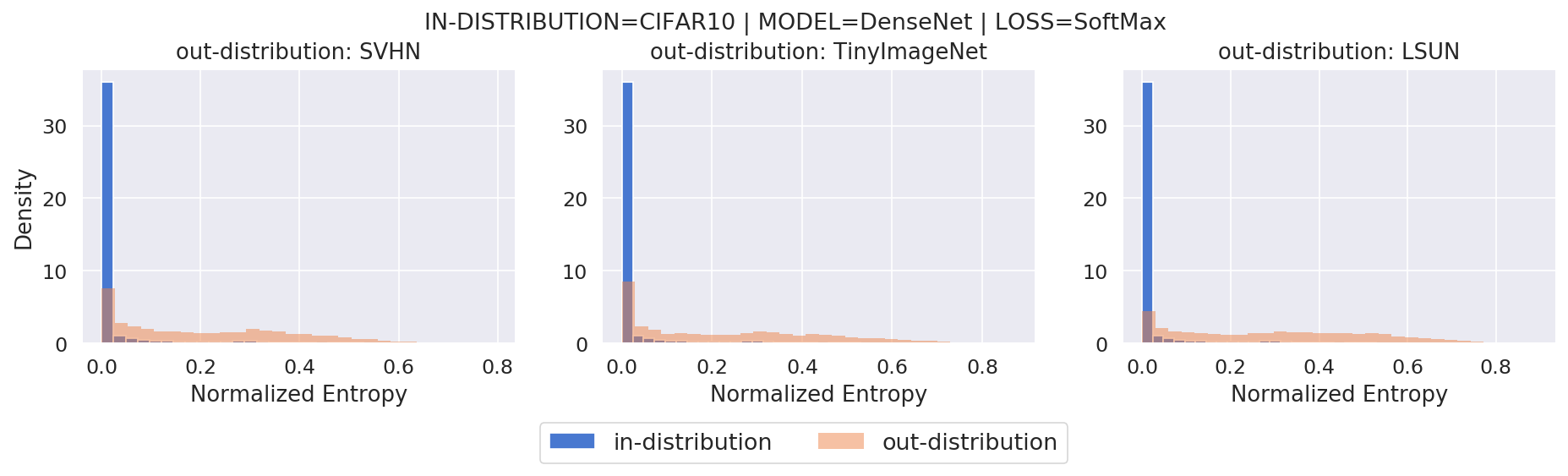}\label{fig:softmax_entropies_histograms}}
\\
\vskip -0.05cm
\subfloat[]{\includegraphics[width=0.8\textwidth]{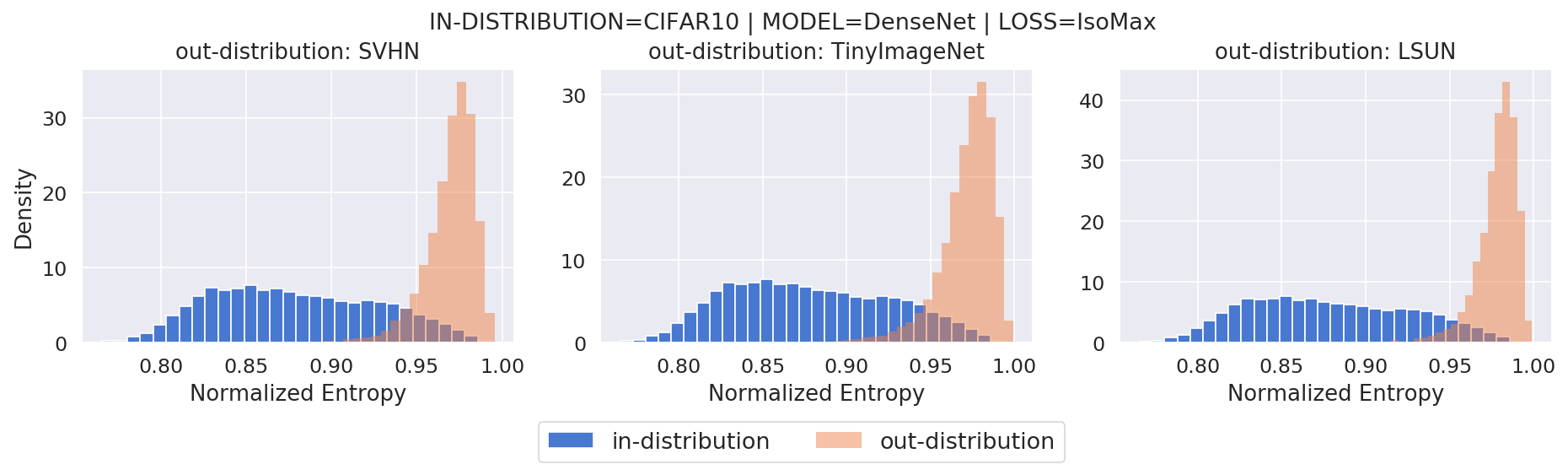}\label{fig:isomax_entropies_histograms}} 
\vskip -0.05cm
\begin{justify}
{Source: The Author (2022). Probabilities and entropies: (a) SoftMax loss produces extremely high confident predictions for in-distribution samples. SoftMax loss usually provides extremely high maximum probabilities, even for OOD samples. (b) IsoMax loss produces less confident predictions for in-distribution samples than SoftMax loss. IsoMax loss commonly produces an even lower maximum probability for OOD samples. (c) SoftMax loss provides extremely low entropy (high confidence) for almost all in-distribution samples and even usually for OOD samples. (d) IsoMax loss produces high entropy for out-distributions. More precise separation between the in-distribution and out-distributions is obtained.}
\end{justify}
\label{fig:maxprobs_entropies_histograms}
\end{figure*}

Fig.~\ref{fig:maxprobs_entropies_histograms} shows that networks trained with SoftMax loss exhibit extremely high maximum probabilities. Sometimes this is true even for OOD samples. For networks trained with IsoMax loss, OOD samples usually present lower maximum probabilities compared with in-distribution samples. Furthermore, it also shows that the networks trained with SoftMax loss are extremely overconfident. It can be observed that the entropy works as a high-quality score to distinguish the in-distribution from the out-distribution in neural networks trained with IsoMax loss.

\clearpage
\subsection{Inference Efficiency}

\begin{table*}
\footnotesize
\caption[Inference Delays, Computational Cost and Energy Consumption]{Inference Delays, Computational Cost and Energy Consumption}
\label{tbl:times}
\begin{tabularx}{\textwidth}{lll|YYY}
\toprule
\multirow{4}{*}{\begin{tabular}[c]{@{}c@{}}\\Model\end{tabular}} & \multirow{4}{*}{\begin{tabular}[c]{@{}c@{}}\\In-Data\\(training)\end{tabular}} & \multirow{4}{*}{\begin{tabular}[c]{@{}c@{}}\\Hardware\\(inference)\end{tabular}} & 
\multicolumn{3}{c}{Out-of-Distribution Detection:}\\
&&& \multicolumn{3}{c}{Inference Delays / Presumed Computational Cost and Energy Consumption Rates.}\\
\cmidrule{4-6}
&&& SoftMax Loss \mbox{(current baseline)} & IsoMax Loss (ours) \mbox{(proposed baseline)} & Input Preprocessing: \mbox{ODIN, Mahalanobis}, Generalized ODIN\\
&&& MPS (ms) [$\downarrow$] / ES (ms) [$\downarrow$] & MPS (ms) [$\downarrow$] / ES (ms) [$\downarrow$] & (ms) [$\downarrow$]\\
\midrule
\multirow{6}{*}{\begin{tabular}[c]{@{}c@{}}DenseNet~~~~~~\end{tabular}}
& \multirow{2}{*}{\begin{tabular}[c]{@{}c@{}}CIFAR10~~~~~~~\end{tabular}} 
&~~~~CPU~~~~~~& 18.1 / 19.4 & 18.0 / 19.2 & 242.4 \bf{($\approx$ 10x slower)}\\
&&~~~~GPU~~~~~~& 11.6 / 13.0 & 11.6 / 11.5 & 39.2 \bf{($\approx$ 4x slower)}\\
\cmidrule{2-6} 
& \multirow{2}{*}{\begin{tabular}[c]{@{}c@{}}CIFAR100\end{tabular}} 
&~~~~CPU~~~~~~& 18.4 / 19.8 & 18.4 / 19.3 & 261.0 \bf{($\approx$ 10x slower)}\\
&&~~~~GPU~~~~~~& 12.9 / 11.4 & 11.8 / 11.5 & 39.6 \bf{($\approx$ 4x slower)}\\
\cmidrule{2-6} 
& \multirow{2}{*}{\begin{tabular}[c]{@{}c@{}}SVHN\end{tabular}} 
&~~~~CPU~~~~~~& 18.1 / 18.6 & 18.3 / 18.6 & 241.5 \bf{($\approx$ 10x slower)}\\
&&~~~~GPU~~~~~~& 11.6 / 11.9 & 11.7 / 11.6 & 39.6 \bf{($\approx$ 4x slower)}\\
\midrule
\multirow{6}{*}{\begin{tabular}[c]{@{}c@{}}ResNet\end{tabular}}
& \multirow{2}{*}{\begin{tabular}[c]{@{}c@{}}CIFAR10\end{tabular}} 
&~~~~CPU~~~~~~& 22.3 / 23.2 & 23.0 / 23.5 & 250.4 \bf{($\approx$ 10x slower)}\\
&&~~~~GPU~~~~~~& 4.5 / 3.8 & 4.2 / 4.1 & 15.4 \bf{($\approx$ 4x slower)}\\
\cmidrule{2-6} 
& \multirow{2}{*}{\begin{tabular}[c]{@{}c@{}}CIFAR100\end{tabular}} 
&~~~~CPU~~~~~~& 23.3 / 23.1 & 23.3 / 23.8 & 252.6 \bf{($\approx$ 10x slower)}\\
&&~~~~GPU~~~~~~& 4.3 / 3.9 & 4.3 / 4.2 & 14.8 \bf{($\approx$ 4x slower)}\\
\cmidrule{2-6} 
& \multirow{2}{*}{\begin{tabular}[c]{@{}c@{}}SVHN\end{tabular}} 
&~~~~CPU~~~~~~& 23.1 / 23.4 & 23.4 / 23.3 & 263.8 \bf{($\approx$ 10x slower)}\\
&&~~~~GPU~~~~~~& 4.2 / 4.0 & 4.0 / 4.0 & 15.7 \bf{($\approx$ 4x slower)}\\
\bottomrule
\end{tabularx}
\small
\begin{justify}
\looseness=-1
Source: The Author (2022). MPS means maximum probability score \citep{hendrycks2017baseline}. ODIN is from \cite{liang2018enhancing}. Mahalanobis is from \cite{lee2018simple}. Generalized ODIN is from \cite{Hsu2020GeneralizedOD}. ES means entropic score. For SoftMax and IsoMax losses (baseline OOD detection approaches), the inference delays combine both classification and detection computation. For the methods based on input preprocessing, the inference delays represent only the input preprocessing phase. All values are in milliseconds. In addition to presenting similar classification accuracy 
and much better OOD detection performance
, IsoMax loss trained networks produce inferences as fast as SoftMax trained networks. Moreover, the entropic score is as fast as the maximum probability score. Using CPU (Intel i7-4790K, 4.00GHz, x64, octa-core) for inference (the case more relevant from a cost point of view), methods based on input preprocessing are about ten times slower than the baseline approaches. Using GPU (Nvidia GTX 1080 Ti) for inference, our approach is about four times faster than the methods based on input preprocessing. The inference delay rates presumably reflect similar computational cost and energy consumption rates. The inference delays presented are the mean value of the inference delay of each image calculated (batch size equals one) over the entire dataset. The standard deviation was below 0.3 for all cases.
\end{justify}
\end{table*}

Table~\ref{tbl:times} presents the inference delays for SoftMax loss, IsoMax loss, and competing methods using CPU and GPU. We observe that neural networks trained using IsoMax loss produce inferences equally fast as the ones produced by networks trained using SoftMax loss, regardless of using CPU or GPU for inference. Additionally, the entropic score is as fast as the usual maximum probability score. Moreover, methods based on input preprocessing were more than ten times slower on CPU and about four times slower on GPU. Finally, those rates presumably apply to the computational cost and energy consumption as well.

At first sight, inference methods (i.e., methods that can be applied to pre-trained models) may be seen as ``low cost'' compared with training methods like our IsoMax loss, as we avoid training or fine-tuning the neural network. However, this conclusion may be misleading, as we have to keep in mind that inference methods (e.g., ODIN~\citep{liang2018enhancing} and Mahalanobis~\citep{lee2018simple}) produce inferences that are much more energy, computation, and time inefficient (Table~\ref{tbl:times}). 

For example, consider initially the rare practical situation where a pretrained model is available, and no fine-tuning to a custom dataset is required. In such cases, an inference method may indeed be applied without requiring any loss function. However, despite avoiding training or fine-tuning a neural network once or a few times, all the subsequent inferences, which are usually performed thousands or millions of times on the field (sometimes even by constrained devices), will be about 6 to 10 times more computational, energy, environment, and time inefficient (Table~\ref{tbl:times}).

Alternatively, consider the case where a pretrained model is not available or fine-tuning to a custom dataset is needed. In this situation, which is much more likely in practice, we cannot avoid training or fine-tuning the neural network, and a training method like ours will be required anyway. In these cases, it would be recommended to train or fine-tune using IsoMax rather than SoftMax loss, as our experiments showed that both training times are the same and IsoMax loss produces considerably higher OOD detection performance.

Hence, inference methods are more inference inefficient because they coexist with a model that was trained with a loss not designed from the start with OOD detection in mind. The drawback is to produce an enormous amount of inefficient inferences on the field that usually uses constrained computational resource devices such as embedded systems.

Nevertheless, suppose the increased inference time, computation, environment damage, and energy consumption required to use an inference method is not a concern from a practical point of view. In that case, the model pretrained or fine-tuned using a training method may be subsequently subjected to the desired inference approach to increase overall OOD detection performance further. In other words, we do not claim that inference methods such as ODIN and Mahalanobis do not increase the OOD detection performance compared with Maximum Score Probability or even the Entropic Score. However, we point out that producing more energy inefficient inferences is one drawback of adopting such inference methods.

In summary, rather than concurrent, the training and inference methods are orthogonal and complementary. Moreover, we see no reason not to train the models with a loss designed to support OOD, regardless of subsequently applying an OOD inference method.

\subsection{Qualitative Study}

\begin{figure*}
\small
\centering
\caption[Out-of-Distribution Detection Qualitative Study]{Out-of-Distribution Detection Qualitative Study}
\subfloat[]{\includegraphics[width=0.8\textwidth]{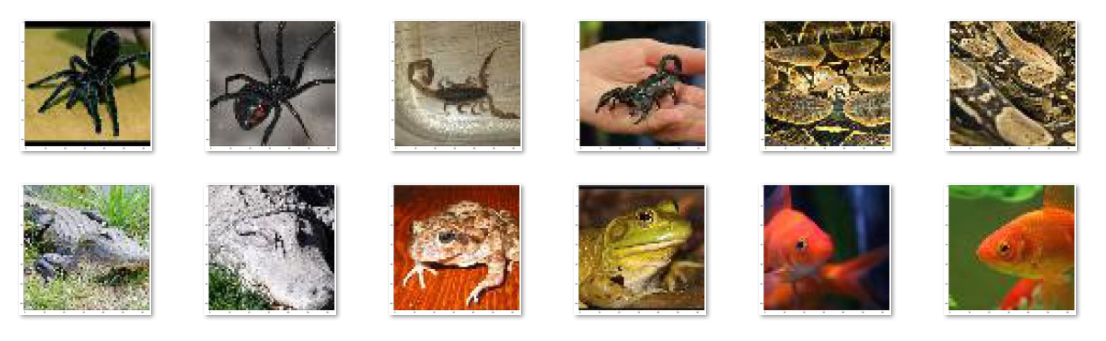}}
\vskip -0.05cm
\subfloat[]{\includegraphics[width=0.8\textwidth]{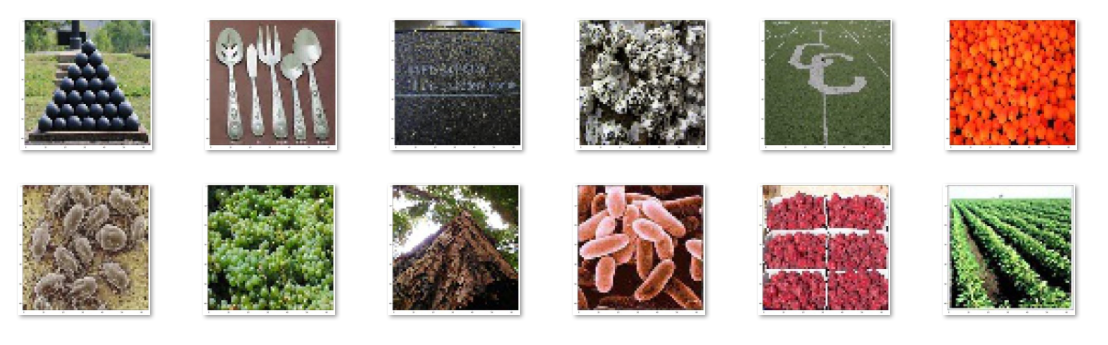}} 
\vskip -0.05cm
\subfloat[]{\includegraphics[width=0.8\textwidth]{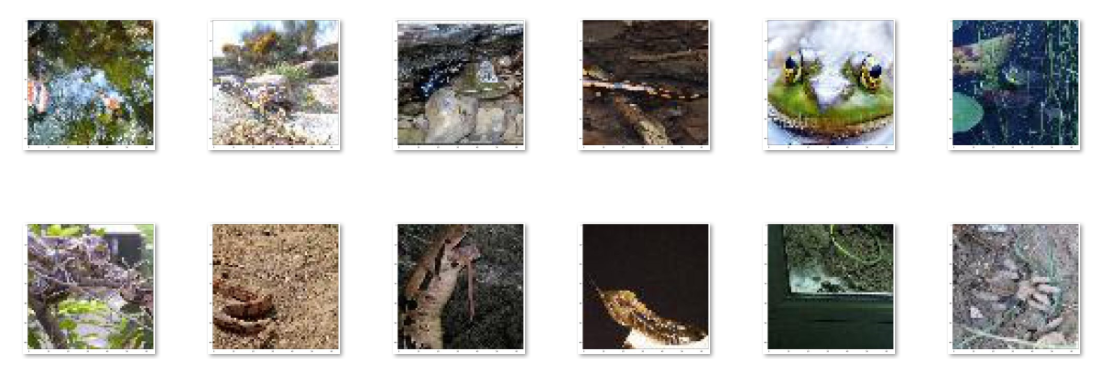}}
\vskip -0.05cm
\subfloat[]{\includegraphics[width=0.8\textwidth]{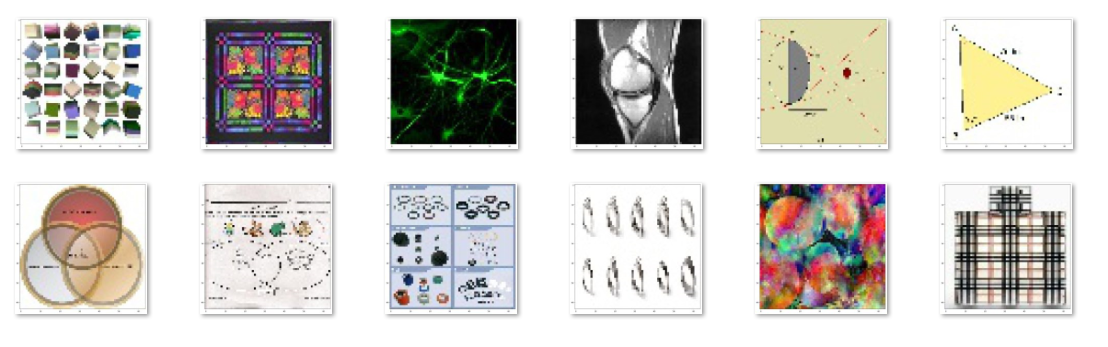}}
\begin{justify}
{Source: The Authors (2022). Out-of-Distribution Detection Qualitative Study: a) In-distribution examples correctly detected as in-distribution. b) Out-of-distribution examples correctly detected as out-of-distribution. c) In-distribution examples wrongly detected as out-of-distribution. d) Out-of-distribution examples wrongly detected as in-distribution.}
\end{justify}
\label{fig:ood_quality_study}
\end{figure*}

\figref{fig:ood_quality_study} presents examples of out-of-distribution detection performed by a ResNet34 trained on TinyImageNet using IsoMax. Resized examples from ImageNet-O \citep{DBLP:journals/corr/abs-1907-07174} were used as OOD examples. Examples in Fig~\ref{fig:ood_quality_study}a are from ID and were correctly detected as such. Moreover, examples from Fig.~\ref{fig:ood_quality_study}b are from OOD and were correctly detected as such. Examples in Fig.~\ref{fig:ood_quality_study}c were wrongly detected as OOD. It may be possibly explained by the fact they visually appear composed of more than one class or no class at all. are not very well-defined as belonging to a particular class. Fig.~\ref{fig:ood_quality_study}d present examples that were wrongly detected as ID. It may have happened because they present images with shapes and textures very different from those used in the training data.

\newpage\section{Enhanced Isotropy Maximization Loss}

To allow standardized comparison, we used the datasets, training procedures, and metrics that were established in \cite{hendrycks2017baseline} and used in many subsequent OOD detection papers \citep{liang2018enhancing, lee2018simple, Hein2018WhyRN}. We trained many 100-layer DenseNetBCs with growth rate $k\!=\!12$ (i.e., 0.8M parameters) \cite{Huang2017DenselyNetworks}, 110-layer ResNets (\emph{correct size proper implementation}) \citep{He_2016}\footnote{\url{https://github.com/akamaster/pytorch_resnet_cifar10}}, and 34-layer ResNets (\emph{overparametrized commonly used implementation}) \citep{He_2016}\footnote{\url{https://github.com/pokaxpoka/deep_Mahalanobis_detector}} on CIFAR10 \citep{Krizhevsky2009LearningImages}, CIFAR100 \citep{Krizhevsky2009LearningImages}, SVHN \citep{Netzer2011ReadingLearning}, and TinyImageNet \citep{Deng2009ImageNetDatabase} datasets with SoftMax, IsoMax, and IsoMax+ losses using the same procedures (e.g., initial learning rate, learning rate schedule, weight decay).

We used SGD with the Nesterov moment equal to 0.9 with a batch size of 64 and an initial learning rate of 0.1. The weight decay was 0.0001, and we did not use dropout. We trained during 300 epochs for CIFAR10, CIFAR100, and SVHN. We used a learning rate decay rate equal to ten applied in epoch numbers 150, 200, and 250 for CIFAR10, CIFAR100, and SVHN.

We used resized images from the TinyImageNet \citep{Deng2009ImageNetDatabase}, the Large-scale Scene UNderstanding dataset (LSUN) \citep{Yu2015LSUNLoop}, CIFAR10, and SVHN \citep{Netzer2011ReadingLearning} to create out-of-distribution samples. We added these out-of-distribution images to the validation sets of in-distribution data to form the test sets and evaluate the OOD detection performance. We evaluated the OOD detection performance using the true negative rate at 95\% true positive rate (TNR@TPR95), the area under the receiver operating characteristic curve (AUROC) and the detection accuracy (DTACC), which corresponds to the maximum classification probability over all thresholds $\delta$:
\begin{align*}
1- \min_{\delta} \big\{ P_{\texttt{in}} \left( o \left( \mathbf{x} \right) \leq \delta \right) P \left(\mathbf{x}\text{ is from }P_{\texttt{in}}\right)\\+ P_{\texttt{out}} \left( o \left( \mathbf{x} \right) >\delta \right) P \left(\mathbf{x}\text{ is from }P_{\texttt{out}}\right)\big\},
\end{align*}
where $o(\mathbf{x})$ is the OOD detection score. It is assumed that both positive and negative samples have an equal probability of being in the test set, i.e., $P \left(\mathbf{x}\text{ is from }P_{\texttt{in}}\right) = P \left(\mathbf{x}\text{ is from }P_{\texttt{out}}\right)$. \hl{This is the way this metric is defined and used in the literature.} All metrics follow the calculation procedures specified in \cite{lee2018simple}.

In this section, we present the results and discussion. We initially show that the enhanced IsoMax produces classification accuracies that are comparable to those of the SoftMax loss function. We then show that the enhanced IsoMax produces much higher OOD detection performance than the IsoMax Loss and the SoftMax Loss.

\subsection{Classification Accuracy}

\begin{table*}
\footnotesize
\centering
\caption[Classification accuracies using SoftMax, IsoMax, and IsoMax+ losses]{Classification accuracies using SoftMax, IsoMax, and IsoMax+ losses}
\vskip -0.25cm
\label{tab:classification_performance}
\begin{tabularx}{\textwidth}{YlYYY}
\toprule
\multirow{2}{*}{\begin{tabular}[c]{@{}c@{}}Model\end{tabular}} & \multirow{2}{*}{\begin{tabular}[c]{@{}c@{}}Data\end{tabular}}
& Train Accuracy (\%) [$\uparrow$] & Test Accuracy (\%) [$\uparrow$]\\
&& \multicolumn{2}{c}{SoftMax Loss / IsoMax Loss / IsoMax+ Loss}\\
\midrule
\multirow{3}{*}{\begin{tabular}[c]{@{}c@{}}DenseNet100\end{tabular}} 
& CIFAR10 & 99.9$\pm$0.1 / 99.9$\pm$0.1 / 99.9$\pm$0.1 & 95.3$\pm$0.2 / 95.2$\pm$0.3 / 95.3$\pm$0.1\\
& CIFAR100 & 99.9$\pm$0.1 / 99.0$\pm$0.1 / 99.9$\pm$0.1 & 77.2$\pm$0.3 / 77.3$\pm$0.4 / 77.0$\pm$0.3\\
& SVHN & 97.0$\pm$0.2 / 97.8$\pm$0.2 / 97.0$\pm$0.3 & 96.5$\pm$0.2 / 96.6$\pm$0.3 / 96.5$\pm$0.2\\
\midrule
\multirow{3}{*}{\begin{tabular}[c]{@{}c@{}}ResNet110\end{tabular}} 
& CIFAR10 & 99.9$\pm$0.1 / 99.9$\pm$0.1 / 99.9$\pm$0.1 & 94.4$\pm$0.3 / 94.5$\pm$0.3 / 94.6$\pm$0.2\\
& CIFAR100 & 99.5$\pm$0.1 / 99.9$\pm$0.1 / 99.8$\pm$0.1 & 72.7$\pm$0.2 / 74.1$\pm$0.4 / 73.9$\pm$0.3\\
& SVHN & 99.8$\pm$0.1 / 99.9$\pm$0.1 / 99.5$\pm$0.1 & 96.6$\pm$0.3 / 96.8$\pm$0.2 / 96.9$\pm$0.3\\
\bottomrule
\end{tabularx}
\small
\begin{justify}
Source: The Authors (2022). Besides preserving classification accuracy compared with SoftMax loss- and IsoMax loss-trained networks, IsoMax+ loss-trained models show higher OOD detection performance. Results are shown as means and standard deviations of five different iterations (see Table~\ref{tbl:fair_odd}).
\end{justify}
\end{table*}

Table~\ref{tab:classification_performance} shows classification accuracies. We see that IsoMax+ loss never exhibits \emph{classification accuracy drop} compared to SoftMax loss or IsoMax loss regardless of the dataset and model. The IsoMax loss variants achieve more than one percent better accuracy than the SoftMax loss when using ResNet110 on the CIFAR100 dataset.

\subsection{Out-of-Distribution Detection}

Table~\ref{tbl:fair_odd} summarizes the results of the \emph{fair} OOD detection comparison. We report the results using the entropic score for SoftMax loss (SoftMax$_{\text{ES}}$) and IsoMax loss (IsoMax$_{\text{ES}}$) because this score always overcame the maximum probability score in these cases. For IsoMax+, we report the values using the minimum distance score (IsoMax+$_{\text{MDS}}$) because this method overcame the maximum probability and the entropic score in this situation.

\looseness=-1
All approaches are accurate (i.e., exhibit no \emph{classification accuracy drop}); fast and power-efficient (i.e., inferences are performed without \emph{input preprocessing}); and no hyperparameter tuning was performed. Additionally, no additional/outlier/background data are required. IsoMax+$_{\text{MDS}}$ always overcomes IsoMax$_{\text{ES}}$ performance, regardless of the model, dataset, and out-of-distribution data.

\begin{figure*}
\centering
\small
\caption[Minimum Distance Score Analyses]{Minimum Distance Score Analyses}
\subfloat[]{\includegraphics[width=0.8\textwidth]{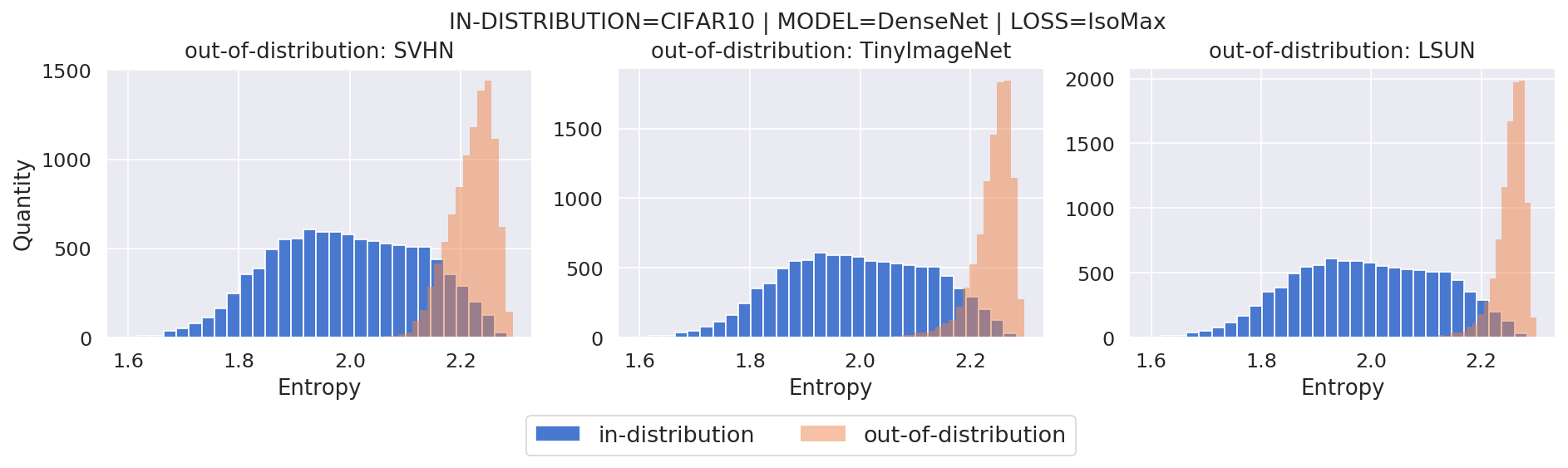}}
\\
\subfloat[]{\includegraphics[width=0.8\textwidth]{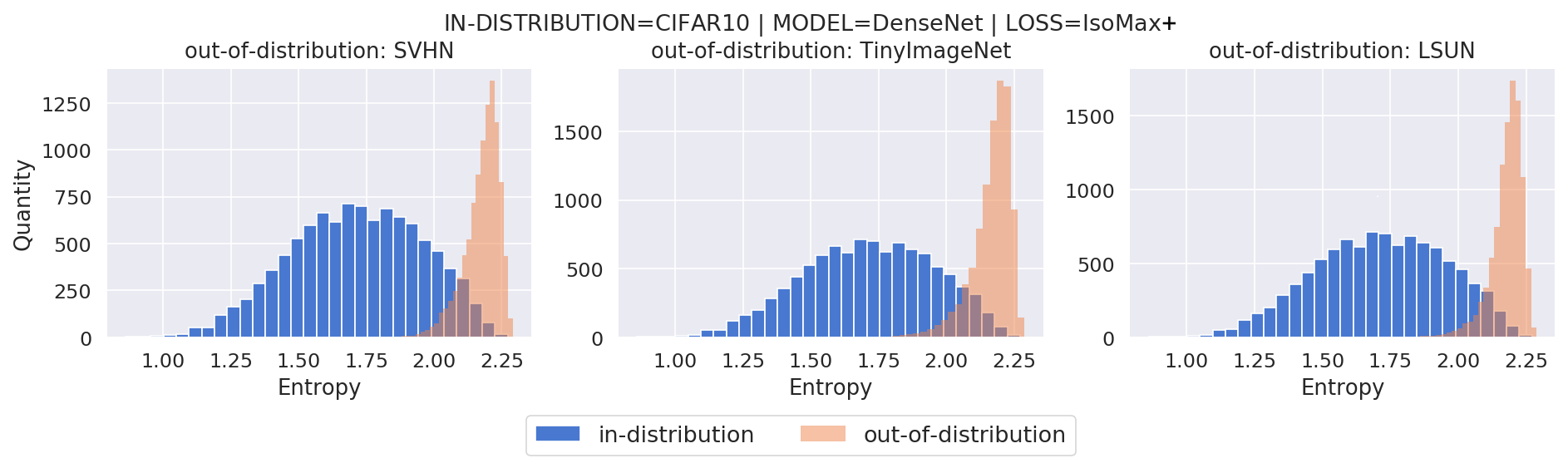}}
\\
\subfloat[]{\includegraphics[width=0.8\textwidth]{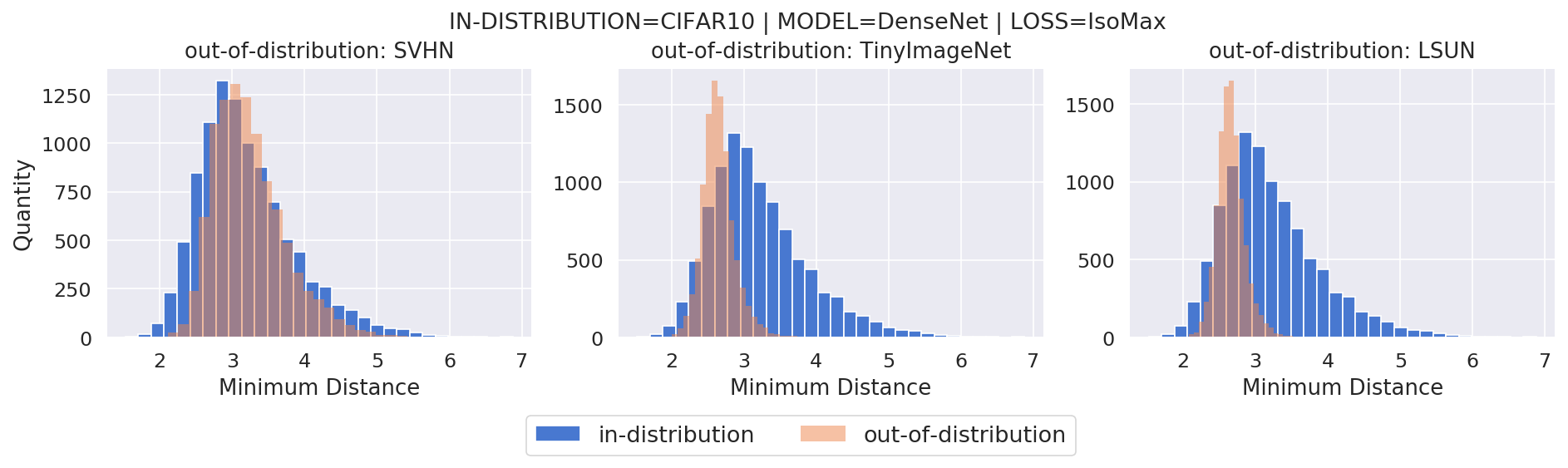}}
\\
\subfloat[]{\includegraphics[width=0.8\textwidth]{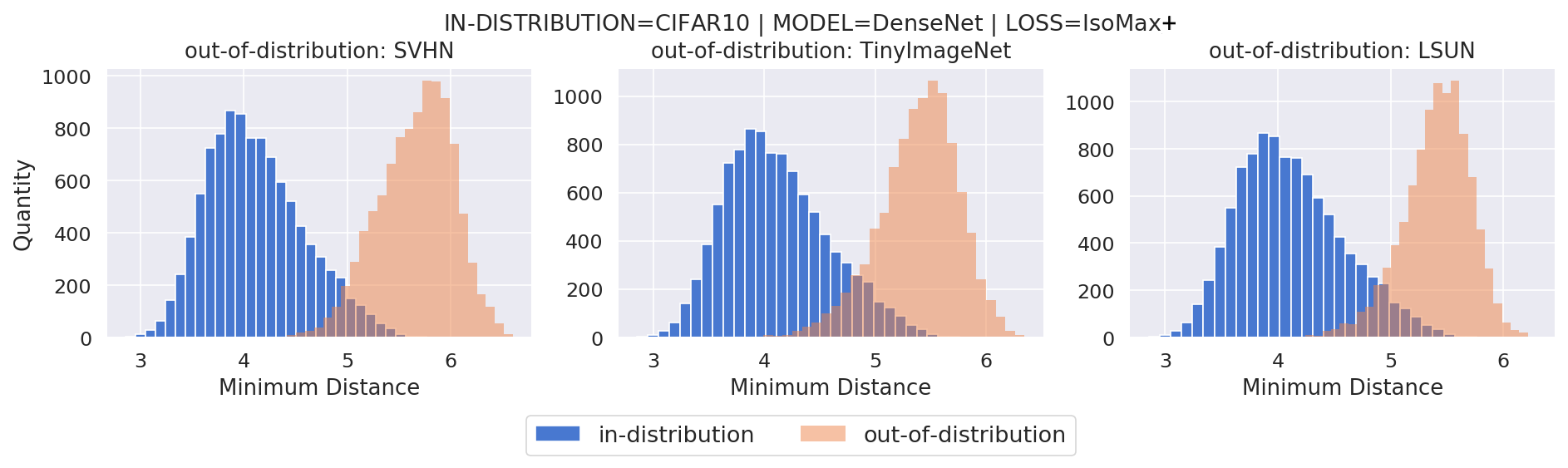}}
\\
\begin{justify}
{Source: The Authors (2022). In agreement with other studies \citep{DBLP:conf/nips/LiuLPTBL20}, (a) IsoMax (b) and IsoMax+ produce higher entropy/uncertainty on out-of-distributions than in-distribution. Therefore, the entropic score produces high OOD detection performance in both cases. (c) However, IsoMax does not make the in-distribution closer to the prototypes than out-of-distributions. (d) Concurrently, by introducing \emph{distance isometrization}, IsoMax+ gets in-distribution closer to the prototypes while pushing out-of-distribution data far away, which is what we expect based on the findings of other recent studies \citep{DBLP:conf/nips/LiuLPTBL20}. This result also explains why the minimum distance score produces high performance when using IsoMax+,  while producing low performance when using IsoMax. See also Table~\ref{tbl:diff_scores}.}
\end{justify}
\label{fig:histograms}
\end{figure*}

\looseness=-2
The minimum distance score produces high OOD detection performance when combined with IsoMax+, which shows that the isometrization of the distances works appropriately in this case. However, the same minimum distance score produced low OOD detection performance when combined with the original IsoMax loss. Fig.~\ref{fig:histograms}, and Table~\ref{tbl:diff_scores} provide an explanation for this important fact.

Table~\ref{tbl:unfair_odd_isomaxplus} summarizes the results of an \emph{unfair} OOD detection comparison because the methods have different requirements and produce distinct side effects. ODIN \citep{liang2018enhancing} and the Mahalanobis\footnote{Considering that the proposed approach outperforms the vanilla Mahalanobis method (i.e., training a Mahalanobis distance-based classifier using features extracted from the neural network and using the Mahalanobis distance as score), we use the term Mahalanobis approach to refer to the full Mahalanobis approach.} \citep{lee2018simple} approaches require adversarial samples to be generated to validate hyperparameters for each combination of dataset and model. These approaches also use \emph{input preprocessing}, \emph{which makes inferences at least four times slower and less energy-efficient} \citep{macdo2019isotropic, DBLP:journals/corr/abs-2006.04005}. Validation using adversarial examples may be cumbersome when performed from scratch on novel datasets because hyperparameters such as optimal adversarial perturbations may be unknown in such cases. IsoMax+$_{\text{MDS}}$ does not have these requirements, and does not produce the side effects.

Additionally, IsoMax+$_{\text{MDS}}$ achieves higher performance than ODIN by a large margin. In addition to the differences between the entropy maximization trick and temperature calibrations present in \cite{macdo2019isotropic, DBLP:journals/corr/abs-2006.04005}, we emphasize that training with an entropic scale affects the learning of all weights, while changing the temperature during inference affects only the last layer. Thus, the fact that the proposed solution overcomes ODIN by a safe margin is evidence that the \emph{entropy maximization trick produces much higher OOD detection performance than temperature calibration}, even when the latter is combined with input preprocessing. Moreover, the entropy maximization trick does not require access to validation data to tune the temperature.

In addition to being seamless and avoiding the drawbacks of the Mahalanobis approach, IsoMax+$_{\text{MDS}}$ typically overcomes it in terms of AUROC and produces similar performance when considering the DTACC.

Table~\ref{tbl:comp_oe} \emph{unfairly} compares the performance of the proposed approach with the outlier exposure solution. Similar to IsoMax variants, the outlier exposure approach does not require hyperparameter tuning and produces efficient inferences. However, it does require collecting outlier data, while our approach does not. We emphasize that outlier exposure may also be combined with IsoMax loss variants to further increase the OOD detection performance \citep{DBLP:journals/corr/abs-2006.04005}. In the table, we present the IsoMax loss variants \emph{without outlier exposure} to show that the outlier exposure-enhanced SoftMax loss typically achieves worse OOD detection than IsoMax+$_{\text{MDS}}$ \emph{even without using outlier exposure}.

The minimum distance score produces high OOD detection performance when combined with IsoMax+, which shows that the isometrization of the distances works appropriately in this case. However, the same minimum distance score produced low OOD detection performance when combined with the original IsoMax loss. Fig.~\ref{fig:histograms}, and Table~\ref{tbl:diff_scores} provide explanations.

\begingroup
\begin{table*}
\footnotesize
\centering
\caption[IsoMax+: OOD Detection Performance]{IsoMax+: OOD Detection Performance}
\vskip -0.25cm
\begin{tabularx}{\textwidth}{lll|YY}
\toprule
\multirow{4}{*}{\begin{tabular}[c]{@{}c@{}}\\Model\end{tabular}} & \multirow{4}{*}{\begin{tabular}[c]{@{}c@{}}\\Data\\(training)\end{tabular}} & \multirow{4}{*}{\begin{tabular}[c]{@{}c@{}}\\OOD\\(unseen)\end{tabular}} & 
\multicolumn{2}{c}{Out-of-Distribution Detection:}\\
&&& \multicolumn{2}{c}{Seamless Approaches.}\\
\cmidrule{4-5}
&&& TNR@TPR95 (\%) [$\uparrow$] & AUROC (\%) [$\uparrow$] \\
&&& \multicolumn{2}{c}{SoftMax$_{\text{ES}}$ / IsoMax$_{\text{ES}}$ / IsoMax+$_{\text{MDS}}$ (ours)}\\
\midrule
\multirow{9}{*}{\begin{tabular}[c]{@{}c@{}}DenseNet100\end{tabular}}
& \multirow{3}{*}{\begin{tabular}[c]{@{}c@{}}CIFAR10\end{tabular}} 
& SVHN & 40.4$\pm$5.3 / 78.6$\pm$9.0 / \bf97.0$\pm$0.7 & 89.1$\pm$2.2 / 96.2$\pm$1.0 / \bf99.4$\pm$0.1\\
&& TinyImageNet & 58.0$\pm$3.7 / 83.9$\pm$3.6 / \bf92.6$\pm$2.4 & 94.0$\pm$0.6 / 97.1$\pm$0.4 / \bf98.5$\pm$0.3\\
&& LSUN & 64.6$\pm$1.7 / 90.3$\pm$1.4 / \bf94.3$\pm$1.4 & 95.2$\pm$0.4 / 98.0$\pm$0.3 / \bf99.1$\pm$0.2\\
\cmidrule{2-5} 
& \multirow{3}{*}{\begin{tabular}[c]{@{}c@{}}CIFAR100\end{tabular}} 
& SVHN & 21.9$\pm$2.8 / 29.6$\pm$3.7 / \bf78.2$\pm$4.1 & 78.4$\pm$3.3 / 88.8$\pm$2.8 / \bf96.3$\pm$1.3\\
&& TinyImageNet & 24.0$\pm$3.0 / 49.3$\pm$4.9 / \bf85.5$\pm$2.8 & 75.5$\pm$2.1 / 90.8$\pm$2.1 / \bf97.5$\pm$0.8\\
&& LSUN & 24.9$\pm$2.9 / 60.6$\pm$6.6 / \bf78.3$\pm$3.9 & 77.2$\pm$3.2 / 93.0$\pm$1.4 / \bf97.2$\pm$1.3\\
\cmidrule{2-5} 
& \multirow{3}{*}{\begin{tabular}[c]{@{}c@{}}SVHN\end{tabular}} 
& CIFAR10 & 85.2$\pm$3.3 / 93.4$\pm$1.1 / \bf95.1$\pm$0.4 & 97.3$\pm$0.3 / 98.4$\pm$0.2 / \bf99.1$\pm$0.1\\
&& TinyImageNet & 90.8$\pm$0.6 / 96.2$\pm$0.9 / \bf98.1$\pm$0.3 & 98.3$\pm$0.2 / 99.0$\pm$0.2 / \bf99.6$\pm$0.1\\
&& LSUN & 87.9$\pm$0.8 / 94.3$\pm$1.7 / \bf97.7$\pm$1.1 & 97.8$\pm$0.3 / 98.7$\pm$0.2 / \bf99.6$\pm$0.2\\
\midrule
\multirow{9}{*}{\begin{tabular}[c]{@{}c@{}}ResNet110\end{tabular}}
& \multirow{3}{*}{\begin{tabular}[c]{@{}c@{}}CIFAR10\end{tabular}} 
& SVHN & 35.6$\pm$5.0 / 68.7$\pm$5.5 / \bf85.3$\pm$3.9 & 88.9$\pm$0.9 / 94.8$\pm$1.5 / \bf98.1$\pm$0.9\\
&& TinyImageNet & 39.0$\pm$4.1 / 65.4$\pm$5.6 / \bf76.2$\pm$2.7 & 88.2$\pm$2.1 / 94.3$\pm$0.4 / \bf96.1$\pm$0.5\\
&& LSUN & 46.9$\pm$5.7 / 80.7$\pm$2.2 / \bf86.9$\pm$2.5 & 91.2$\pm$1.7 / 96.5$\pm$0.3 / \bf97.9$\pm$0.7\\
\cmidrule{2-5} 
& \multirow{3}{*}{\begin{tabular}[c]{@{}c@{}}CIFAR100\end{tabular}} 
& SVHN & 16.6$\pm$1.5 / 20.8$\pm$5.9 / \bf41.0$\pm$6.4 & 73.6$\pm$3.7 / 85.3$\pm$1.3 / \bf88.9$\pm$1.2\\
&& TinyImageNet & 15.6$\pm$2.1 / 22.8$\pm$1.9 / \bf44.7$\pm$5.9 & 71.5$\pm$1.7 / 81.3$\pm$2.3 / \bf88.0$\pm$2.9\\
&& LSUN & 16.5$\pm$3.3 / 22.9$\pm$3.5 / \bf46.1$\pm$4.3 & 72.6$\pm$4.2 / 83.3$\pm$2.0 / \bf89.1$\pm$2.4\\
\cmidrule{2-5} 
& \multirow{3}{*}{\begin{tabular}[c]{@{}c@{}}SVHN\end{tabular}} 
& CIFAR10 & 80.9$\pm$3.5 / 84.3$\pm$1.3 / \bf88.4$\pm$2.3 & 95.1$\pm$0.3 / 96.5$\pm$0.2 / \bf97.4$\pm$0.3\\
&& TinyImageNet & 84.0$\pm$3.2 / 87.4$\pm$2.2 / \bf93.4$\pm$1.8 & 96.6$\pm$0.7 / 96.7$\pm$0.6 / \bf98.5$\pm$0.7\\
&& LSUN & 81.4$\pm$2.3 / 84.7$\pm$2.1 / \bf90.2$\pm$2.2 & 96.1$\pm$0.3 / 95.8$\pm$0.4 / \bf97.8$\pm$0.3\\
\bottomrule
\end{tabularx}
\small
\begin{justify}
Source: The Authors (2022). Fair comparison of seamless out-of-distribution detection approaches: No hyperparameter tuning, no additional/outlier/background data, no classification accuracy drop, and no slow inferences. SoftMax$_{\text{ES}}$ means training using SoftMax loss and performing OOD detection using the entropic score (ES). IsoMax$_{\text{ES}}$ means training using IsoMax loss and performing OOD detection using the entropic score (ES). IsoMax+$_{\text{MDS}}$ means training using IsoMax+ loss and performing OOD detection using minimum distance score (MDS). Results are shown as the mean and standard deviation of five iterations. The best results are shown in bold. See Table~\ref{tab:classification_performance}.
\end{justify}
\label{tbl:fair_odd}
\end{table*}
\endgroup

\begingroup
\begin{table*}
\footnotesize
\centering
\caption[IsoMax+: Unfair Comparison with No Seamless Approaches]{IsoMax+: Unfair Comparison with No Seamless Approaches}
\vskip -0.25cm
\begin{tabularx}{\textwidth}{lll|YY}
\toprule
\multirow{4}{*}{\begin{tabular}[c]{@{}c@{}}\\Model\end{tabular}} & \multirow{4}{*}{\begin{tabular}[c]{@{}c@{}}\\Data\\(training)\end{tabular}} & \multirow{4}{*}{\begin{tabular}[c]{@{}c@{}}\\OOD\\(unseen)\end{tabular}} & 
\multicolumn{2}{c}{Comparison with approaches that use}\\
&&& \multicolumn{2}{c}{input preprocessing and adversarial validation.}\\
\cmidrule{4-5}
&&& AUROC (\%) [$\uparrow$] & DTACC (\%) [$\uparrow$] \\
&&& \multicolumn{2}{c}{ODIN / Mahalanobis / IsoMax+$_{\text{MDS}}$ (ours)}\\
\midrule
\multirow{6}{*}{\begin{tabular}[c]{@{}c@{}}DenseNet100\end{tabular}}
& \multirow{3}{*}{\begin{tabular}[c]{@{}c@{}}CIFAR10\end{tabular}} 
& SVHN & 92.1$\pm$0.2 / 97.2$\pm$0.3 / \bf99.4$\pm$0.1 & 86.1$\pm$0.3 / 91.9$\pm$0.3 / \bf96.3$\pm$0.4\\
&& TinyImageNet & 97.2$\pm$0.3 / 97.7$\pm$0.2 / \bf98.5$\pm$0.3 & 91.9$\pm$0.3 / {\bf94.3$\pm$0.5} / \bf93.9$\pm$0.6\\
&& LSUN & 98.5$\pm$0.3 / 98.6$\pm$0.2 / \bf99.1$\pm$0.2 & 94.3$\pm$0.3 / {\bf95.7$\pm$0.4} / \bf95.3$\pm$0.5\\
\cmidrule{2-5} 
& \multirow{3}{*}{\begin{tabular}[c]{@{}c@{}}CIFAR100\end{tabular}} 
& SVHN & 88.0$\pm$0.5 / 91.3$\pm$0.4 / \bf96.3$\pm$1.3 & 80.0$\pm$0.6 / 84.3$\pm$0.4 / \bf90.3$\pm$0.5\\
&& TinyImageNet & 85.6$\pm$0.5 / {\bf96.7$\pm$0.3} / \bf97.5$\pm$0.8 & 77.6$\pm$0.5 / {\bf91.0$\pm$0.4} / {\bf91.3$\pm$0.3}\\
&& LSUN & 85.7$\pm$0.6 / {\bf97.1$\pm$1.9} / {\bf97.2$\pm$1.3} & 77.5$\pm$0.4 / {\bf92.5$\pm$0.8} / \bf91.7$\pm$0.7\\
\midrule
\multirow{6}{*}{\begin{tabular}[c]{@{}c@{}}ResNet34\end{tabular}}
& \multirow{3}{*}{\begin{tabular}[c]{@{}c@{}}CIFAR10\end{tabular}} 
& SVHN & 86.0$\pm$0.3 / 95.0$\pm$0.3 / \bf98.0$\pm$0.4 & 77.1$\pm$0.4 / 88.7$\pm$0.3 / \bf93.5$\pm$0.4\\
&& TinyImageNet & 92.6$\pm$0.3 / {\bf98.3$\pm$0.4} / 95.3$\pm$0.3 & 86.5$\pm$0.5 / {\bf94.8$\pm$0.3} / 90.0$\pm$0.4\\
&& LSUN & 93.0$\pm$0.4 / {\bf98.8$\pm$0.3} / 96.3$\pm$0.4 & 86.3$\pm$0.4 / {\bf96.8$\pm$0.4} / 92.1$\pm$0.5\\
\cmidrule{2-5} 
& \multirow{3}{*}{\begin{tabular}[c]{@{}c@{}}CIFAR100\end{tabular}} 
& SVHN & 71.0$\pm$0.4 / 84.0$\pm$0.6 / \bf88.0$\pm$0.7 & 68.0$\pm$0.5 / 77.3$\pm$0.7 / \bf82.1$\pm$0.4\\
&& TinyImageNet & 83.1$\pm$0.3 / 87.3$\pm$0.5 / \bf90.5$\pm$0.4 & 76.2$\pm$0.4 / {\bf84.0$\pm$0.4} / \bf84.2$\pm$0.5\\
&& LSUN & 81.0$\pm$0.3 / 82.0$\pm$0.5 / \bf88.6$\pm$0.6 & 75.2$\pm$0.3 / 79.3$\pm$0.5 / \bf82.5$\pm$0.4\\
\bottomrule
\end{tabularx}
\small
\begin{justify}
Source: The Authors (2022). Unfair comparison with approaches that use input preprocessing and produce slow/inefficient inferences in addition to requiring validation using adversarial examples. ODIN and Mahalanobis were applied to models trained using SoftMax loss. These approaches compute at least four times slower and less power efficient inferences \cite{macdo2019isotropic} because they use input preprocessing. Their hyperparameters were validated using adversarial examples. \textcolor{black}{Additionally, Mahalanobis requires feature extraction for training ad-hoc models to perform OOD detection. Finally, feature ensemble was also used.} IsoMax+$_{\text{MDS}}$ (ours) means training using IsoMax+ loss and performing OOD detection using minimum distance score (MDS). \textcolor{black}{No hyperparameter tuning is required when using IsoMax+ loss for training and the MDS for OOD detection.} \textcolor{black}{The IsoMax+ loss OOD detection performance shown in this table may be increased further without relying on input preprocessing or hyperparameter tuning by replacing the minimum distance score with the Mahalanobis \cite{lee2018simple} or the energy score~\cite{DBLP:journals/corr/abs-2010-03759}.} Results are the mean and standard deviation of five runs. The best results are shown in bold.
\end{justify}
\label{tbl:unfair_odd_isomaxplus}
\end{table*}
\endgroup

\begin{table*}
\footnotesize
\renewcommand{\arraystretch}{1.25}
\centering
\caption[IsoMax+: Unfair Comparison with Outlier Exposure]{IsoMax+: Unfair Comparison with Outlier Exposure}
\vskip -0.25cm
\begin{tabularx}{\textwidth}{ll|YY}
\toprule
\multirow{4}{*}{\begin{tabular}[c]{@{}c@{}}\\Model\end{tabular}} & 
\multirow{4}{*}{\begin{tabular}[c]{@{}c@{}}\\Data\\(training)\end{tabular}} & 
\multicolumn{2}{c}{Comparison of IsoMax loss variants without using additional data}\\
&& \multicolumn{2}{c}{with outlier exposure-enhanced SoftMax loss.}\\
\cmidrule{3-4}
&& TNR@TPR95 (\%) [$\uparrow$] & AUROC (\%) [$\uparrow$]\\
&& \multicolumn{2}{c}{SoftMax$^{\text{OE}}_{\text{ES}}$ / IsoMax$_{\text{ES}}$ / IsoMax+$_{\text{MDS}}$ (ours)}\\
\midrule
\multirow{2}{*}{\begin{tabular}[c]{@{}c@{}}DenseNet100\end{tabular}}
& \multirow{1}{*}{\begin{tabular}[c]{@{}c@{}}CIFAR10\end{tabular}} 
& {\bf94.4$\pm$1.4} / 84.2$\pm$4.6 / \bf94.6$\pm$1.5 & 98.0$\pm$0.3 / 97.1$\pm$0.5 / \bf99.0$\pm$0.2\\
& \multirow{1}{*}{\begin{tabular}[c]{@{}c@{}}CIFAR100\end{tabular}} 
& 36.4$\pm$9.4 / 46.5$\pm$5.0 / \bf80.6$\pm$3.6 & 83.8$\pm$5.6 / 90.8$\pm$2.1 / \bf97.0$\pm$1.1\\
\midrule
\multirow{2}{*}{\begin{tabular}[c]{@{}c@{}}ResNet110\end{tabular}}
& \multirow{1}{*}{\begin{tabular}[c]{@{}c@{}}CIFAR10\end{tabular}} 
& {\bf82.4$\pm$2.1} / 71.6$\pm$4.4 / \bf82.8$\pm$3.0 & {\bf96.8$\pm$0.7} / 95.2$\pm$0.7 / \bf97.3$\pm$0.7\\
& \multirow{1}{*}{\begin{tabular}[c]{@{}c@{}}CIFAR100\end{tabular}} 
& 29.1$\pm$4.5 / 22.1$\pm$3.7 / \bf43.9$\pm$5.5 & 80.5$\pm$2.8 / 83.3$\pm$1.8 / \bf88.6$\pm$2.1\\
\bottomrule
\end{tabularx}
\small
\begin{justify}
Source: The Author (2022). Unfair comparison of outlier exposure-enhanced SoftMax loss with IsoMax loss and IsoMax+ loss without using additional data. SoftMax$^{\text{OE}}_{\text{ES}}$ means training using SoftMax loss enhanced during training using outlier exposure \citep{hendrycks2018deep}, which requires the collection of outlier data, and performing OOD detection using the entropic score. We used the same outlier data used in \cite{hendrycks2018deep}. We collected the same amount of outlier data as the number of training examples present in the training set used to train SoftMax$^{\text{OE}}$. Despite being possible \citep{DBLP:journals/corr/abs-2006.04005}, the IsoMax loss and IsoMax+ loss were not enhanced by outlier exposure to keep the solution seamless. IsoMax$_{\text{ES}}$ means training using IsoMax loss and performing OOD detection using the entropic score. IsoMax+$_{\text{MDS}}$ (ours) means training using IsoMax+ loss and performing OOD detection using minimum distance score (MDS). The values of the performance metrics TNR@TPR95 and AUROC were averaged over all out-of-distribution (SVHN, TinyImageNet, and LSUN). Results are shown as the mean and standard deviation of five iterations. The best results are shown in bold.
\end{justify}
\label{tbl:comp_oe}
\end{table*}

\begingroup
\begin{table*}
\footnotesize
\centering
\caption[IsoMax Variants using Different Scores]{IsoMax Variants using Different Scores}
\vskip -0.25cm
\begin{tabularx}{\textwidth}{ll|YY}
\toprule
\multirow{4}{*}{\begin{tabular}[c]{@{}c@{}}\\Model\end{tabular}} & 
\multirow{4}{*}{\begin{tabular}[c]{@{}c@{}}\\Data\\(training)\end{tabular}} & 
\multicolumn{2}{c}{Comparison of IsoMax loss variants}\\
&& \multicolumn{2}{c}{using different scores.}\\
\cmidrule{3-4}
&& TNR@TPR95 (\%) [$\uparrow$] & AUROC (\%) [$\uparrow$]\\
&& \multicolumn{2}{c}{IsoMax$_{\text{ES}}$ / IsoMax$_{\text{MDS}}$ / IsoMax+$_{\text{ES}}$ / IsoMax+$_{\text{MDS}}$ (ours)}\\
\midrule
\multirow{2}{*}{\begin{tabular}[c]{@{}c@{}}DenseNet100\end{tabular}}
& \multirow{1}{*}{\begin{tabular}[c]{@{}c@{}}CIFAR10\end{tabular}} 
& 84.2$\pm$4.6 / 0.9$\pm$0.5 / 89.3$\pm$2.3 / \bf94.6$\pm$1.5 & 97.1$\pm$0.5 / 65.5$\pm$6.0 / 97.9$\pm$0.3 / \bf99.0$\pm$0.2\\
& \multirow{1}{*}{\begin{tabular}[c]{@{}c@{}}CIFAR100\end{tabular}} 
& 46.5$\pm$5.0 / 6.2$\pm$6.1 / 63.7$\pm$8.0 / \bf80.6$\pm$3.6 & 90.8$\pm$2.1 / 50.1$\pm$7.4 / 94.0$\pm$1.3 / \bf97.0$\pm$1.1\\
\midrule
\multirow{2}{*}{\begin{tabular}[c]{@{}c@{}}ResNet110\end{tabular}}
& \multirow{1}{*}{\begin{tabular}[c]{@{}c@{}}CIFAR10\end{tabular}} 
& 71.6$\pm$4.4 / 0.1$\pm$0.1 / 74.8$\pm$3.5 / \bf82.8$\pm$3.0 & 95.2$\pm$0.7 / 83.7$\pm$4.1 / 95.2$\pm$0.6 / \bf97.3$\pm$0.7\\
& \multirow{1}{*}{\begin{tabular}[c]{@{}c@{}}CIFAR100\end{tabular}} 
& 22.1$\pm$3.7 / 1.6$\pm$2.0 / 22.3$\pm$5.3 / \bf43.9$\pm$5.5 & 83.3$\pm$1.8 / 61.9$\pm$7.2 / 84.4$\pm$0.8 / \bf88.6$\pm$2.1\\
\bottomrule
\end{tabularx}
\small
\begin{justify}
Source: The Author (2022). Comparison of IsoMax variants using different scores. IsoMax$_{\text{ES}}$ means training using IsoMax loss and performing OOD detection using the entropic score (ES). IsoMax$_{\text{MDS}}$ means training using IsoMax loss and performing OOD detection using the minimum distance score (MDS). IsoMax+$_{\text{ES}}$ means training using IsoMax+ loss and performing OOD detection using entropic score (ES). IsoMax+$_{\text{MDS}}$ (ours) means training using IsoMax+ loss and performing OOD detection using minimum distance score (MDS). The values of the performance metrics TNR@TPR95 and AUROC were averaged over all out-of-distribution (SVHN, TinyImageNet, and LSUN). Results are shown as the mean and standard deviation of five iterations. The best results are shown in bold. See Fig.~\ref{fig:histograms}.
\end{justify}
\label{tbl:diff_scores}
\end{table*}
\endgroup

\newpage\section{Distinction Maximization Loss}

To allow standardized comparison, we used the datasets, training procedures, and metrics that were established in \cite{hendrycks2017baseline} and used in many subsequent papers \citep{liang2018enhancing, lee2018simple, Hein2018WhyRN}. We trained many 100-layer DenseNetBCs with growth rate $k\!=\!12$ (i.e., 0.8M parameters) \citep{Huang2017DenselyNetworks}, 34-layer ResNets \cite{He_2016}, and 28-layer WideResNets (widening factor $k\!=\!10$) \citep{DBLP:conf/bmvc/ZagoruykoK16} on the CIFAR10 \citep{Krizhevsky2009LearningImages} and CIFAR100 \citep{Krizhevsky2009LearningImages} datasets with SoftMax, IsoMax+, and DisMax losses using the same procedures (e.g., initial learning rate, learning rate schedule, weight decay). We used DisMax for DenseNetBC100 trained on CIFAR10 because this is a very small model and the mentioned dataset has too many examples per class; therefore, no augmentation is needed. We used DisMax\textsuperscript{$\dagger$} with $\alpha\!=\!1$ for all cases.

We used SGD with Nesterov moment equal to 0.9 with a batch size of 64 and an initial learning rate of 0.1. The weight decay was 0.0001, and we did not use dropout. We trained during 300 epochs. We used a learning rate decay rate equal to ten applied in epoch numbers 150, 200, and 250. 
We used resized images from TinyImageNet \citep{Deng2009ImageNetDatabase}, the Large-scale Scene UNderstanding dataset (LSUN) \citep{Yu2015LSUNLoop}, CIFAR10 \citep{Krizhevsky2009LearningImages}, CIFAR100 \citep{Krizhevsky2009LearningImages}, and SVHN \citep{Netzer2011ReadingLearning} to create OOD samples.

We added these out-of-distribution images to the validation sets of the ID data to form the test sets and evaluate the OOD detection performance. We evaluated the accuracy (ACC) to assess classification performance. We evaluated the OOD detection performance using the \gls{auroc}, the \gls{aupr}, and the true negative rate at a 95\% true positive rate (TNR@TPR95). We used the \gls{ece} \citep{DBLP:conf/aaai/NaeiniCH15,Guo2017OnCO,minderer2021revisiting} for uncertainty estimation performance. The results are the mean and standard deviation of five runs. Two methods are considered to produce the same performance if their mean performance difference is less than the sum of the error margins.

\begingroup
\begin{table}[!t]
\footnotesize
\centering
\caption[DisMax: Ablation Study]{DisMax: Ablation Study}
\label{tab:dismax_ablation}
\resizebox{\textwidth}{!}{%
\begin{tabular}{@{}clccccc@{}}
\\
\multicolumn{7}{c}{\small{CIFAR10}}\\
\toprule
\multirow{5}{*}{Model} & \multirow{5}{*}{Method} & \multirow{3}{*}{\begin{tabular}[c]{@{}c@{}}Score\end{tabular}} & \multicolumn{4}{c}{Out-of-Distribution Detection} \\
\cmidrule{4-7} 
&  & & Near & \multicolumn{2}{c}{Far} & Very Far \\
\cmidrule{4-7} 
&  &  & CIFAR100 & TinyImageNet & LSUN & SVHN \\
\cmidrule{3-7} 
& & MPS,MDS & TNR@95TPR & TNR@95TPR & TNR@95TPR & TNR@95TPR\\
& & MPS/MMLES & (\%) [$\uparrow$] & (\%) [$\uparrow$] & (\%) [$\uparrow$] & (\%) [$\uparrow$]\\
\midrule
\multirow{5}{*}{\shortstack{DenseNetBC100\\(small size)}}
& SoftMax (baseline) & MPS & 39.5$\pm$2.1 & 53.1$\pm$7.8 & 62.1$\pm$6.2 & 41.2$\pm$3.6\\
& IsoMax+ & MDS & \bf{57.3$\pm$1.1} & 86.9$\pm$0.4 & 91.4$\pm$0.3 & 96.4$\pm$0.6\\
& IsoMax+ with CutMix & MDS & 41.4$\pm$0.8 & 73.8$\pm$8.5 & 87.4$\pm$3.2 & 59.7$\pm$9.3\\
\cmidrule{2-7}
& \multirow{2}{*}{DisMax (ours)} & MDS & \bf{56.2$\pm$0.5} & \bf{88.0$\pm$0.3} & \bf{92.2$\pm$0.3} & 97.5$\pm$0.3\\
& & MPS/MMLES & 54.2$\pm$1.0 & \bf{89.0$\pm$1.0} & \bf{92.4$\pm$1.1} & \bf{98.3$\pm$0.3}\\
\midrule
\multirow{5}{*}{\shortstack{ResNet34\\(medium size)}}
& SoftMax (baseline) & MPS & 40.0$\pm$1.6 & 46.4$\pm$4.9 & 53.6$\pm$4.7 & 44.1$\pm$9.3\\
& IsoMax+ & MDS & 55.1$\pm$1.3 & 71.0$\pm$6.4 & 81.5$\pm$4.4 & 82.4$\pm$8.8\\
& IsoMax+ with CutMix & MDS & 40.2$\pm$8.0 & 62.4$\pm$4.3 & 77.4$\pm$4.5 & 78.6$\pm$7.2\\
\cmidrule{2-7}
& \multirow{2}{*}{DisMax\textsuperscript{$\dagger$} (ours)} & MDS & \bf{60.4$\pm$0.7} & \bf{92.0$\pm$1.5} & 97.2$\pm$0.4 & \bf{91.1$\pm$2.9}\\
& & MPS/MMLES & \bf{60.0$\pm$0.5} & \bf{93.3$\pm$1.1} & \bf{98.0$\pm$0.3} & \bf{91.2$\pm$2.7}\\
\midrule
\multirow{5}{*}{\shortstack{WideResNet2810\\(big size)}}
& SoftMax (baseline) & MPS & 44.9$\pm$0.7 & 53.4$\pm$3.3 & 59.2$\pm$3.6 & 50.1$\pm$5.2\\
& IsoMax+ & MDS & 61.5$\pm$0.4 & 80.2$\pm$4.2 & 87.4$\pm$3.0 & \bf{96.3$\pm$1.4}\\
& IsoMax+ with CutMix & MDS & 37.4$\pm$8.0 & 76.3$\pm$8.2 & 85.8$\pm$6.3 & 60.0$\pm$9.8\\
\cmidrule{2-7}
& \multirow{2}{*}{DisMax\textsuperscript{$\dagger$} (ours)} & MDS & 60.2$\pm$1.3 & \bf{98.4$\pm$0.4} & \bf{99.4$\pm$0.1} & \bf{93.8$\pm$1.7}\\
& & MPS/MMLES & \bf{62.9$\pm$0.5} & \bf{98.6$\pm$0.2} & \bf{99.5$\pm$0.1} & 92.8$\pm$2.0\\
\bottomrule
\\
\multicolumn{7}{c}{\small{CIFAR100}}\\
\toprule
\multirow{5}{*}{Model} & \multirow{5}{*}{Method} & \multirow{3}{*}{\begin{tabular}[c]{@{}c@{}}Score\end{tabular}} & \multicolumn{4}{c}{Out-of-Distribution Detection} \\
\cmidrule{4-7} 
&  & & Near & \multicolumn{2}{c}{Far} & Very Far \\
\cmidrule{4-7} 
&  &  & CIFAR10 & TinyImageNet & LSUN & SVHN \\
\cmidrule{3-7} 
& & MPS,MDS & TNR@95TPR & TNR@95TPR & TNR@95TPR & TNR@95TPR\\
& & MPS/MMLES & (\%) [$\uparrow$] & (\%) [$\uparrow$] & (\%) [$\uparrow$] & (\%) [$\uparrow$]\\
\midrule
\multirow{5}{*}{\shortstack{DenseNetBC100\\(small size)}}
& SoftMax (baseline) & MPS & 17.6$\pm$1.1 & 18.1$\pm$1.7 & 18.7$\pm$2.0 & 19.8$\pm$2.9\\
& IsoMax+ & MDS & 17.2$\pm$0.7 & 71.6$\pm$6.5 & 66.8$\pm$9.4 & \bf67.1$\pm$3.0\\
& IsoMax+ with CutMix & MDS & \bf{19.8$\pm$2.1} & 60.8$\pm$9.9 & 57.6$\pm$9.8 & 57.9$\pm$3.5\\
\cmidrule{2-7}
& \multirow{2}{*}{DisMax\textsuperscript{$\dagger$} (ours)} & MDS & 16.6$\pm$0.6 & 97.7$\pm$0.3 & 98.5$\pm$0.4 & 57.9$\pm$3.6\\
& & MPS/MMLES & \bf{22.1$\pm$1.1} & \bf{99.0$\pm$0.5} & \bf{99.4$\pm$0.3} & \bf66.6$\pm$2.6\\
\midrule
\multirow{5}{*}{\shortstack{ResNet34\\(medium size)}}
& SoftMax (baseline) & MPS & 19.4$\pm$0.5 & 20.6$\pm$2.4 & 21.3$\pm$3.4 & 17.1$\pm$5.0\\
& IsoMax+ & MDS & 18.0$\pm$0.7 & 43.3$\pm$4.3 & 41.5$\pm$5.7 & 43.6$\pm$3.5\\
& IsoMax+ with CutMix & MDS & 18.9$\pm$1.7 & 46.3$\pm$9.6 & 46.5$\pm$9.0 & 35.6$\pm$3.9\\
\cmidrule{2-7}
& \multirow{2}{*}{DisMax\textsuperscript{$\dagger$} (ours)} & MDS & 20.8$\pm$0.4 & 79.9$\pm$1.5 & 81.5$\pm$1.4 & 43.7$\pm$1.6\\
& & MPS/MMLES & \bf{22.0$\pm$0.5} & \bf{85.4$\pm$1.7} & \bf{86.4$\pm$1.3} & \bf{48.5$\pm$2.0}\\
\midrule
\multirow{5}{*}{\shortstack{WideResNet2810\\(big size)}}
& SoftMax (baseline) & MPS & 21.8$\pm$0.7 & 26.7$\pm$5.9 & 28.7$\pm$6.7 & 15.8$\pm$5.5\\
& IsoMax+ & MDS & 19.0$\pm$0.7 & 66.9$\pm$3.9 & 67.9$\pm$3.3 & 61.8$\pm$1.9\\
& IsoMax+ with CutMix & MDS & 21.5$\pm$1.9 & 52.5$\pm$9.4 & 52.0$\pm$9.1 & 33.3$\pm$9.2\\
\cmidrule{2-7}
& \multirow{2}{*}{DisMax\textsuperscript{$\dagger$} (ours)} & MDS & 22.4$\pm$0.2 & 92.3$\pm$1.3 & 95.2$\pm$0.4 & 56.8$\pm$1.8\\
& & MPS/MMLES & \bf{24.6$\pm$0.3} & \bf{96.3$\pm$1.2} & \bf{97.8$\pm$0.9} & \bf{65.6$\pm$1.2}\\
\bottomrule
\end{tabular}%
}
\small
\begin{justify}
Source: The Author (2022). MPS \citep{hendrycks2017baseline} means Maximum Probability Score (i.e., the standard for SoftMax loss). MDS indicates Minimum Distance Score (i.e., the standard for IsoMax+ loss \citep{macedo2021enhanced}). MMLES means Max-Mean Logit Entropy Score (i.e., the standard for DisMax loss for (very) far OOD detection). CutMix is from \cite{DBLP:conf/iccv/YunHCOYC19}. We used MPS for near OOD detection for DisMax, as this score provided the best results in this particular case. We emphasize that the MPS for DisMax is based on \emph{logits+ rather than usual logits}. The best performances are bold. All results can be reproduced using the provided code.
\end{justify}
\end{table}
\endgroup

\begingroup
\setlength{\tabcolsep}{8pt} 
\begin{table}
\footnotesize
\centering
\caption[DisMax: Classification, Efficiency, Uncertainty, and OOD Detection]{DisMax: Classification, Efficiency, Uncertainty, and OOD Detection}
\label{tab:dismax-comparative-results}
\resizebox{\textwidth}{!}{%
\begin{tabular}{@{}clccccccc@{}}
\multicolumn{9}{c}{\normalsize{CIFAR10}}\\
\toprule
\multirow{5}{*}{Model} & \multirow{5}{*}{Method} & \multirow{3}{*}{Classification} & \multirow{3}{*}{\begin{tabular}[c]{@{}c@{}}Inference\end{tabular}} & \multirow{3}{*}{\begin{tabular}[c]{@{}c@{}}Uncertainty\\ Estimation\end{tabular}} & \multicolumn{4}{c}{Out-of-Distribution Detection} \\ \cmidrule{6-9} 
&  &  &  &  & Near & \multicolumn{2}{c}{Far} & Very Far \\ \cmidrule{6-9} 
&  &  &  &  & CIFAR100 & TinyImageNet & LSUN & SVHN \\ 
\cmidrule{3-9} 
&  & ACC & Efficiency & ECE & AUPR & AUROC & AUROC & AUPR\\
&  & (\%) [$\uparrow$] & (\%) [$\uparrow$] & [$\downarrow$] & (\%) [$\uparrow$] & (\%) [$\uparrow$] & (\%) [$\uparrow$] & (\%) [$\uparrow$]\\
\midrule
\multirow{5}{*}{\shortstack{DenseNetBC100\\(small size)}}
& SoftMax (baseline) & \bf{95.2$\pm$0.1} & \bf100.0 & \bf{0.0043$\pm$0.0008} & 86.2$\pm$0.5 & 92.9$\pm$1.6 & 94.7$\pm$0.9 & 93.7$\pm$3.3\\
& Scaled Cosine & \textcolor{black}{94.9$\pm$0.1} & \bf100.0 & - & - & \bf{98.8$\pm$0.3} & \bf{99.2$\pm$0.2} & -\\
& GODIN with preprocessing & \textcolor{black}{95.0$\pm$0.1} & \textcolor{black}{26.0} & - & - & \bf{99.1$\pm$0.1} & \bf{99.4$\pm$0.1} & -\\
& IsoMax+ & \bf{95.1$\pm$0.1} & \bf100.0 & \bf{0.0043$\pm$0.0012} & \bf{90.4$\pm$0.3} & 97.6$\pm$0.9 & 98.3$\pm$0.5 & 99.7$\pm$0.1\\
& DisMax (ours) & \bf{95.1$\pm$0.1} & \bf100.0 & \bf{0.0045$\pm$0.0021} & \bf{90.0$\pm$0.2} & 98.0$\pm$0.5 & 98.4$\pm$0.3 & \bf{99.9$\pm$0.1}\\
\midrule
\multirow{4}{*}{\shortstack{ResNet34\\(medium size)}}
& SoftMax (baseline) & 95.6$\pm$0.1 & \bf100.0 & \bf{0.0060$\pm$0.0013} & 85.3$\pm$0.4 & 89.7$\pm$2.8 & 92.4$\pm$1.6 & 94.9$\pm$1.0\\
& GODIN & \textcolor{black}{95.1$\pm$0.1} & \bf100.0 & - & - & 95.6$\pm$0.5 & 97.6$\pm$0.2 & -\\
& IsoMax+ & 95.5$\pm$0.1 & \bf100.0 & \bf{0.0053$\pm$0.0007} & \bf{90.1$\pm$0.3} & 95.1$\pm$1.0 & 96.9$\pm$0.6 & \bf{98.7$\pm$0.6}\\
& DisMax\textsuperscript{$\dagger$} (ours) & \bf{96.7$\pm$0.2} & \bf100.0 & \bf{0.0058$\pm$0.0008} & \bf{90.3$\pm$0.2} & \bf{98.3$\pm$0.3} & \bf{99.5$\pm$0.1} & \bf{99.1$\pm$0.3}\\
\midrule
\multirow{7}{*}{\shortstack{WideResNet2810\\(big size)}}
& SoftMax (baseline) & 96.2$\pm$0.1 & \bf100.0 & \bf{0.0038$\pm$0.0005} & 87.5$\pm$0.3 & 92.6$\pm$0.9 & 94.0$\pm$0.7 & 95.3$\pm$0.9\\
& Deep Ensemble & 96.6$\pm$0.1 & \textcolor{black}{10.3} & 0.0100$\pm$0.0010 & 88.8$\pm$1.0 & - & - & 96.4$\pm$1.0\\
& DUQ & \textcolor{black}{94.7$\pm$0.1} & \textcolor{black}{45.0} & 0.0340$\pm$0.0020 & 85.4$\pm$1.0 & - & - & 97.3$\pm$1.0\\
& SNGP & \textcolor{black}{95.9$\pm$0.1} & \textcolor{black}{62.5} & 0.0180$\pm$0.0010 & 90.5$\pm$1.0 & - & - & 99.0$\pm$1.0\\
& Scaled Cosine &  \textcolor{black}{95.7$\pm$0.1} & \bf100.0 & - & - & 97.7$\pm$0.7 & 98.6$\pm$0.3 & -\\
& IsoMax+ & 96.0$\pm$0.1 & \bf100.0 & \bf{0.0034$\pm$0.0009} & \bf{91.8$\pm$0.1} & 96.6$\pm$0.6 & 97.7$\pm$0.4 & \bf{99.7$\pm$0.3}\\
& DisMax\textsuperscript{$\dagger$} (ours) & \bf{97.0$\pm$0.1} & \bf100.0 & \bf{0.0043$\pm$0.0008} & 90.1$\pm$0.3 & \bf{99.7$\pm$0.1} & \bf{99.9$\pm$0.1} & \bf{99.3$\pm$0.3}\\
\bottomrule
\\
\multicolumn{9}{c}{\normalsize{CIFAR100}}\\
\toprule
\multirow{5}{*}{Model} & \multirow{5}{*}{Method} & \multirow{3}{*}{Classification} & \multirow{3}{*}{\begin{tabular}[c]{@{}c@{}}Inference\end{tabular}} & \multirow{3}{*}{\begin{tabular}[c]{@{}c@{}}Uncertainty\\ Estimation\end{tabular}} & \multicolumn{4}{c}{Out-of-Distribution Detection} \\ \cmidrule{6-9} 
&  &  &  &  & Near & \multicolumn{2}{c}{Far} & Very Far \\ \cmidrule{6-9} 
&  &  &  &  & CIFAR10 & TinyImageNet & LSUN & SVHN \\
\cmidrule{3-9} 
&  & ACC & Efficiency & ECE & AUPR & AUROC & AUROC & AUPR\\
&  & (\%) [$\uparrow$] & (\%) [$\uparrow$] & [$\downarrow$] & (\%) [$\uparrow$] & (\%) [$\uparrow$] & (\%) [$\uparrow$] & (\%) [$\uparrow$]\\
\midrule
\multirow{5}{*}{\shortstack{DenseNetBC100\\(small size)}}
& SoftMax (baseline) & 77.3$\pm$0.4 & \bf100.0 & 0.0155$\pm$0.0026 & 71.3$\pm$0.8 & 71.8$\pm$2.2 & 73.1$\pm$2.4 & 87.5$\pm$1.5\\
& Scaled Cosine & \textcolor{black}{75.7$\pm$0.1} & \bf100.0 & - & - & 97.8$\pm$0.5 & 97.6$\pm$0.8 & -\\
& GODIN with preprocessing & \textcolor{black}{75.9$\pm$0.1} & \textcolor{black}{24.0} & - & - & 98.6$\pm$0.2 & 98.7$\pm$0.0 & -\\
& IsoMax+ & 76.9$\pm$0.3 & \bf100.0 & \bf{0.0108$\pm$0.0017} & 71.3$\pm$0.4 & 95.1$\pm$1.1 & 94.2$\pm$1.7 & \bf{97.4$\pm$0.6}\\
& DisMax\textsuperscript{$\dagger$} (ours) & \bf{79.4$\pm$0.2} & \bf100.0 & 0.0154$\pm$0.0006 & \bf{74.4$\pm$0.2} & \bf{99.8$\pm$0.1} & \bf{99.9$\pm$0.1} & \bf{96.4$\pm$0.8}\\
\midrule
\multirow{5}{*}{\shortstack{ResNet34\\(medium size)}}
& SoftMax (baseline) & 77.7$\pm$0.3 & \bf100.0 & 0.0268$\pm$0.0015 & 73.3$\pm$0.1 & 79.0$\pm$2.1 & 79.6$\pm$1.7 & 86.3$\pm$3.3\\
& GODIN & \textcolor{black}{75.8$\pm$0.2} & \bf100.0 & - & - & 91.8$\pm$1.1 & 92.0$\pm$0.7 & -\\
& GODIN with dropout & \textcolor{black}{77.2$\pm$0.1} & \bf100.0 & - & - & \textcolor{black}{87.0$\pm$1.1} & \textcolor{black}{87.0$\pm$2.2} & -\\
& IsoMax+ & \textcolor{black}{76.5$\pm$0.3} & \bf100.0 & 0.0190$\pm$0.0025 & 72.1$\pm$0.4 & 89.7$\pm$1.0 & 89.8$\pm$1.3 & \bf{94.5$\pm$0.6}\\
& DisMax\textsuperscript{$\dagger$} (ours) & \bf{80.6$\pm$0.3} & \bf100.0 & \bf{0.0116$\pm$0.0014} & \bf{74.2$\pm$0.6} & \bf{97.6$\pm$0.5} & \bf{97.7$\pm$0.6} & \bf{94.8$\pm$1.0}\\
\midrule
\multirow{7}{*}{\shortstack{WideResNet2810\\(big size)}}
& SoftMax (baseline) & 79.9$\pm$0.2 & \bf100.0 & 0.0272$\pm$0.0032 & 75.4$\pm$0.5 & 81.7$\pm$2.3 & 82.7$\pm$2.2 & 86.0$\pm$2.6\\
& Deep Ensemble & 80.2$\pm$0.1 & \textcolor{black}{12.3} & 0.0210$\pm$0.0040 & 78.0$\pm$1.0 & - & - & 88.8$\pm$1.0\\
& DUQ & \textcolor{black}{78.5$\pm$0.1} & \textcolor{black}{79.9} & 0.1190$\pm$0.0010 & 73.2$\pm$1.0 & - & - & 87.8$\pm$1.0\\
& SNGP & 79.9$\pm$0.1 & \textcolor{black}{74.9} & 0.0250$\pm$0.0120 & \bf{80.1$\pm$1.0} & - & - & 92.3$\pm$1.0\\
& Scaled Cosine & \textcolor{black}{78.5$\pm$0.3} & \bf100.0 & - & - & 95.8$\pm$0.7 & 95.2$\pm$0.8 & -\\
& IsoMax+ & \textcolor{black}{79.5$\pm$0.1} & \bf100.0 & 0.0188$\pm$0.0016 & 73.0$\pm$0.8 & 94.2$\pm$2.1 & 94.6$\pm$2.0 & \bf{96.7$\pm$1.7}\\
& DisMax\textsuperscript{$\dagger$} (ours) & \bf{83.0$\pm$0.1} & \bf100.0 & \bf{0.0143$\pm$0.0027} & 76.0$\pm$1.0 & \bf{99.4$\pm$0.2} & \bf{99.6$\pm$0.1} & \bf{97.0$\pm$1.5}\\
\bottomrule
\end{tabular}%
}
\small 
\begin{justify}
{Source: The Author (2022). In this table, \mbox{efficiency} represents the inference speed (i.e., the inverse of the inference delay) calculated as a percentage of the performance of a single deterministic neural network trivially trained. For a fair comparison, we also calibrated the temperature of the SoftMax loss and IsoMax+ loss approaches using the same procedure that we defined for DisMax loss. Considering that input preprocessing can be applied indistinctly to improve the OOD detection performance of all methods compared \citep{Hsu2020GeneralizedOD} (at the cost of making their inferences approximately four times less efficient \citep{DBLP:journals/corr/abs-2006.04005}), unless explicitly mentioned otherwise, all results are presented without using input preprocessing. The methods that present the best performances are bold. SoftMax (baseline) was proposed in \cite{hendrycks2017baseline} and IsoMax+ in \cite{macedo2021enhanced}. Results for Scaled Cosine are from Scaled Cosine paper \citep{techapanurak2019hyperparameterfree}. Results for GODIN are from GODIN paper \citep{Hsu2020GeneralizedOD}. Results for Deep Ensemble \citep{lakshminarayanan2017simple}, DUQ \citep{Amersfoort2020SimpleAS}, and SNGP are from SNGP paper \citep{DBLP:conf/nips/LiuLPTBL20}.}
\end{justify}
\end{table}
\endgroup

\subsection{Ablation Study}

\looseness=-1
Table \ref{tab:dismax_ablation} shows that logits+ and FPR often improve the accuracy and OOD detection performance compared to IsoMax+ even when using MDS. It also shows that replacing MDS with the \emph{composite} score MMLES often increases OOD detection. These conclusions are essentially true regardless of the model, in-distribution, and (near, far, and very far) out-of-distribution.

Finally, we performed experiments \emph{combining IsoMax+ with CutMix}. However, adding CutMix to IsoMax+ did \emph{not} significantly increase the OOD detection performance. Often, the performance actually \emph{decreased}. Therefore, DisMax easily outperformed IsoMax+ even when the latter was combined with CutMix.

\clearpage
\subsection{Classification, Efficiency, Uncertainty, and OOD Detection Results}

Table \ref{tab:dismax-comparative-results} compares DisMax with major approaches such as Scaled Cosine \citep{techapanurak2019hyperparameterfree}, GODIN \citep{Hsu2020GeneralizedOD}, Deep Ensemble \citep{lakshminarayanan2017simple}, DUQ \citep{Amersfoort2020SimpleAS}, and SNGP \citep{DBLP:conf/nips/LiuLPTBL20} regarding classification accuracy, inference efficiency, uncertainty estimation, and (near, far, and very far) out-of-distribution detection. Unlike other approaches, DisMax is as inference efficient as a trivially trained neural network using the usual SoftMax loss. Furthermore, DisMax often outperforms other approaches simultaneously in all evaluated metrics.

\subsection{Max-Mean Logit Entropy Score Analyses}

Fig.~\ref{fig:histograms} shows the distribution of \emph{mean} logits+ under some scenarios. We see that prototypes are, on average, usually closer to in-distribution examples than out-of-distribution examples, which explains why the \emph{mean} enhanced logit improves OOD detection performance when combined with the maximum logit+ and the negative entropy to compose the MMLES. In other words, even prototypes \emph{that are not associated with the class of a given in-distribution example} are usually closer to it than they are to out-of-distribution examples. 

\begin{figure*}
\small
\centering
\caption[Max-Mean Logit Entropy Score Analyses]{Max-Mean Logit Entropy Score Analyses}
\includegraphics[width=\textwidth]{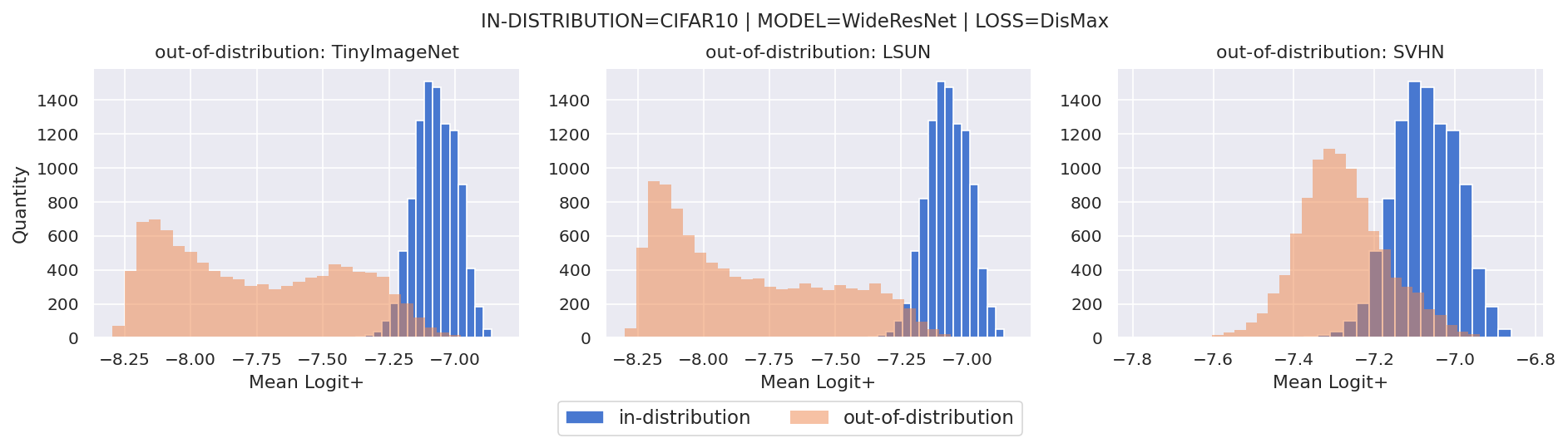}
\begin{justify}
{Source: The Author (2022). In the feature space, the mean distance from an in-distribution image to \emph{all} prototypes is usually smaller than the mean distance from an out-of-distribution image to \emph{all} prototypes. For example, consider a given class present in CIFAR10. This figure shows that even prototypes associated with classes \emph{other than the selected class} are usually closer to images of the assumed class (in-distribution in blue) than images that do not belong to CIFAR10 at all (out-of-distributions in orange). This explains why the \emph{mean value} of logits+ considering all prototypes contributes to the OOD detection performance. Therefore, not only the distance to the nearest prototype is used in the mentioned task.}
\end{justify}
\label{fig:mmles_analyses}
\end{figure*}

\newpage
\subsection{Loss Landscape Study}

\begin{figure*}
\small
\centering
\caption[Loss Surface Study]{Loss Surface Study}
\subfloat[]{\includegraphics[height=3.5cm]{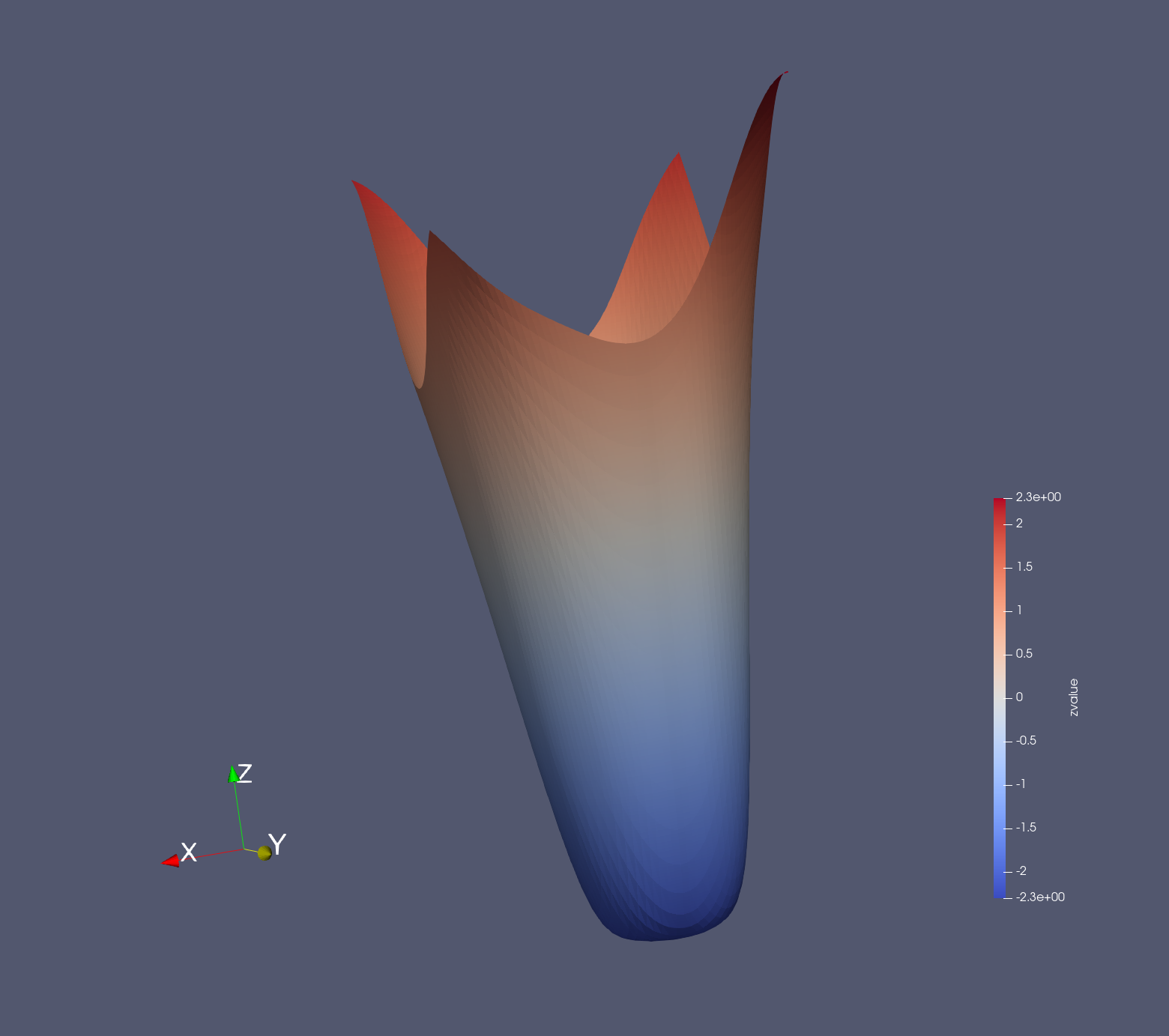}}
\hspace{0.05cm}
\subfloat[]{\includegraphics[height=3.5cm]{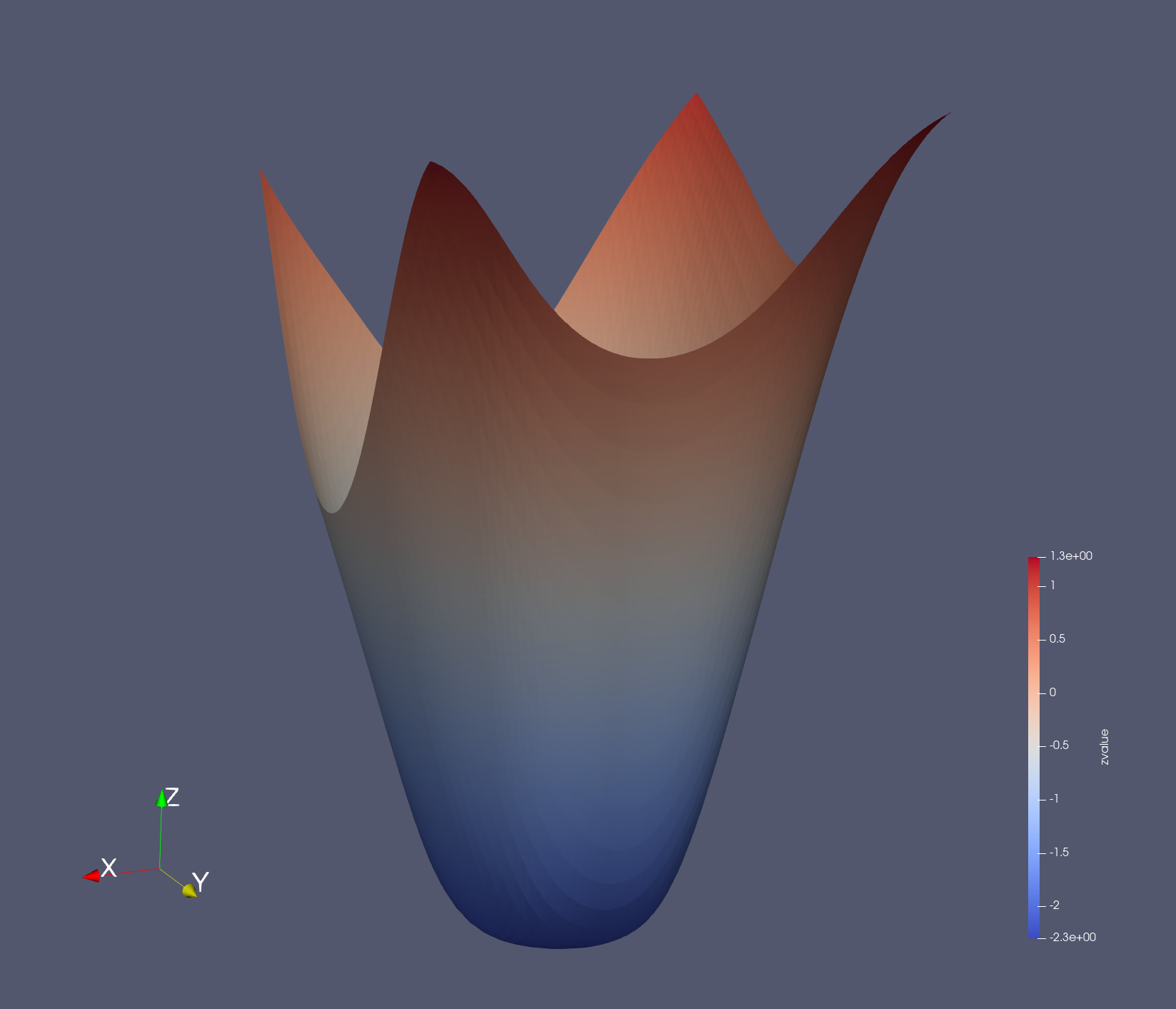}}
\hspace{0.05cm}
\subfloat[]{\includegraphics[height=3.5cm]{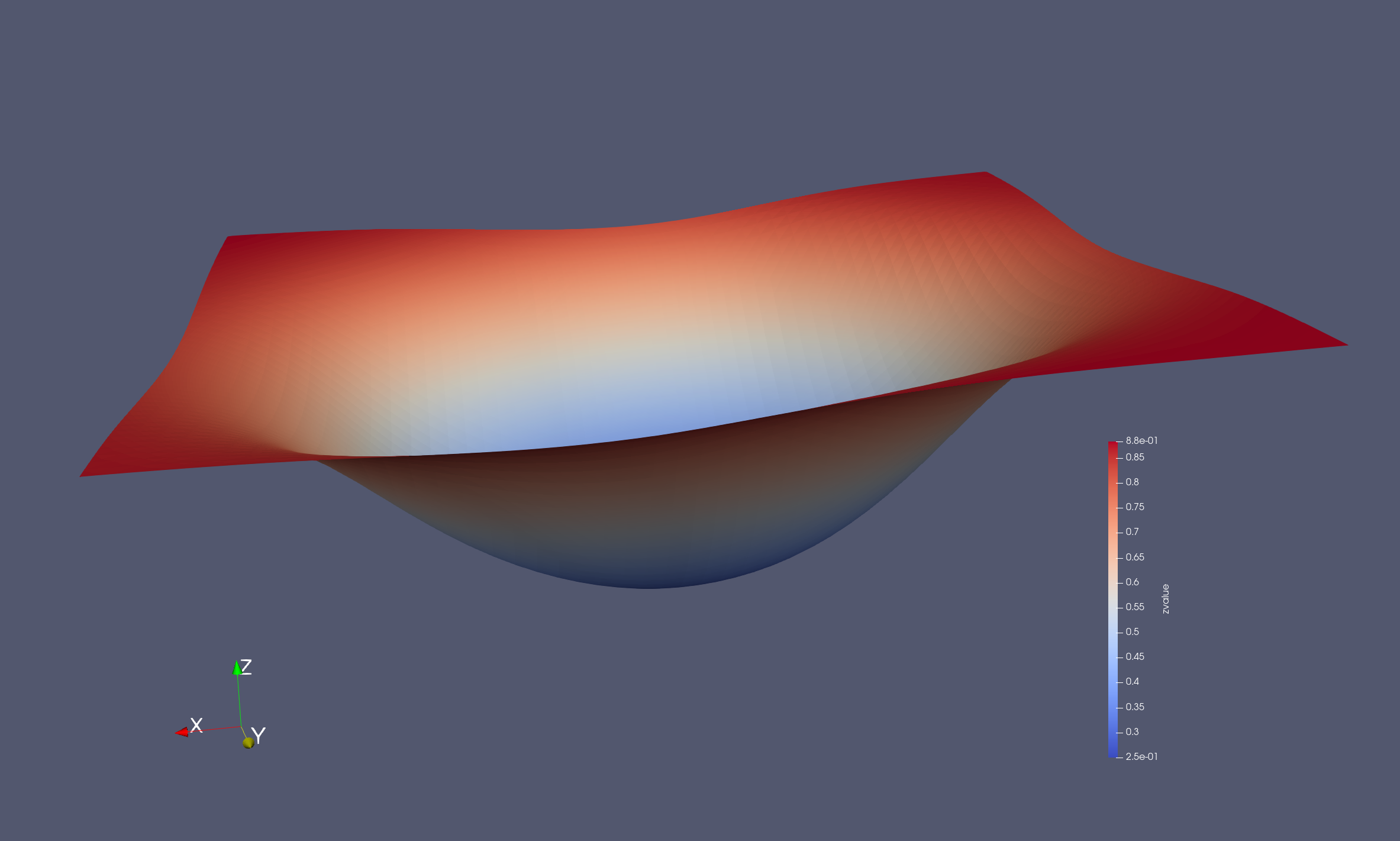}}
\\
\subfloat[]{\includegraphics[width=0.33\textwidth]{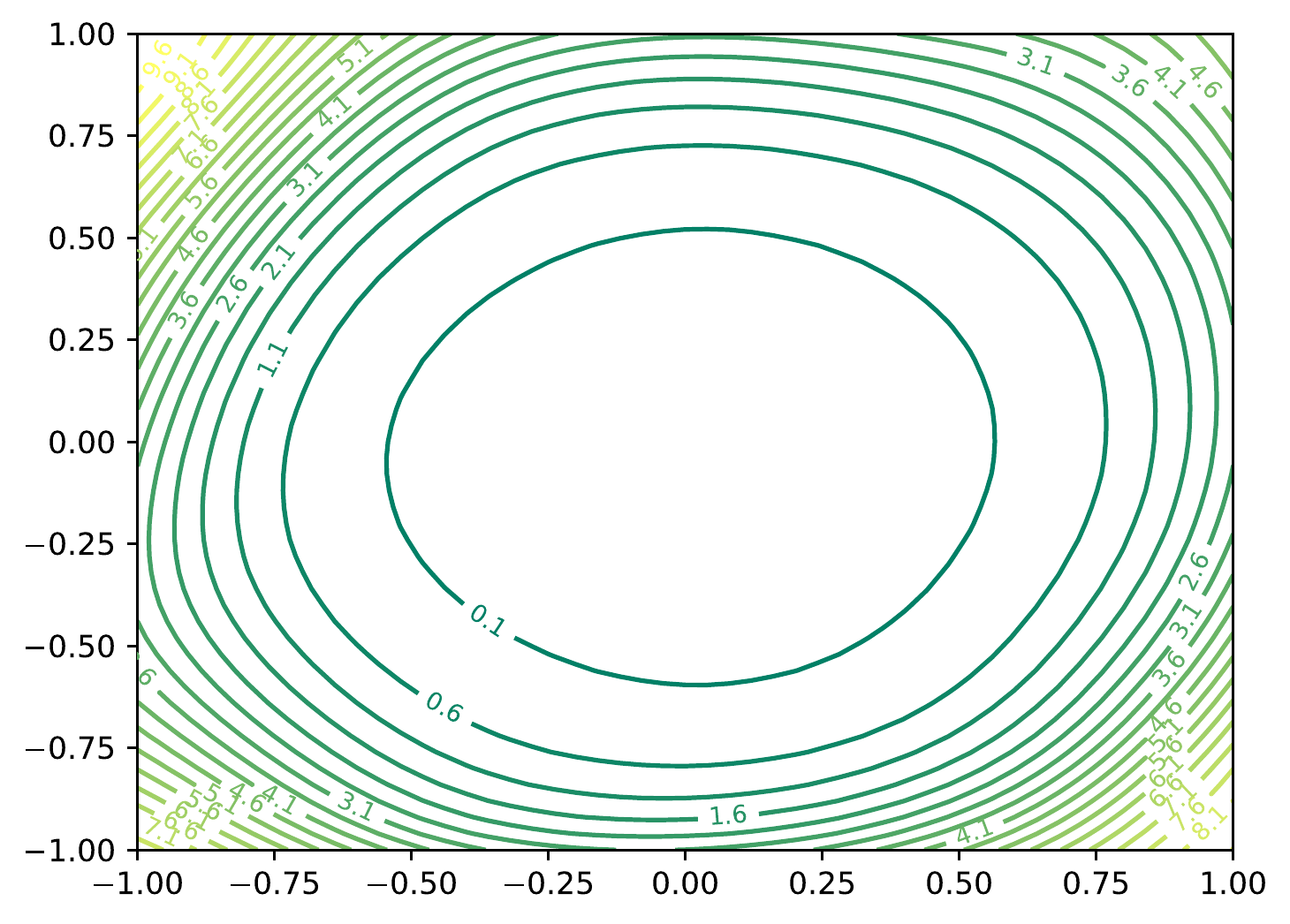}}
\subfloat[]{\includegraphics[width=0.33\textwidth]{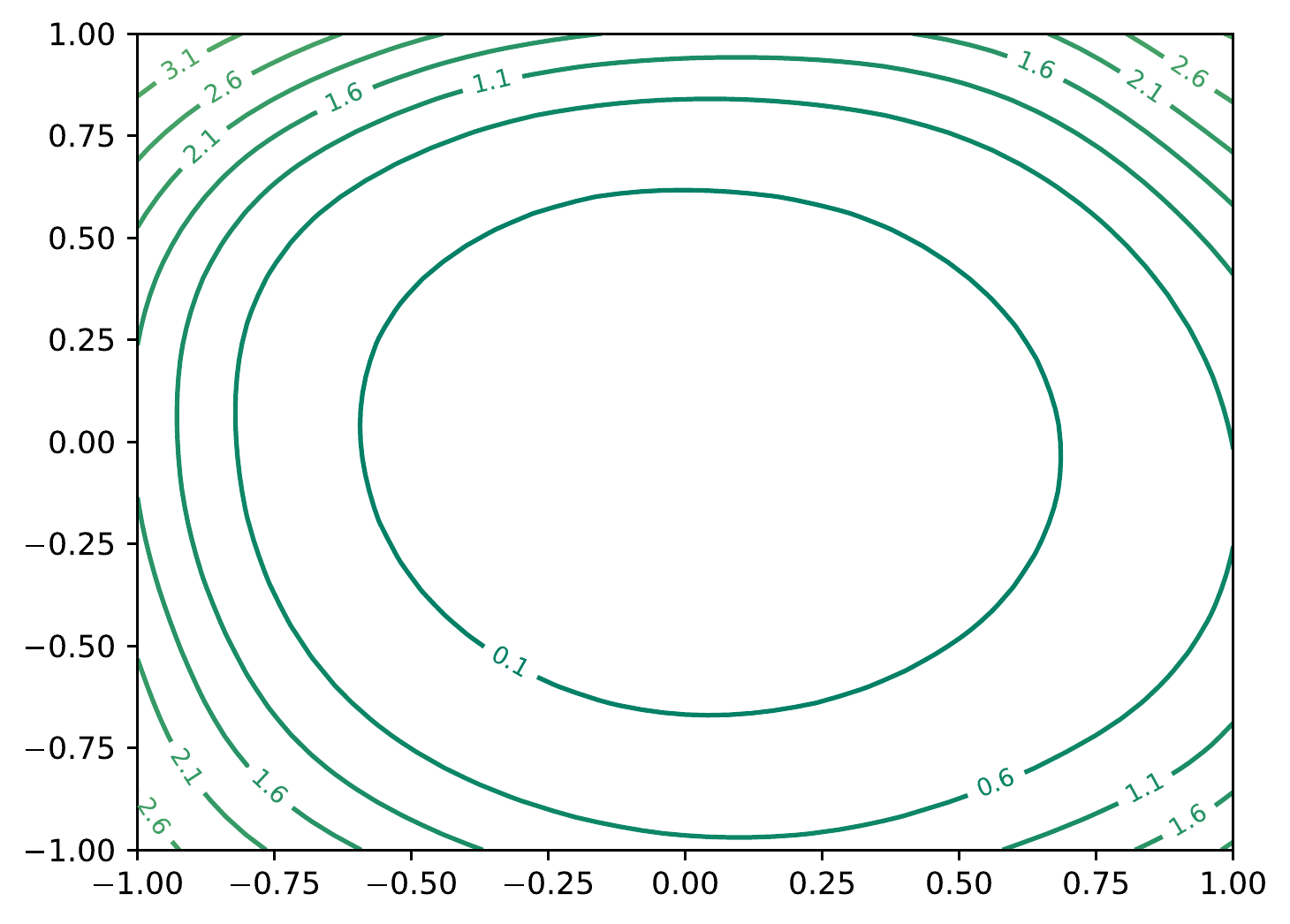}}
\subfloat[]{\includegraphics[width=0.33\textwidth]{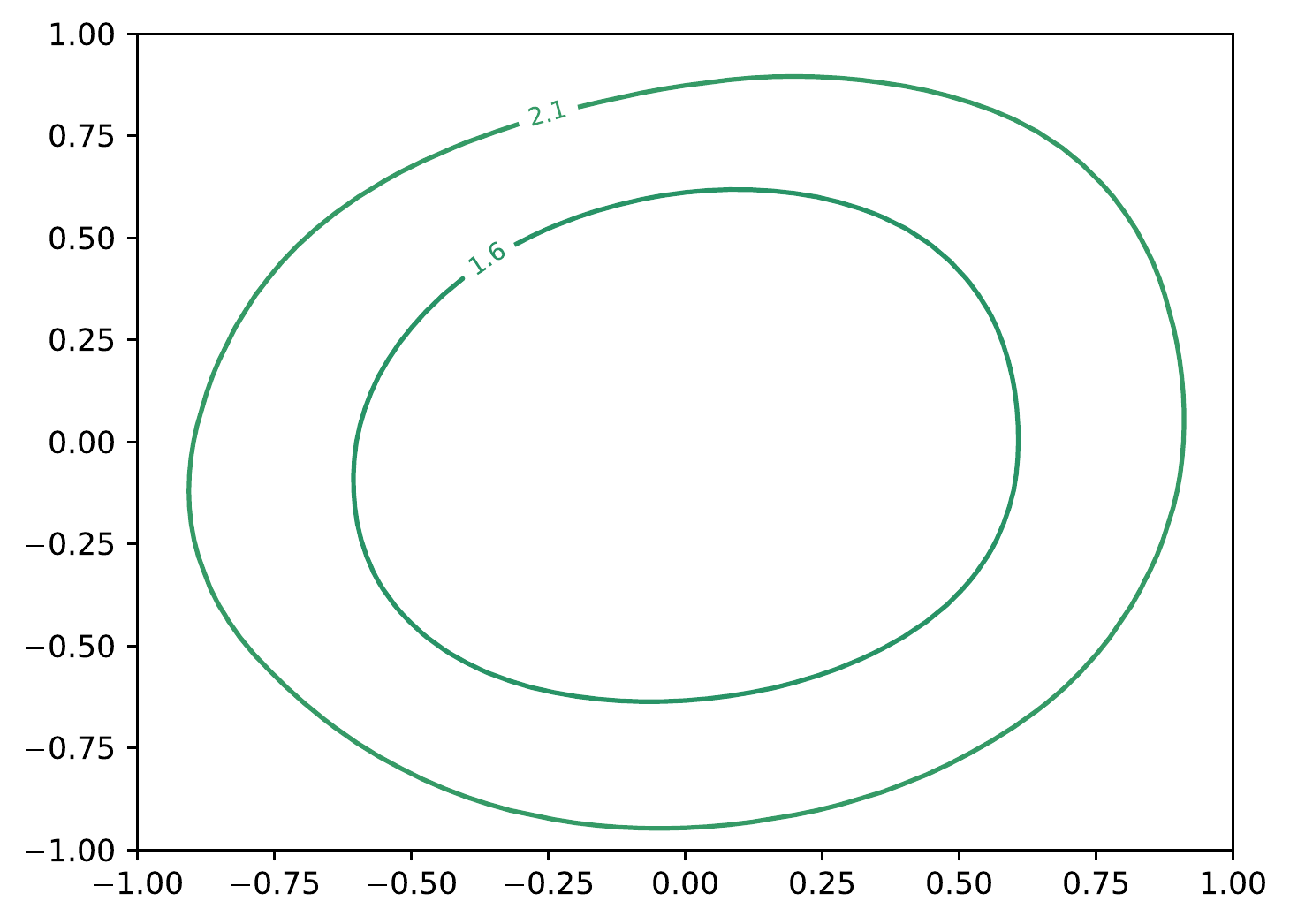}}
\begin{justify}
{Source: The Author (2022). 3D loss surfaces and 2D loss contours as proposed in \cite{DBLP:conf/nips/Li0TSG18}. Loss landscapes for ResNet34 trained on CIFAR10. (a, d) SoftMax; (b, e) IsoMax+; and (c, f) DisMax\textsuperscript{$\dagger$}. Considering that IsoMax+ outperforms SoftMax and DisMax\textsuperscript{$\dagger$} outperforms IsoMax+, a less steep 3D inclination (i.e. a lower 2D contour concentration) provides increased robustness.}
\end{justify}
\label{fig:loss_landscape}
\end{figure*}

\looseness=-1
Fig.~\ref{fig:loss_landscape} presents the 3D loss surfaces and 2D loss contours as studied in \cite{DBLP:conf/nips/Li0TSG18}. Considering that IsoMax+ overcomes SoftMax and that DisMax\textsuperscript{$\dagger$} outperforms IsoMax+, we conclude that improved robustness is obtained by less steep 3D inclination (i.e. a lower 2D contour concentration).

\hl{The above-mentioned conclusion may be somewhat unexpected, as we always were conducted to believe that losses with high inclinations are ``better''. It is the case probably because we were not used to thinking that we need losses that produce robust models rather than only losses that build accurate ones.}

%% file: chapters/5.conclusion.tex
\chapter{Conclusion}\label{chap:conclusion}

\begin{quotation}[Official Trailer]{The Lord of the Rings: The Return of the King}
``There can be no triumph without loss.\\
No victory without suffering. No freedom without sacrifice.''
\end{quotation}

\begin{quotation}[]{Friedrich Nietzsche}
``Become who you are!''
\end{quotation}


\begin{quotation}[Odes]{Horace}
``Seize the present, trust tomorrow e'en as little as you may.''
\end{quotation}



\looseness=-1
In this work, we mentioned that AI is the subfield of computer science that handles problems not well-fitted for the classical algorithmic programming paradigm. We also showed the relevance and limitations of machine learning. We presented the revolution that deep learning represents for the machine learning community. Finally, we showed the deep learning strengths and current major limitations such as reasoning, causal inference, interpretability, and robustness.

\looseness=-1
We exemplified how current deep networks present extremely confident predictions even when they are wrong. We also gave examples of how hard it is to apprehend when the system cannot reliably predict. Subsequently, we presented the \gls{ood} detection task, which is one of the challenges that best outlines the issue of reliability in deep learning. 
\\

\newpage
Finally, after identifying the SoftMax loss limitations, we present our proposals to tackle the \gls{ood} detection problem. We started from deep learning foundations such as loss function design and the principle of maximum entropy to develop solutions that do not rely on ad hoc techniques currently used to attack this problem. By doing this, we were able to achieve state-of-the-art \gls{ood} detection performance, avoiding drawbacks of previously proposed methods.

\section{Contributions}\label{sec:contributions}

\subsection{Isotropy Maximization Loss}\label{sec:contri_isomax}

Initially, we proposed IsoMax, a loss that is isotropic (exclusively distance-based) and produces high entropy posterior probability distributions in agreement with the maximum entropy principle. We additionally proposed the entropic score, which is a fast and efficient way to perform OOD detection using the information provided by all neural network output probabilities rather than just one, usually the case in current OOD detection approaches.

The proposed approach avoids the techniques, requirements, and side effects used by current methods. Networks trained using IsoMax loss produce accurate predictions, as no classification accuracy drop is observed compared to networks trained with SoftMax loss. Additionally, the models trained using IsoMax loss provide inferences that are fast and have energy and computational efficiency equivalent to models trained with SoftMax loss, making our solution viable from an economical and environmental point of view \citep{Schwartz2019GreenA}.

Moreover, our approach does not require \emph{hyperparameter tuning}, which means it does not require validation using unrealistic design-time access to OOD samples or the generation of adversarial examples. Regarding this point, we remember that \emph{all} experiments performed in this work always used the same constant value for the Entropic Scale.

Indeed, no hyperparameter tuning is required because the entropic scale is a global constant that is kept equal to ten for all combinations of datasets and models. Even if we call the entropic scale hyperparameter, the IsoMax loss does not involve hyperparameter tuning because the same constant value of entropic scale is used in all cases. This result is possible because it was shown in \cite{macdo2019isotropic,DBLP:journals/corr/abs-2006.04005} that the OOD detection performance exhibits a \emph{well-behaved dependence} on the entropic scale regardless of the dataset and model.

Furthermore, our solution does not require feature extraction, metric learning, or hyperparameter tuning. In other words, no extra procedures other than typical neural network training are required. Considering our approach avoids ad hoc techniques and associated troublesome requirements and undesired side effects, we say that IsoMax loss works as a SoftMax loss \emph{drop-in replacement} and that the overall solution is seamless. 

\newpage
We provide theoretical foundations based on the maximum entropy principle to explain why our \emph{seamless and principled approach} works. Substantial experimental evidence confirms our theoretical assumptions and shows that our solution presents \emph{competitive} OOD detection performance in addition to avoiding limitations of current methods.

Considering that the replacement of the SoftMax loss by the IsoMax loss improves OOD detection performance, we conclude that the general low performance of networks in this regard is due to SoftMax loss drawbacks rather than limitations of the models.


Nevertheless, when the cited requirements and side effects are not a concern for a particular real-world application, the mentioned (or other) techniques may be combined with IsoMax loss to try to achieve even higher OOD detection performance.

For example, loss enhancement techniques, such as outlier exposure or background samples, may be readily adapted to work with the IsoMax loss. Another promising approach could be using recent data augmentation techniques \citep{NIPS2019_9540, DBLP:conf/iccv/YunHCOYC19} or strategies based on pretrained models \citep{Sastry2019DetectingOE}.

\subsection{Enhanced Isotropy Maximization Loss}

We improved the IsoMax loss function 
by replacing its original distance with what we call the \emph{isometric distance}. Additionally, we proposed a zero computational cost minimum distance score. Experiments showed that these modifications achieve higher OOD detection performance while maintaining the desired benefits of IsoMax loss (i.e., absence of hyperparameters to tune, no reliance on additional/outlier/background data, fast and power-efficient inference, and no classification accuracy drop).

Similar to IsoMax loss, after training using the proposed IsoMax+ loss, we may apply inference-based approaches (e.g., ODIN, Mahalanobis, Gram matrices, outlier exposure, energy-based) to the pretrained model to eventually increase the overall OOD detection performance even more. Thus, the IsoMax+ loss is a replacement for SoftMax loss but not for OOD methods that may be applied to pretrained models, which may be used to improve even more the OOD detection performance of IsoMax+ loss pretrained~networks.

There is no drawback in training a model using IsoMax+ loss instead of SoftMax loss or IsoMax loss, regardless of planning to subsequently use an inference-based OOD detection approach to further increase OOD detection performance. Therefore, instead of competitors, the OOD detection approaches that may be applied to pretrained models are actually complementary to the proposed approach \citep{macdo2019isotropic, DBLP:journals/corr/abs-2006.04005}. IsoMax+ loss achieves a better baseline performance than SoftMax loss or IsoMax loss to construct future OOD detection methods.

\subsection{Distinction Maximization Loss}

We proposed DisMax by improving the IsoMax+ with the \emph{enhanced} logits and the \emph{Fractional Probability Regularization}. We also presented a novel \emph{composite} score called MMLES for OOD detection by combining the maximum logit+, the \emph{mean} logit+, and the negative entropy of the network output. We present a \emph{simple and fast temperature scaling procedure} performed after training that makes DisMax produce a high-performance uncertainty estimation. Our experiments showed that the proposed method commonly outperforms the current approaches simultaneously in classification accuracy, inference efficiency, uncertainty estimation, and out-of-distribution detection.

\newpage\section{Benefits of Entropic Losses}

\hlf{Regarding the first research question, we showed that it is possible to perform state-of-the-art OOD replacing just the loss, and we also provided many options for scores. Concerning the second question, we showed that the maximum entropy principle offers solid theoretical motivation for the proposed approaches. Finally, to answer the third question and to summarize our achievements, here we present some features that are common to the proposed solutions (i.e., the entropic losses and the novel scores):}

\begin{enumerate}

\item Entropic losses do not present classification accuracy drop: The experiments show that training with entropic losses does not produce a classification accuracy drop. Other training-based approaches like Generalized ODIN (G-ODIN) require losing some training data for validation, consequently producing a classification accuracy drop compared with training the SoftMax loss. Losing training data for validation is particularly harmful in the critical low labeled data regime. Classification accuracy drop is usually extremely undesired in many cases in practice.

\item Models trained using entropic losses do not produce slow and energy inefficient inferences: Approaches such as ODIN, full Mahalanobis, and G-ODIN incorporate input preprocessing, making the solution at least four times slower and energy inefficient. We should prefer environment-friendly and low-energy consumption solutions for real-world, large-scale adoption.

\item Entropic losses do not require additional data collection: Solutions like outlier exposure require extra data collection, while the proposed approach does not. If additional data is available, outlier exposure (or similar data-driven techniques) may be used to enhance the performance of our method.

\item IsoMax, IsoMax+ and DisMax do not require hyperparameter tuning: It makes the solution readily available to be fast adopted in practice and significantly increases the overall baseline OOD performance in many areas and applications.

\looseness=-1
\item Scalability: considering the solution consists of a SoftMax loss drop-in replacement, we have strong reasons to believe that the proposed approach scales well for large image datasets.

\item Domain-Agnostic: Considering that IsoMax, IsoMax+, and DisMax work at loss level and avoids data augmentation, it may potentially be applied to text.

\item Easy of use: considering that the proposed solutions work as seamless SoftMax loss drop-in replacement, using it is as simple as replacing two lines of code.

\item Compatibility with existing inference-based approaches: inference-based approaches (e.g., ODIN, vanilla or full Mahalanobis, outlier exposure, Gram matrices) may be applied to our losses pretrained model just like they are applied to SoftMax loss pretrained models to achieve even higher OOD detection performance. Unlike inference-based approaches that usually increase the computational cost of performing inferences or OOD detection, the proposed solutions produce inferences as fast and computationally efficient as pure SoftMax loss-trained models.

\item Competitive state-of-the-art performance even operation under more restrictive conditions and producing no side effects: Our results show that the proposed approaches overcome ODIN despite avoiding inefficient inference. It also usually overcomes full Mahalanobis, avoiding inefficient inferences and validations using adversarial examples. Moreover, it is competitive or overcomes outlier exposure without relying on extra data. 

\item Simplicity: The simplicity and solid foundations (e.g., distance-based loss, the principle of maximum entropy, isometric distances, minimum distance score) suggest that the proposed solutions generalize well.

\item Compatibility with existing Bayesian, ensemble, and uncertainty estimation techniques: As the entropic losses work as a SoftMax loss drop-in replacement, all available Bayesian, ensemble, and uncertainty estimation techniques may be immediately combined with it to start from a much better baseline than SoftMax loss.

\item Considering that the minimum distance is computed during the classification phase and that this value is reused as the score to perform OOD detection, we say that this is a zero computational cost out-of-distribution detection. Hence, unlike other approaches (e.g., Gram matrices), no extra computation is required.

\item Do not require feature extraction or training additional models: After the neural network training, the solution is readily available. It is unnecessary to perform feature extraction or additional train models on the extracted features.
\end{enumerate}

\newpage\section{Future Works}\label{sec:future_work}

In future work, we intend to analyze the behavior of our solutions when dealing with natural corruptions and perturbations. Some studies have shown that the performance of deep neural networks deteriorates when submitted to corruptions and perturbations that may occur naturally \citep{DBLP:conf/iclr/HendrycksD19}. Therefore, in some applications, it is important to design robust solutions in these cases. Robustness in the presence of distribution shifts\footnote{\url{https://wilds.stanford.edu/}} is a relevant case that we also intend to study.

Besides the natural corruptions and perturbations, we also plan to evaluate the performance of our solutions when dealing with intentionally manipulated corruptions and perturbations. In other words, we intend to analyze the behavior of our solutions when submitted to the so-called adversarial attacks \citep{DBLP:conf/icic/WangWL21}. Making the solutions robust against adversarial examples may allow its usage in critical scenarios in which this is a real concern.

We also plan to study the behavior of the proposed losses when dealing with other types of media (e.g., audio), structured data (tabular and temporal data), and tasks (e.g., object detection \citep{DBLP:journals/corr/abs-2104-11892}). We may broaden their practical applications by showing that the proposed losses also present satisfactory behavior in such cases.

Finally, we intend to verify the performance of our approach using transformer-based models, regardless of being pretrained or fine-tuned using the proposed losses and scores. Considering that vision transformers are recently becoming a reality, evaluate the performance of our proposals in such cases is becoming important.

%% file: appendix/entropic_out-of-distribution_detection_ijcnn2021.tex
\chapter*{Appendix A - Entropic Out-of-Distribution Detection}
\label{ap:entropic_ood}
\addcontentsline{toc}{chapter}{Appendix A - Entropic Out-of-Distribution Detection}

\includepdf[pages=-,pagecommand={\thispagestyle{plain}}]{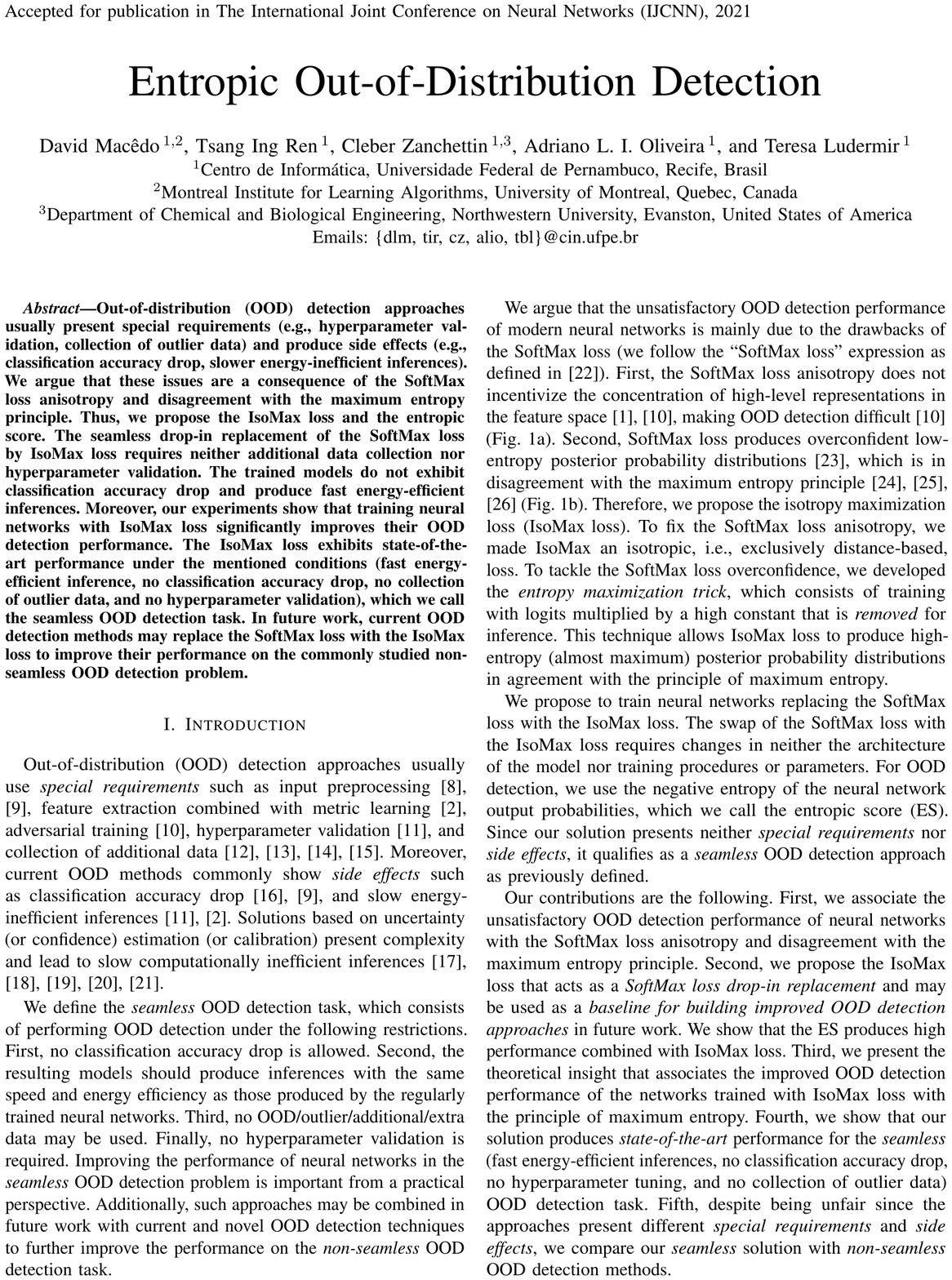}

%% file: appendix/entropic_out-of-distribution_detection_ieee2021.tex
\chapter*{Appendix B - Entropic Out-of-Distribution Detection: Seamless Detection of Unknown Examples}
\label{ap:entropic_ood_seamless_detection}
\addcontentsline{toc}{chapter}{Appendix B - Entropic Out-of-Distribution Detection: Seamless Detection of Unknown Examples}



\includepdf[pages=-,pagecommand={\thispagestyle{plain}}]{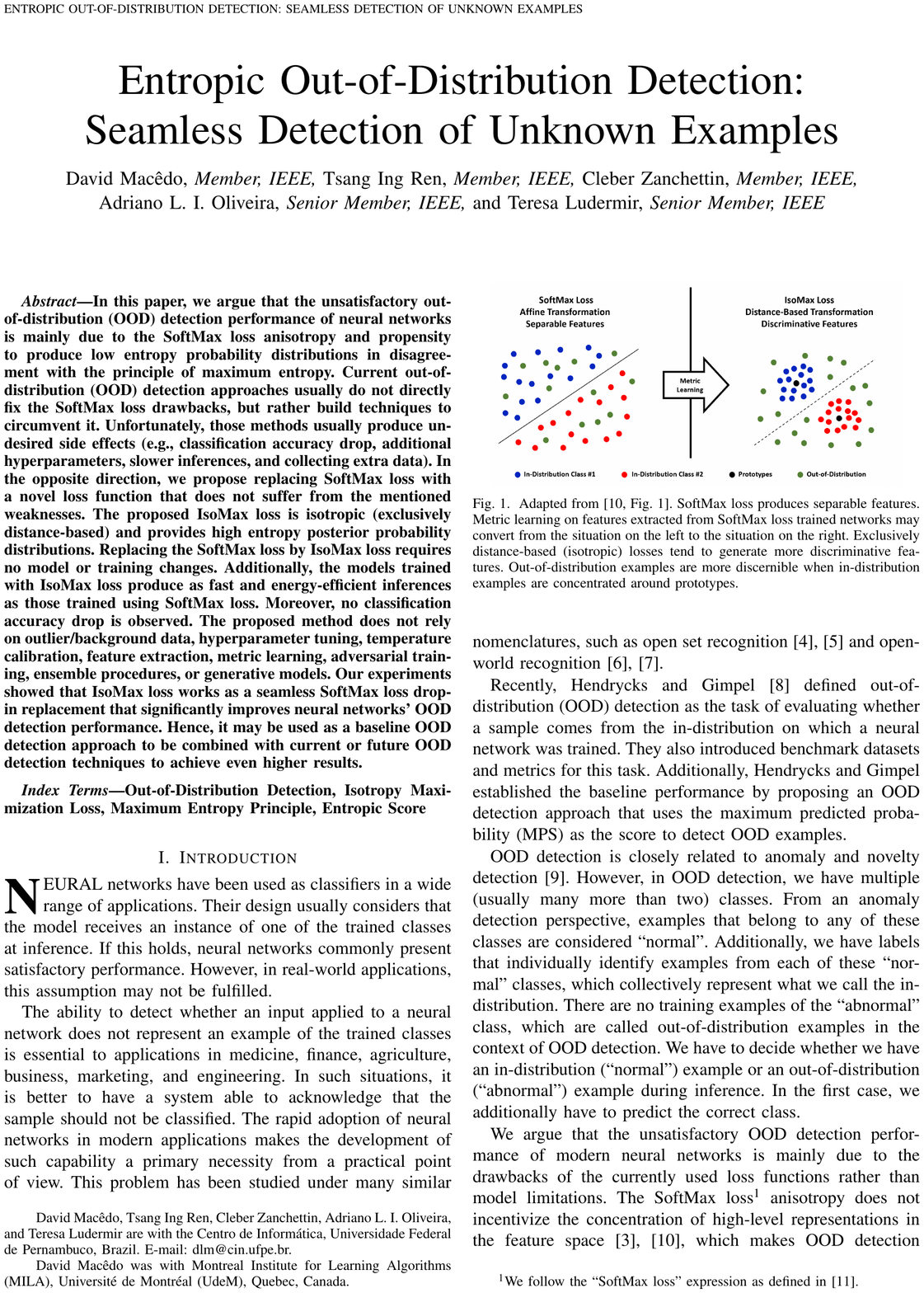}

%% file: appendix/enhanced_isotropy_maximization_loss.tex
\chapter*{Appendix C - Enhanced Isotropy Maximization Loss}
\label{ap:enhanced_isotropy_maximization_loss}
\addcontentsline{toc}{chapter}{Appendix C - Enhanced Isotropy Maximization Loss}

\includepdf[pages=2-12,pagecommand={\thispagestyle{plain}}]{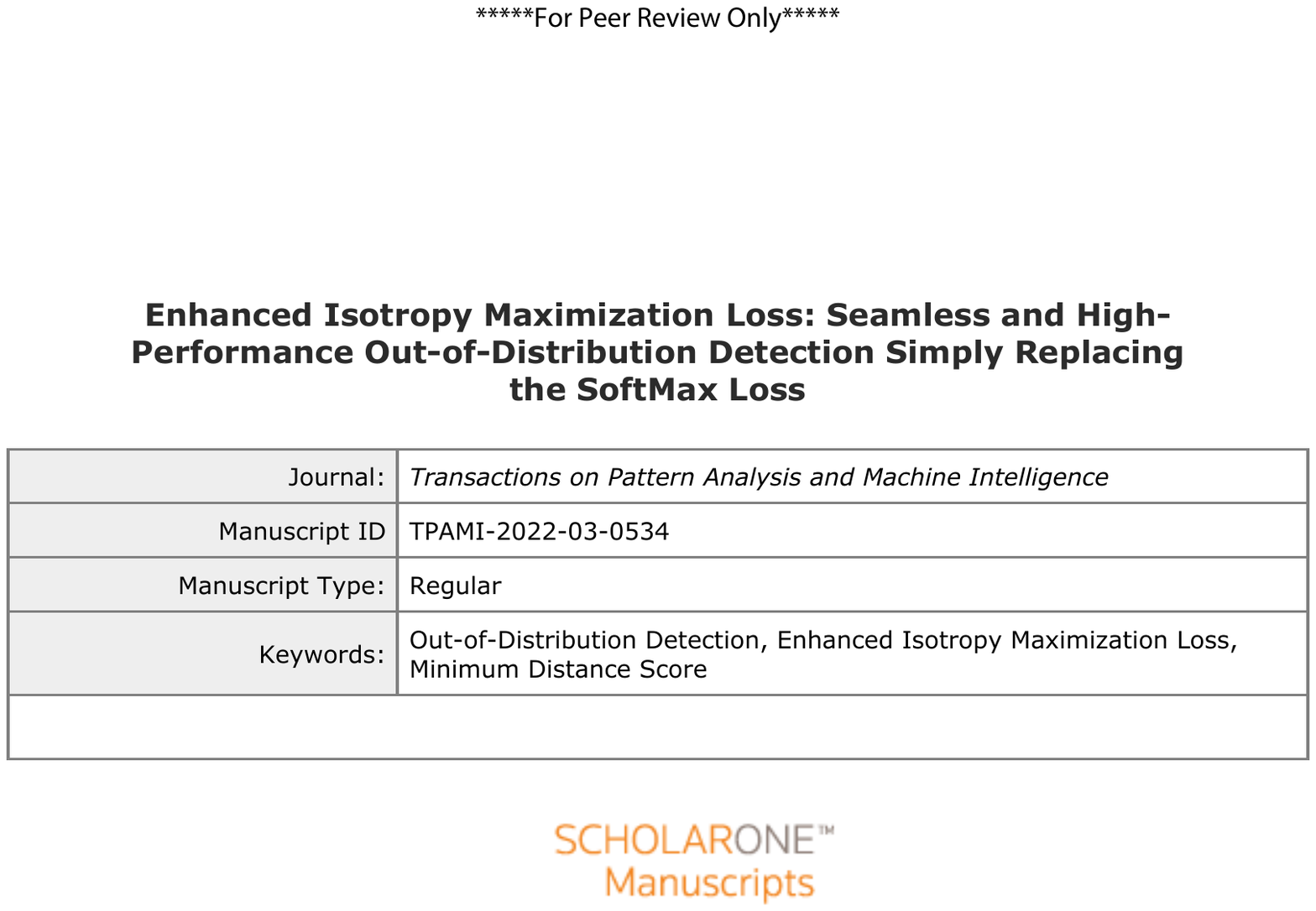}

%% file: appendix/distinction_maximization_loss.tex
\chapter*{Appendix D - Distinction Maximization Loss}
\label{ap:dismax}
\addcontentsline{toc}{chapter}{Appendix D - Distinction Maximization Loss}

\includepdf[pages=-,pagecommand={\thispagestyle{plain}}]{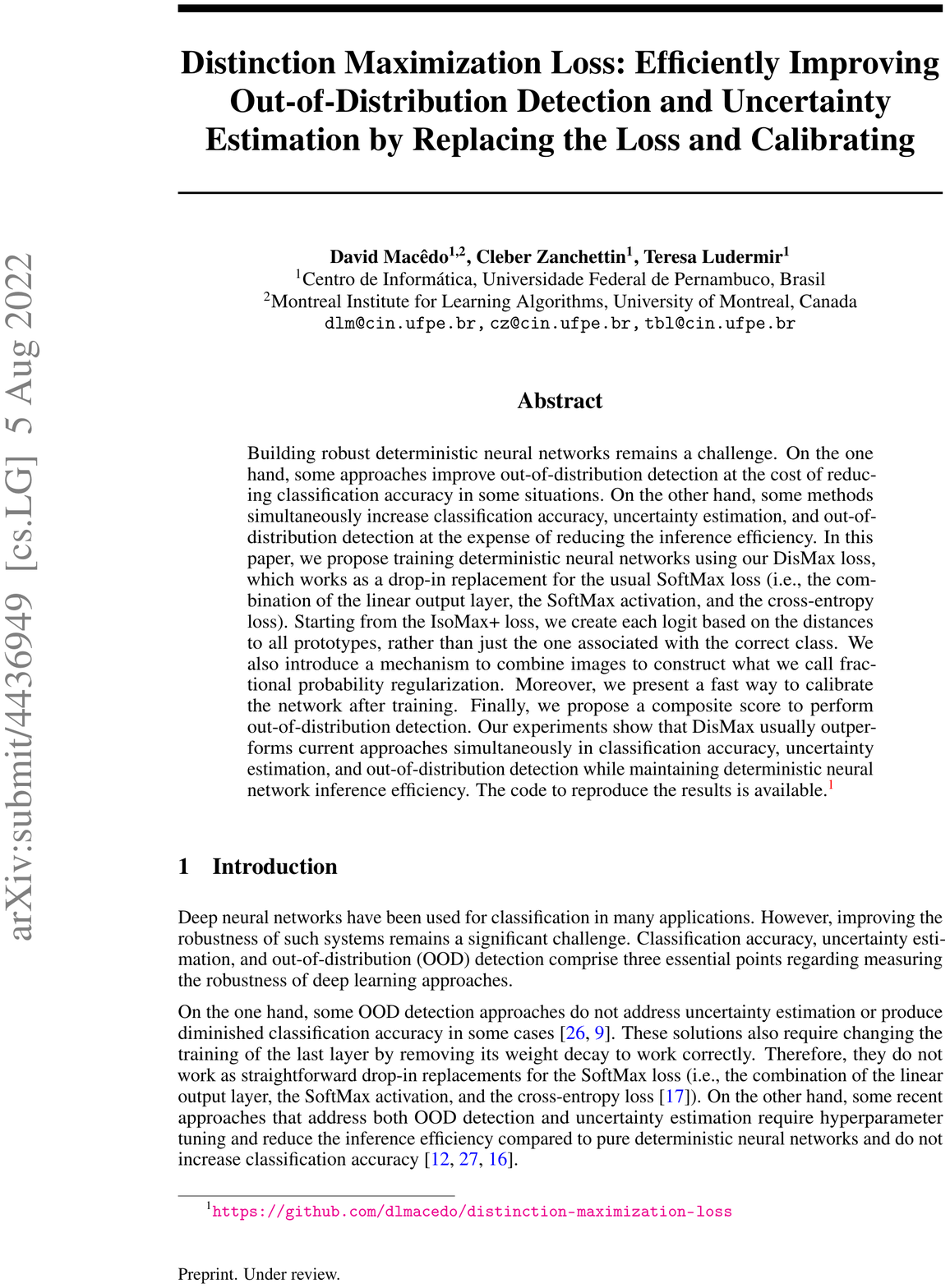}